\documentclass[twoside]{aghdpl}
\usepackage[utf8]{inputenc}
\usepackage[noend]{algpseudocode}
\usepackage{graphicx}
\usepackage{wrapfig}
\usepackage{listings}
\usepackage{multirow}
\usepackage{multicol}
\usepackage{amssymb}
\usepackage{amsmath}
\usepackage{txfonts}
\usepackage{adjustbox}
\usepackage[caption=false]{subfig}
\usepackage[titlenumbered,ruled]{algorithm2e}
\usepackage{times}
\usepackage{latexsym}
\usepackage{array}
\usepackage{empheq}
\usepackage{longtable}
\usepackage{cancel}
\usepackage{appendix}
\usepackage{hyperref}
\usepackage{xspace}
\usepackage[cal=boondoxo]{mathalfa}

\newcommand{\appendixsubsection}[1]{
    \stepcounter{subsection}
    \subsection*{\Alph{section}.\arabic{subsection}\hspace{1em}{#1}}
}
\newcommand{\appref}[1]{
    \hyperref[#1]{\Alph{section}.\arabic{subsection}}
}

\newtheorem{lemma}{Lemma}
\newtheorem{definition}{Definition}

\titleformat{\paragraph}
{\normalfont\normalsize\bfseries}{\theparagraph}{1em}{}
\titlespacing*{\paragraph}
{0pt}{3.25ex plus 1ex minus .2ex}{1.5ex plus .2ex}

\author{Bartosz Minch}
\shortauthor{Minch, B.}

\titlePL{W poszukiwaniu najbardziej wydajnej i oszczędzającej pamięć wizualizacji danych wielowymiarowych}
\titleEN{In search of the most efficient and memory-saving visualization of high dimensional data}

\shorttitlePL{W poszukiwaniu najbardziej wydajnej i oszczędzającej pamięć wizualizacji danych wielowymiarowych}
\shorttitleEN{In search of the most efficient and memory-saving visualization of high dimensional data.}

\disciplinePL{Nauki Inżynieryjno-Techniczne}
\disciplineEN{Engineering and Technology}
\fieldPL{Informatyka Techniczna i Telekomunikacja}
\fieldEN{Information and Communication Technology}
\thesistypePL{Rozprawa doktorska}
\thesistypeEN{Doctoral thesis}

\authorPL{mgr inż. Bartosz Minch}
\supervisorPL{prof. dr hab. inż. Witold Dzwinel}
\assistantSupervisorPL{dr Dariusz Jamróz}
\authorEN{Bartosz Minch}
\supervisorEN{Witold Dzwinel, Ph.D, Professor}
\assistantSupervisorEN{Dariusz Jamróz, Ph.D}

\universityPL{Akademia G\'{o}rniczo-Hutnicza}
\universityEN{AGH University of Science and Technology}

\date{2023}

\departmentPL{Instytut Informatyki}
\departmentEN{Institute of Computer Science}

\facultyPL{Wydział Informatyki, Elektroniki i Telekomunikacji}
\facultyEN{Faculty of Computer Science, Electronics and Telecommunications}

\abstract{abstract}

\begin{document}

\titlepages

\chapter*{Abstract}

Interactive visual exploration of large, high-dimensional datasets plays a very important role in various fields of science, which requires aggregated information about mutual relationships between numerous objects. It enables not only to recognize their important structural features and forms, such as clusters of vertices and their connectivity patterns, but also to assess their mutual relationships in terms of position, distance, shape, and connection density. The structural properties of these large datasets can be scrutinized throughout their interactive visualization.  We argue that the visualization of very high-dimensional data is well approximated by the two-dimensional ($2D$) problem of embedding undirected \textit{k}NN-graphs. In the advent of the big data era, the size of complex networks (datasets) $G(V,E)$ ($|V|$$=$$M$$\sim$$10^{6+}$) represents a great challenge for today's computer systems and still requires more efficient $ND$$ \rightarrow$$2D$ dimensionality reduction (DR) algorithms. The existing DR methods, which involve more computational and memory complexities than $O(M)$, are too slow for interactive manipulation on large networks that involve millions of vertices. We show that high-quality embeddings can be produced with minimal time\&memory complexity. Very efficient IVHD (interactive visualization of high-dimensional data) and IVHD-CUDA algorithms are presented and then compared to the state-of-the-art DR methods (both CPU and GPU versions): t-SNE, UMAP, TriMAP, PaCMAP, BH-SNE-CUDA, and AtSNE-CUDA. We show that the memory and time requirements for IVHD are radically lower than those for the baseline codes. For example, IVHD-CUDA is almost 30 times faster in embedding (without the $kNN$ graph generation procedure, which is the same for all methods) of one of the largest datasets used, YAHOO ($M$$=$$1.4\cdot10^6$), than AtSNE-CUDA. We conclude that at the expense of a minor deterioration of embedding quality, compared to baseline algorithms, IVHD well preserves the main structural properties of $ND$ data in $2D$ for a significantly lower computational budget. We also present a meta-algorithm that enables using any unsupervised DR method in supervised fashion and as a result allows for flexible control of global and local properties of the embedding. Thus, our methods can be a good candidate for an interactive visualization of truly big data ($M$$=$$10^{8+}$) and can be further used to inspect and interpret relationships between alternative representations of observations learned by artificial neural networks (ANN). Additionally, we have provided a framework for testing the trade-off between preservation of global structure and local structure of DR.

\clearpage
\blankpage

\chapter*{Streszczenie}

Interaktywna, wizualna eksploracja dużych, wielowymiarowych zbiorów danych odgrywa bardzo ważną rolę w różnych dziedzinach nauki, która wymaga zagregowanej informacji o wzajemnych relacjach między wieloma obiektami. Umożliwia ona nie tylko rozpoznanie ich istotnych cech i form strukturalnych, takich jak skupiska wierzchołków i ich wzorce połączeń, ale także ocenę ich wzajemnych relacji w zakresie położenia, odległości, kształtu i gęstości połączeń. Twierdzimy, że wizualizacja wielowymiarowych danych ($ND$) jest dobrze przybliżana przez problem dwuwymiarowego ($2D$) osadzania nieukierunkowanych grafów najbliższych sąsiadów (ang. $k$NN graphs). W dzisiejszych czasach, rozmiar złożonych sieci (zbiorów danych) $G(V,E)$ ($|V|$$=$$M$$\sim$$10^{6+}$) stanowi duże wyzwanie dla dzisiejszych systemów komputerowych i wciąż wymaga bardziej wydajnych algorytmów osadzania danych wielowymiarowych. Istniejące metody redukcji wymiarowości danych, które wymagają większej złożoności obliczeniowej i pamięciowej niż $O(M)$, są zbyt wolne do interaktywnej manipulacji na dużych sieciach obejmujących miliony wierzchołków. Pokazujemy, że osadzenia wysokiej jakości mogą być produkowane przy minimalnej złożoności czasowej i pamięciowej. Przedstawiamy bardzo wydajne algorytmy IVHD oraz IVHD-CUDA, a następnie porównujemy je z najnowszymi i najpopularniejszymi metodami redukcji wymiarowości (zarówno w wersji dla CPU, jak i GPU): t-SNE, UMAP, TriMAP, PaCMAP, BH-SNE-CUDA oraz AtSNE-CUDA. Pokazujemy, że wymagania pamięciowe i czasowe dla IVHD są radykalnie niższe niż dla kodów bazowych. Na przykład, IVHD-CUDA jest prawie 30 razy szybsza w osadzaniu (bez procedury generowania grafu najbliższych sąsiadów, która jest taka sama dla wszystkich metod) jednego z największych użytych zbiorów danych, YAHOO ($M$$=$$1.4$$\cdot$$10^6$), niż AtSNE-CUDA. Stwierdzamy, że kosztem niewielkiego pogorszenia jakości osadzania, w porównaniu do algorytmów bazowych, IVHD dobrze zachowuje główne własności strukturalne danych $ND$ w $2D$ przy znacznie niższym budżecie czasowym. Przedstawiamy również meta-algorytm, który umożliwia wykorzystanie dowolnej nienadzorowanej metody osadzania danych w sposób nadzorowany i w rezultacie pozwala na elastyczną kontrolę globalnych i lokalnych własności osadzenia. Dzięki temu, nasze metody mogą być dobrym kandydatem do interaktywnej wizualizacji naprawdę dużych zbiorów danych ($M$=$10^{8+}$) i mogą być dalej wykorzystywane do inspekcji i interpretacji zależności pomiędzy alternatywnymi reprezentacjami obserwacji wyuczonymi przez sztuczne sieci neuronowe (ANN).

\clearpage
\blankpage

\chapter*{Acknowledgements}

I express my deepest gratitude to Professor Witold Dzwinel for guiding and encouraging me during my research efforts. I also thank Dr. Dariusz Jamróz for helpful discussions and invaluable advice that greatly improved this thesis.

I dedicate this work to my parents, my family and my friends.

The research for this thesis was supported by:

\begin{itemize}
    \item Funds allocated to the AGH University of Science and Technology by the Polish Ministry of Science and Higher Education. 
    \item Infrastructure and computing resources made available by ACK Cyfronet PL-Grid.
\end{itemize}

\clearpage
\blankpage

\chapter*{Table of Contents}
\tableofcontents

\chapter*{Glossary of symbols}
\addcontentsline{toc}{chapter}{Glossary of symbols}

\begin{table}[h!]
    \begin{tabular}{ l l }
    \hspace{1cm} $\mathnormal{x}$ & a scalar \\
    \hspace{1cm} $\mathbf{x}$ & a vector \\
    \hspace{1cm} $\mathbf{X}$ & a matrix (dataset) \\
    \hspace{1cm} $\mathbb{R}$ & the set of real numbers \\
    \hspace{1cm} $\mathbb{R}^{N}$ & high-dimensional space ($N$$>>$$n$) \\
    \hspace{1cm} $\mathbb{R}^{n}$ & low-dimensional space ($n$$<<$$N$) \\
    \hspace{1cm} $|\mathbb{X}|$ & number of elements in $\mathbb{X}$ \\
    \hspace{1cm} $\mathbf{X}^{T}$ & transpose of matrix $\mathbf{X}$ \\
    \hspace{1cm} $x_{i}$ & i-th element of vector $\mathbf{x}$ \\
    \hspace{1cm} $\mathbf{X}_{i\cdot}$ & i-th row of matrix $\mathbf{X}$ \\
    \hspace{1cm} $\mathbf{X}_{\cdot j}$ & j-th row of matrix $\mathbf{X}$ \\
    \hspace{1cm} $x_{ij}$ & element $(i,j)$ of matrix $\mathbf{X}$ \\
    \hspace{1cm} $||\mathbf{x}||_{1}$ & $\mathcal{l}_{1}$ norm of vector $\mathbf{x}$ \\
    \hspace{1cm} $||\mathbf{x}||_{2}$ & $\mathcal{l}_{2}$ norm of vector $\mathbf{x}$ \\
    \hspace{1cm} $\frac{\delta x}{\delta y}$ & partial derivative of $y$ with respect to $x$ \\
    \hspace{1cm} $\nabla_{x}y$ & gradient of $y$ with respect to $x$ \\
    \hspace{1cm} $G(V,E)$ & graph with sets of $\mathbb{V}$ vertices and $\mathbb{E}$ edges \\
    \hspace{1cm} $\mathcal{N} (m, s^{2})$ & a Gaussian distribution with mean $m$ and variance $s^{2}$ \\
    \hspace{1cm} $\mathcal{M}$ & Riemannian manifold
    \end{tabular}
\end{table}

\clearpage
\blankpage

\chapter{Introduction}

In the era of big data, most of the data generated every day is represented directly as: 1) \textit{structured data} - $ND$ vectors, or as 2) \textit{unstructured data} - embedded in $ND$ Euclidean space (e.g. pictures, graphs, text). It is common for popular deep learning approaches to use data augmentation to satisfy the need to train a huge number of parameters without overfitting, the growing amount of data requires some key data reduction methods for different motivations. In general, dimensionality reduction (DR) is useful for: 1) better storage efficiency, 2) shorter computation time, 3) creating better data representation, 4) removing outliers, and 5) better recognition performance. Data reduction approaches are divided into two categories (Fig. 1.1), that is, dimensionality reduction and numerosity reduction, which reduce the dimensionality and sample size of the data, respectively.

\begin{figure}[ht]
    \begin{center}
        \includegraphics[clip,width=0.5\textwidth]{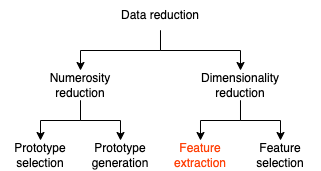}
        \label{fig:data_reduction_taxonomy}
        \caption{Taxonomy of data reduction. This dissertation tackles the red part.}
    \end{center}
\end{figure}

\section{Motivation}

In this work, we investigate the ways in which dimensionality reduction (DR) methods can benefit from a common framework that would allow us to determine the reliability of low-dimensional mapping (embedding). We focus on visualization using the scatter plot technique. This type of data visualization is useful in illustrating the relationships that exist between variables and can be used to identify trends or correlations in the data. Other types of data visualization (such as charts, heat maps, graphs) have been omitted from our considerations. Furthermore, to visualize the structured data, we only need a distance matrix between consecutive objects. For unstructured data, we first need to create a vector representation in Euclidean space, which allows for: 1) not using complicated metrics to calculate distances, and 2) such data can be further used to train a neural network. Additionally, it would allow one to trace the changes in embedding quality for modern algorithms used in the visualization of high-dimensional information, which is obtained by reducing the amount of information each individual object (that is being embedded) has about the entire data set and reducing the computational complexity of modern DR algorithms.  In particular, we focus on two main areas where, as we have found, DR methodologies have the biggest procedural holes.

First, we attempt to establish a common framework that would allow for a proper quality evaluation of different embeddings. We show how this can be useful in the problem of interpreting data that are represented by sparse, unstructured, high-dimensional feature vectors. This type of data often arises in fields such as social networks, web indexing, gene sequencing, and biomedical analysis. Interactive visualization allows for: 1) instant verification of a number of hypotheses, 2) precise matching of data mining tools to the properties of the data investigated, 3) adapting the optimal parameters to machine learning algorithms, and 4) selecting the best data representation. 

Second, we investigate how to considerably decrease the time\&memory complexity of visualization of high-dimensional data\footnote{While the term "high-dimensional" is sometimes used to refer to data described by at least four features, here we consider a feature vector to be high-dimensional when its dimensionality is on the order of at least $10^{2}-10^{4}$} with minimal decrease of the embedding quality. This would enable one to analyze visually radically larger datasets than those of the state-of-the-art visualization algorithms. The 2D data embedding would allow for both insight into the large data structure and its interactive exploration through direct manipulation of all or part of the data set. In this way, the shapes and mutual locations of single data and classes can be observed, irrelevant data samples can be removed, and outliers can be identified. The multiscale structure can be explored visually by changing data embedding strategies and visualization modes (e.g., the type of the loss function) and zooming in and out selected fragments of 2D (3D) data mapping. We also investigated a centroid meta-procedure, which utilizes state-of-the-art clustering algorithms \cite{clustering_algorithms}, allowing for even better parameterization of the final visualization in terms of its local and global properties.

Finally, using the research described in the first two paragraphs, our objective is to solve another well-recognized challenge, namely to interpret a model prediction when training and analyzing deep learning models \cite{zhang2021survey}. Although these techniques are useful, few works have been published \cite{rauber2016,schulz_deepview_2020,xianglin_dvi_2022} to explain how model predictions are developed during the training process. Obtaining training information, which evolves over time, can be useful, but it is difficult to abstract the evolving part of the underlying model. The following subsidiary questions arise: 1) How does the training process incrementally improve the model? 2) How does the model gradually make trade-offs to fit some samples while sacrificing others? 3) How does the model handle matching and learning difficult samples? To answer these questions, we visualize the relationship between learned representations of observations and the relationship between artificial neurons. We show how visualization can provide highly valuable feedback for network designers.

\section{The thesis and goals}
The overarching purpose of this dissertation is to investigate an optimal DR method that could effectively and interactively visualize truly large and high-dimensional datasets, when very strict time\&memory performance regimes are implemented. To this end, we follow these crucial assumptions.

\begin{enumerate}
    \item We focus on embedding high-dimensional data into low-dimensional spaces. For this purpose, each object is represented as a point in order to visualize the structure of the dataset, choose the appropriate representation of the data, choose the appropriate metric to represent the manifold, and match meta-parameters of machine learning algorithms. 
    \item The method is expected to enable direct analysis of data (through interactive intervention in the data set, selection of loss functions, etc.).
    \item We are not interested in preprocessing the data beforehand, i.e., we work on the final vector representation of the data and the defined distance metric.
\end{enumerate}

In the dissertation, we also refer to supervised embedding methods and present the application of the proposed method to the analysis and interpretation of the performance of neural networks.

The thesis of the dissertation is as follows. 

\textit{Computational complexity of modern DR algorithms for embedding $N$ high-dimensional data vectors into low-dimensional spaces might be reduced to $O(kN)$. It can be achieved by: a) drastically reducing the neighbor information of each object, b) introducing the binary distance between objects, and c) using an efficient loss function minimization method, which does not drastically degrade the embedding quality.}

The main contributions of this dissertation are as follows:

\begin{itemize}
    \item An extensive overview and analysis of existing DR methods and the challenges they face.
    \item Highlight the advantages and disadvantages of the simplest method (IVHD) in a variety of possible contexts compared to far more sophisticated approaches.
    \item GPU implementation, which allowed one to visualize large high-dimensional datasets ($N$$\sim$$10^{6+}$) in a reasonable amount of time.
    \item Introducing significant improvements to IVHD that improve the global and local properties of visualization.
    \item Implementation of the DR library \cite{viskit}, which allows us to efficiently measure the reliability of low-dimensional (embedded) data representations.
    \item Comparison of methods for complex low-dimensional data (in addition to medium and large datasets).
    \item The proposition of a centroid-based meta-procedure that allows any unsupervised method to be used in supervised fashion.
    \item Application of the IVHD method for inspection and interpretation of inter-epoch and inter-layer DNN behavior.
\end{itemize}

All the research described above has made it possible to develop an optimal method that can be used to visualize very large datasets and investigate neural networks in an interactive way.

\section{The structure of dissertation}

\noindent The dissertation is organized as follows. 

Chapter 2 introduces the state-of-the-art division of DR methods and describes the classical ones. Then, we focus on two currently dominant DR methods: UMAP and t-SNE. Both are currently the fundamental algorithms that are employed as the basis for many new ideas in high-dimensional data visualization.

In Chapter 3 we focus on reducing the dimensionality reduction problem to a graph visualization. We introduce the IVHD method, which is a modification of the classical MDS method and verify how binary and Euclidean distances affect the quality of DR algorithms. Additionally, optimization methods are presented and compared to force-directed method implemented in IVHD. In the last subsection, we present the improvements that have been made to the IVHD method that have improved its local and global properties.

Chapter 4 introduces the common framework for conducting DR experiments. We define the quality measures that are used for all of the methods. Furthermore, we describe the baseline methods used for IVHD comparisons. The experiments are divided into two subsections, depending on the size of analyzed datasets.

In Chapter 5, IVHD-CUDA is introduced. It is a CUDA implementation of the IVHD method. The chapter includes benchmarks for the IVHD-CUDA and SOTA methods implemented in a CUDA environment (AtSNE and BH-SNE-CUDA).

Chapter 6 presents meta-platform for supervised visualization of high-dimensional data. It allows any unsupervised DR method to be used in supervised fashion.

In Chapter 7 we evaluate our research by applying it to inspection and interpretation of Artificial Neuron Networks (ANNs). We investigate how to visualize the relationships between learned representations and between neurons in networks.

Finally, in Chapter 8 we conclude the dissertation and discuss further directions for research.
\chapter{Dimensionality reduction}
\label{chapter:dr}

In this chapter, we introduce several important baseline DR methods and explain the technical background. It also explains
the open problems in data reduction which are tackled in this thesis. In Section \ref{sec:structured_unstructured_data} we focus on structured and unstructured data that can be embedded with DR methods. In Section \ref{sec:dimensionality_reduction} we introduce the reasons and related work for dimensionality reduction. We present baseline methods for specific subcategories of dimensionality reduction types. Sections \ref{sec:umap} and \ref{sec:sne_heuristics} introduce stochastic neighbor embedding heuristics and uniform manifold approximation and projection (UMAP) \cite{umap}, which are the state-of-the-art methods for data visualization.

\begin{figure}[ht]
    \begin{center}
        \includegraphics[clip,width=0.75\textwidth]{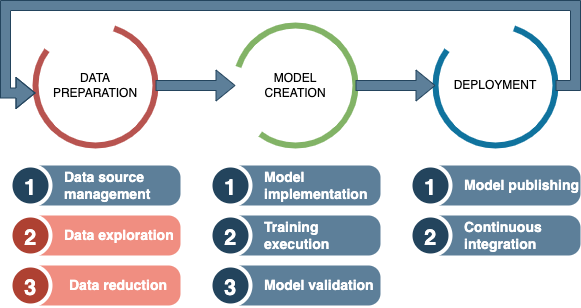}
        \label{fig:dr_in_ml_scheme}
        \caption{A taxonomy of AI systems infrastructure. This dissertation focuses on two key stages of the data preparation step: data exploration and data reduction, which are shown as the red parts of the diagram. Data reduction taxonomy was described in the first chapter of this thesis.}
    \end{center}
\end{figure}

Dimensionality reduction methods can be divided into three categories: 1) spectral methods, 2) probabilistic methods and 3) methods based on neural networks \cite{ghojogh2020}, which have a geometric, probabilistic, and information-theoretic viewpoint on dimensionality reduction. These categories are based on the generalized eigenvalue problem, latent variables, and neural networks, respectively. In feature extraction for dimensionality reduction, required for interactive visualization of high-dimensional data, which this thesis focuses on, a new set of features is found for better representation or data discrimination.

\begin{definition} 
Let $\mathbf{X}=\{x_i\}^{M}_{i=1}$ and $\mathbf{Y}=\{y_i\}^{M}_{i=1}$ be, respectively, sets of M instances in a high-dimensional space ($\mathbf{X}\subset{\mathbb{R}}^{N}$) and corresponding embeddings in a low-dimensional space ($\mathbf{Y}\subset{\mathbb{R}}^{n}$), where $N$$>>$$n$.
\end{definition}

\begin{definition}
In supervised learning, there are $\{(x_i, l_i)\}_{i=1}^{M}$ meaning that every instance $x_i$ has a corresponding label $l_i \subset \mathbb{R}^{\mathcal{l}}$, where $\mathcal{l}$ is the dimensionality of the label.
We can then form the label matrix $L = [l_1, \dots, l_i] \subset \mathbb{R}^{M x \mathcal{l}}$. In classification, each instance belongs to one of $|\mathnormal{C}|$ where $\mathnormal{C}$ is the set that includes labels from classes. The cardinality of the set of instances in class $c$ is denoted by $n_c$.
\end{definition}

\begin{definition}
Dimensionality reduction (DR) is defined as a mapping:

\begin{center}
    \textbf{B}: $\mathbf{X}$$\rightarrow$$\mathbf{Y}$
\end{center}

$\textbf{B}$ can be perceived as a lossy compression of the data. It is carried out by minimizing a loss function $E(\|\mathbf{X}-\mathbf{Y}\|)$, where $\|.\|$ is a measure of the topological dissimilarity between $\mathbf{X}$ and $\mathbf{Y}$.  Due to the high complexity of the low-dimensional manifold, immersed in the $N$D feature space and occupied by data samples $x_{i}$, perfect embedding of $\mathbf{X}$ in the $n$D space is possible only for trivial cases.
\end{definition}

Dimensionality reduction (DR) tools in data visualization is a double-edged sword in understanding the geometric and neighborhood structures of data sets. Having the ability to efficiently visualize data sets can provide an understanding of the cluster structure and provide an intuition of distributional characteristics. However, it is well-known that DR results can be misleading, showing cluster structures that are simply not present in the original data, or showing that the observations are far apart in the projection space when they are close together in the original space. Thus, if we were to run several DR algorithms, we could get different results. It is not clear how we would determine which of these results, if any, give a reliable representation of the original data distribution. 

\section{Structured and unstructured data}
\label{sec:structured_unstructured_data}
From the point of view of interactive data visualization, the following properties of the datasets are important because they determine the calculation of the distance matrix, which is later used as input to the dimensionality reduction algorithms \cite{mishra2017}.

\textbf{Data dimensionality.} A popular and intuitive way to represent a given data set is to use a vector space model \cite{salton1979}. In a vector space model, observations are represented by a matrix $\mathnormal{M}$$\times$$\mathnormal{N}$ called a design matrix, in which each of the rows $M$ corresponds to an observation that is described by attributes $N$ (also called characteristics or variables). Interpreting the attributes depends, of course, on the nature of the data set. In the case of an image collection, an observation refers to an image that is defined by a list of pixel intensities or higher-order features, while text, for example, is often represented as a multi-collection of its words, the so-called bag-of-words (\textit{BOW}) representation. Regardless of the interpretation of the features, their $N$ number or the dimensionality of the data play an important role in determining the applicability of machine learning, data mining and data embedding methods. High-dimensionality is an ubiquitous property of modern datasets. Data with hundreds or even millions of features appear in various application domains such as 2D/3D digital image processing, bioinformatics, e-commerce, web crawling, social networks, mass spectrometry, text analytics, and speech processing. 

\textbf{Data sparsity.} It is a common property of many high-dimensional data sets. It is defined as the number of elements with zero values in a matrix $M$$\times$$N$ divided by the total number of elements $MN$. However, when working with highly sparse datasets, a more convenient term is data density, which is equal to one minus sparsity \cite{herlocker2004}. The zeros in the computational matrix may simply represent missing measurements, also denoted as null values or "NaN" values. 

\textbf{Data structure.} Structured data are data that have been predefined and formatted to a set structure before being placed in data storage, which is often called schema-on-write. The best example of structured data is the relational database: the data have been formatted into precisely defined fields, such as credit card numbers or addresses. Unstructured data are data stored in its native format and not processed until it is used, which is known as a schema-on-read. They come in a myriad of file formats, including emails, social media posts, presentations, chats, IoT (Internet of Things) sensor data, audio, video, and satellite imagery. The big data industry is growing rapidly, and so are the data it produces. The biggest challenge is to effectively harness the latent power of unstructured data, especially with the speed of information creation continuing to accelerate. 

\textbf{Distance matrix.} It is a non-negative, square matrix with elements $-\frac{1}{2}d_{ij}^{2}$ corresponding to estimates of some pairwise distance between the sequences in a set. The distance matrix $\mathbf{D}$ is Euclidean, when the $\frac{1}{2}N(N-1)$ quantities $d_{ij}$ can be generated as distances between a set of $M$ points, $\mathbf{X} (M \times N)$, in an Euclidean space of dimension $N$. The dimensionality of $\mathbf{D}$ is defined as the least value of $p=rank(\mathbf{X})$ of any generating $\mathbf{X}$. The most important properties of the Euclidean distance matrix follow from the fact that the Euclidean distance is a metric. Thus, the Euclidean matrix $\mathbf{D}$ has the following properties: 
\begin{itemize}
    \item all elements on the diagonal $d_{ii}=0$, hence $trace(\mathbf{D})=0$,
    \item $\mathbf{D}$ is symmetric ($d_{ij}=d_{ji}$),
    \item $\sqrt{d_{ij}} \leq \sqrt{a_{ik}} + \sqrt{a_{kj}}$ (by the triangle inequality),
    \item $d_{ij} \geq 0$,
    \item in dimension $k$, an Euclidean distance matrix has rank less than or equal to $k+2$.
\end{itemize}

Additionally, it is worth mentioning that distances do not have to meet the formal definition of distance. Dimensionality reduction methods often use the so-called proximity matrix, which measures similarity or dissimilarity between data.

\section{Manifold learning}
\label{sec:dimensionality_reduction}

Manifold learning methods \cite{roychowdhury2011} play a prominent role in non-linear dimensionality reduction and other tasks involving high-dimensional data sets
with low intrinsic dimensionality. Many of these methods are graph-based: they associate a vertex with each data point and a weighted edge with each
pair. 

A manifold is a generalization of curves and surfaces to higher dimensions. It is locally Euclidean (Def. \ref{def:locally_euclidean_manifold}) in that every point has a neighborhood, called a chart, homeomorphic (Def. \ref{def:homeomorphism}) to an open subset of $\mathbb{R}^{N}$. The coordinates on a chart allow one to perform computations as if in Euclidean space, so many concepts from $\mathbb{R}^{N}$, such as differentiability, point derivations, tangent spaces, and differential forms, carry over to a manifold \cite{loring2011}. A good example of a manifold is the Earth. Locally, at each point on the surface of the Earth, we have a 3-D coordinate system: two for location and the last one for altitude. Globally, it is a 2-D sphere in a 3-D space.

\begin{definition}
A continuous map $F: X \rightarrow Y$ is a homeomorphism if it is bijective and its inverse $F^{-1}$ is also continuous. If two topological spaces admit a homeomorphism between them, we say that they are homeomorphic: they are essentially the same topological space.
\label{def:homeomorphism}
\end{definition}

\begin{definition}
A topological space $\mathcal{M}$ is locally Euclidean of dimension $N$ if every point $p$ in $\mathcal{M}$ has a neighborhood $\mathcal{U}$ such that there is a homeomorphism $\phi$ from $\mathcal{M}$ onto an open subset of $\mathbb{R}^N$. We call the pair $(\mathcal{U}, \phi : \mathcal{U} \rightarrow \mathbb{R}^N)$ a chart, $\mathcal{U}$ a coordinate neighborhood or a coordinate open set, and $\phi$ a coordinate map or a coordinate system in $\mathcal{U}$. We say that a chart $(\mathcal{U}, \phi)$ is centered at $p \in \mathcal{U}$ if $\phi(p) = 0$.
\label{def:locally_euclidean_manifold}
\end{definition}

Geometric and topological relationships are fundamental to essentially every data analysis and machine learning task, since we use geometry to identify similarities and distinctive characteristics in the data. In classification, for example, data points that are similar (close to each other) are assigned to the same class, and data points that are significantly different (i.e., far apart in one or many of the feature dimensions) are assigned to different labels. We usually consider data as a finite set of $M$-dimensional vectors $\mathbf{X}$, sometimes called a point cloud. However, geometric and topological structures, such as metric spaces and manifolds, are continuous, not discrete. To discover any geometric or topological properties of the data, we fit a continuous shape to the data, and we must make certain assumptions about the underlying mathematical space we are working in. It is often simply assumed that our data lie in the standard Euclidean space with the typical Euclidean metric, and the data is analyzed by referencing a global, external coordinate system. However, many interesting and important structures that arise are actually non-Euclidean, and by fitting a continuous shape (such as a manifold) to the data, we can translate our data analysis task from an external, global coordinate system (possibly having very high dimensionality) into the intrinsic coordinate system defined by the assumed manifold structure itself. Manifolds offer a powerful framework for DR and there are several motivations for manifold learning and interactive data visualization as a result.
 
\begin{itemize}
    \item According to the manifold hypothesis \cite{fefferman2013}, the data usually exist in a subspace or submanifold (unless it is a random noise). Therefore, the entire $N$-dimensional space is not required and a large part of it is redundant information. We can find the best $n$-dimensional subspace to represent the data with the smallest possible reconstruction error.
    \item Manifold learning methods can provide more efficient feature extraction later used for classication, representation, clustering, or revealing patterns in data.
\end{itemize}

\section{Spectral dimensionality reduction}

Spectral DR methods come down to eigenvalue decomposition and the generalized eigenvalue problem \cite{ghojogh2019}. They use a geometric approach and unfold the manifold into a lower-dimensional subspace. The most well-known unsupervised methods with the spectral approach are the following: PCA \cite{pca}, classical MDS \cite{mds}, Sammon mapping \cite{sammon}, LLE \cite{roweis2000}, Isomap \cite{tenenbaum2001}.

\subsection{Unsupervised methods}

Unsupervised visualization refers to visualization based only on input data without corresponding output variables, or labels. The goal is for the system to generate its own model of the underlying structure or distribution.

\subsubsection{Principal Component Analysis (PCA)}
PCA \cite{pca}, one of the most widely used tools in data analysis and data mining, is also one of the most popular linear-dimensionality reduction methods. 

Assume that we have a matrix of centered data observations:
\begin{equation}
    \mathbf{X} = [x_1 - \mu, \dots, x_N - \mu]
\end{equation}

where $\mu$ denotes the mean vector. $\mathbf{X} \in \mathbb{R}^{N x M}$, where $N$ is the number of dimensions (dimensionality) and $M$ is the number of observations (numerosity). Their covariance matrix is given by:

\begin{equation}
    \mathbf{S}_{t} = \frac{1}{M} \sum_{i=1}^{M}(x_1 - \mu)(x_1 - \mu)^{T} = \frac{1}{M}\mathbf{X}\mathbf{X}^{T}
\end{equation}
\vspace{0.15cm}

In Principal Component Analysis (PCA), we aim to maximize the variance of each dimension by:

\begin{equation}
    \begin{aligned}
        \mathbf{W}_0 = & {\arg \max}_{\mathbf{W}} \;\; tr(\mathbf{W}^{T} \mathbf{S}_{t} \mathbf{W}) \\
                       & \text{subject to} \;\; \mathbf{W}^{T} \mathbf{W} = \mathbf{I}
    \end{aligned}
\end{equation}
\vspace{0.15cm}

The solution of Eq. 3 can be derived by solving:

\begin{equation}
    \mathbf{S}_{t} \mathbf{W} = \mathbf{W} \mathbf{\Lambda}
\end{equation}

Thus, we need to perform an eigenanalysis on $ \mathbf{S}_{t}$. If we want to keep $d$ principal components, the computational cost of the above operation is $O(dN^{2})$.

\begin{lemma}
    Let us assume that $\mathbf{B} = \mathbf{X}\mathbf{X}^T$ and $\mathbf{C} = \mathbf{X}^T\mathbf{X}$. It can be proven that $\mathbf{B}$ and $\mathbf{C}$ have the same positive eigenvalues $\mathbf{\Lambda}$ and, assuming that $M<N$, then the eigenvectors $\mathbf{U}$ of $\mathbf{B}$ and the eigenvectors $\mathbf{V}$ of $\mathbf{C}$ are related as $\mathbf{U} = \mathbf{X}\mathbf{V}\mathbf{\Lambda}^{-\frac{1}{2}}$.
    \label{lemma:pca}
\end{lemma}

Using Lemma \ref{lemma:pca} we can compute the eigenvectors $\mathbf{U}$ of $\mathbf{S}_{t}$ in $O(N^3)$. The eigenanalysis of $\mathbf{X}^{T}\mathbf{X}$ is denoted by:
\begin{equation}
    \mathbf{X}^{T} \mathbf{X} = \mathbf{V} \mathbf{\Lambda} \mathbf{V}^{T}
\end{equation}

Given that $\mathbf{V}^T\mathbf{V} = \mathbf{I}$ and $\mathbf{V}\mathbf{V}^T \neq \mathbf{I}$ the covariance matrix of $\mathbf{Y}$ is:
\begin{equation}
    \mathbf{Y} \mathbf{Y}^{T} = \mathbf{U}^T \mathbf{X} \mathbf{X}^{T} \mathbf{U} = \mathbf{\Lambda}
\end{equation}

The final solution of Eq. 2.3 is given as the projection matrix:
\begin{equation}
    \mathbf{W} = \mathbf{U} \mathbf{\Lambda}^{-\frac{1}{2}}
\end{equation}

\begin{algorithm}
    \caption{Principal Component Analysis scheme.} \label{alg:pca}
    \SetAlgoLined
    \SetKwInOut{Input}{Input}
    
    \Input{data matrix $\mathbf{X}$.}
    \textbf{procedure} PCA: \\
    \hskip1cm 1. Compute dot product matrix: $\mathbf{X}^{T} \mathbf{X}$. \\
    \hskip1cm 2. Eigenanalysis: $\mathbf{X}^{T} \mathbf{X} = \mathbf{V} \mathbf{\Lambda} \mathbf{V}^{T}$. \\
    \hskip1cm 3. Compute eigenvectors: $\mathbf{U} = \mathbf{X}\mathbf{V}\mathbf{\Lambda}^{-\frac{1}{2}}$. \\
    \hskip1cm 4. Keep specific numbers of first components: $\mathbf{U}_d = [u_1, \dots, u_d]$. \\
    \hskip1cm 5. Compute $d$ features: $\mathbf{Y} = \mathbf{U}_d^{T} \mathbf{X}$.
\end{algorithm}

PCA has been applied to a variety of applied problems, such as image processing, statistics, text mining, and facial recognition. However, there are obvious drawbacks to the method, clearly, one being that the centralized data are required to lie on a linear subspace or something very close to it. This is a very strong assumption, since it assumes that the variables of the data are correlated in a linear fashion, which is not true in many applications. The most common variations of PCA are the following:

\begin{itemize}
    \item Kernel PCA \cite{ghojogh2019_pca}, which increases the dimensionality of the data by mapping them to the feature space with a higher dimensionality in hopes that they fall on a linear manifold in that feature space.
    \item Supervised PCA \cite{ghojogh2019_pca}, which utilizes information concerning classes.
\end{itemize}

Furthermore, it is worth mentioning SiMultaneous PCA (SMPCA) \cite{liu2007}, ProGressive PCA (PGPCA) \cite{liu2007}, Successive PCA (SCPCA) \cite{liu2007} and PRioritized PCA (PRPCA) \cite{liu2007}.

\subsubsection{Multidimensional Scaling (MDS)}

MDS \cite{mds} aims to preserve the similarity (later distances) of the data points in the embedding space and in the input space. There are different types of multidimensional scaling technique, and in this section we will examine the classical MDS that was first introduced by Torgerson and Gower \cite{torgenson1958}. Classical MDS measures similarity using the Euclidean distance and the mapping $\boldsymbol{B}:\mathbf{Y}\rightarrow \mathbf{X}$ by minimizing the following cost function (stress), where for cMDS $k=1$ and $m=2$.

\begin{equation}\label{eq:cost_function_mds}
E(\|\mathbf{D}-\mathbf{d}\|)=\sum_{ij}^M w_{ij}\left(\delta_{ij}^k - d_{ij}^k\right)^m = ||\mathbf{D^X}-\mathbf{D^Y}||_{F}^{2} = ||(-\frac{1}{2}\mathbf{H}\mathbf{D^X}\mathbf{H}) - \mathbf{Y^T}\mathbf{Y}||^{2}_{F} ,
\end{equation}

\noindent which represents the error between the dissimilarities $\textit{D}$ and the corresponding distances $\textit{d}$, where: $i,j=\{1,\dots M\}$, $w_{ij}$ are weights, $||\cdot||_{F}$ is the Frobenius norm, and $k$, $m$ are the parameters.

\begin{algorithm}
    \caption{Multidimensional Scaling scheme.} \label{alg:mds}
    \SetAlgoLined
    \SetKwInOut{Input}{Input}
    
    \Input{proximity matrix $\mathbf{D} = [d_{ij}^{2}]$.}
    \textbf{procedure} MDS: \\
    \hskip1cm 1. Apply double centering: $\mathbf{B}=-\frac{1}{2}\mathbf{H}\mathbf{D}\mathbf{H}$ using the centering matrix \\
    \hskip1cm $\mathbf{H}$$=$$\mathbf{I}-\frac{1}{M}1_{M}1_{M}^T$. \\ 
    \hskip1cm 2. Find the $n$ largest eigenvalues $\lambda_1, \dots, \lambda_{n}$ and corresponding \\
    \hskip1cm eigenvectors $(v_1, \dots, v_n)$ of $\mathbf{B}$.\\
    \hskip1cm 3. $\mathbf{Y}$$=$$\mathbf{U}_{n}\Lambda_{n}^{\frac{1}{2}}$$=$$(\sqrt{\lambda_{1}}v_{1}, \dots, \sqrt{\lambda_{n}}v_{n})$$=$$(y_{1}, \dots, y_{M})^{T}$. \\
\end{algorithm}

The MDS algorithm described here is used in a wide variety of applications, such as surface matching \cite{bronstein2006}, psychometrics \cite{takane2006}, and more \cite{borg2005}. The major drawback of classical MDS is its sensitivity to noise and the fact that it does not work well when the underlying structure of the data is non-linear. For these reasons, many variations of MDS were developed later. Few of those are:
\begin{itemize}
    \item Generalized classical MDS \cite{ghojogh2020_mds}.
    \item Metric MDS \cite{ghojogh2020_mds} tries to preserve the distances of the points
in the embedding space rather than similarities.
    \item Non-metric MDS \cite{ghojogh2020_mds}, which rather than using a distance metric,
    $d_y(x_i, x_j)$, for the distances between points in the embedding space, uses $f(d_y(x_i, x_j))$ where $f(\cdot)$ is a non-parametric monotonic function.
    \item Sammon mapping \cite{ghojogh2020_mds} is a special case of metric MDS. Introduce changes to the MDS optimization formulation.
\end{itemize}

Additionally, MDS has inspired the non-linear manifold learning technique Isomap, which we will cover next.

\subsubsection{Isometric mapping (Isomap)}

Isomap \cite{tenenbaum2001} is a special case of the generalized classical MDS, which gives a closed form solution to the dimensionality reduction problem and uses the Euclidean distance as the similarity metric, while Isomap uses an approximation of geodesic distances. 

\begin{algorithm}
    \caption{Isomap scheme.} \label{alg:isomap}
    \SetAlgoLined
    \SetKwInOut{Input}{Input}
    
    \Input{\textit{k}-nearest neighbor graph $G(X,E)$.}
    \textbf{procedure} Isomap: \\
    \hskip1cm 1. Compute the shortest path distances between all pairs of vertices \\
    \hskip1cm $x_i$ and $x_j$ and store them in $\mathbf{S}_{ij}$. \\
    \hskip1cm 2. Create dissimilarity matrix  $\mathbf{D}$$=$$\mathbf{S}_{ij}^2$.\\
    \hskip1cm 3. Apply the MDS algorithm using $\mathbf{D}$ as input.
\end{algorithm}

The geodesic distance is the length of the shortest path between two points on the possibly curvy manifold. It is ideal to use the geodesic distance; however, calculating the geodesic distance is very difficult because it requires traversing from one point to another point on the manifold.  Isomap approximates the geodesic distance by pairwise Euclidean distances and finds the k-nearest neighbors ($k$NN) graph of the dataset. Then, the shortest path between two points, through their neighbors, is found using a shortest-path algorithm such as for example the Dijkstra algorithm \cite{cormen2009} or the Floyd-Warshall algorithm \cite{cormen2009} ($O(N^3)$).

Most Isomap variants are heavily focused on technical aspects of implementation to increase computation speed \cite{lei2010,taiguo2015}.

\subsubsection{Laplacian eigenmaps}

Laplacian eigenmaps \cite{laplacian_eigenmaps,laplacian_eigenmaps_2} attempt to capture information about the local geometry and reconstruct the global geometry from the local information. The locality-preserving character of the Laplacian eigenmap algorithm makes it relatively insensitive to outliers and noise. It is also not prone to short circuits, as only the local distances are used. The algorithm constructs a weighted graph with $k$ nodes,
one for each point, and a set of edges that connect neighboring points. The embedding map is provided by computing the eigenvectors of the graph Laplacian. There are two options for choosing the edge weights:

\begin{enumerate}
    \item \textbf{Heat kernel}: If $x_{i}$ and $x_{j}$ are connected by an edge, then set the edge weight as:
    \begin{equation}
        W_{ij} = \exp^{-\frac{||x_{i}-x_{j}||^{2}}{t}}
    \end{equation}
    otherwise, set $W_{ij}$$=$$0$.
    \item \textbf{Combinatorial}: $W_{ij}$$=$$1$ if the vertex $i$ and $j$ are connected by an edge and $W_{ij}$$=$$0$ otherwise. This choice of weights avoids the need to choose a parameter $t$, making it easier to apply.
\end{enumerate}

\begin{algorithm}
    \caption{Laplacian eigenmaps scheme.} \label{alg:laplacian_eig}
    \SetAlgoLined
    \SetKwInOut{Input}{Input}
    
    \Input{data matrix $\mathbf{X}$.}
    \textbf{procedure} Laplacian eigenmaps: \\
    \hskip1cm 1. Constructs a weighted graph $G(X, E)$. \\
    \hskip1cm 2. Choose weights of the edges $E$. \\
    \hskip1cm 3. Compute eigenvalues and eigenvectors $\mathbf{L} \phi^{i} = \lambda_{i} \mathbf{D} \phi^{i}$. \\
    \hskip1cm 4. Calculate embedding $\mathbf{Y}$: $y_{i} \mapsto (\phi_{1}^{1}, \dots, \phi_{n}^{n})$. \\
\end{algorithm}

Let $G(V, E)$ be the graph constructed according to the previous two steps. Furthermore, assume that $G$ is a connected graph; otherwise, apply the current step to each connected component of $G$. Compute eigenvalues and eigenvectors of the generalized eigenvalue problem $\mathbf{L}f = \lambda \mathbf{D} f$, where $\mathbf{D}$ is a diagonal weight matrix called a degree matrix, and its entries are $D_{ii} = \sum_{j \in N(x_{i})} W_{ji}$. We call the Laplacian matrix graph $\mathbf{L}$$=$$\mathbf{D}$$-$$\mathbf{W}$. By the spectral theorem, we know that the eigenvalues are real. We order the eigenvalues in increasing order $\lambda_{0} \leq \lambda_{1} \leq \dots \leq \lambda_{n}$, and let $\phi^{i}$ be the corresponding eigenvectors such that $\mathbf{L} \phi^{i} = \lambda_{i} \mathbf{D} \phi^{i}$. We leave out the zeroth eigenvector (since it is constant) and proceed with the embedding using the following map:
\begin{equation}
    \hspace{0.3cm} y_{i} \mapsto (\phi_{1}^{1}, \dots, \phi_{n}^{n}).
\end{equation}
where $\phi_{i}^{j}$ stands for $i$-th component of the $j$-th eigenvector. The idea behind the method is to map close together points to close together points in the new dimension reduced space.

There have been recent developments on the basic Laplacian eigenmap. Some examples are Laplacian Eigenmaps Latent Variable Model (LELVM) \cite{carreira2007}, robust Laplacian eigenmap \cite{roychowdhury2011} and Laplacian forest \cite{lombaert2014}.

\subsubsection{Locally Linear Embedding (LLE)}

LLE \cite{roweis2000} consists of three steps. First, it finds the k-nearest neighbors ($k$NN) graph of all training points. Then, it tries to find weights for reconstructing every point by its neighbors, using linear combination. Using the same weights, it embeds every point using a linear combination of its embedded neighbors. The main idea of LLE is to use the same reconstruction weights in the low-dimensional embedding space as in the high-dimensional input space. 

\begin{enumerate}
    \item A $k$NN graph is formed using pairwise Euclidean distance
    between the data points
    \item Linear reconstruction by the neighbors: We find the weights for the linear reconstruction of every point by its $k$NN. The optimization for this linear reconstruction in the high-dimensional input space is formulated as follows:
    \begin{equation}
        \epsilon (\Tilde{W}) := \sum_{i=1}^{M} || x_i - \sum_{j=1}^k \Tilde{w}_{ij}x_{ij}||_{2}^{2}
    \end{equation}
    where $\mathbb{R}^{Mxk} \ni \Tilde{W} := [\Tilde{w_{1}}, \dots, \Tilde{w_{n}}]^{T}$ includes the weights, $\mathbb{R}^{k} \ni \Tilde{w_i} := [\Tilde{w_{i1}}, \dots, \Tilde{w_{ik}}]^{T}$ includes the weights of linear reconstruction of the $i$-th data point using its $k$ neighbors, and $x_{ij} \in \mathbb{R}^{N}$ is the $j$-th neighbor of the $i$-th data point.
    \item Linear embedding: We embed the data in the low-dimensional embedding space using the same weights as in the input space. This linear embedding can be formulated as the following optimization problem:
    \begin{equation}
        \text{minimize} \sum_{i=1}^{M} ||y_{i} - \sum_{j=1}^{M} w_{ij}y_{j}||_{2}^{2} 
    \end{equation}
\end{enumerate}

\begin{algorithm}
    \caption{LLE scheme.} \label{alg:lle}
    \SetAlgoLined
    \SetKwInOut{Input}{Input}
    
    \Input{\textit{k}-nearest neighbor graph $G(X,E)$.}
    \textbf{procedure} Laplacian eigenmaps: \\
    \hskip1cm 1. Constructs a weighted graph $G(X, E)$. \\
    \hskip1cm 2. Linear reconstruction of each point by its $k$ neighbors. \\
    \hskip1cm 3. Linear embedding. \\
\end{algorithm}

The most known variations of LLE are the following:
\begin{itemize}
    \item Kernel LLE \cite{ghojogh2020_lle}, 
    \item Incremental LLE \cite{kouropteva2005} to handle online data by embedding new received data using the already embedded data.
    \item Landmark LLE \cite{vladymyrov2013} used in big data, that approximates the embedding of all points. using the embedding of some landmark.
\end{itemize}

Currently, the most widely used nonlinear dimension reduction algorithms (described in Section 1.4) are: 1) t-distributed Stochastic Neighbor Embedding (t-SNE) and 2) Uniform Manifold Approximation and Projection (UMAP). UMAP produces similar or better representations, as it preserves more global features of the data, and the performance of the algorithm itself, measured by the Procrustean measure \cite{gower2004} (a form of statistical shape analysis), is more stable. Furthermore, UMAP, in terms of both dimensionality and data size, is more efficient than t-SNE. However, to gain a good understanding of the modern algorithms that have emerged in recent years, we need to go back to the classical algorithms. An advantage of probabilistic methods is that they are relatively robust to noise because of their stochastic behavior.

\begin{figure}[ht]
    \begin{center}
        \includegraphics[clip,width=\textwidth]{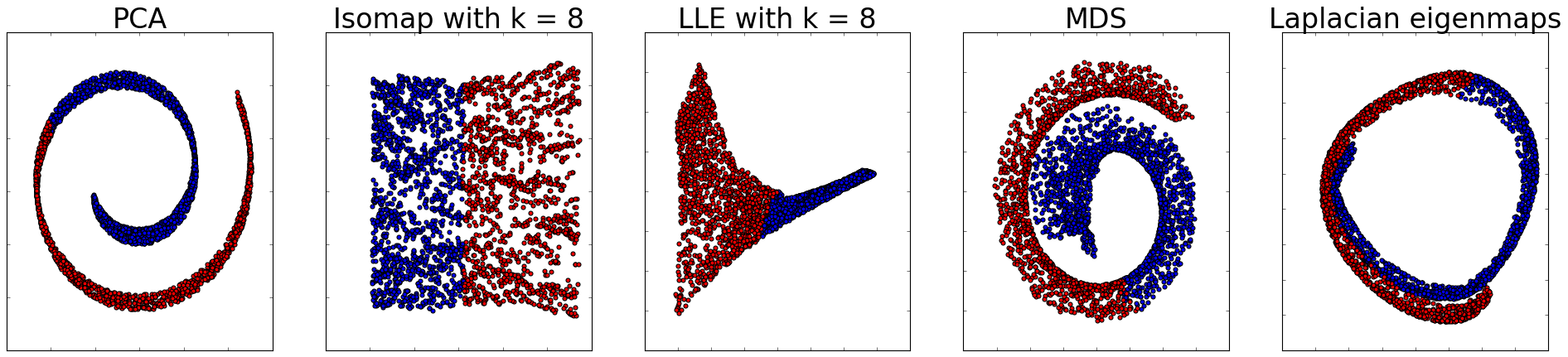}
        \label{fig:swiss_roll}
        \caption{Comparison of baseline state-of-the-art spectral methods using "swiss roll" dataset \cite{swissroll}.}
    \end{center}
\end{figure}

Figure 2.2 presents visualizations for the \textit{swiss roll} dataset created by the unsupervised spectral methods mentioned above. As shown, LLE and Isomap completely distort the overall structure of the dataset while reducing to 2-D. Best results are achieved by MDS and PCA.

\subsection{Supervised methods}

Unlike unsupervised visualization, supervised variants visualize the underlying structure or distribution of the data using corresponding output variables, or labels (algorithms know both input and output).

\subsubsection{Linear Discriminant Analysis (LDA)}

LDA \cite{tharwat2017}, unlike PCA, is a supervised method and computes linear directions that maximize separation between multiple classes. This is mathematically expressed as maximizing

\begin{equation}
    \begin{aligned}
        \mathbf{W}_0 = & {\arg \max}_{\mathbf{W}} \;\; tr(\mathbf{W}^{T} \mathbf{S}_{b} \mathbf{W}) \\
                       & \text{subject to} \;\; \mathbf{W}^{T} \mathbf{S}_{\mathbf{W}} \mathbf{W} = \mathbf{I}
    \end{aligned}
\end{equation}
\vspace{0.15cm}

$\mathbf{S}_{\mathbf{W}}$ being scatter matrix within-class and $\mathbf{S}_{b}$ being the scatter matrix between classes.

The solution of Eq. 2.8 is given from the generalized eigenvalue problem:
\begin{equation}
    \mathbf{S}_{b} \mathbf{W} = \mathbf{S}_{\mathbf{W}} \mathbf{W} \mathbf{\Lambda}
\end{equation}

\begin{algorithm}
    \caption{Linear Discriminant Analysis scheme.} \label{alg:lda}
    \SetAlgoLined
    \SetKwInOut{Input}{Input}
    
    \Input{data matrix $\mathbf{X}$.}
    \textbf{procedure} LDA: \\
    \hskip1cm 1. Find eigenvectors of of $\mathbf{S}_{\mathbf{W}}$ that correspond to non-zero eigenvalues \\ 
    \hskip1cm by performing eigen-analysis to $(\mathbf{I} - \mathbf{M})\mathbf{X}^T\mathbf{X}(\mathbf{I}-\mathbf{M})$ and computing.\\
    \hskip1cm 2. $\mathbf{U} = \mathbf{X}(\mathbf{I} - \mathbf{M})\mathbf{V}_{\mathbf{W}}\mathbf{\Lambda}_{\mathbf{W}}^{-1}$. \\
    \hskip1cm 3. Project the data as $\mathbf{\Tilde{X}_b} = \mathbf{U}^T \mathbf{X} \mathbf{M}$. \\
    \hskip1cm 4. Perform PCA on $\mathbf{\Tilde{X}_b}$ to find $\mathbf{Q}$. \\
    \hskip1cm 5. The total transform is $\mathbf{W} = \mathbf{U} \mathbf{Q}$.
\end{algorithm}

Given the properties of the scatter matrix \cite{oja2006}, the objective function of Eq. 2.8 can be expressed as:
\begin{equation}
    \begin{aligned}
        \mathbf{W}_0 = & {\arg \max}_{\mathbf{W}} \;\; tr(\mathbf{W}^{T} \mathbf{X}\mathbf{M}\mathbf{M}\mathbf{X}^T \mathbf{W}) \\
                       & \text{subject to} \;\; \mathbf{W}^{T} \mathbf{X}(\mathbf{I} - \mathbf{M})(\mathbf{I}-\mathbf{M})\mathbf{X}^T \mathbf{W} = \mathbf{I}
    \end{aligned}
\end{equation}
\vspace{0.15cm}

The optimization of this problem involves a procedure called Simultaneous Diagonalization. Let us assume that the final transform matrix has the form:
\begin{equation}
    \mathbf{W} = \mathbf{U} \mathbf{Q}
\end{equation}

\subsubsection{Supervised Multidimensional Scaling (SMDS)}
SMDS \cite{witten2011} compared to its unsupervised variant only changes the criterion, which is minimized. It includes two steps: 1) evaluating pairwise distances among entities based on their labels and constructing a new space based on the distance matrix using a projection strategy similar to MDS, and 2) establishing an explicit
linear relationship between the feature space and the new space.

The first step aims to construct a new space in which the distances among the entities approximate the distances among their labels. Unlike classic MDS, which establishes a low-dimensional new space based on the distance matrix generated from the features, SMDS establishes a new space based on the distance matrix generated from the labels. By replacing the distance matrix $\mathbf{D^X}$ in Eq. \ref{eq:cost_function_mds} with the distance matrix $\mathbf{D^{Label}}$, a new space represented by $\mathbf{Y^{Label}}$ can also be obtained as follows:

\begin{equation}\label{eq:chapter_2cost_function_smds}
\min_{Y^{Label}} (||-\frac{1}{2}HD^{Label}H - ((Y^{Label})^{T}Y^{Label})||_{F}^{2})
\end{equation}

The second step aims to establish a linear model $\mathbf{Y^{Label}} = \mathbf{W} \mathbf{X^{Feature}}$, which projects the high-dimensional features of each training entity (indicated by $\mathbf{X^{Feature}}$) into the new space obtained above (indicated by $\mathbf{Y^{Label}}$).

\begin{equation}\label{eq:chapter_2cost_function_smds_2}
\min_{W} (||\mathbf{Y^{Label}} - \mathbf{W} \mathbf{X^{Feature}}||_{F}^{2})
\end{equation}

\subsubsection{Enhanced Supervised Isomap (ES-Isomap)}

The ES-Isomap \cite{ribeiro2008} is based on the dissimilarity matrix constructed as:

\vspace{0.5cm}
\begin{equation}
    D(x_{i}|x_{j}) =
    \begin{dcases}
        (\frac{a-1}{a})^\frac{1}{2}, & \text{if} \; c_{i} = c_{j} \\
        a^{\frac{1}{2}} - d_0, & \text{if} \; c_{i} \ne c_{j}
    \end{dcases},
    \label{eq:es_isomap_diss_matrix}
\end{equation}
\vspace{0.5cm}

where $a = 1 / e^{-d_{ij}^{2} / \sigma}$, $\sigma$ is a smoothing parameter (set according to the 'density' of the data), $d_0$ is a
constant $(0 \leq d_0 \leq 1)$ and $c_i, c_j$ are the labels of the data. 

\begin{algorithm}
    \caption{ES-Isomap scheme.} \label{alg:es-isomap}
    \SetAlgoLined
    \SetKwInOut{Input}{Input}
    
    \Input{data matrix $\mathbf{X}$.}
    \textbf{procedure} ES-Isomap: \\
    \hskip1cm 1. Compute dissimilarity matrix using class labels in the distance matrix. \\
    \hskip1cm 2. Run Isomap using dissimilarity matrix from step 1. \\
    \hskip1cm 3. Learn the embedded mapping.\\
    \hskip1cm 4. SVM testing on new points data.
\end{algorithm}

\subsubsection{Supervised Laplacian Eigenmaps}

Similarly as in ES-Isomap, a crucial difference between supervised and unsupervised variants of Laplacian eigenmaps is how the neighbor graph is being calculated. In S-LapEig \cite{jiang2009} the dissimilarity distance between two points x$_i$ and $x_j$ is defined as:

\vspace{0.5cm}
\begin{equation}
    D(x_{i}|x_{j}) =
    \begin{dcases}
        \sqrt{1 - exp(-d_{ij}^2) / \beta}, & \text{if} \; c_{i} = c_{j} \\
        \sqrt{exp(d_{ij}^2)}, & \text{if} \; c_{i} \ne c_{j}
    \end{dcases},
    \label{eq:sle_diss_matrix}
\end{equation}
\vspace{0.5cm}

where $d_{ij}$ denotes the Euclidean distance between $x_i$ and $x_j$, $\beta$ is set to the average Euclidean
distance between all pairs of data points and $c_i, c_j$ are the labels of the data. 

\subsubsection{Supervised LLE (SLLE)}

SLLE \cite{kouropteva2003} was introduced to deal with datasets containing multiple (often disjoint) manifolds, corresponding to classes. For fully disjoint manifolds, the local neighborhood of a sample $x_i$ from class $c$ $(1 \leq c \leq C)$ should be composed of samples belonging to the same class only. This can be achieved by artificially increasing the precalculated distances between samples belonging to different classes, but leaving them unchanged if samples are from the same class:
\begin{equation}
    G' = G + \alpha \max (G) \Lambda, \; \alpha \in [0,1],
\end{equation}

where $G$ is the square distance matrix, $max(G)$ is the maximum entry of $G$, $\Lambda_{ij=1}$ if $x_i$ and $x_j$ belong to the same class and 0 otherwise. When $\alpha=0$, one obtains unsupervised LLE; when $\alpha=1$, the result is fully supervised.

\section{Probabilistic dimensionality reduction}

Probabilistic dimensionality reduction methods assume that there is a low-dimensional emedding influenced and caused by its high-dimensional representation. Probabilistic methods try to infer and discover this link. The advantage of the probabilistic approach over spectral methods is the handling of missing data.

\subsection{Uniform Manifold Approximation and Projection (UMAP) as a general approach to data embedding and visualization}
\label{sec:umap}

UMAP \cite{umap} is a dimension reduction technique that can be used for visualization and also for general non-linear dimension reduction. The algorithm is founded on three crucial assumptions about the data:

\begin{enumerate}
    \item The data is uniformly distributed on a Riemannian manifold \cite{parametric_umap}.
    \item The Riemannian metric is locally constant (or can be approximated as such).
    \item The manifold is locally Euclidean (Def. \ref{def:locally_euclidean_manifold}).
\end{enumerate}

On the basis of these assumptions, it is possible to model the manifold using a fuzzy topological structure. The embedding is found by searching for a low-dimensional projection of the data that has the closest possible equivalent fuzzy topological structure.  UMAP uses local approximations of manifolds and combines their local representations of fuzzy simplicial sets to construct a topological representation of high-dimensional data. Given some low-dimensional representation of the data, a similar process can be used to construct an equivalent topological representation. UMAP then optimizes the layout of the data representation in a low-dimensional space to minimize the cross-entropy between the two topological representations. The construction of fuzzy topological representations can be divided into two problems: the approximation of the manifold on which the data are inherently based and the construction of a fuzzy representation of the simplicial sets of this approximated manifold.

\subsubsection{Uniform distribution of data on a manifold and geodesic approximation}

The first step of the UMAP algorithm is to approximate the manifold in which the data lie (approximately). The manifold may be known a priori (simply $\mathbb{R}^{N}$) or may need to be inferred from the data. Suppose that the manifold is not known in advance and that we wish to approximate the geodesic distance on it. Let the input data be $X = [x_{1}, \dots, x_{N}] \in \mathbb{R}^{MxN}$, where $M$ is the sample size, and $N$ is the dimensionality. As shown in the work of Belkin and Niyogi on Laplacian eigenmaps \cite{laplacian_eigenmaps_2,laplacian_eigenmaps}, for theoretical reasons, it is beneficial to assume that the data are uniformly distributed in the manifold. Additionally, if we assume that the manifold has a Riemannian metric not inherited from the ambient space, we can find a metric such that the data are approximately uniformly distributed regarding that metric. Formally, let $\mathcal{M}$ be the data manifold on which to lie, and let $g$ be the Riemannian metric on $\mathcal{M}$. For each point $p \in \mathcal{M}$, we have $g_p$, which is an inner product in the tangent space $\mathcal{T}_{p}\mathcal{M}$.

\begin{lemma}
    Let $(\mathcal{M}, g)$ be a Riemannian manifold in ambient $\mathbb{R}^{N}$, and let $p$$\in$$M$ be a point. If $g$ is locally constant about $p$ in an open neighborhood $U$ such that $g$ is a constant diagonal matrix in the ambient coordinates, then in a ball $B \subseteq U$ centered on $p$ with volume $\frac{\pi^{n/2}}{\Gamma (n/2+1)}$ with respect to $g$, the geodesic distance from $p$ to any point $q \in B$ is $\frac{1}{r}d_{\mathbb{R}^{N}}(p,q)$, where $r$ is the radius of the ball in the ambient space and $d_{\mathbb{R}^{N}}$ is the existing metric in the ambient space.
\end{lemma}

If we assume that the data are uniformly distributed on $\mathcal{M}$ (with respect to $g$), then away from any boundaries, any ball of fixed volume should contain approximately the same number of points of $X$ regardless of where it is centered in the manifold.

\textbf{Data Graph in the Input Space} UMAP inspired by t-SNE \cite{tsne} uses the Gaussian or Radial Basis Function (RBF) kernel to measure the similarity between points in the input space. The probability that a point $x_{i}$ has the point $x_{j}$ as its neighbor can be calculated by the similarity of these points:

\vspace{0.5cm}
\begin{equation}
    p_{j|i} =
    \begin{dcases}
        \exp(- \frac{||x_{i}, x_{j}||_{2} - \rho_{i}}{\sigma_{i}}), & \text{if} \; x_{j} \in \mathcal{N}_{i} \\
        0, & \text{otherwise}
    \end{dcases},
    \label{eq:umap_probability}
\end{equation}
\vspace{0.5cm}

where $||.||_{2}$ denotes the norm $\mathcal{l}_{2}$. The $\rho_i$ is the distance from $x_i$ to its nearest neighbor:

\begin{equation}
    \rho_i = min{||x_{i}-x_{i,j}||_{2}} \; | \; 1 \leq j \leq k.
    \label{eq:umap_distance_1}
\end{equation}

 $\sigma_{i}$ is the scale parameter calculated so that the total similarity of the point $x_i$ to its nearest neighbors $k$ is normalized. By binary search, we find $\sigma_{i}$ to satisfy:
\vspace{0.1cm}
\begin{equation}
    \sum^{k}_{j=1}exp(- \frac{||x_{i} - x_{i,j}||_{2} - \rho_{i}}{\sigma_{i}}) = \log_{2}(k).
    \label{eq:umap_distance_2}
\end{equation}
\vspace{0.1cm}


The Eq. \ref{eq:umap_probability} is a measure of directional similarity. To have a symmetric measure with respect to $i$ and $j$, UMAP symmetrizes it as: 

\begin{equation}
    \mathbb{R} \ni p_{ij} := p_{j|i} + p_{i|j} - p_{j|i}p_{i|j}.
    \label{eq:umap_probability_input_space}
    \vspace{0.3cm}
\end{equation}

\textbf{Data Graph in the Embedding Space}  Let the embeddings of the points be $\mathbf{Y}$$=$$[y_{1}, \dots, y_{M}] \in \mathbb{R}^{nxM}$, where $n$ is the dimensionality of the embedding space. In the embedding space, the probability that a point $y_{i}$ has the point $y_{j}$ as its neighbor can be calculated by the similarity of these points: 
\begin{equation}
    \mathbb{R} \ni q_{ij} := (1 + a||y_{i} - y_{j}||_{2}^{2b})^{-1},
    \label{eq:umap_probability_embedding_space}
\end{equation}

which is symmetric with respect to $i$ and $j$. The variables $a$$>$$0$ and $b$$>$$0$ are hyperparameters determined by the user.

\subsubsection{Fuzzy topological representation}

UMAP uses functors between the relevant categories to convert metric spaces to fuzzy topological representations. This will provide a means of merging incompatible local views of the data. The topological structure of choice is that of simplicial sets, which are a means to construct topological spaces out of simple combinatorial components. This allows one to reduce the complexity of dealing with the continuous geometry of topological spaces to the task of relatively simple combinatorics and counting. This method of taking geometry and topology is fundamental to UMAP topological data analysis. 

The first step is to provide some simple combinatorial building blocks called \textit{simplices}. Geometrically, a simplex is a very simple way to build a $k$-dimensional object. A $k$ dimensional simplex is called a $k$ simplex and is formed by taking the convex hull of $k$+1 independent points. Thus, a 0-simplex is a point, a 1-simplex is a line segment, a 2-simplex is a triangle, and a 3-simplex is a tetrahedron. Such a simple construction allows for easy generalization to arbitrary dimensions and provides a basic building block. Formally, simplicial sets are most easily defined purely abstractly in the language of category theory.

\begin{definition}
The category $\Delta$ has as objects the finite-order sets $[n] = \{1, \dots, n\}$ with morphims given by (nonstrictly) order-preserving maps.
Following the standard category-theoretic notation, $\Delta^{op}$ denotes the category with the same objects as $\Delta$ and the morphisms given by the morphisms of $\Delta$ with the direction (domain and codomain) reversed.
\end{definition}

\begin{definition}
A simplicial set is a functor from $\Delta^{op}$ to \textbf{Sets}, the category of sets; that is, a contravariant functor from $\Delta$ to \textbf{Sets}.
\end{definition}

However, to construct complex topological spaces, we need to be able to combine simplices. A \textit{simplicial complex} $\mathcal{K}$ is a set of simplices such that any face of any simplex in $\mathcal{K}$ is also in $\mathcal{K}$ and the intersection of any two simplices in $\mathcal{K}$ is a face of both simplices. To construct a \textit{simplicial complex} from a topological space, UMAP uses the Čech or Vietoris-Rips complex \cite{algebraic_topology} given an open cover of a topological space. The key difference between two complexes is that Vietoris-Rips is entirely determined by the 0 and 1 simplices.

\subsubsection{Optimizing a low-dimensional representation}

In contrast to the source data, where we want to estimate a manifold on which the data are uniformly distributed, a target manifold for $Y$ is chosen a priori (usually this will simply be $\mathbb{R}^{n}$ itself, but other choices such as $n$-spheres or $n$-tori are certainly possible). Therefore, we know the manifold and the manifold metric a priori, and we can compute the fuzzy topological representation directly. In particular, we still want to incorporate the distance to the nearest neighbor according to the local connectivity requirement. This can be achieved by providing a parameter that defines the expected distance between the nearest neighbors in the embedded space. Given fuzzy simplicial set representations of X and Y , a means of comparison is required. If we consider only the 1-skeleton of fuzzy simplicial sets, we can describe each as a fuzzy graph, or, more specifically, a fuzzy set of edges. To compare two fuzzy sets, UMAP uses the fuzzy set cross-entropy.

\begin{definition}
The cross entropy C of two fuzzy sets $(A, \mu)$ and $(A,\nu)$ is defined as:
\begin{equation}
    c_{1} := \sum_{i=1}^{n} \sum_{j=1, j \neq i}^{n}(p_{ij} \ln (\frac{p_{ij}}{q_{ij}}) + (1 - p_{ij}) \ln (\frac{1-p_{ij}}{1-q_{ij}})),
    \label{eq:umap_cost_function}
    \vspace{0.2cm}
\end{equation}
\end{definition}

The first term in Eq. \ref{eq:umap_cost_function} is the attractive force that attracts the embeddings of neighboring points toward each other. This term should only appear when $p_{ij}$$\neq$$0$, which means that $x_{j}$ is a neighbor of $x_{i}$, or $x_{i}$ is a neighbor of $x_{j}$, or both. The second term in Eq. \ref{eq:umap_cost_function} is the repulsive force that repulses the embeddings of non-neighbor points away from each other. As the number of all permutations of non-neighbor points is very large, computation of the second term is non-tractable in big data.

Inspired by Word2Vec \cite{mikolov_2013} and LargeVis \cite{largevis}, UMAP uses negative sampling \cite{umap} where, for every point $x_{i}$ , $m$ points are randomly sampled from the training dataset and treated as non-negative (negative) points for $x_{i}$. As the dataset is usually large, the sampled points will be actual negative points with high probability. The summation over the second term in Eq. \ref{eq:umap_cost_function} is computed only on these negative samples, rather than on all negative points. UMAP changes the data graph in the embedding space to make it similar to the data graph in the input space. Eq. \ref{eq:umap_cost_function} is the cost function minimized in UMAP where the optimization variables are $\{y_{i}\}_{i=1}^{n}$:

\vspace{1cm}
\begin{equation}
    \begin{aligned}
        \min_{\{y_{i}\}_{i=1}^{n}} c_{1} := & \min_{\{y_{i}\}_{i=1}^{n}} \sum_{i=1}^{n} \sum_{j=1, j\neq i}^{n} (p_{ij} \ln (p_{ij}) - p_{ij} \ln (q_{ij}) \\\\
        & + (1-p_{ij}) \ln (1 - p_{ij}) - (1 - p_{ij}) \ln (1-q_{ij})) \\\\
        & = min_{\{y_{i}\}_{i=1}^{n}} - \sum_{i=1}^{n} \sum_{j=1, j\neq i}^{n} (p_{ij} \ln(q_{ij}) + (1-p_{ij}) \ln (1-q_{ij}))
    \end{aligned}
\end{equation}
\vspace{1cm}

According to Eqs. \ref{eq:umap_probability}, \ref{eq:umap_probability_input_space}, and \ref{eq:umap_probability_embedding_space}, in contrast to $q_{ij}$, $p_{ij}$ is independent of the optimization variables ${\{y_{i}\}_{i=1}^{n}}$. Hence, we can drop the constant terms to revise the cost function:

\begin{equation}
    c_{2} := - \sum_{i=1}^{n} \sum_{j=1, j\neq i}^{n} (p_{ij} \ln(q_{ij}) + (1-p_{ij}) \ln (1-q_{ij})),
\end{equation}
\vspace{0.2cm}

which should be minimized.

Similarly to t-SNE, UMAP can optimize the embedding $Y$ with respect to the cross entropy of the fuzzy set $C$ using stochastic gradient descent. However, this requires a differentiable fuzzy singular set functor. If the expected minimum distance between points is zero, the fuzzy singular set functor is differentiable for these purposes; however, for any nonzero value, we need to make a differentiable approximation (chosen from a suitable family of differentiable functions). The complete algorithm presents as follows: By using manifold approximation and patching together local fuzzy simplicial set representations, UMAP constructs a topological representation of the high-dimensional data. Then, it optimizes the layout of data in a low-dimensional space to minimize the error between the two topological representations. The whole process can be extended to the comparison of $\mathcal{L}$-skeleta with fuzzy simplicial sets instead of the 1-skeleton. Then, the cost function is defined as follows:

\begin{equation}
    C_{\mathcal{L}}(X, Y) = \sum_{i=1}^{\mathcal{L}} \lambda_{i} C(X_{i}, Y_{i}),
\end{equation}

where $X_{i}$ denotes the fuzzy set of $i$-simplices of $X$ and $\lambda_{i}$ are appropriately chosen real-valued weights. While such an approach captures the overall topological structure more accurately, it comes at a non-negligible computational cost due to the increasingly large number of higher-dimensional simplices. For this reason, current implementations restrict themselves to the 1-skeleton.

\subsubsection{A computational view of UMAP}

UMAP can be ultimately described in terms of weighted graph construction and operations. In particular, this situates UMAP in the class of k-neighbor based graph learning algorithms such as Laplacian Eigenmaps, Isomap and t-SNE. As with other k-neighbor graph-based algorithms, UMAP can be described in two phases. In the first phase, a particular weighted k-neighbor graph is constructed. In the second phase, a low-dimensional layout of this graph is computed. The differences between all algorithms in this class amount to specific details in how the graph is constructed and how the layout is computed. From previous sections, UMAP assumes these axioms to be true:
\begin{enumerate}
    \item There exists a manifold on which the data would be uniformly distributed.
    \item The underlying manifold of interest is locally connected.
    \item The primary goal is to preserve the topological structure of this manifold.
\end{enumerate}

Any algorithm that attempts to use a mathematical structure similar to a k-neighbor graph to approximate a manifold must follow a similar basic structure.
\begin{itemize}
    \item Graph Construction
    \begin{enumerate}
        \item Construct a weighted $k$NN graph.
        \item Apply some transform on the edges to the ambient local distance. 
        \item Deal with the inherent asymmetry of the $k$NN graph.
    \end{enumerate}
    \item Graph Layout
    \begin{enumerate}
        \item Define an objective function that preserves the desired characteristics of this $k$NN graph.
        \item Find a low-dimensional representation that optimizes this objective function. 
    \end{enumerate}
\end{itemize}

\begin{algorithm}
    \caption{UMAP scheme.} \label{alg:umap}
    \SetAlgoLined
    \SetKwInOut{Input}{Input}
    
    \Input{$k$NN graph $G(X,E)$.}
    \textbf{procedure} UMAP: \\
    \hskip0.5cm Initialize $\textbf{Y}$ using Laplacian eigenmap. \\
    \hskip0.5cm Calculate $p_{ij}, q_{ij} \; \forall_{i,j} \in \{1, \dots, n\}$ (Eqs. \ref{eq:umap_probability_input_space} and \ref{eq:umap_probability_embedding_space}).\\
    \hskip0.5cm $\eta \leftarrow 1, \nu \leftarrow 1$ \\
    \hskip0.5cm \textbf{while} not converged \textbf{do} \\
    \hskip1cm \textbf{for} i from 1 to $M$ \textbf{do} \\
    \hskip1.5cm \textbf{for} j from 1 to $M$ \textbf{do}\\
    \hskip2cm $u \sim U(0,1)$\\
    \hskip2cm \textbf{if} $u \leq p_{ij}$ \textbf{then}\\
    \hskip2.5cm $y_{i} \leftarrow y_{i} - \eta \frac{\delta c_{i,j}^{a}}{\delta y_{i}}$\\
    \hskip2.5cm $y_{j} \leftarrow y_{j} - \eta \frac{\delta c_{i,j}^{a}}{\delta y_{j}}$\\
    \hskip2.5cm \textbf{for} it interations \textbf{do}\\
    \hskip3cm $l \sim U\{1, \dots, n\}$\\
    \hskip3cm $y_{i} \leftarrow y_{i} - \eta \frac{\delta c_{i,l}^{r}}{\delta y_{i}}$\\
    \hskip1cm $\eta \leftarrow 1 - \frac{\nu}{\nu_{max}}$ \\
    
    \hskip0.5cm \textbf{return} embedding $\mathbf{Y}$.\\
\end{algorithm}

\subsection{Stochastic Neighborhood Embedding heuristics}
\label{sec:sne_heuristics}
 It was originally developed in 2002 by Sam Roweis and Geoffrey Hinton \cite{sne_hinton_roweis_2003}. SNE starts by converting the high-dimensional Euclidean distances between data points into conditional probabilities that represent similarities. The similarity of the data point $x_{j}$ to the data point $x_{i}$ is the conditional probability $p_{j|i}$, which $x_{i}$ would choose $x_{j}$ as its neighbor if the neighbors were chosen proportionally to their probability density under a Gaussian center at $x_{i}$. For nearby datapoints, $p_{j|i}$ is relatively high, while for widely separated datapoints, $p_{j|i}$ will be almost infinite (for reasonable values of Gaussian variance). Mathematically, the conditional probability $p_{j|i}$ is given by:

\begin{equation}
    p_{ij} = \frac{\exp(-||x_{i} - x_{j}||^{2}/2\sigma_{i}^{2})}{\sum_{k\neq i}\exp(-||x_{i} - x_{k}||^{2}/2\sigma_{i}^{2})}
    \label{eq:pji_sne}
    \vspace{0.4cm}
\end{equation}

where $\sigma_{i}$ is the Gaussian variance that is centered on the data point $x_{i}$. For the low-dimensional counterparts $y_{i}$ and $y_{j}$ of the high-dimensional data points $x_{i}$ and $x_{j}$, it is possible to compute a similar conditional probability, which is denoted by $q_{j|i}$. The similarity of the map point $y_{j}$ to the map point $y_{i}$ is given by the following:

\begin{equation}
    q_{ij} = \frac{\exp(-||y_{i} - y_{j}||^{2})}{\sum_{k\neq i}\exp(-||y_{i} - y_{k}||^{2})}
    \label{eq:qji_sne}
    \vspace{0.4cm}
\end{equation}

If the map points $y_i$ and $y_j$ correctly model the similarity between the high-dimensional data points $x_i$ and $x_j$, the conditional probabilities $p_{j|i}$ and $q_{j|i}$ will be equal. Motivated by this observation, the SNE aims to find a low-dimensional data representation that minimizes the mismatch between
$p_{j|i}$ and $q_{j|i}$. A natural measure of the faithfulness with which $q_{j|i}$ models $p_{j|i}$ is the Kullback-Leibler divergence. SNE minimizes the sum of Kullback-Leibler divergences over all data points using the gradient descent method. The cost function C is given by:

\begin{equation}
    C =\sum_{i} KL(P_{i} || Q_{i}) = \sum_i\sum_j p_{ij}\log(\frac{p_{ij}}{q_{ij}}),
    \label{eq:cost_function_sne}
    \vspace{0.4cm}
\end{equation}

in which $P_{i}$ represents the conditional probability distribution over all other data points given the data point $x_{i}$, and $Q_{i}$ represents the conditional probability distribution over all other map points given the map point $y_{i}$. Because the Kullback-Leibler divergence is not symmetric, different types of error in the pairwise distances in the low-dimensional map are not weighted equally.

The remaining parameter to be selected is the Gaussian variance $\sigma_{i}$ that is centered on each high-dimensional datapoint, $x_{i}$. It is not likely that there is a single value of $\sigma_{i}$ that is optimal for all data points in the data set, because it is likely that the density of the data will vary. In dense regions, a smaller value of $\sigma_{i}$ is generally more appropriate than in sparse regions. Any particular value of $\sigma_{i}$ induces a probability distribution, $P_{i}$, on all other data points. This distribution has an entropy that increases as $\sigma_{i}$ increases. SNE performs a binary search for the value of $\sigma_{i}$ that produces a $P_{i}$ with a fixed perplexity specified by the user. Perplexity is defined as follows:

\begin{equation}
    Perp(P_{i}) = 2^{H(P_{i})},
    \label{eq:perp-sne}
\end{equation}

where $H(P_{i})$ is the Shannon entropy of $P_{i}$ measured in bits:

\begin{equation}
    H(P_{i}) = - \sum_{j} p_{j|i} \log_{2} p_{j|i}
    \vspace{0.2cm}
\end{equation}

Perplexity can be interpreted as a smooth measure of the effective number of neighbors.

Minimizing the cost function $C$  is performed using a gradient descent method:

\begin{equation}
    \frac{\delta C}{\delta y_{i}} = 2 \sum_{j} (p_{ij} - q_{ij} + p_{ji} - q_{ji})(y_{i}-y_{j})
    \vspace{0.4cm}
    \label{eq:derivative_cost_function_sne}
\end{equation}

Physically, the gradient can be interpreted as the resultant force created by a set of springs between the point on the map $y_{i}$ and all other points on the map $y_{j}$. All springs exert a force along the direction $y_{i}-y_{j}$. The spring between $y_{i}$ and $y_{j}$ repels or attracts the points on the map, depending on whether the distance between the two on the map is too small or too large to represent the similarity between the two high-dimensional data points. The force exerted by the spring between $y_{i}$ and $y_{j}$ is proportional to its length and proportional to its stiffness, which is the mismatch ( $p_{j|i} - q_{j|i} + p_{i|j} - q_{i|j}$ ) between the pairwise similarities of the data points and the points on the map.

\subsubsection{Symmetric SNE, t-SNE, bh-SNE}

In symmetric SNE \cite{tsne}, we consider a Gaussian probability around every point $x_i$. The probability that the point $x_i \in \mathbb{R}^N$ takes $x_j \in \mathbb{R}^N$ as its neighbor is:

\begin{equation}
    p_{ij} = \frac{\exp(-||x_{i} - x_{j}||^{2}/2\sigma_{i}^{2})}{\sum_{k\neq l}\exp(-||x_{k} - x_{l}||^{2}/2\sigma_{k}^{2})}
    \label{eq:symmetric-tsne}
    \vspace{0.4cm}
\end{equation}

Note that the denominator of Eq. \ref{eq:symmetric-tsne} for all points is fixed and, thus, it is symmetric for $i$ and $j$. Compare this with Eq. \ref{eq:pji_sne} which is not symmetric. The $\sigma_{i}^{2}$ is the variance that we consider for the Gaussian distribution used for the $x_{i}$. It can be set to a fixed number or by a binary search to make the distribution entropy a specific value \cite{sne_hinton_roweis_2003}. The Eq. (15) has a problem with outliers. If the point $x_{i}$ is an outlier, its $p_{ij}$ will be extremely small because the denominator is fixed for every point and the numerator will be small for the outlier. However, if we use Eq. \ref{eq:pji_sne} for $p_{ij}$ , the denominator for all points is not the same, and therefore the denominator for an outlier will also be large, excluding the problem of a large numerator. For this problem mentioned, we do not use Eq. \ref{eq:symmetric-tsne} and rather we use:

\begin{equation}
    p_{j|i} = \frac{p_{i|j} + p_{j|i}}{2n}
    \label{eq:symmetric-tsne-pij}
\end{equation}

where:

\begin{equation}
    p_{j|i} = \frac{\exp(-||x_{i} - x_{j}||^{2}/2\sigma_{i}^{2})}{\sum_{k\neq i}\exp(-||x_{i} - x_{k}||^{2}/2\sigma_{i}^{2})}
    \label{eq:pji_symmetric_sne}
    \vspace{0.4cm}
\end{equation}

is the probability that $x_{i} \in \mathbb{R}^N$ takes $x_{j} \in \mathbb{R}^N$ as its neighbor. In the low-dimensional embedding space, we consider a Gaussian probability distribution for the point $x_{i} \in \mathbb{R}^N$ to take $y_{i} \in \mathbb{R}^n$ as its neighbor and make it symmetric (fixed denominator for all points):

\begin{equation}
    q_{ij} = \frac{\exp(-||y_{i} - y_{j}||^{2})}{\sum_{k\neq l}\exp(-||y_{k} - y_{l}||^{2})}
    \label{eq:qji_symmetric_sne}
    \vspace{0.4cm}
\end{equation}

Note that the Eq. \ref{eq:qji_symmetric_sne} does not have the problem of outliers as in Eq. \ref{eq:symmetric-tsne} because even for an outlier, the embedded points are initialized close to each other and not far away.

We want the probability distributions in both the input and embedded spaces to be as similar as possible; therefore, the cost function to be minimized can be a summation of the Kullback-Leibler (KL) divergences:

\begin{equation}
    C =\sum_{i} KL(P_{i} || Q_{i}) = \sum_i\sum_j p_{ij}\log(\frac{p_{ij}}{q_{ij}}),
    \label{eq:cost_function_symmetric_sne}
\end{equation}
\vspace{0.1cm}

Minimizing the cost function $C$  is performed using a gradient descent method:

\begin{equation}
    \frac{\delta C}{\delta y_{i}} = 4 \sum_{j} (p_{ij} - q_{ij})(y_{i}-y_{j})
    \label{eq:derivative_cost_function_symmetric_sne}
\end{equation}

\subsubsection{t-distributed Stochastic Neighbor Embedding (t-SNE)}

The SNE cost functions previously presented are difficult to optimize and are the root cause of why visualizations are not resistant to the phenomenon \textit{crowding}. In 2008, to address this, Laurens van der Maaten created t-SNE \cite{tsne}. Implement a modified cost function such that (1) uses a symmetrized version of the SNE cost function with simpler gradients and (2) uses a student t-distribution rather than a Gaussian to compute the similarity between two points in the low-dimensional space.

Let $\textbf{D}=[D_{ij}]$ be the distance table in $Y$ and $D_{ij}$ be the distances between the feature vectors $i$ and $j$ of $y_{i}$ and $y_{j}$, while $\textbf{d}=[d_{ij}]$ is the respective distance matrix in $X$. Then the loss function $C$=$E(\textbf{D},\textbf{d})$ is defined by the Kullback–Leibler (KL) divergence:
	
\begin{equation}
    C = E(\textbf{D},\textbf{d})=\sum_{i} KL(P_{i} || Q_{i}) = \sum_i\sum_j p_{ij}\log(\frac{p_{ij}}{q_{ij}}),
    \label{eq:tsne_cost_function}
\end{equation}

\noindent where, for the t-SNE algorithm, $p_{ij}$ is approximated by Gaussian $\mathcal{N}(y_i,\sigma)$, while $q_{ij}$ is defined by the Cauchy distribution \cite{tsne}. Then $p_{ij}$ and $q_{ij}$ are defined as follows:

\begin{multicols}{2}
\begin{equation}
    p_{ij} = \frac{exp(-D_{ij}^2 / 2\sigma_{i}^2)}{\sum_{k \ne l} exp(-D_{kl}^2 / 2\sigma_{i}^2)}
    \label{eq:tsne_pij}
\end{equation}

\begin{equation}
    q_{ij} = \frac{(1 + d_{ij}^{2})^{-1}}{\sum_{k \ne l} (1 + d_{kl}^2)^{-1}}
    \label{eq:tsne_qij}
\end{equation}
\end{multicols}

To minimize KL divergence, t-SNE uses modern optimal gradient descent optimization schemes \cite{optimizers}. The gradient of the loss function $C(.)$ (Eq. \ref{eq:tsne_cost_function}) is as follows:
\begin{equation}
    \label{eq:sne_optimization}
    \frac{\delta C}{\delta y_i} = 4 \sum_{j}(p_{ij} - q_{ij})(y_i - y_j).
\end{equation}

\begin{algorithm}
    \caption{t-SNE scheme.} \label{alg:tsne}
    \SetAlgoLined
    \SetKwInOut{Input}{Input}
    \Input{data matrix $\mathbf{X}$.}
    \textbf{procedure} t-SNE: \\
    \hskip0.5cm 1. Compute pairwise affinities $p_{ij}$ with defined \textit{perplexity}. \\
    \hskip0.5cm 2. Set $p_{ij} = \frac{p_{j|i} + p_{i|j}}{2}$. \\
    \hskip0.5cm 3. Sample initial solution $\mathbf{Y}^{(0)}$. \\
    \hskip0.5cm \textbf{for} t from 1 to $T$ \textbf{do} \\
    \hskip1cm Compute low-dimensional affinities $q_{ij}$. \\
    \hskip1cm Compute gradient $\frac{\delta C}{\delta Y}$. \\
    \hskip1cm Set $\mathbf{Y}^{(t)}$. \\
\end{algorithm}

\subsubsection{Barnes-Hut-SNE (BH-SNE)}

This variant \cite{bhsne} of SNE uses metric trees to approximate $P$ by a sparse distribution in which only the values of $O(uN)$ are non-zero and approximate the gradients $\frac{\delta C}{\delta y_{i}}$ using a Barnes-Hut algorithm.

As input similarities are computed using a (normalized) Gaussian kernel, the probabilities $p_{ij}$ corresponding to dissimilar input objects $i$ and $j$ are (nearly) infinitesimal. Therefore, a sparse approximation of probability $p_{ij}$ can be used without a substantial negative effect on the quality of the final embeddings. In particular, bh-SNE computes the sparse approximation by finding the nearest neighbors $[3u]$ of each of the N data objects and redefining pairwise similarities $p_{ij}$ as:

\vspace{0.5cm}
\begin{equation}
    p_{j|i} =
    \begin{dcases}
        \frac{\exp{(-d(x_i, x_j)^2 / 2 \sigma_{i}^2})}{\sum_{k \in \mathcal{N}_i} \exp{(-d(x_i, x_k)^2 / 2 \sigma_{i}^2)}}, & \text{if $j \in \mathcal{N}_{i}$}\\\\
        0 & otherwise
    \end{dcases}
\end{equation}
\vspace{0.5cm}

where $\mathcal{N}_{i}$ represents the set of nearest neighbors $[3u]$ of $x_i$, and $\sigma_i$ is set so that the perplexity of the conditional distribution is equal to $u$. The nearest-neighbor sets are found in $O(uN \log N)$ time by building a vantage point tree on the data set.

Some examples of probabilistic dimensionality reduction are factor analysis, whose non-linear extension is the variational autoencoder \cite{kingma2013}, probabilistic PCA \cite{bishop2001}, probabilistic LDA \cite{ioffe2006}. Some other examples are SNE \cite{sne_hinton_roweis_2003} and t-SNE \cite{tsne}, where the Gaussian and Student-t distributions are considered for the embedded space, respectively. A recent successful method is UMAP \cite{umap}, which optimizes the probability of closeness of the graphs in the input and embedded spaces.

\subsection{State-of-the-art supervised and unsupervised improvements and simplifications. UMAP and t-SNE.}

Both t-SNE \cite{tsne} and UMAP \cite{umap} have become very popular among the DR community. This has motivated the design of their variants, such as, for example, parametric extensions \cite{crecchi2020,sainburg2020}. In addition, both algorithms comprise the same two broad steps: 1) construct a graph of local relationships between datasets, 2) optimize an embedding in a low-dimensional space that preserves the structure of the graph.

\subsubsection{t-SNE improvements}
t-SNE requires the user to choose an approximation to adjust the width of its Gaussian HD neighborhoods. Although such a single-scale method does a good job of preserving neighborhood sizes close to perplexity, but without achieving similar performance for other neighborhoods, multi-scale approaches typically recover both local and global HD structures much better \cite{lee2015}. Additionally, in its original formulation, t-SNE \cite{tsne} is a non-parametric manifold learner. The primary limitation of non-parametric manifold learners is that they do not provide a parametric mapping between the high-dimensional data space and the low-dimensional latent space, making it impossible to embed new data points without retraining the model. For this reason, there is plenty of room for improvement.

\subsubsection{Parametric t-SNE}
The parametric t-SNE \cite{vandermaaten2009_parametric_tsne} parameterizes the nonlinear mapping $f: \mathbf{X} \rightarrow \mathbf{Y}$ by means of a feed-forward neural network with weights $W$. It uses a deep neural network (DNN) because a neural network with sufficient hidden layers (with non-linear activation functions) is capable of parametrizing arbitrarily complex non-linear functions. The neural network is trained in such a way as to preserve the local structure of the data in the latent space. The DNN weights are learned to minimize the Kullback-Leibler (KL) divergence $C_{t-SNE}$. Since t-SNE optimization requires normalization over the embedding distribution
in the projection space, gradient descent can only be performed after calculating the edge probabilities over the entire dataset. However, projecting the entire dataset onto a neural network; between each gradient descent step, would be too computationally expensive for optimization. The trick that parametric t-SNE proposes for this problem is to divide the dataset into large batches (e.g., 5000 data points in the original paper \cite{vandermaaten2009_parametric_tsne}), which are used to compute separate graphs that are independently normalized and used continuously during training, meaning that the relationships between elements in different batches are not explicitly preserved.

\subsubsection{Perplexity-free Parametric t-SNE}

The perplexity-free Parametric t-SNE updates the original parametric t-SNE neural network using $\sigma_{ij}$ to compute high-dimensional similarities, in a multi-scale fashion. Moreover, it replaces logistic activation functions with piecewise-linear ones (i.e., ReLUs), which do not saturate during training. This simple architectural choice allowed one to simplify the training procedure by eliminating the unsupervised pre-training step introduced in the original implementation.

\subsubsection{q-Gaussian SNE}

The q-SNE \cite{qsne2020} in theory leads to a more powerful and flexible visualization of the 2 dimension mapping than t-SNE and SNE using a q-Gaussian distribution as the low-dimensional data distribution. The q-Gaussian distribution includes the Gaussian distribution and the t-distribution as special cases with $q$$=$$1$ and $q$$=$$2$. Therefore, q-SNE can also express t-SNE and SNE by changing the parameter $q$, which allows the best visualization by choosing the parameter $q$. 

The q-Gaussian distribution is derived by maximizing the Tsallis entropy under appropriate constraints and is a generalization of the Gaussian distribution. Let $s$ be a 1-dimensional observation. The q-Gaussian distribution for the observation $s$ is defined as follows:

\begin{equation}
    P_{q}(s; \mu, \sigma^2) = \frac{1}{Z_{q}} (1+\frac{q-1}{3-q}\frac{(s-\mu)^{2}}{\sigma^{2}})^{-\frac{1}{q-1}},
    \vspace{0.4cm}
    \label{eq:qsne_q_gaussian_distribution}
\end{equation}

where $\mu$ and $\sigma$ are the mean and variance, respectively. The normalization factor $Z_q$ is given by:

\vspace{0.4cm}
\begin{equation}
    Z_{q} =
    \begin{dcases}
        \sqrt{\frac{3-q}{q-1}} \cdot B(\frac{3-q}{2(q-1)}, \frac{1}{2}) \cdot \sigma, & \;\; 1 \leq q < 3 \\\\
        \sqrt{\frac{3-q}{1-q}} \cdot B(\frac{2-q}{1-q}, \frac{1}{2}) \cdot \sigma, & \;\; q < q \\
    \end{dcases}
    ,
    \label{eq:qsne_normalization_factor}
\end{equation}
\vspace{0.6cm}

where $B$ is the beta function. It is known that the q-Gaussian distribution defined by Equation \ref{eq:qsne_q_gaussian_distribution} always satisfies the inequality: 
\begin{equation}
    1 + \frac{q-1}{3-q} \frac{(s-\mu)^2}{\sigma^2} \geq 0.
\end{equation}

Similarly to symmetric SNE or t-SNE, q-SNE uses the local Gaussian distribution in a high-dimensional void. The joint probability in the low-dimensional space is defined as:

\vspace{0.3cm}
\begin{equation}
    r_{ij} = \frac{(1+\frac{q-1}{3-q}||y_i - y_j||^{2})^{-\frac{1}{q-1}}}{\sum_{l}^{N} \sum_{k \neq l}^{N} (1+\frac{q-1}{3-q}||y_l - y_k||^{2})^{-\frac{1}{q-1}}},
\end{equation}
\vspace{0.3cm}

where $q$ is the hyperparameter, $r_{ii}=0$ and $r_{ij}$$=$$r_{ji}$ $\forall_{i,j}$. The Kullback-Leibler divergence and the optimization update rule are the same as in the SNE equations \ref{eq:cost_function_sne} and \ref{eq:derivative_cost_function_sne}. The gradient for $y_i$ is given as

\vspace{0.2cm}
\begin{equation}
    \frac{\delta C}{\delta y_{i}} = \frac{4}{3-q} \sum_{j}^{N} (p_{ij} - r_{ij})(y_{i} - y_{j})(1 + \frac{q-1}{3-q} ||y_{i} - y_{j}||^{2})^{-1}.
\end{equation}
\vspace{0.2cm}

Since the q-Gaussian distribution is an extension of the Gaussian distribution and the t-distribution with the parameter $q$, q-SNE can generate the same low-dimensional embedded space with SNE or t-SNE by changing the parameter $q$.

\subsubsection{Tree-SNE}

Tree-SNE\cite{treesne} is a hierarchical clustering method based on a one-dimensional t-SNE with decreasing values of $\alpha$ and perplexity at each level. It allows us to visualize and elucidate high-dimensional hierarchical structures by creating t-SNE embeddings with increasingly heavy tails to reveal increasingly fine-grained structures and then stacking these embeddings to create a tree-like structure. Then, it performs spectral clustering on each one-dimensional embedding, computationally determining the number of distinct clusters in the embedding. The number of clusters will increase as the value of $\alpha$ decreases. Tree-SNE defines alpha-clustering of data as the cluster assignment that is stable over the largest range of $\alpha$ values.

The method starts with a standard one-dimensional t-SNE embedding ($\alpha$=$1$) with a high perplexity, by default equal to the square root of the number of data points. A high initial perplexity increases the effective number of neighbors used by t-SNE, which means that larger clusters will tend to form, capturing more global structures in the data. Using a high error count at the start, a t-SNE tree can show the entire spectrum of data organization, from global structures at the base of the tree to very fine structures. The exact initial value of the error count does not appear to be important in many trials, as long as it is high, because t-SNE is quite robust to small changes in the error count, and most of the interesting features of the data emerge in further embeddings from adjustments in the error count and $\alpha$.

\subsubsection{Local Interpolation with Outlier CoNtrol t-SNE}
The LION-tSNE\cite{lion_tsne} uses a random sampling method for the design of the tSNE model, creating an initial visual environment. Then, new data points are added to this environment using the local-IDW \cite{shepard_1968} (inverse distance weighting) interpolation method. 
\begin{algorithm}
    \caption{LION-tSNE scheme.} \label{alg:lion_tsne}
    \SetAlgoLined
    \SetKwInOut{Input}{Input}
    \Input{data point $x$.}
    \textbf{procedure} LION-tSNE: \\
    \hskip0.5cm Find neighbors in radius $r_{x}$. \\
    \hskip0.5cm \textbf{if} $\text{found neighbors} > 1$ \textbf{then}\\
    \hskip1cm Perform $IDW$.\\
    \hskip0.5cm \textbf{else if} $\text{found neighbors} == 1$ \textbf{then}\\
    \hskip1cm Perform single neighbor placement.\\
    \hskip0.5cm \textbf{else}\\
    \hskip1cm Perform outlier placement.\\
    \hskip0.5cm \textbf{return} $y$.\\
\end{algorithm}

The randomly selected sample data often suffer from non-representativeness of the entire data, which creates inconsistency in the t-SNE environment. To overcome this problem, two new sampling methods were proposed in \cite{lion_tsne}, which are based on graph update properties of $k$-NN. It has been empirically shown that the proposed methods outperform the existing LION-tSNE method with 0.5 to 2\% more precision in k-NN and the results are more consistent. LION t-SNE combines outlier detection and locality into a single approach: it can uses local interpolation when a new sample $x$ has neighbors in a certain radius $r_{x}$.

\subsubsection{Capacity Preserving Mapping}

In CPM \cite{2019capacity}, the low-dimensional embedding $\mathbf{Y}$ is the minimizer of the following optimization problem with the well-defined Capacity Adjusted Distance $\mathbf{\tilde{D}}$:

\begin{equation}
    \mathbf{Y} = arg_{Y_{i}} min \sum_{i,j} p_{ij} \log \frac{p_{ij}}{q_{ij}},
    \label{eq:cpm_optimization}
\end{equation}

where

\begin{equation}
    p_{ij} = \frac{(\epsilon + \mathbf{\tilde{D}}^2 ||x_i - x_j||)^{-1}}{\sum_{k,l, k \neq l} (\epsilon + \mathbf{\tilde{D}}^2 ||x_k - x_l||)^{-1}}, \;\;\; q_{ij} = \frac{(1 +||y_i - y_j||)^{-1}}{\sum_{k,l, k \neq l} (1 +||y_k - y_l||)^{-1}},
    \vspace{0.4cm}
\end{equation}

where $||\cdot||$ is the distance measure in high-dimensional space, and $\epsilon > 0$ is some small constant used to avoid taking the reciprocal of $0$. In addition to the KL divergence, one can use other (dis)similarity measures and obtain different formulations of the optimization problems. Optimization Eq. \ref{eq:cpm_optimization} is trying to match the KL divergence between the network probabilities of the original and embedded graphs. The small positive constant $\epsilon$ in the expression of $p_{ij}$ and the 1 in the expression of $q_{ij}$ are used to avoid dividing by 0. Although the formulation of Eq. \ref{eq:cpm_optimization} looks similar to that of t-SNE (Eq. \ref{eq:tsne_cost_function}), their performances are completely different. The biggest upside of CPM compared to t-SNE is that it: 1) lowers over-stretching, 2) does not have tuning parameters, and 3) preserves more geometry.

\subsubsection{Class-aware t-SNE} 

Cat-SNE \cite{cat_sne} explicitly accounts for the class labels in t-SNE to improve the accuracy of KNN in the LD embedding. For this purpose, it modifies the t-SNE adjustment of the individual radius of the normalized Gaussian neighborhood around each datum. Instead of targeting a fixed neighborhood entropy provided by the user through the perplexity, it adjusts the neighborhood radius for neighbors with the same class to cumulate a dominant fraction of the probability distribution. This results in smaller HD neighborhoods near class boundaries than in their bulk, and therefore tends to stretch the former and shrink the latter.

Let $c_i$ be the class associated with $x_{i}$ ( $x_i$ being a point in the HD space $X$, $y_i$ in the LD space $Y$). Cat-SNE defines a condition on the weighted proportion $t_i$ of neighbors $x_j$ that share the same class as $x_i$, that is,

\begin{equation}
    t_i = \sum_{j \in Y \\ \{i\}|c_j = c_i} \sigma_{ij} > 0.
    \vspace{0.3cm}
\end{equation}

where $t_i \in [0, 1]$ as $\sum_{j \in Y \\ \{i\}|c_j = c_i} \sigma_{ij} = 1$ and the hyper-parameter $\theta$ is in $[0.5, 1]$ to ensure the majority of the class $c_i$. Therefore, the perplexity metaparameter in t-SNE is replaced by the threshold $\theta$. No other changes are made to t-SNE. \\

\subsubsection{Supervised t-SNE} 

St-SNE \cite{cheng2021} incorporates the outcome information, while calculating the $KL(P||Q)$ by defining a similarity measure $o_{ij}$ in the outcome space $O$ as follows:
\begin{equation}
    o_{j|i} = \frac{exp(-||y_i - y_j||^2)}{\sum_{k \neq i}exp(-||y_i - y_k||^2)}
\end{equation}.

The low-dimensional representation by minimizing the following cost function $C_S$:

\begin{equation}
    C_S = \rho KL(P||Q) + (1-\rho)KL(O||Q) = \rho \sum_{i \neq j} p_{ij} log \frac{p_{ij}}{q_{ij}} + (1 - \rho) \sum_{i \neq j} o_{ij} log \frac{o_{ij}}{q_{ij}},
\end{equation}

where $\rho \in [0,1]$ controls the level of supervision in the learning process. Larger $\rho$ reflect less supervision.

\begin{algorithm}
    \caption{St-SNE scheme.} \label{alg:sttsne}
    \SetAlgoLined
    \SetKwInOut{Input}{Input}
    \Input{data matrix $\mathbf{X}$.}
    \textbf{procedure} St-SNE: \\
    \hskip0.5cm Generate $z_1^{(0)}, \dots, z_n^{(0)}$ using multivariate normal distribution with mean zero.\\
    \hskip0.5cm Calculate $p_{ij}$, $q_{ij}$. \\
    \hskip0.5cm \textbf{for} t from 1 to $T$ \textbf{do} \\
    \hskip1cm Calculate gradient $\frac{\delta C_{S}}{\delta z}$. \\
    \hskip1cm $z^{(t)} = z^{(t-1)} - \eta \frac{\delta C_{S}}{\delta z} + \alpha \delta^{(t-1)}$. \\
        \hskip1cm $\delta^{(t)} = z^{(t)} - z^{(t-1)}$. \\
    \hskip0.5cm \textbf{return} $z = z^{T}.$
\end{algorithm}

\subsubsection{Other improvements}
Here, we just list some of the recent improvements of t-SNE and do not explain them in detail for the sake of brevity. 

\begin{itemize}
    \item Dense t-SNE \cite{narayan2021}, which tackles the problem of local density information for points in denser regions, which have a smaller $\sigma_{i}^2.$
    \item Parametric kernel t-SNE \cite{gisbrecht2015}.
    \item Fast Interpolation-based t-SNE \cite{linderman2019}, which accelerates the t-SNE procedure.
    \item d-SNE \cite{xu2019} used for domain adaptation in neural network training, which uses SNE and a novel modified-Hausdorff distance.
    \item Approximated and User Steerable tSNE (A-tSNE) \cite{pezzotti2015} is a controllable tSNE approximation, which trades off speed and accuracy, to enable interactive data exploration.
\end{itemize}

\subsubsection{UMAP improvements and simplifications.}

UMAP is an emerging dimensionality reduction technique that offers better versatility and stability than t-SNE. Although UMAP is also more efficient than t-SNE, it still suffers from an initial delay of several minutes to produce the first projection, which limits its use in interactive data mining. It was developed in 2018, so not as many new variants were created as in the case of the t-SNE method.
 
\subsubsection{Parametric UMAP}

In general, parametric approaches use deep neural networks to preserve the structure of the dataset. The parametric form of UMAP \cite{sainburg2020}, through the use of negative sampling, can in principle be trained on batches as small as a single edge, making it suitable for training the mini-patches needed for memory-intensive neural networks trained on full graphs on large data sets, as well as for online learning. Given these design features, UMAP loss can be used as regularization in typical deep learning paradigms using stochastic gradient descent, without the batch trick on which parametric t-SNE relies (desribed in previous Section).

\subsubsection{Progressive UMAP}
Progressive UMAP \cite{kerren2020} allows users to feed small batches of data points incrementally into UMAP to obtain the desired latency between intermediate projection outputs. To this end, it identifies sequential procedures in the original UMAP algorithm and transforms them into progressive procedures.

\textbf{Computing} $\mathcal{N}_{i}$: To build and maintain the k-nearest neighbor graph, Progressive UMAP leverages the $k$-NN lookup table from PANENE \cite{panene}. PANENE uses randomized kd-trees \cite{randomized_kdtree} to approximate and update the k-nearest neighbors of an increasing number of data points. PANENE accepts a parameter called $ops$, which indicates the allowed number of tasks per iteration that can be controlled to find the balance between latency and accuracy.

\textbf{Computing} $\rho_{i}$ and $\sigma_{i}$: For every data point in $\mathbf{X}_{updated}$ and $\mathbf{X}_{new}$, Progressive UMAP recomputes $\rho_{i}$ and $\sigma_{i}$. For space efficiency, UMAP used the Coordinate List (COO), which stores only row, column, and value information as a list of tuples. Progressive UMAP updates the COO by recalculating $v_{j|i}$ for the selected points – $\mathbf{X}_{updated}$ and $\mathbf{X}_{new}$ – and changing the corresponding values if there is a change from the previous ones.

\begin{equation}
    \rho_{i} = min_{j \in \mathcal{N}_{i}} \{ d(x_i, x_j) | d(x_i, x_j) > 0\},
    \label{eq:chapter3_umap_rho}
\end{equation}

\begin{equation}
    \sum_{j \in \mathcal{N}_{i}} exp(\frac{-max(0, d(x_i, x_j)) - \rho_i}{\sigma_i}) = \log_2(k).
    \label{eq:chapter3_umap_sigma}
\end{equation}
\vspace{0.3cm}

Using $\rho_i$ and $\sigma_i$ it can calculate (the same as UMAP) $v_{j|i}$, the weight of the edge from a point $x_i$ to another point $x_j$:

\begin{equation}
    v_{j|i} = exp(\frac{-max(0, d(x_i, x_j)) - \rho_i}{\sigma_i}).
    \label{eq:chapter3_umap_edge_weight}
\end{equation}
\vspace{0.2cm}

\textbf{Layout Initialization}: Although spectral embedding produces an effective initial projection, its quadratic time complexity causes severe delay. Progressive UMAP initializes the positions of newly inserted points in two stages. For the first batch of points, 1) it runs the algorithm with a large value of ops (e.g., 15000), using the same spectral embedding technique as UMAP. Because it starts with a relatively small number of points, this would take much less time than the original UMAP spectral embedding. Hereafter, 2) we lower the value of ops (e.g., 1000) not to focus on the appending process, but to obtain an optimized projection output fast. 

\textbf{Layout Optimization}: Similarly, Progressive UMAP goes through two stages for layout optimization. As it affects the overall convergence time and increases the stability of the final output, it is very important to position the first batch of points well so that the clusters are unambiguously separated. To this end, 1) it runs more iterations (e.g., 40) in the first batch so that each cluster can settle its position. Afterwards, 2) it runs fewer iterations (e.g., 4) to focus on attaining the projection result fast.

\subsubsection{Other improvements}
Here, we just list some of the recent improvements of UMAP and do not explain them in detail for the sake of brevity. 
\begin{itemize}
    \item Hierarchical UMAP \cite{humap2021}, preserves the mental map while requesting more cluster details, while balancing the trade-off between global and local relationships.
    \item UMAP \cite{umap} by design allows for using categorical label information to do supervised dimension reduction.
    \item DensMAP \cite{narayan2020} computes estimates of the local density and uses those estimates as a regularizer in the optimization of the low-dimensional representation.
\end{itemize}

\begin{algorithm}[ht!]
    \caption{Progressive UMAP scheme.} \label{alg:progressive_umap}
    \SetAlgoLined
    \SetKwInOut{Input}{Input}
    
    \Input{New batch of training data $\mathbf{X}_{new}$.}
    \textbf{procedure} Progressive UMAP: \\
    \hskip0.5cm Update $k$NN graph by \textit{PANENE} method $\rightarrow$ $\mathbf{X}_{new}$, $\mathbf{X}_{updated}$.\\
    \hskip0.5cm \textbf{if} is \textit{initial batch} \textbf{then}\\
    \hskip1cm Initialize $\textbf{Y}$ using Laplacian eigenmap. \\
    \hskip0.5cm \textbf{else}\\
    \hskip1cm \textbf{foreach} $\mathbf{x}_{i} \in \mathbf{X}_{new}$ \textbf{do} \\
    \hskip1.5cm Find nearest neighbor to previously accumulated data. \\
    \hskip1.5cm Initialize $\mathbf{y}_{i}$ to the embedding of nearest neighbor point.\\

    \hskip0.5cm \textbf{foreach} $\mathbf{x}_{i} \in \mathbf{X}_{new}, \mathbf{X}_{updated}$ \textbf{do} \\
    \hskip1cm Calculate $\rho_i$, $\sigma_i$ (Eqs. \ref{eq:umap_distance_1} and \ref{eq:umap_distance_2}). \\
    \hskip1cm Calculate $p_{ij}, q_{ij} \; \forall_{i,j} \in \{1, \dots, n\}$ (Eqs. \ref{eq:umap_probability_input_space} and \ref{eq:umap_probability_embedding_space}).\\
    
    \hskip0.5cm \textbf{while} not converged \textbf{do} \\
    \hskip1cm \textbf{foreach} $\mathbf{x}_{i} \in \mathbf{X}_{new}, \mathbf{X}_{updated}$ \textbf{do} \\
    \hskip1.5cm \textbf{for} j from 1 to $M$ \textbf{do}\\
    \hskip2cm $u \sim U(0,1)$\\
    \hskip2cm \textbf{if} $u \leq p_{ij}$ \textbf{then}\\
    \hskip2.5cm $y_{i} \leftarrow y_{i} - \eta \frac{\delta c_{i,j}^{a}}{\delta y_{i}}$\\
    \hskip2.5cm \textbf{for} it interations \textbf{do}\\
    \hskip3cm $l \sim U\{1, \dots, n\}$\\
    \hskip3cm $y_{i} \leftarrow y_{i} - \eta \frac{\delta c_{i,l}^{r}}{\delta y_{i}}$\\

    \hskip0.5cm \textbf{return} embedding $\mathbf{Y}$ for new and updated points.\\
\end{algorithm}

\section{Neural network based dimensionality reduction}

The category of DR based on neural networks has an approach based on information theory, where the center of a neural network or autoencoder is viewed as an information bottleneck that retains only useful and important information. Some examples are the Restricted Boltzmann machine (RBM) and the Deep Belief Network (DBN) \cite{hinton2006}, which are fundamental DR methods in a network structure. They have been proposed to avoid the problem of vanishing gradient \cite{lee2013}. Another example is autoencoder, in which the latent embedding space is encoded by the middle layer of a possibly deep autoencoder. There is also deep-metric learning \cite{kaya2019}, which encodes data in the embedding space trained by a deep neural network and attempts to increase and decrease the interclass and intraclass variances of the data, respectively \cite{ghojogh2020}. Note that metric learning can be seen as a DR, as it can be viewed as a linear or non-linear projection onto the embedding space and then applying the Euclidean distance to that space. In a variational autoencoder \cite{kingma2013}, the latent space has a specific distribution, such as a Gaussian distribution. Another example is the adversarial autoencoder \cite{makhzani2015}, which uses game theory to optimize the encoding.

More recently, several deep learning metric methods have been proposed that focus on maximizing and minimizing interclass and intraclass variance in data \cite{ghojogh2020}. A Siamese network is a collection of several networks (usually two or three) that share weights with each other \cite{schroff2015}. The weights are trained using loss based on anchor, neighbor (positive) and distant (negative) samples, where anchor and neighbor belong to the same class, but anchor and distant instances belong to different classes. The loss functions used to train a Siamese network typically use anchor, neighbor and distant samples, trying to attract the anchor and neighbor to each other while pushing the anchor and distant tiles away from each other. Two different loss functions that are used to train Siamese networks are triple loss \cite{schroff2015} and contrast loss \cite{hadsell2006} for networks with three and two subnets, respectively.

\subsubsection{Autoencoders}

Autoencoders \cite{hinton2006} can learn non-linear mappings that are required to embed highly non-linear real-world data in the latent space. They primarily focus on maximizing the variance of the data in the latent space and as a result, autoencoders are less successful in retaining the local structure of the data in the latent space compared to manifold learners.

\begin{figure}[ht]
    \begin{center}
        \includegraphics[clip,width=\textwidth]{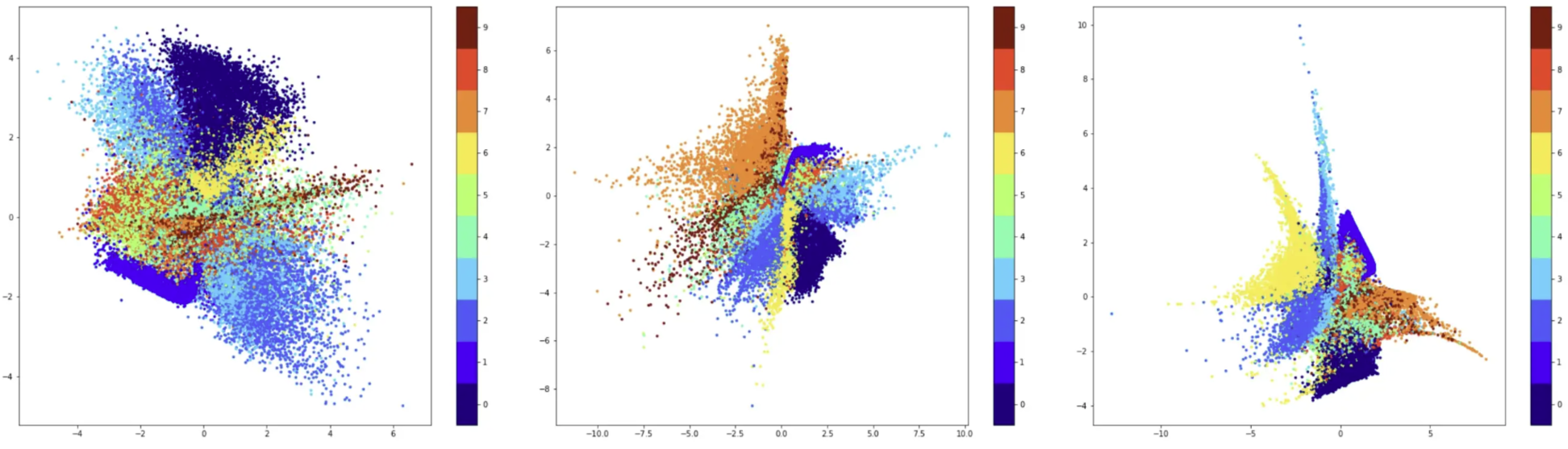}
        \label{fig:cvae_mnist}
        \caption{Visualization of the latent spaces generated by three different variational autoencoder (VAE) instances trained on the same MNIST data.}
    \end{center}
\end{figure}

Autoencoders can be used to reduce the dimensionality of data, but in our work, we do not consider cases where the dataset has been preprocessed with an autoencoder. Such datasets already have a structure, which contains global information about itself, thus it will strongly affect the final visualization generated by the embedding method (e.g. t-SNE or UMAP).

In Figure 2.3 we can observe that in all cases VAE tends to produce a very smooth latent space and preserves global structure very well. In contrast, the strong clustering of UMAP, for example, very clearly separates similar digits, while in the case of VAE they smoothly transition into each other.
\chapter{Towards optimal NN graph visualization}

Many successful techniques have been reported recently that first compute a similarity structure of the data points and then project them into a low-dimensional space with the structure preserved \cite{chao2019}. These two steps suffer from considerable computational costs, preventing state-of-the-art methods (such as the SNE variants) from scaling to large-scale and high-dimensional data (e.g., millions of data points and hundreds of dimensions). 

\begin{figure}[ht]
    \begin{center}
    \includegraphics[width=12cm]{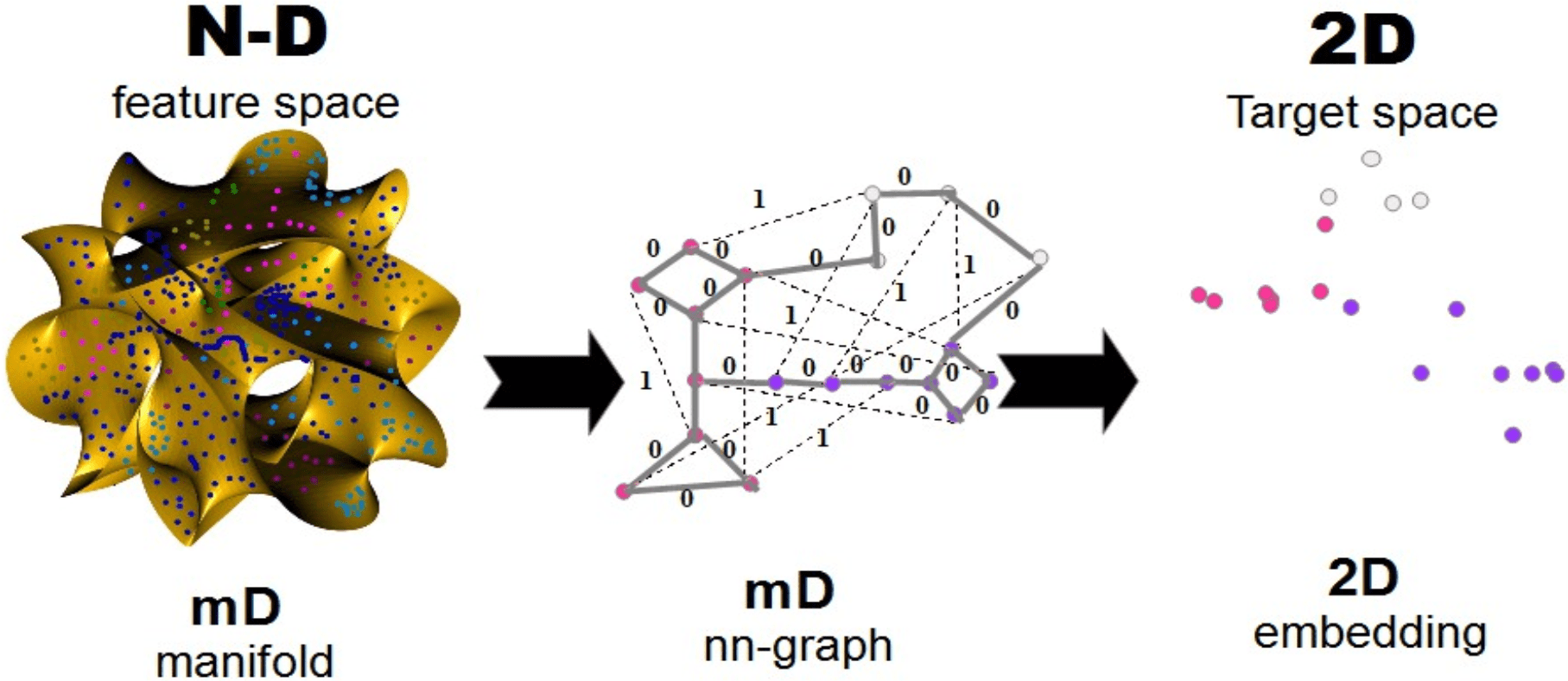}
    \end{center}
    \caption{The general idea of data embedding by means of the nearest neighbor graph. Source: \cite{dzwinel2017ivga}.}
    \label{fig:general_idea_of_embedding_by_nn_graph}
\end{figure}

The construction of $k$NN graphs from high-dimensional data is critical for many applications, such as similarity search, collaborative filtering, manifold learning, and network analysis. Although the exact computation of a $k$NN has a complexity of $O(M^{2}N)$ (with the number of data points $M$ and the number of dimensions $N$) that is too costly, existing approaches use roughly three categories of techniques: space partition trees \cite{bentley_1975,bentley_1977,hartley_2008,dasgupta_2008}, locality-sensitive hashing techniques \cite{datar_2004,zhao_2014}, and neighbor exploration techniques \cite{dong_2011}. Space partitioning methods divide the entire space into different regions and organize the regions into different tree structures, e.g., $k$$-$d trees \cite{bentley_1975,bentley_1977}, $vp$-trees \cite{yianilos_1993}, cover trees \cite{beygelzimer_2006} and random projection trees \cite{dasgupta_2008}. Once the trees have been constructed, the nearest neighbors of each data point can be found by traversing the trees. Locality-sensitive hashing techniques \cite{datar_2004} deploy multiple hashing functions to map data points to different buckets, and data points in the same buckets are likely to be similar to each other. Neighbor-exploring techniques, such as NN descent \cite{dong_2011}, are built on the intuition that "my neighbors neighbor are likely to be my neighbors as well". Starting from an initial nearest-neighbor graph, the algorithm iteratively refines the graph by exploring the neighbors of neighbors defined according to the current graph.

\section{LargeVis}

Technique that first constructs an accurately approximated $k$NN graph from the data and then layouts the graph in a low-dimensional space. Compared to t-SNE, LargeVis significantly reduces the computational cost of the graph construction step and employs a principled probabilistic model for the visualization step, the objective of which can be effectively optimized through asynchronous stochastic gradient descent with linear time complexity. The whole procedure thus easily scales to millions of high- dimensional data points. Existing approaches generally first compute the similarities of all pairs of ${x_i,x_j}$ and then preserve the similarities in the low-dimensional transformation. As computing pairwise similarities is too expensive (i.e. $O(N^{2}d)$), approaches such as the t-SNE or UMAP construct a graph of $k$-nearest neighbor instead and then project the graph into the $2D$ space. LargeVis follows this procedure, but uses a very efficient algorithm for the construction of the $k$-nearest neighbor graph and a principled probabilistic model for graph visualization. It is worth mentioning that LargeVis visualizes high-dimensional information and is not used to visualize graphs.

\subsection{\textit{k}NN graph construction}

LargeVis uses the most common Euclidean metric to compute the graph of $k$NN in a high-dimensional space. The same approach is used by t-SNE. The algorithm starts by partitioning the entire space and building a tree. Specifically, for each non-leaf node of the tree, the algorithm chooses a random hyperplane to divide the subspace corresponding to the non-leaf node into two, which become the children of that node. The hyperplane is selected by randomly sampling two points from the current subspace and then taking a hyperplane that is equally distant from those two points. This process continues until the number of nodes in the subspace reaches a threshold. After building a random projection tree, each data point can traverse the tree to find the corresponding leaf node. The points in the subspace of this leaf node will be treated as candidates for the nearest neighbors of the input data point. In practice, multiple trees can be built to improve the accuracy of the nearest neighbors. After finding the nearest neighbors of all data points, a $k$NN graph is built. However, to construct a very accurate $k$NN graph requires the construction of many trees, which significantly affects the efficiency. This dilemma has been a bottleneck in applying random projection trees to visualization. LargeVis proposes a new solution: instead of building a large number of trees to obtain a highly accurate $k$NN graph, it uses neighbor exploration techniques to improve the accuracy of a less accurate graph \cite{dong_2011}. Specifically, it builds a few random projection trees to construct an approximate $k$NN graph, whose accuracy may not be that high. Then, for each node of the graph, it searches for neighbors of its neighbors, which are also likely to be candidates of its nearest neighbors. This process can be repeated for multiple iterations to improve the accuracy of the graph. For the edge weights in the $k$NN graph, LargeVis uses the same approach as t-SNE. The conditional probability of data $x_i$ and $x_j$ is first calculated as:

\begin{equation}
    p_{j|i} = \frac{\exp{(-d(x_i, x_j)^2 / 2 \sigma_{i}^2})}{\sum_{(i, k) \in E} \exp{(-d(x_i, x_k)^2 / 2 \sigma_{i}^2)}}, \hspace{1cm} p_{i|i} = 0
    \vspace{0.2cm}
\end{equation}

where the parameter $\sigma_i$ is chosen by setting the perplexity of the conditional distribution $p_{\cdot | i}$ equal to the perplexity $u$. The graph is then symmetrized by setting the weight between $x_i$ and $x_j$ as:

\begin{equation}
    w_{ij} = \frac{p_{j|i} + p_{i|j}}{2N}.
\end{equation}

\subsection{Probabilistic model for graph visualization}

After constructing the $k$NN graph, LargeVis only needs to project the graph nodes into a 2D / 3D space to visualize the data. For this purpose, it introduces an essential probabilistic model for this purpose. The idea is to preserve the similarity of the vertices in a low-dimensional space. In other words, it wants to keep similar vertices close to each other and dissimilar vertices far apart in a low-dimensional space. Given a pair of vertices $(v_{i},v_{j})$, the probability of observing a binary edge $e_{ij} = 1$ between $v_{i}$ and $v_{j}$ is first determined as follows:

\begin{equation}
    P(e_{ij} = 1) = f(||y_{i} - y_{j}||)
    \label{eq:chapter_4_largevis_probability_of_observing_binary_edge}
\end{equation}

where $y_{i}$ is the embedding of the vertex $v_{i}$ in the low-dimensional space, $f(\cdot)$ is a probabilistic function with respect to the distance of the
vertex $y_i$ and $y_j$, that is, $d = ||y_i - y_j||$. When $y_i$ is close to $y_j$ in low-dimensional space (that is, $d$ is small), there is
a high probability of observing a binary edge between the two vertices. In reality, many probabilistic functions can be used, such as $f(x) = \frac{1}{1 + ax^2}$ or $f(x) = \frac{1}{1 + exp{(x^2)}}$. Different probabilistic functions are compared in \cite{largevis}.

Equation \ref{eq:chapter_4_largevis_probability_of_observing_binary_edge} only determines the probability of observing a binary
edge between a pair of vertices. To further extend it to general weighted edges, LargeVis defines the probability of observing a weighted edge $e_{ij} = w_{ij}$ as follows:

\begin{equation}
    P(e_{ij} = w_{ij}) =  P(e_{ij} =1)^{w_{ij}}
    \label{eq:chapter_4_largevis_probability_of_observing_weighted_edge}
\end{equation}

With the above definition, given a weighted graph $G=(V,E)$, the likelihood of the graph can be calculated as follows:

\begin{equation}
    O = \prod_{(i,j) \in E} (p(e_{ij} = 1)^{w_{ij}} \prod_{(i,j) \in \Tilde{E}}(1 - (p(e_{ij} = 1))^\gamma
    \vspace{0.2cm}
    \label{eq:chapter4_largevis_cost_function}
\end{equation}

in which $\Tilde{E}$ is the set of pairs of vertices that are not observed and $gamma$ is the unified weight assigned to the negative edges. The first part of the above equation models the probability of observed edges, and by maximizing this part, similar data points will be kept close together in the low-dimensional space; the second part models the probability of all pairs of vertices without edges, i.e. negative edges. By maximizing this part, the dissimilar data will move away from each other.

\subsubsection{Optimization}

Direct optimization of Eq. \ref{eq:chapter4_largevis_cost_function} is computationally expensive because the number of negative edges is
squared to the number of nodes. Inspired by negative sampling techniques, instead of using all negative edges, LargeVis randomly selects all negative edges to optimize the model. For each vertex $i$, it randomly selects some
vertices $j$ according to the noisy distribution $P_{n}(j)$ and treats $(i,j)$ as negative edges. It used the noisy distribution in the field: $P_{n}(j) ^{0.75}$, where $d_j$ is the degree of the vertex $j$. Let $M$ be the number of negative samples for each positive edge; the objective function can be redefined as:

\begin{equation}
    O = \sum_{(i,j) \in E} w_{ij}(\log p(e_{ij} = 1) + \sum_{k=1}^{M} E_{j_{k} \sim P_n(j)} \gamma \log (1 - p(e_{ij_{k}} = 1)))
    \vspace{0.2cm}
\end{equation}

A simple approach to optimizing the above equation is to use the stochastic gradient descent, although it is problematic. 
This happens because when the edges $(i,j)$ are sampled to update the model, the edge weight $w_{ij}$ will be multiplied by the gradient. 
When the values of the weights are divergent (e.g., from 1 to thousands), the norms of the gradient are also divergent, and in 
this case it is very difficult to choose an appropriate learning rate. Therefore, LargeVis adopts the edge sampling approach proposed in \cite{tang_2015}. 
It randomly samples edges with a probability proportional to their weights and then treats the sampled edges as binary edges. With this edge sampling technique, the objective function remains the same and the learning process will not be affected by the variance of the edge weights.

\subsection{Differences between UMAP, SNE and LargeVis methods}
\label{sec:sne_vs_umap_vs_largevis}

As an aid to illustrate the similarities of UMAP with other neighbor embedding methods (t-SNE and LargeVis), we review the main equations used in these methods and then present the equivalent UMAP expressions. Other 

\subsubsection{Probabilities comparison}

First, consider how the probabilities between two objects \textit{i} and \textit{j} are defined in the high-dimensional input space \textit{X} and the low-dimensional space \textit{Y}. These are normalized and symmetrized in various ways. In a typical implementation, these pairwise quantities are stored and manipulated as (potentially sparse) matrices. Quantities with the subscript $ij$ are symmetric, that is, $v_{ij} = v_{ji}$. Extending it to the conditional probability notation used in SNE, $j | i$ indicates an asymmetric similarity, that is, $v_{j|i} \neq v_{i|j}$. In t-SNE, the probabilities in the input and embedding spaces (Eqs. \ref{eq:tsne_pij}, \ref{eq:tsne_qij}) can be computed for the $k$NN graph, where $p_{j|i}$ is set to zero for non-neighbor points in the $k$NN graph. LargeVis uses the same $p_{ij}$ probabilities as t-SNE but approximates the $k$NN to compute it very quickly and become more efficient. In LargeVis (described in detail in Chapter 4), the probability in the embedding space is:

\begin{equation}
    q_{ij} := (1 + ||y_{i} - y_{j}||^2_{2})^{-1}.
    \label{eq:largevis_qij}
\end{equation}

Comparing Eqs. \ref{eq:umap_probability} and \ref{eq:pji_symmetric_sne} show that UMAP, t-SNE, and LargeVis all use a Gaussian or RBF kernel for probabilities in the input space. Comparing Eqs. \ref{eq:umap_probability_input_space} and \ref{eq:symmetric-tsne-pij} show that UMAP and t-SNE/LargeVis use different approaches to symmetrize probabilities in the input space. Comparing Eqs. \ref{eq:umap_probability_embedding_space}, \ref{eq:tsne_qij}, and \ref{eq:largevis_qij} show that, in contrast to t-SNE, UMAP, and LargeVis do not normalize the probabilities in the embedding space by all pairs of points. This advantage makes UMAP much faster than t-SNE and also makes it more suitable for mini-batch optimization in deep learning \cite{ghojogh_2021}. Comparing Eqs. \ref{eq:umap_probability_embedding_space}, \ref{eq:tsne_qij}, and \ref{eq:largevis_qij} also show that all three methods use the Cauchy distribution for probabilities in the embedding space.

\subsubsection{Cost functions comparison}

LargeVis uses a similar approach to Barnes-Hut t-SNE when approximating $p_{ij}$, but further improves efficiency by only requiring approximate nearest neighbors for each point. For low-dimensional coordinates, it completely abandons the normalization of $w_{ij}$. Rather than using the Kullback-Leibler divergence, it optimizes a likelihood function, and hence is maximized, not minimized:

\begin{equation}
    C_{LV} = \sum_{i \neq j} p_{ij}\ln (q_{ij}) + \gamma \sum_{i \neq j} \ln(1 - q_{ij})
    \vspace{0.4cm}
    \label{eq:largevis_cost_function}
\end{equation}

$p_{ij}$ and $q_{ij}$ are defined as in Barnes-Hut t-SNE (apart from the use of approximate nearest neighbors for $p_{ij}$ , and the fact that, in implementation, LargeVis does not normalize $p_{ij}$ by $N$) and $\gamma$ is a positive constant chosen by the user that weights the strength of repulsive contributions (second term) relative to the attractive contribution (first term). 

Ignoring the two constant terms, the UMAP cost function has a very similar form to that of LargeVis, but without a $\gamma$ term to weight the repulsive component of the cost function and without requiring matrix-wise normalization in the high-dimensional space. The cost function for UMAP can be optimized (in this case, minimized) with stochastic gradient descent in the same way as LargeVis. Comparison of Eqs. \ref{eq:tsne_cost_function}, \ref{eq:largevis_cost_function}, and \ref{eq:umap_cost_function} show that UMAP, t-SNE, and LargeVis have similar but not exactly equal cost functions. The first term in all these cost functions is responsible for the attractive forces and the second term is for the repulsive forces; therefore, they can all be considered as neighbor embedding methods \cite{bohm_2020,damrich2022}.

\begin{figure}[ht!]
     \centering
        \subfloat[]{\includegraphics[width= 0.33\textwidth]{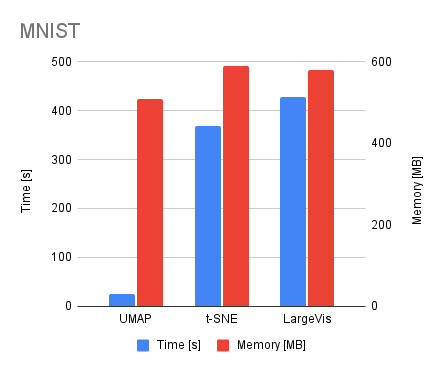}}
        \subfloat[]{\includegraphics[width= 0.33\textwidth]{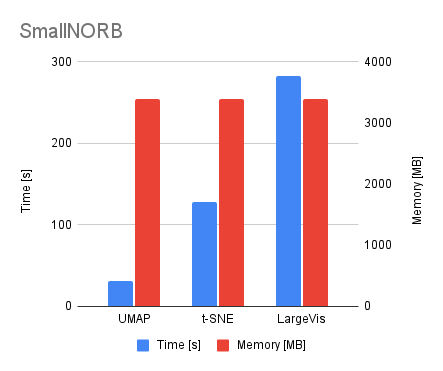}}
        \subfloat[]{\includegraphics[width= 0.33\textwidth]{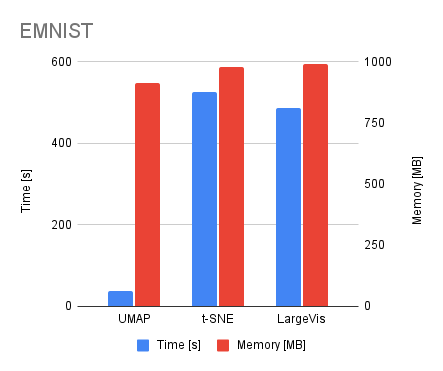}}
    \caption{Time efficiency and memory load of UMAP methods compared to t-SNE and LargeVis.}
    \label{fig:umap_timings_and_memory}
\end{figure}

As shown in Fig. \ref{fig:umap_timings_and_memory}, an experiment was conducted for different datasets in which the results were averaged to show that UMAP clearly leads the way when it comes to embedding time. It is almost an order of magnitude faster than its competitors. On the contrary, all methods have comparable memory occupancy, which is entirely dependent on the size of the dataset.

\section{t-SNE with Euclidean and binary distances}

To evaluate whether the data visualization method can be perceived as embedding of an undirected graph, we create a variation of t-SNE that uses Euclidean and binary distances instead of the probability matrix $p_{ij}$. With this operation, we would be able to parameterize the t-SNE method by the number of nearest neighbors $k$ instead of the perplexity. Other t-SNE calculations remain unchanged. All visualizations were created for a 10\% subset of the MNIST dataset ($M$$=$$7$$\cdot$$10^3$).

\begin{figure}[ht!]
     \centering
        \subfloat[Original t-SNE.]{\includegraphics[width=0.22\textwidth]{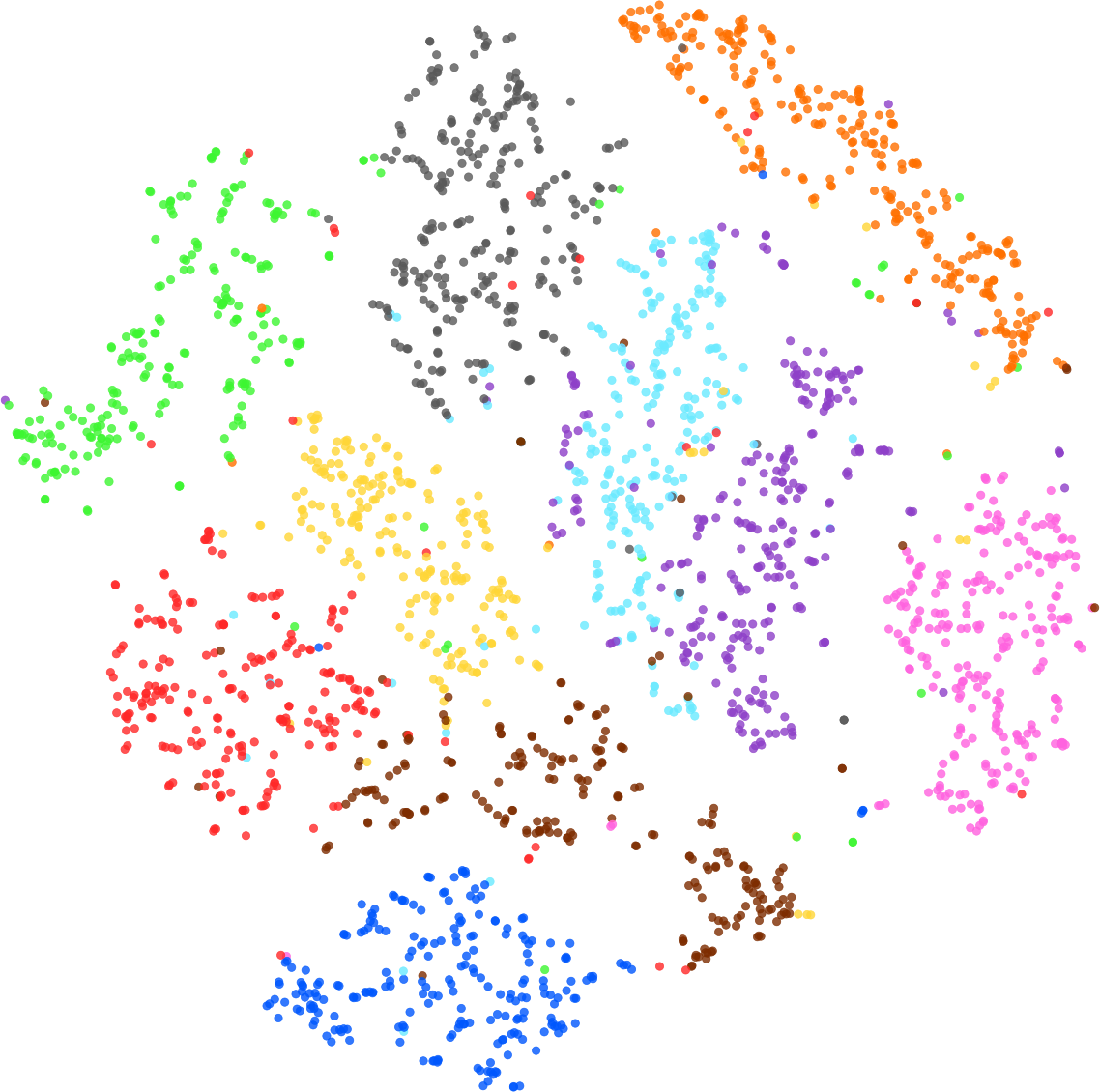}}
        \hspace{0.15cm}
        \subfloat[Binary distances (k=10).]{\includegraphics[width=0.22\textwidth]{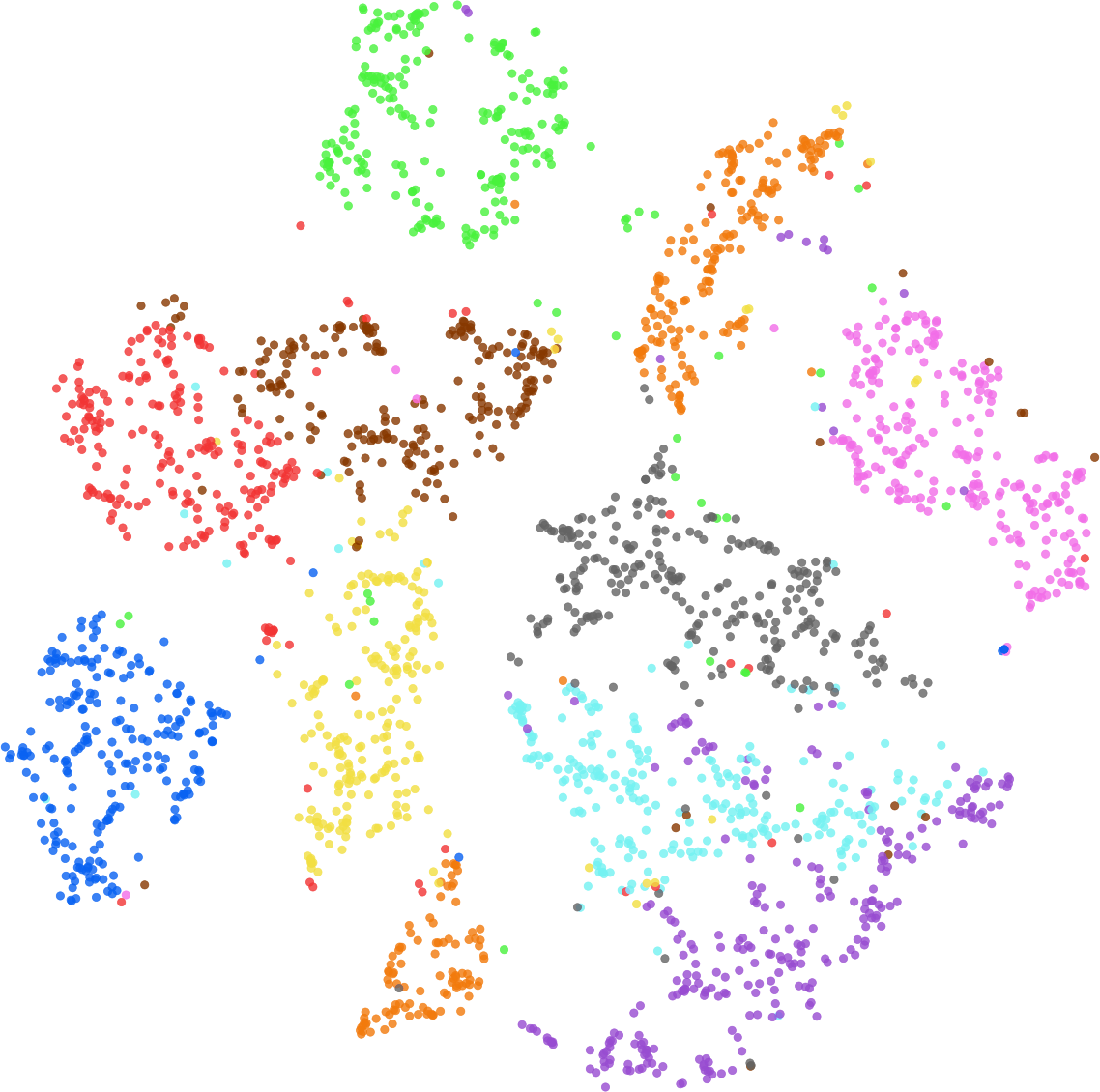}}
        \hspace{0.15cm}
        \subfloat[Binary distances (k=20).]{\includegraphics[width=0.22\textwidth]{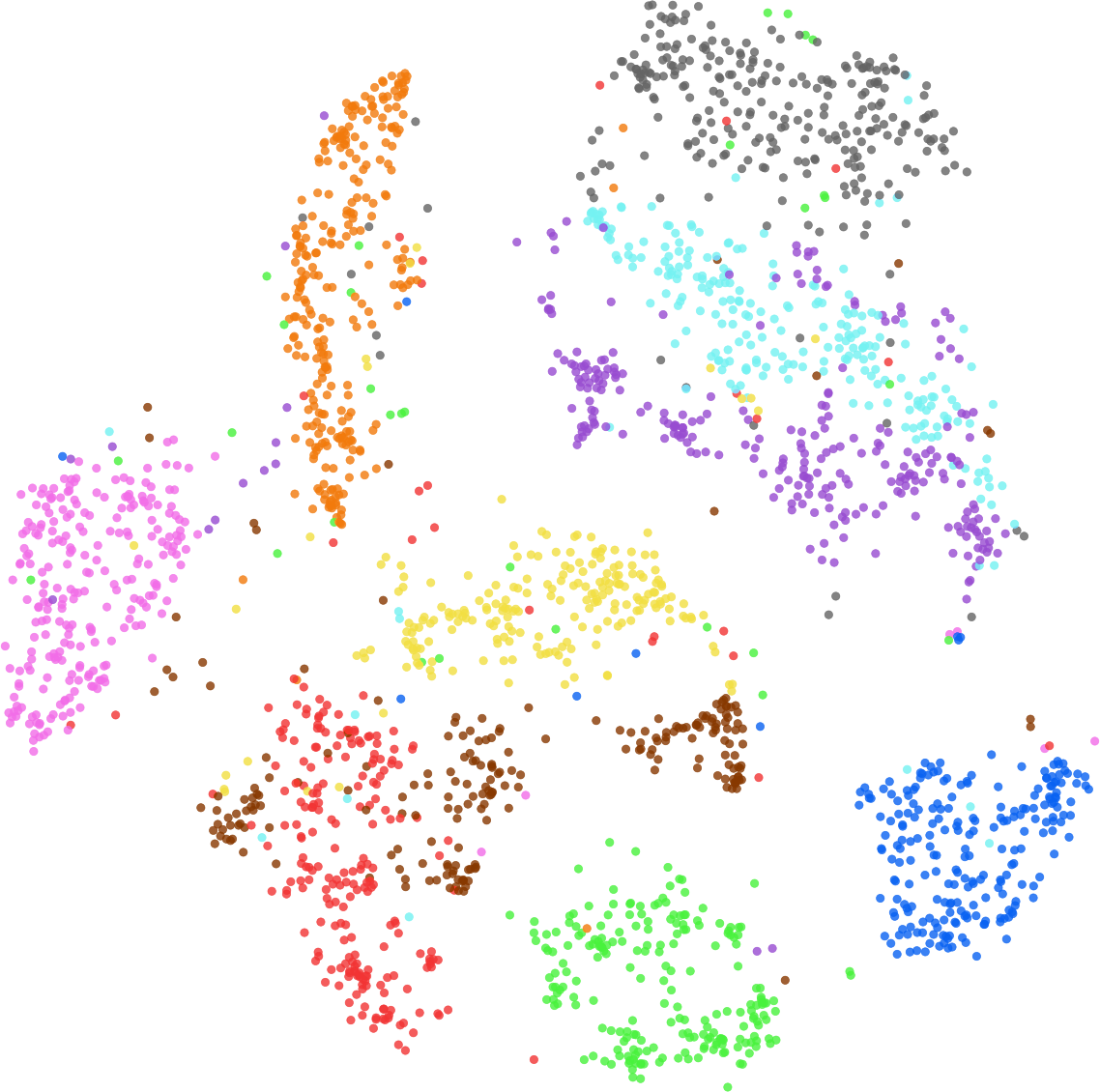}}
        \hspace{0.15cm}
        \subfloat[Binary distances (k=50).]{\includegraphics[width=0.22\textwidth]{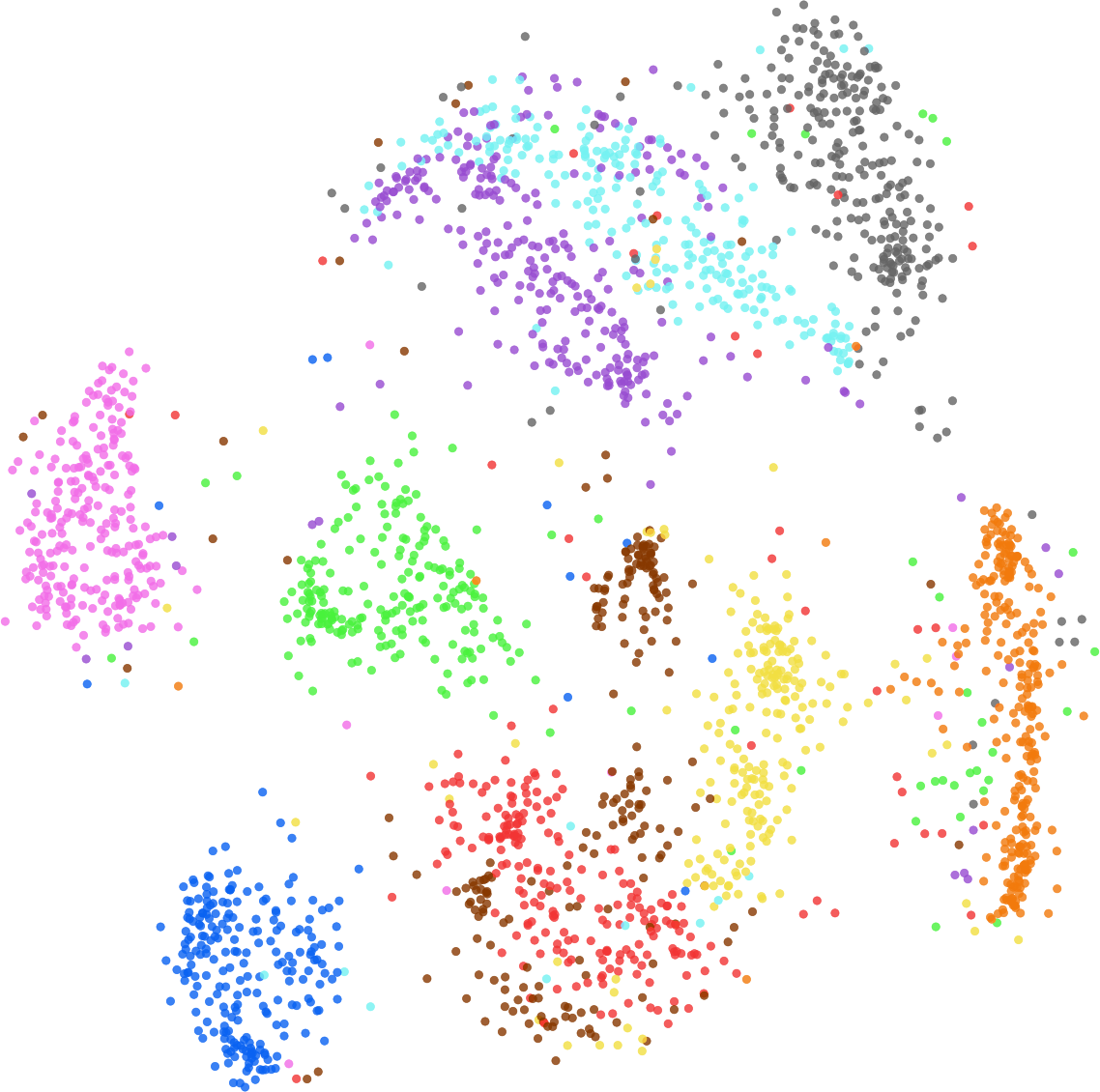}}
        \hfill
        \subfloat[DR quality.]{\includegraphics[width=0.45\textwidth]{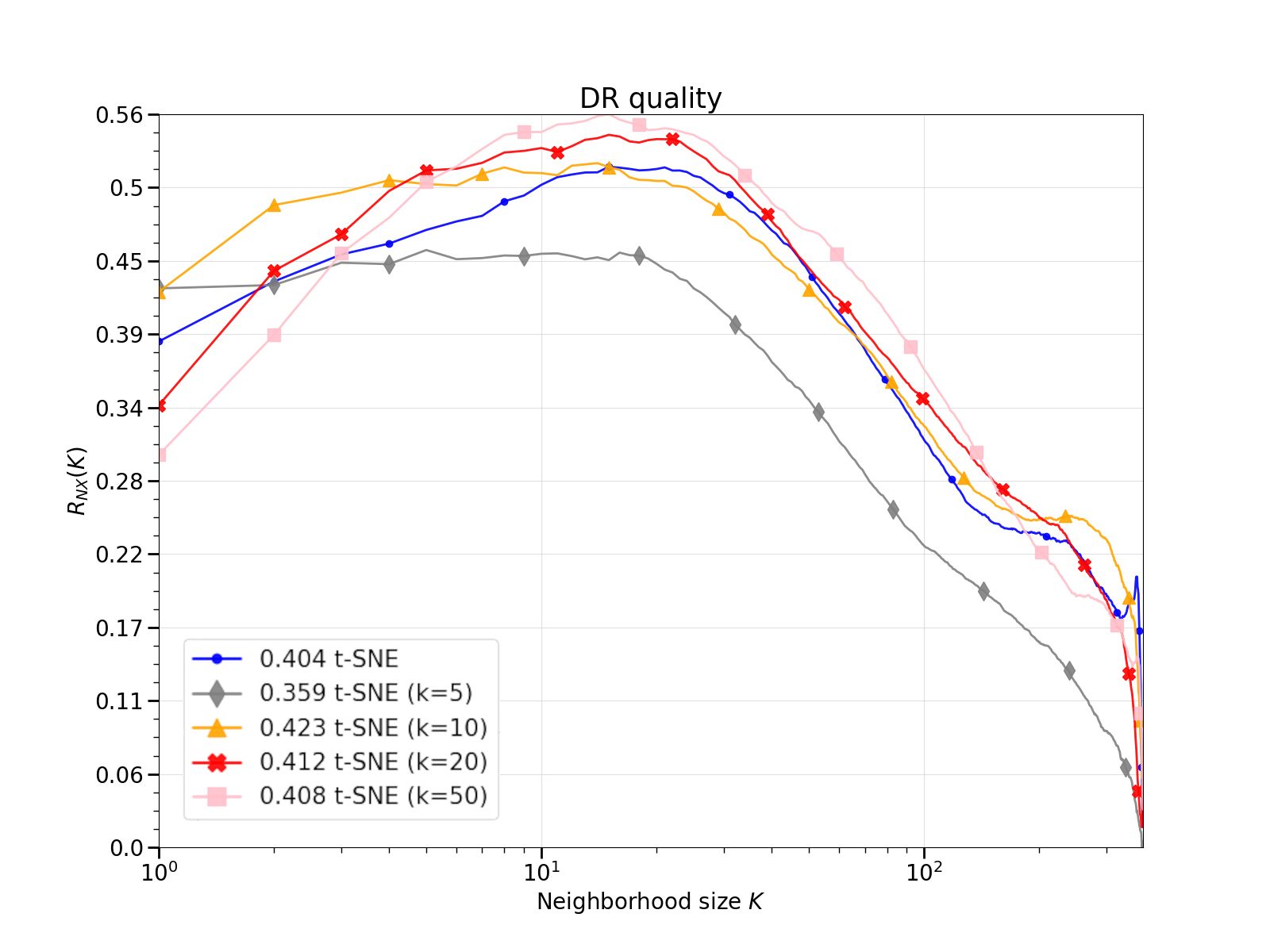}}
        \subfloat[KNN gain.]{\includegraphics[width=0.45\textwidth]{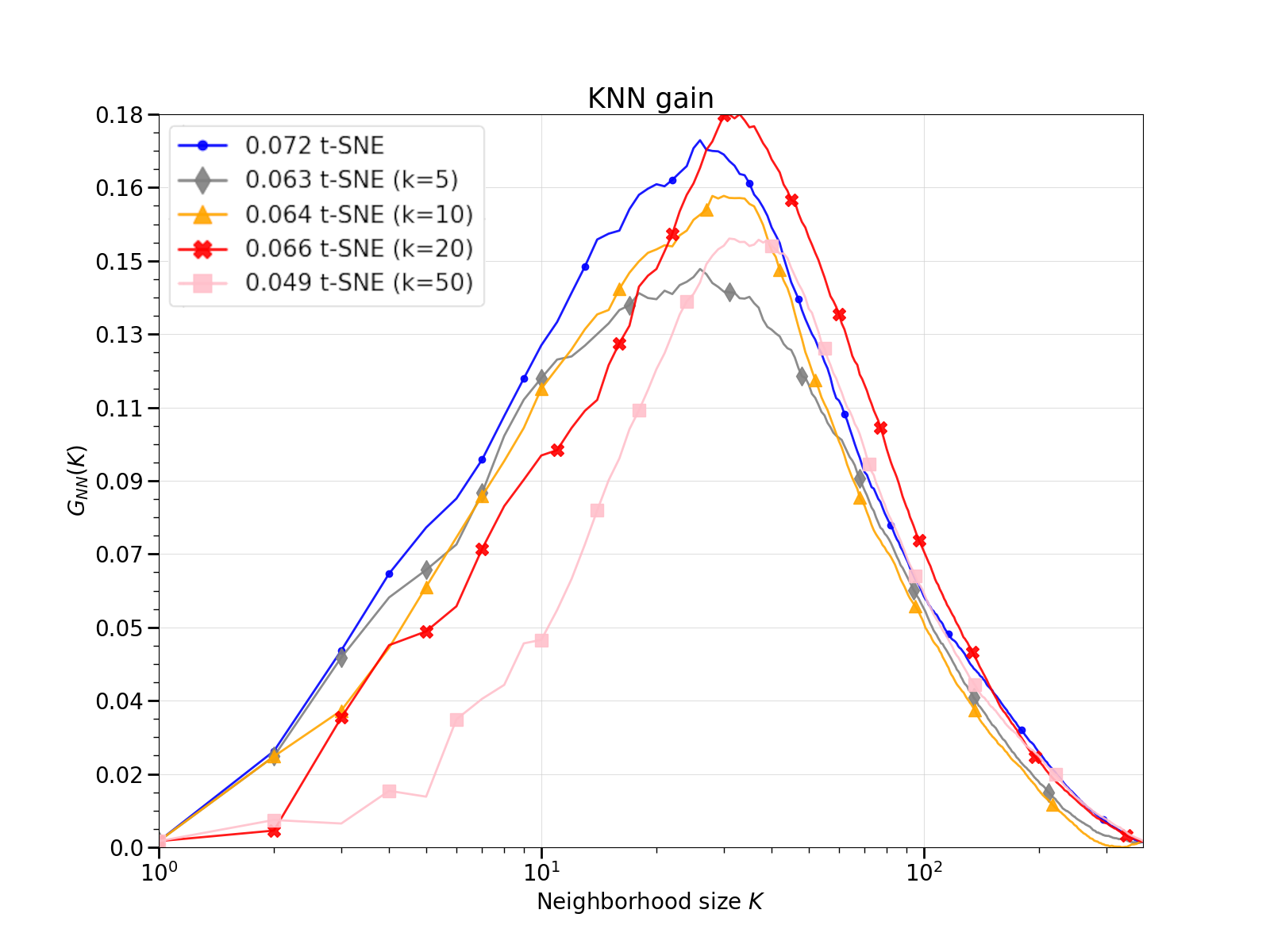}}
        \caption{Visualizations and metrics obtained for binary distances instead of probabilities. Parameterized by the number of nearest neighbors $k$.}
    \label{fig:tsne_binary_distances}
\end{figure}

In case of binary distance, we simply create a binary matrix. We insert 1 where we have determined the nearest neighbors and 0 otherwise. The same mechanism is used for the Euclidean distances, but instead of 1 we insert the actual Euclidean distance into the \textit{probability} matrix.

\begin{figure}[ht!]
     \centering
        \subfloat[Original t-SNE.]{\includegraphics[width=0.22\textwidth]{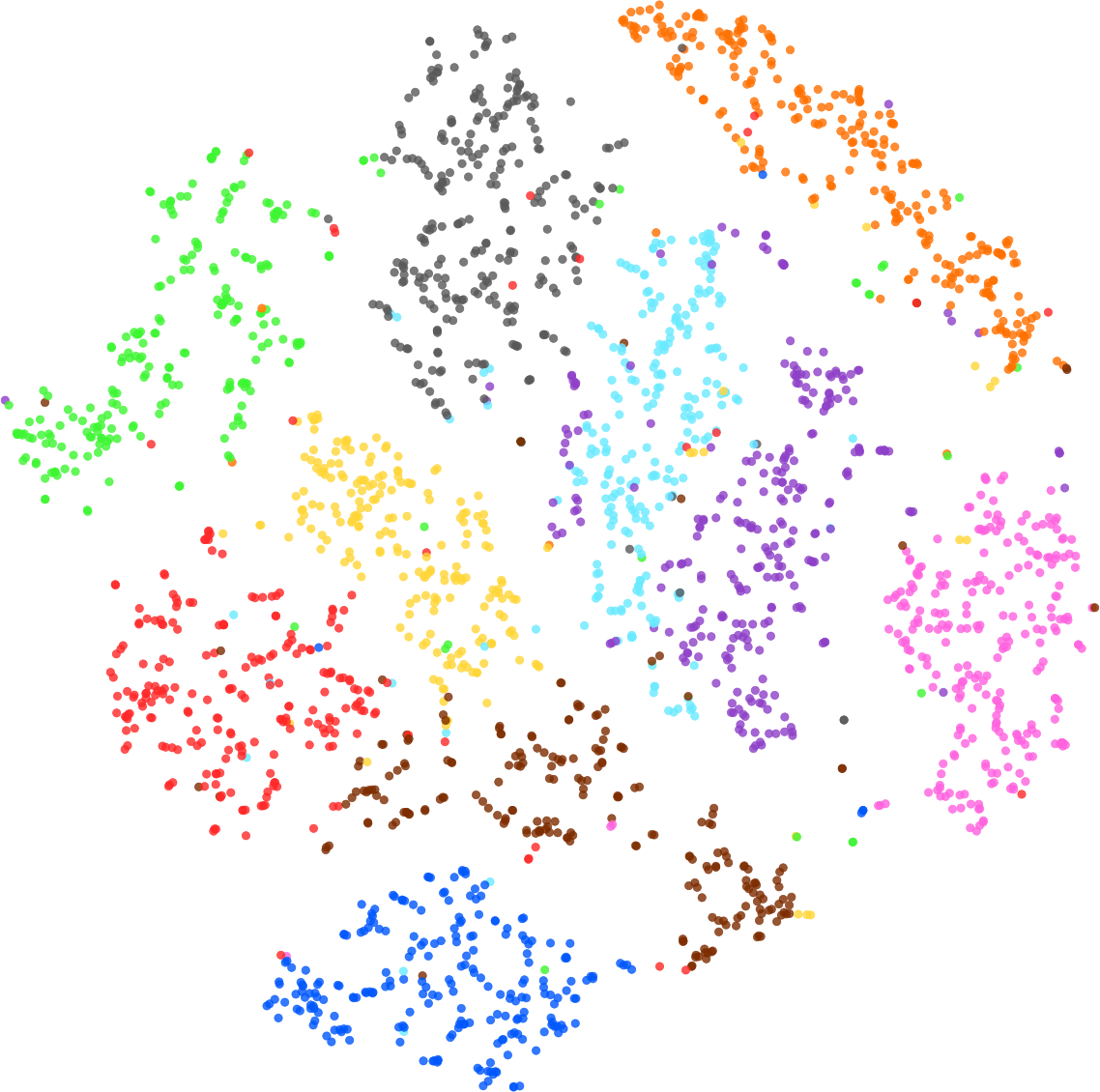}}
        \hspace{0.15cm}
        \subfloat[Euclidean distances (k=10).]{\includegraphics[width=0.22\textwidth]{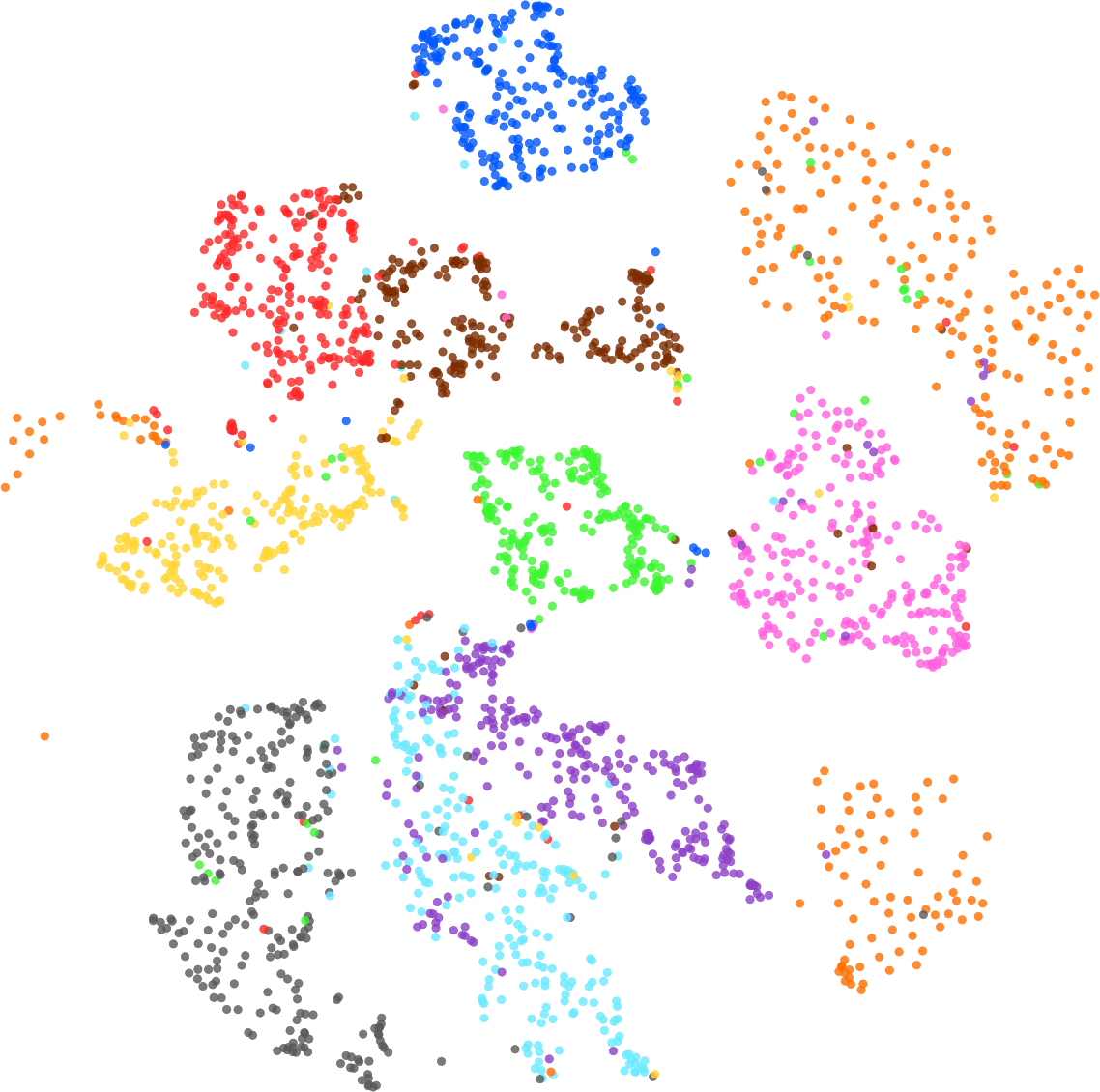}}
        \hspace{0.15cm}
        \subfloat[Euclidean distances (k=20).]{\includegraphics[width=0.22\textwidth]{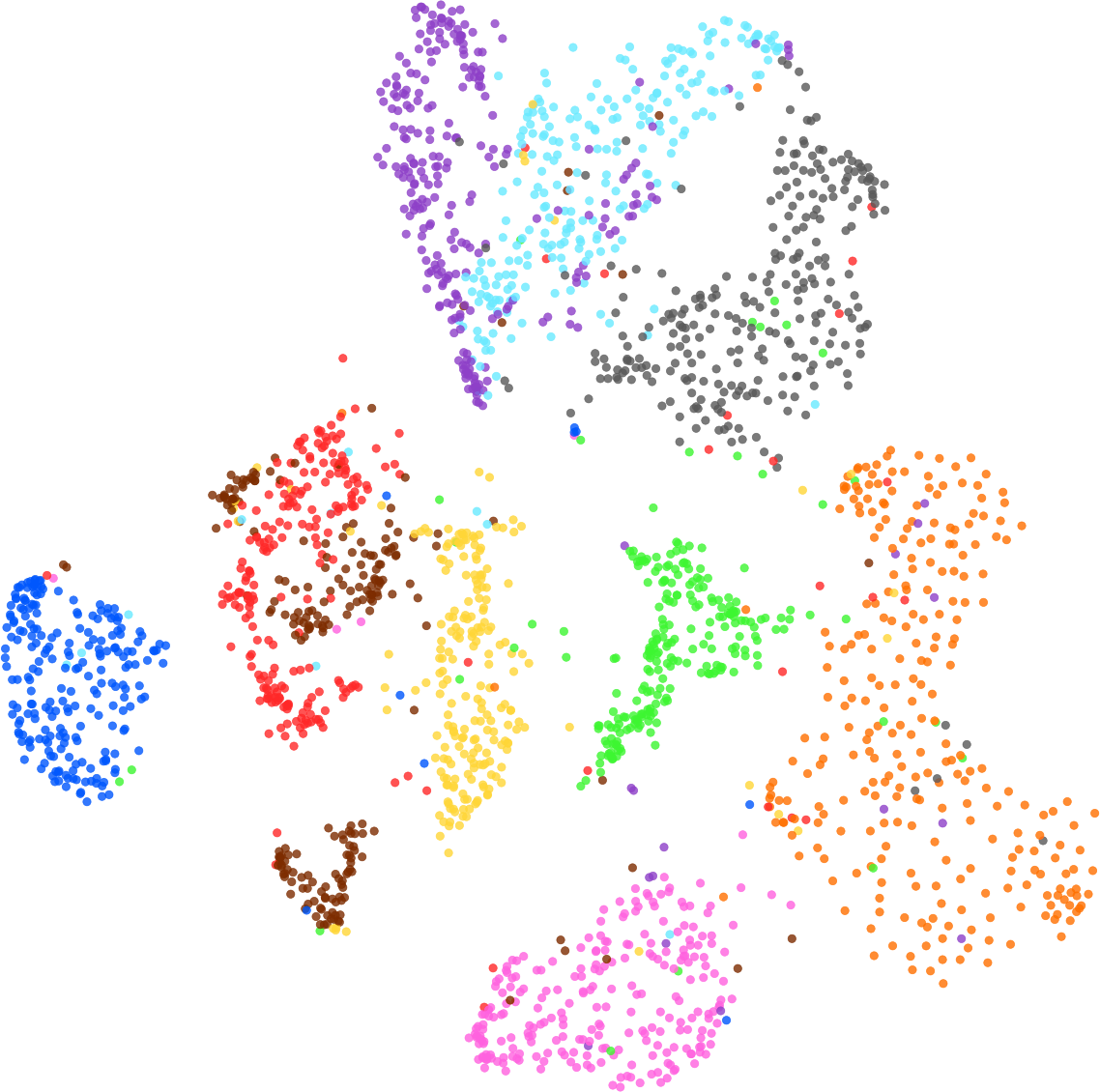}}
        \hspace{0.15cm}
        \subfloat[Euclidean distances (k=50).]{\includegraphics[width=0.22\textwidth]{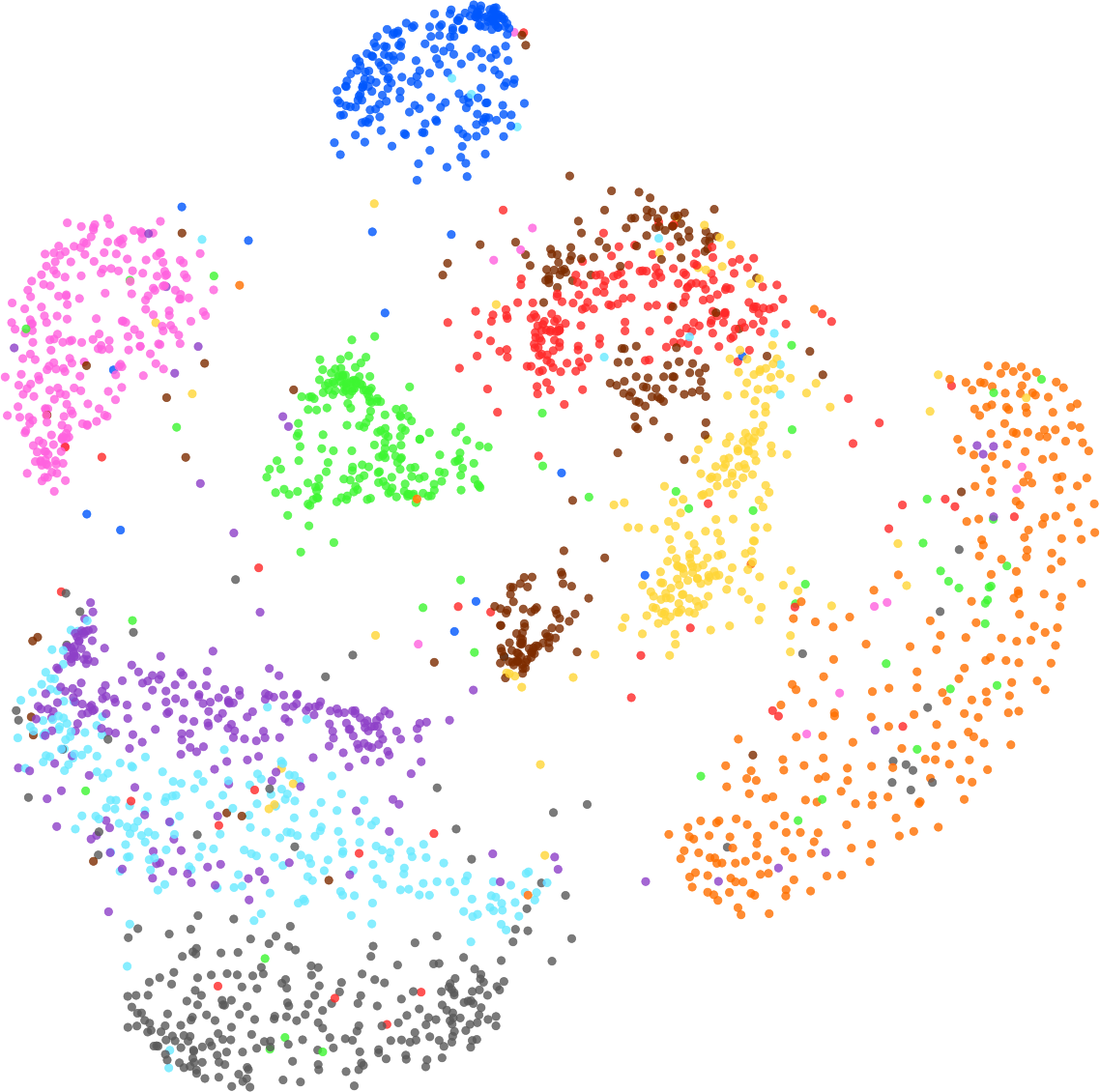}}
        \hfill
        \subfloat[DR quality.]{\includegraphics[width=0.45\textwidth]{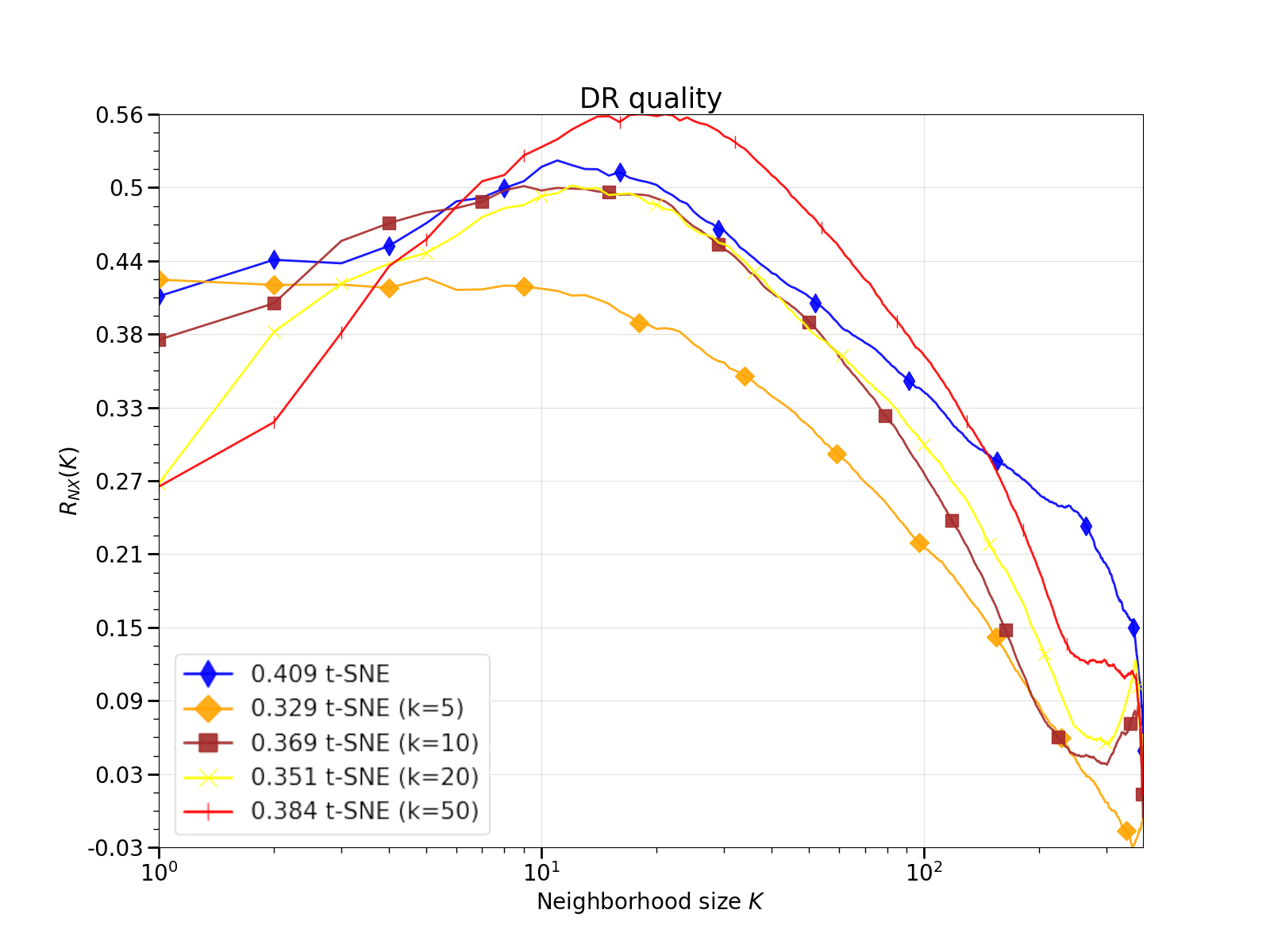}}
        \subfloat[KNN gain.]{\includegraphics[width=0.45\textwidth]{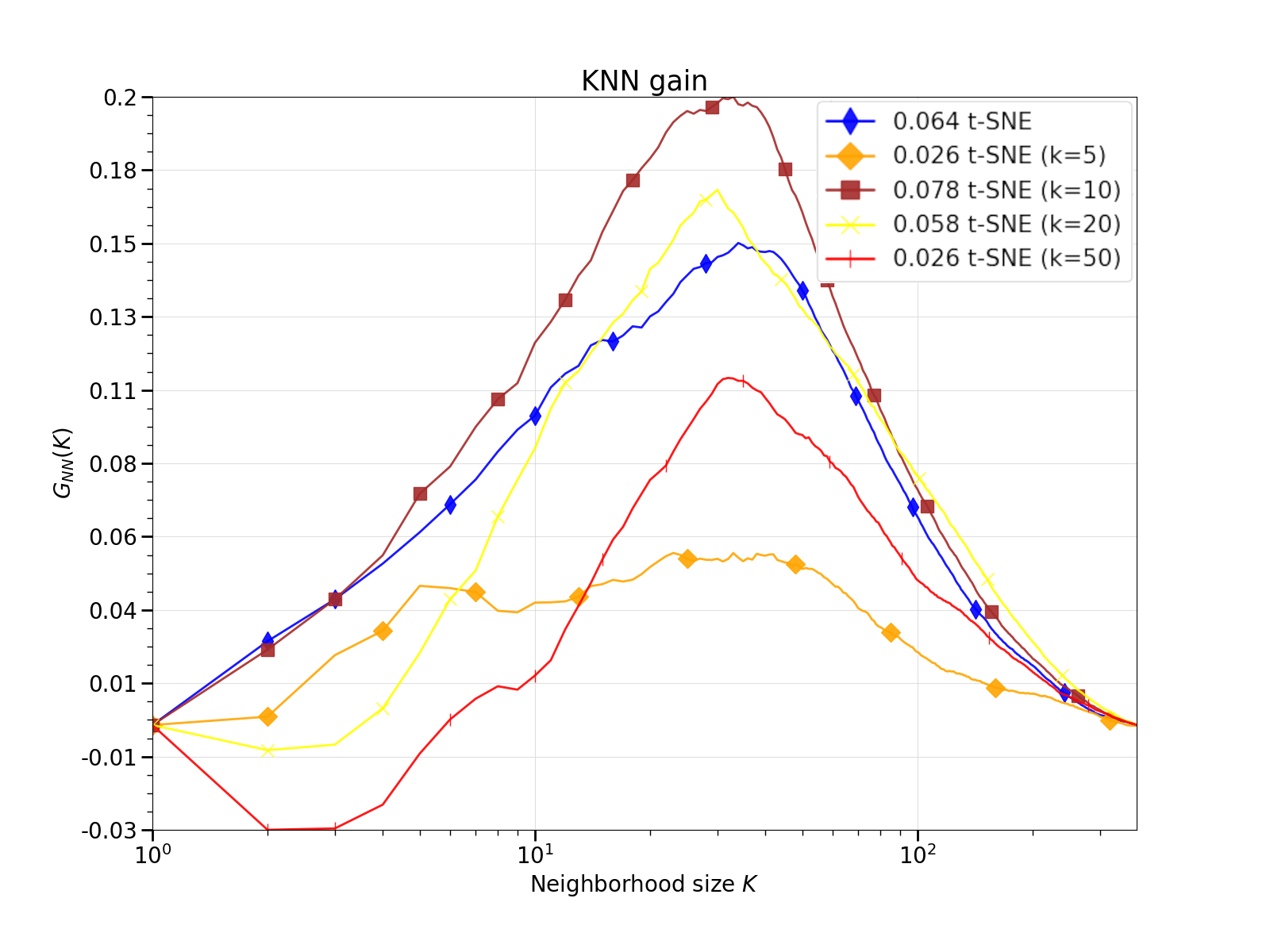}}
        \caption{Visualizations and metrics obtained for Euclidean distances instead of probabilities. Parameterized by the number of nearest neighbors $k$.}
    \label{fig:tsne_euclidean_distances}
\end{figure}

In both cases, we show that for $k$$=$$\{10,20\}$, our t-SNE embedding variant achieves the same data reduction (DR) quality as the original visualization of t-SNE. It is verified in Figures \ref{fig:tsne_binary_distances} and \ref{fig:tsne_euclidean_distances}, where \textit{DR quality} is presented for binary and Euclidean distances. In both cases, we want the curves to at least overlap, which means that the visualization quality is comparable to the original t-SNE implementation. We can observe this effect for $k=10$ using binary distances and for $k=10$ and $k=20$ for Euclidean distances. Intuitively, the more neighbors that are used, the better the quality of visualization, which is reflected by the increasingly higher position of the curves in both graphs (resulting in a higher AUC value).

This observation confirms that we are able to modify the t-SNE method step by step in a way that will ultimately result in the obtaining of the simplest IVHD (Interactive visualization of high-dimensional data) method described in the next Section. The next step would be to swap the part of the algorithm responsible for optimizing the Kullback-Leibler divergence. This is an important feature of data visualization algorithms that rely on neighbor embedding. Their cost function is defined in a way that defines two members. The first member is responsible for the forces of attraction and the second for the forces of repulsion. For this reason, we can think of them as methods described using a common mathematical apparatus imposed by UMAP.

\section{Interactive visualization of high-dimensional data}
\label{sec:ivhd}

As in previous chapters, we assume that each $ND$ feature vector ${x}_{i}$$\in$$\mathbf{X}$$\subset$$\varmathbb{R}^N$ is represented in a 2-D Euclidean space by a corresponding point $\textit{y}_i$$\in$$\mathbf{Y}$$\subset$$\varmathbb{R}^2$ and $i=\{1,\dots, M\}$. We define two distance matrices: $\mathbf{D}=\{\delta_{ij}\}_{M\times M}$ and $\mathbf{d}=\{d_{ij}\}_{M\times M}$, in the spaces {\it source} $\varmathbb{R}^N$ and {\it target} $\varmathbb{R}^2$, respectively.

The value of $\delta_{ij}$ is a measure of dissimilarity (proximity) between the feature vectors $\textit{x}_i$ and $\textit{x}_j$ while in the 2D Euclidean space: $d_{ij}=\sqrt{\|\textit{y}_i - \textit{y}_j\|}$. In general, the source space does not have to be Cartesian and can be solely defined by a proximity matrix $\mathbf{D}$ between any two data objects of a (possibly) different data representation from the vector one (e.g., shapes, graphs, etc.).

Classical MDS performs the mapping as described in Eq. \ref{eq:cost_function_mds}. For IVHD, we assume that $m=2$ and $k=1$. However, there are many other forms of this stress function, which are defined, for example, in~\cite{yang2005,vandermaaten2009,mds}. One can easily adopt our approach to particular ($k$, $m$) parameters choice. However, finding the global minimum of this multidimensional and multimodal cost function (\ref{eq:cost_function_mds}) with respect to $\textit{Y}$ is not a trivial problem. To this end, we use the force-directed approach presented in~\cite{dzwinel1997virtual,pawliczek2013interactive,dzwinel2014,pawliczek2015visual} and described in detail in a later section of this chapter.

We assume that the set of points $\mathbf{Y}$ is treated as an ensemble of interacting particles $\textit{y}_i$. Particles evolve in $\varmathbb{R}^2$ space with discrete time $t$ scaled by time step $\Delta t$, according to Newtonian equations of motion discretized using the {\it leapfrog} numerical scheme. We use their simplified form~\cite{dzwinel2017ivga}:

\begin{equation}\label{eq:ivhd_force_1}
\Delta\textit{y}_i \leftarrow a{\cdot}\Delta\textit{y}_i+b{\cdot}\textit{f}_i,
\end{equation}

\begin{equation}\label{eq:ivhd_force_2}
\textit{f}_i^{\;t} = -\nabla\left(\sum_{j=1}^M\left(\delta_{ij}^t - d_{ij}^t\right)^2  \right),
\end{equation}

\begin{equation}\label{eq:ivhd_position_update}
\textit{y}_i \leftarrow \textit{y}_i + \Delta\textit{y}_i,
\end{equation}

\noindent what resembles well-known momentum minimization method. Aforementioned Equations and differences to momentum method are described in details in Section \ref{subsec:optimization_methods}. The value of $a\in [0,1]$ represents friction and is equal to 1 in the absence of energy dissipation. Meanwhile, $b$ parameterizes the forces between the particles. The proper balance of $a$ and $b$ is crucial for the convergence speed of the particle system to a stable and good (close to the global one) minimum of the stress function (Eq. \ref{eq:cost_function_mds}). It is demonstrated in~\cite{dzwinel1997virtual,pawliczek2013interactive,dzwinel2014,pawliczek2015visual}  that this formulation of the classical MDS algorithm produces acceptable embeddings for low-dimensional ($N<10$) and rather small datasets ($M\sim 10^3$) in sublinear time complexity, that is, similar to widely used stochastic gradient descent (SGD) algorithms and its clones employed, e.g., in the original implementation of t-SNE~\cite{tsne}. 
The main problems with MDS for $N>10$ are both the effect of \textit{curse of dimensionality}, which produces poor quality embeddings, and the high computational and storage complexity for $M\sim 10^{5+}$.

The linear time and memory complexity of the embedding of the data can be achieved by using only a limited number of distances from \textit{D} and the corresponding distances from \textit{d} (such as in~\cite{dzwinel1997virtual,pawliczek2013interactive,dzwinel2014,pawliczek2015visual,ingram2009glimmer}). In the context of the so-called theory of structural rigidity~\cite{thorpe1999rigidity}, all the distances between the samples in a $n$-D dataset are not needed to retain the rigidity of its shape in a $n$-D space. The term \textit{rigidity} can be understood as a property of a $n$-D structure made of rods (distances) and joints (data vectors) such that it does not bend or flex under an applied force. Therefore, to ensure the rigidity of \textit{X} (and its 2-D embedding \textit{Y}), only a fraction of the distances (joints) from \textit{D} (and~\textit{d}) is required. What is the minimum number of distances for which the original \textit{X} and target \textit{Y} data sets remain rigid?

As shown in~\cite{thorpe1999rigidity}, a minimal $n$ rigid structure, which consists of $M\geq n$ joints (vectors) in a $n$-dimensional space, requires at least:

\begin{equation}\label{eq:eq10}
    L(n)_{\min}=n{\cdot} M - \frac{n{\cdot}(n+1)}{2}
\end{equation}

\noindent rigid rods (distance). This means that in the source space, the number of distances $L(N)_{\min}$ can ensure a lossless reconstruction of the data structure in $N$-D. 
In particular, in 2-D the structural rigidity can be preserved for $L(2)_{\min}=2{\cdot} M-3$. However, $L_{\min}$ defines only the lowest bound of $L$ required to maintain structural rigidity. Therefore, IVHD answers the following questions:

\begin{enumerate}
    \item What minimum number of distances in \textit{X} should be known to obtain a reasonable reconstruction of the data in \textit{Y}, simultaneously, preserving the structural rigidity of \textit{Y}? Is it closer to $N$ or rather to 2?
    \item What distances should be retained?
\end{enumerate}

In general, $\mathbf{D}$ cannot be an Euclidean matrix. It may represent the proximity of samples in an abstract space. In particular, samples $\textit{x}_i$$\in$$\textbf{X}$ can occupy a complicated $n$-D manifold for which $n\ll N$ (see Figure~\ref{fig:general_idea_of_embedding_by_nn_graph}). Then we can assume that only the distances of each sample $\textit{x}_{i}$ from its nearest neighbors ($nn$) are Euclidean. 
As shown in~\cite{tenenbaum2001,dzwinel2017ivga}, the $k$-nearest neighbor graph ($k$NN graph), in which the vertices represent the data samples and the edges represent the connections to their nearest neighbors, can be treated as an approximation of this low-dimensional manifold, immersed in a high-dimensional feature space (see Figure~\ref{fig:tehtrahedron_embedding}). In the CDA and Isomap~\cite{yang2005,tenenbaum2001} DR algorithms, the proximities of the more distant graph vertices (samples) are calculated as the lengths of the shortest paths between them in a respective $k$NN graph. However, although in this way we can more precisely approximate the real distances in the low-dimensional manifold, the full distance matrix \textit{D} has to be calculated by the very time-consuming $O(M^3)$ Dijkstra algorithm (or Floyd-Warshall~\cite{floyd1962algorithm}) and, even more demanding, has to be stored in operational memory. Calculating the complete \textit{D} is not necessary, and the problem of data embedding can be replaced by the congruent problem of visualization of the $k$NN graph.

Graph visualization (GV) and DR methodology share many common concepts. For example, the metaphor of the particle system and the force-directed method were used to minimize the cost function, which was introduced independently into GV and DR~\cite{dzwinel1997virtual,fruchterman1991graph}. In summary, as shown in Figure~\ref{fig:general_idea_of_embedding_by_nn_graph}, the first step of the {\bf IVHD} DR method consists of the construction of a $k$NN graph, which approximates the $n$ non-Cartesian dimensional manifold immersed in $\varmathbb{R}^N$. Then, we can use the fast procedure for graph visualization in the 2-D space presented in~\cite{dzwinel2017ivga}. 

Let us assume that $G(V,E)$ is the $k$NN graph for the high-dimensional dataset $\mathbf{X}$. 
The data samples \mbox{$\textit{x}_i \in \mathbf{X}$} correspond to the graph vertices  $v_i$$\in$$V$, while $E$ is the set of edges connecting each $\textit{x}_{i}$ with their $k$ nearest neighbors. Because we have assumed (see Figure~\ref{fig:general_idea_of_embedding_by_nn_graph}) that this graph is an approximation of the $n$-dimensional manifold, consequently, we assume that its topology-preserving embedding in 2-D Euclidean space will produce similar results as DE algorithms. According to the definition formulated in~\cite{shaw2009structure}:

\begin{definition}
The topology is preserved if a connectivity algorithm, such as the $k$NN, can easily recover the edges of the input graph from the coordinates of the nodes after embedding.
\end{definition}

This means that unlike DR algorithms, which tend to retain the order of neighbors, the ordering of nearest neighbors (usually a few) is irrelevant in this case. The requirement of preserving the order of a small number of nearest neighbors $nn$ for each $\textit{x}_i \in \mathbf{X}$, which are subject to uncertainty and measurement errors, is secondary in terms of visualization of large data. The crucial problem in formulating hypotheses \textit{ad hoc} and deciding on the use of particular machine learning tools in further data analysis is revealing the data topology, that is, its multiscale cluster structure.


\subsection{Improving classical MDS}

In addition to the assumptions of the previous chapter, we assume that by imposing a high contrast on these two types of distance, we will be able to preserve in $\mathbf{Y}$ the multiscale cluster structure of $\mathbf{X}$, by employing MDS mapping only on those binary distances $L$. This way we could decrease both the computational complexity of data embedding to $O(a{\cdot} M)$ with \mbox{$a\sim n_{vi}$} and its computational load to only $nn{\cdot} M$ integers.

To obtain the 2-D embedding $\mathbf{Y}$ of $G(\mathbb{V},\mathbb{E})$, IVHD minimize the following stress function:

\begin{equation}
    \label{eq:mds_stress_function}
    E(\|\mathbf{D}-\mathbf{d}\|)=
    \sum_i
    \ \sum_{j \in O_{\text{\it nn}}(i)\cup O_{\text{\it rn}}(i)}
    \!\!\!\!\!\!\!\!\!\!\!b\left(\delta_{ij}-d_{ij}\right)^2.
    \vspace{0.3cm}
\end{equation}

Consequently, the interparticle force $\textit{f}_i$ from Eq.~(\ref{eq:ivhd_force_2}) in $E(.)$ minimization procedure simplifies to:

\begin{equation}
    \label{eq:eq13}
    \textit{f}_i^n=
    -\sum_{j \in O_{\text{\it nn}}(i)}^{nn}\textit{y}_{ij}^n
    -c\!\!\!\!\sum_{k \in O_{\text{\it rn}}(i)}^{rn}
    \!\!\left(1-d_{ik}^n\right){\cdot}\ \frac{\textit{y}_{ik}^n}{d_{ik}^n},
    \quad \textit{y}_{ik}^n = \textit{y}_i^n-\textit{y}_k^n.
    \vspace{0.3cm}
\end{equation}

To explain this concept in terms of data embedding, let us use a toy example.

\begin{figure}[ht!]
    \begin{center}
    \includegraphics[width=12cm]{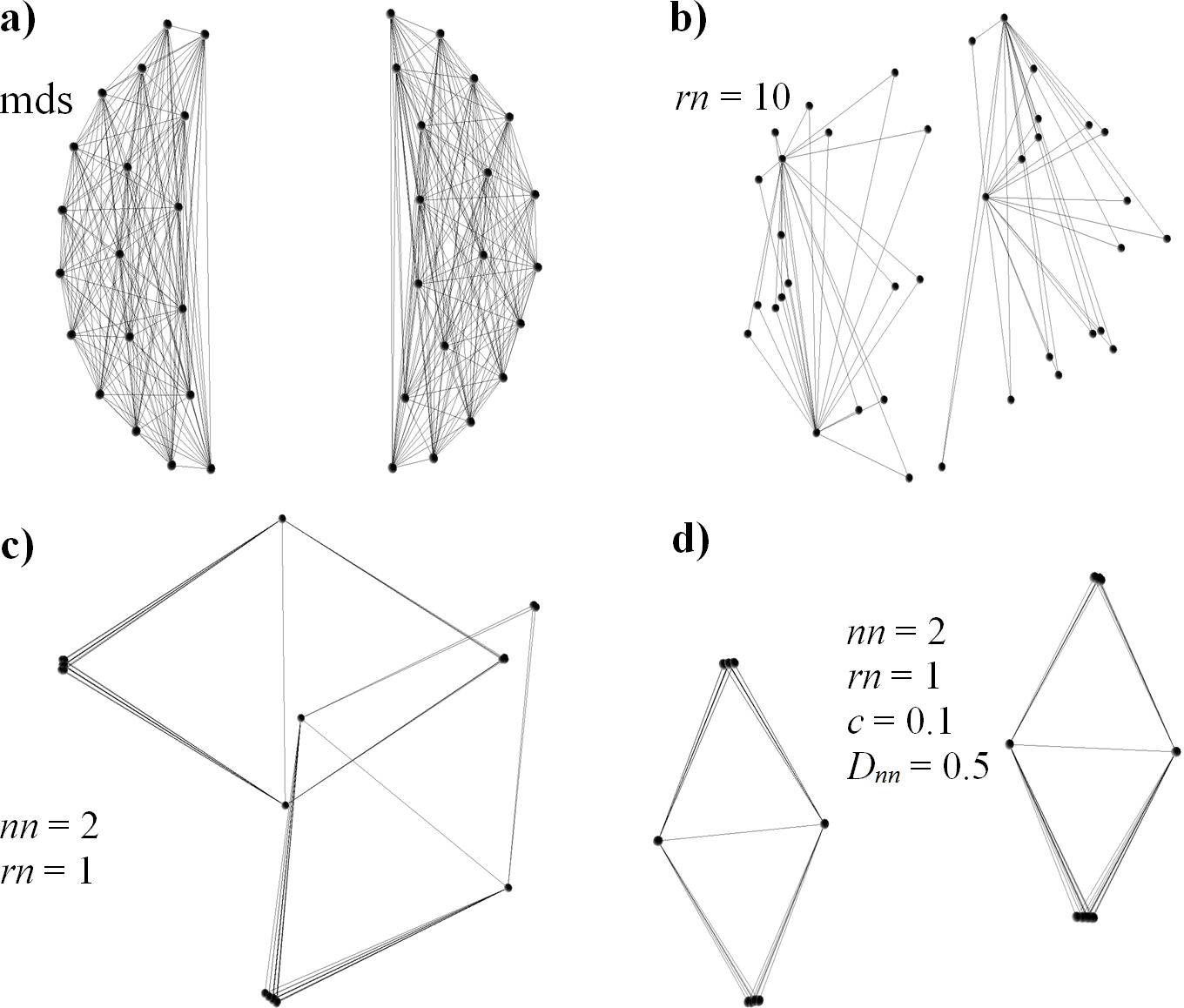}
    \end{center}
    \caption{The results of 2-D embedding of two identical regular hypertetrahedrons for (a) original MDS setting and (b-d) highly reduced number of distances in the cost function (1). Source: \cite{dzwinel2017}.}
    \label{fig:tehtrahedron_embedding}
\end{figure}

Assume that $\mathbf{X}$ consists of $M=38$ samples located in the vertices of two identical but translated and mutually rotated regular 18-dimensional hypertetrahedrons. In fact, the data create an unconnected graph where the vertices are joined by the hypertetrahedron edges. As shown in Figure~\ref{fig:tehtrahedron_embedding}a, by employing classical MDS with the cost function (\ref{eq:cost_function_mds}) and the data set defined by the complete distance matrix $\mathbf{D}$ ($L=703$ distances), we have obtained the embedding $\mathbf{Y}$ shown in Figure~\ref{fig:tehtrahedron_embedding}a. One can observe that the result is not satisfactory.  Although it shows the separability of the data well, the local structure of $\mathbf{Y}$ remains very unclear. Now, assume that we know only $rn=10$ distances from $\mathbf{D}$ to random neighbors for every $\textit{x}_i \in \mathbf{X}$ ($L=300$ distance). As expected, the final embedding from Figure~\ref{fig:tehtrahedron_embedding}b shows that data separation is still visible, but lacks any fine-grained structure. As shown in Figure~\ref{fig:tehtrahedron_embedding}c, this flaw can be partially eliminated by also taking into account the distances to the nearest neighbors $nn$ of $\textit{x}_{i}$, that is, in this particular case $nn=2$ and $rn=1$ were established ($L=103$ distances). Furthermore, we have reduced the strong bias caused by long-range interactions between the $\textit{x}_{i}$ samples and their random neighbors, by decreasing the scaling factor of the respective \textit{forces} ($c=0.1$ in Eq.~(\ref{eq:eq13})) 10 times. Unlike in Figure~\ref{fig:tehtrahedron_embedding}b, in Figure~\ref{fig:tehtrahedron_embedding}c we observe a much better reconstruction of the local structure \textit{Y}, but at the cost of worse data separation. The embeddings of two regular hypertetrahedrons overlap each other.

Now, assume that all the distances between all the samples $M$ are equal to 1. To increase the contrast between the $nn$ nearest and $rn$ random neighbors of $\textit{y}_{i}$, we assume that $D_{\text{\it nn}}=0.5$ and $D_{\text{\it rn}}=1$. Since now we use binary distances $\{0.5, 1\}$, we do not need to store floating point matrix $\mathbf{D}$ but just remember the $nn$ indices (integers) of each $\textit{y}_{i}$ to their nearest neighbors (here 70 integers in total).  As shown in Figure~\ref{fig:tehtrahedron_embedding}d, the 2-D embedding clearly reconstructs the local and global structure of the $nn$-graph representing the original 18-dimensional data. Because $M$ and $N$ were small in this example, the value of $D_{\text{\it nn}}=0.5$, was sufficient to properly contrast the distances $nn$ and $rn$.
However, as we show above (see Eq.~(\ref{eq:mds_stress_function})), for larger $M$ and $N$, $D_{\text{\it nn}}$ should be equal to 0 to reduce the effect of \textit{curse of dimensionality}. It is worth mentioning here that the neighborhood relation is not symmetrical; thus, $(n=nn+rn){\cdot} M$ can be smaller than $L(2)_{\min}$ for $n=2$. Consequently, it may produce non-rigid and deformed (2-D) embeddings. Furthermore, assuming that $nn=1$, we can obtain meaningless 2-D embeddings, although $rn$ is much greater than 2. In summary, we can state the following conjectures.

\begin{enumerate}
    \item The number of distances from $\mathbf{D}$, sufficient to obtain the 2-D embedding $\mathbf{Y}$ of the original $N$-D dataset $\mathbf{X}$, which preserves the local and global properties of $\mathbf{X}$, can be surprisingly small and close to $\sim 3{\cdot} M$.
    \item For each $\textit{x}_i \in \mathbf{X}$, its vicinity consisting of a few nearest neighbors $nn$ should be preserved, while the global cluster structure is controlled by a few (often one) $rn$ randomly selected neighbors.
    \item The distances between $\textit{x}_i \in \mathbf{X}$ and their nearest and random neighbors, and the respective forces in the optimization procedure, should be properly contrasted. For high-dimensional data, the binary distance can be used. This can drastically reduce the memory load from $M{\cdot} (M-1)$ floating points (two distance arrays $\mathbf{D}$ and $\mathbf{d}$) to $nn{\cdot} M$ integers, where $nn\sim 3$.
\end{enumerate}

Many types of force-directed methods and algorithms have been used for both vector visualization and data embedding problems \cite{gibson2013,largevis}. Neglecting technical details, all of them are based on N-body dynamics, where the low-dimensional embedding of a graph (or a data set) is represented by an ensemble of interacting particles evolving in 2-D (or 3- D) Euclidean space. However, to adapt it directly to graph visualization, a precise definition of the graph in terms of a dissimilarity matrix is required. The stress function, similar to that of MDS, could then be applied as the cost function in graph visualization. However, the number and kind of distances that IVHD employs are radically different from those of the classical MDS algorithm.

\begin{figure}[ht]
    \begin{center}
        \includegraphics[clip,width=\columnwidth]{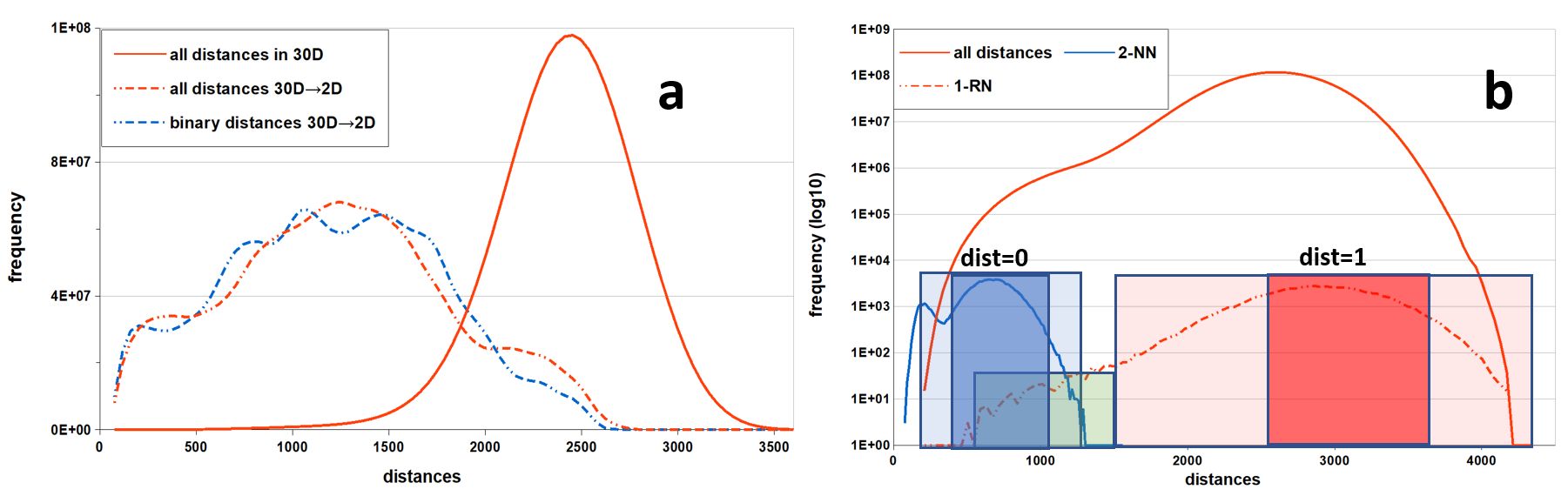}%
        \caption{The envelopes of histograms for MNIST dataset (after PCA transformation $784D\rightarrow30D$). \textit{Red solid line}:  all $\textbf{D}$  distances (a - linear, b - logarithmic scale); a) \textit{dashed lines}: all $\textbf{d}$ distances; b) \textit{blue solid line}: $\textbf{D}$  distances only between samples and their 2-NNs, and \textit{red dashed line}: $\textbf{D}$  distances only between samples and one random neighbor. Source: \cite{minch2020}.}
    \end{center}
\end{figure}

In Fig. 3.6, we show the envelopes of the distance histograms $\textbf{D}$ and $\textbf{d}$ for the MNIST dataset before and after IVHD embedding. Although the MNIST dataset has a varied structure, the envelope of the $\textbf{D}$ histogram in linear coordinates (Fig.3.6a) is perfectly bell-shaped, while that of $\textbf{d}$ is more deformed, but still resembles the Cauchy distribution. To increase distance diversification (see Fig.3.6b), instead of all $M(M-1)/2$ floating point distances, we can consider binary distances for only a few ($nn$) nearest neighbors and just one ($rn$) randomly selected neighbor. This is because most of the real distances ($95\%$) from the nearest and random neighbors are located in separated and rather distant intervals (darker blue and red boxes in Fig.3.6b). For higher dimensions, the random neighbors are almost equidistant from $y_{i}$ due to the "curse of dimensionality" effect. The overlap region (green) contains only $0.3\%$ distances. Therefore, we can also assume that $O_{\text{\it nn}}(i) \cap O_{\text{\it rn}}(i)=\emptyset$. However, this assumption is superfluous because the probability of choosing the nearest neighbor as a random neighbor is negligibly small for large $N$.  As shown in Fig.3.6a, for non-binarized and binarized source distances, their histograms for respective 2D embeddings are very similar.  Thus, let $O_{\text{\it nn}}(i)$ and $O_{\text{\it rn}}(i)$ will be the sets of indices of the $nn$ nearest (connected) neighbors and the $rn$ (unconnected) random neighbors of a feature vector $y_i$ in \textit{k}NN-graph, respectively. We define the binary dissimilarity measure as follows:

\vspace{0.3cm}
\begin{equation}
    \forall x_i \in \mathbf{X}:D_{ij}=
    \begin{dcases}
        0 & \text{\it if} \hspace{0.3cm} j \in O_{\text{\it nn}}(i) \\
        1 & \text{\it if} \hspace{0.3cm} j \in O_{\text{\it rn}}(i)
    \end{dcases}
    .
    \label{eq:ivhd_binary_distances}
\end{equation}
\vspace{0.3cm}

Thus, unlike the UMAP and t-SNE algorithms, IVHD is not interested in even an approximate ordering of $k$NN for each $x_i\in \mathbf{X}$. This is justified for small $nn$, because the distances to the first few NNs, in general, cannot differ too much (see the blue plot in Fig. 4.5b), and the ordering of $NN$ can result from measurement errors. We assume that the number of neighbors $nn$ must meet two conditions. First, the $k$NN-graph should be fully connected (or approximately, that is, the size of the largest component should be comparable to the size of the entire graph). Second, the $kNN$-graph augmented with approximately $rn$ edges should be at least a minimal $n$-rigid graph (in 2D: 2-rigid). The lower band of the number of connections $L$, required to make the 2-rigid augmented \textit{k} NN-graph, is $L$$\sim$$2$${\cdot}$$M$. Meanwhile, the augmented $k$NN-graph has approximately $L$$\sim$$n_{vi}$${\cdot}$$M$ edges, where $n_{vi}$$=$$nn$$+$$rn>2$ \cite{ivhd3}. As our experience shows, the probability that the largest connected component is rigid (or approximately rigid) is very high. In summary, to obtain the largest connected component approximately equal to the full $k$NN-graph, the number of nearest neighbors $nn$ can be very low (mostly $nn=2$, but for some specific datasets with very similar samples, it can be a bit larger). Assuming additionally that $rn=1$, we can obtain a stable and rigid 2-D embedding of the $k$NN-graph. 

This way, instead of the $O(M^2)$ floating point $\textbf{D}$ matrix, we have as input data $O(nn\cdot M)$ integers - the list of edges of $k$NN graph. The indices of $rn$ random neighbors can be generated {\it ad hoc} during the embedding process. Thus, embedding high-dimensional data reduces to embedding of the corresponding sparse $k$NN graph. To this end, we minimize the following stress function:

\vspace{0.3cm}
\begin{equation}
\label{eq:ivhd_cost_function}
    E(\|\textbf{D}-\textbf{d}\|) = \sum_i \ \sum_{j\in O_{\text{\it nn}}(i)\cup O_{\text{\it rn}}(i)}
    \begin{dcases}
        d_{ij}^{2} & \text{\it if} \hspace{0.3cm} j \in O_{\text{\it nn}}(i) \\
        c\cdot(1-d_{ij})^{2} & \text{\it if} \hspace{0.3cm} j \in O_{\text{\it rn}}(i) \\
    \end{dcases}
    ,
\end{equation}
\vspace{0.3cm}

\noindent which represents the error between the dissimilarities $D_{ij}\in \{0,1\}$ and the corresponding Euclidean distances $d_{ij}$, where: $i,j=\{1,\dots, M\}$, and $c\in(0,1)$ is the scaling factor for random neighbors.

\subsection{Optimization methods}
\label{subsec:optimization_methods}

Consider the continuous cost function $f$, such that $f: \Omega \subset \mathbb{R}^{N} \to \mathbb{R}$. Then, the general solution to the optimization problem is to find $x_{min} \in \Omega$ such that: 

\begin{equation}
    f(x_{min})  \leq f(x) \;\; \forall \;\; x \in \Omega.
\end{equation}

\subsubsection{Stochastic Gradient Descent}

The direction opposite to the direction of the gradient of the function $f$, $\nabla f$, is equal to the direction of the steepest descent. Therefore, minimizing $f$ is useful because it tells us how to change the argument to obtain a small improvement in the value of $f$. We can formulate this idea using an iterative algorithm that starts from a selected point $x_{0}$ and updates its value according to the formula:

\begin{equation}
    x_i = x_{i-1} - \alpha \cdot \nabla f(x_{i-1}),  \:\:\:\:\:\: \alpha \in \mathbb{R},
\end{equation}

where $\alpha$ is \textit{learning rate}. This is a gradient descent (GD) method of optimization. Stochastic gradient descent (SGD) is a probabilistic approximation of the above method. At each step, the algorithm calculates the gradient for one observation picked at random, rather than calculating the gradient for the entire dataset. It is much faster and more suitable for large-scale datasets.

\subsubsection{Adadelta}
\textit{Adadelta} tries to improve \textit{Gradient Descent}, by calculating \textit{learning rate} in each iteration based on the previous values $x$ and $\nabla f(x)$. 
\begin{equation}
    x_i = x_{i-1} - \frac{RMS(\Delta x_{i-1})}{\epsilon + RMS(\nabla f(x_{i-1}))} \nabla f(x_{i-1})
\end{equation}

By $RMS(x)$ we refer to the decaying root mean square of the time series of x \cite{algorithms_for_optimization_book_2019}.

\subsubsection{Momentum}
It is a heuristic algorithm. Its name derives from an analogy to physics. The matrix $v$ can be interpreted as a matrix of velocities, the gradient as gravity, and $\beta$ as a coefficient of friction.

\begin{equation}
    v_i = \beta v_{i-1} + \alpha \nabla f(x_{i-1})
    \label{eq:momentum_update}
    \vspace{-0.3cm}
\end{equation}
\begin{equation}
    x_i = x_{i-1} + v_i
\end{equation}

\subsubsection{Adam}
The name \textit{Adam} derives from the phrase \textit{adaptive moments} \cite{deep_learning_book_2017}. It is a stochastic gradient descent method that is based on adaptive estimation of first-order and second-order moments. The iteration of \textit{Adam} consists of five steps:
\begin{equation}
    v_i = \gamma_v v_{i-1} + (1-\gamma_v) \nabla f(x_{i-1})
    \vspace{-0.3cm}
\end{equation}
\begin{equation}
    s_i = \gamma_s s_{i-1} + (1-\gamma_s) (\nabla f(x_{i-1}) \cdot \nabla f(x_{i-1}))
    \vspace{-0.3cm}
\end{equation}
\begin{equation}
    \hat{v_i} = \frac{v_i}{(1-\gamma_v)^{i-1}}
    \vspace{-0.3cm}
\end{equation}    
\begin{equation}
    \hat{s_i} = \frac{s_i}{(1-\gamma_s)^{i-1}}
    \vspace{-0.3cm}
\end{equation}    
\begin{equation}    
    x_i = x_{i-1} - \alpha \frac{\hat{v_i}}{\epsilon + \sqrt{\hat{s_i}}}
    \vspace{0.3cm}
\end{equation}

It can be seen as a combination of the \textit{RMSProp} and \textit{Momentum} algorithms and is known for its computational efficiency, low memory requirements, and invariant to diagonal gradient rescaling.

\subsubsection{Nesterov}
Algorithm that is very similar to the \textit{Momentum} (in fact it is sometimes called \textit{Nesterov Momentum}). The only difference between the two is that \textit{Nesterov} uses a gradient at the \textit{future} position.
\begin{equation}
    v_i = \beta v_{i-1} + \alpha \nabla f(x_{i-1} + \beta v_{i-1})
    \vspace{-0.3cm}
\end{equation}
\begin{equation}
    x_i = x_{i-1} + v_i
\end{equation}

\begin{figure}[ht]
     \centering
        \subfloat[MNIST dataset.]{\includegraphics[width=2.6in]{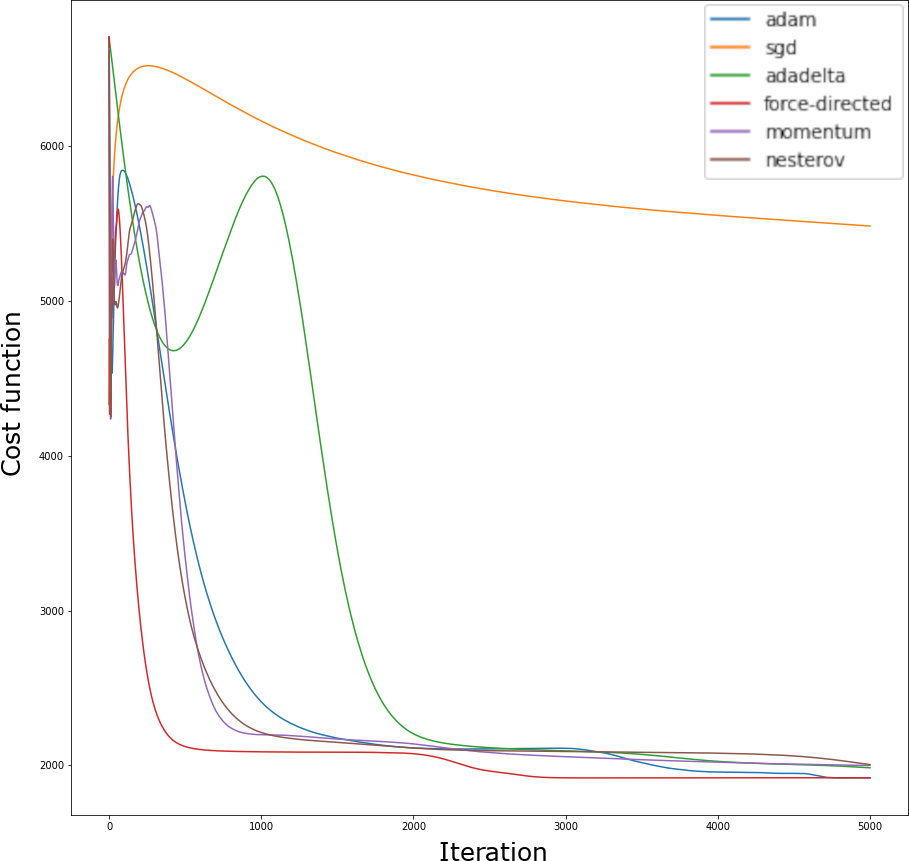}}
        \subfloat[FMNIST dataset.]{\includegraphics[width=2.6in]{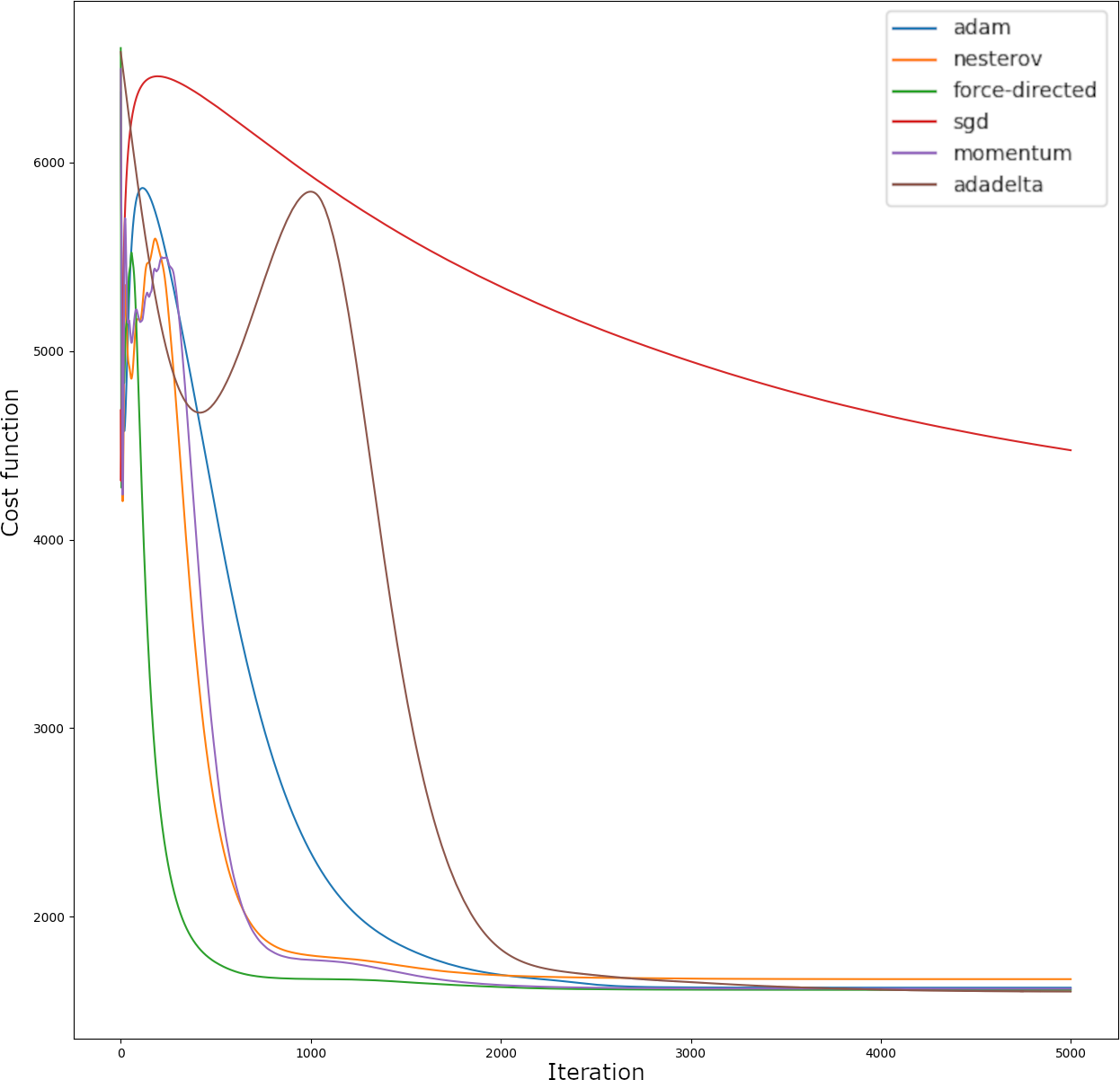}}
        \caption{Speed of convergence to solution (measured as the value of the cost function in successive iterations) for different optimization methods used in IVHD.}
    \label{fig:ivhd_optimizer_comparison}
\end{figure}

\subsection{Force directed}

To find the minimum of $E(.)$ IVHD uses mainly the force-directed approach in the form of the interacting particle method presented in \cite{dzwinel1999} (although it can also utilize state-of-the-art optimization methods). The set of vertices $v_{i}$ in $\mathbb{R}^2$ is treated as an ensemble of interacting particles. Particles evolve in $\mathbb{R}^2$ space with time according to Newtonian equations of motion starting from a random setup. Newtonian equations are discretized using the leap-frog numerical scheme with a timestep length equal to $\Delta t$. The force $f_{i}^{n}$$=$$-\nabla E_{i}$ coming from the particles corresponding to the vertices of the neighborhoods $O_{nn}(i)$ and $O_{rn}(i)$ is calculated for each particle $i$ at each time step $n$. Furthermore, the friction force proportional to the velocity of the particles $-\lambda\textbf{v}_{n}^{i}$ and acting against the direction of movement of the particles is used to dissipate the energy of the particle system. The following equations present the discretized version of the Newtonian equations, which is used to calculate the particle velocities $v_{i}^{n}$ and their positions $r_{i}^{n}$:

\begin{equation}
    v_{i}^{n + \frac{1}{2}} = v_{i}^{n - \frac{1}{2}} + (2 k_{nn}f^{n}_{i} - \lambda v_{i}^{n}) \cdot \Delta t
\end{equation}

\begin{equation}
    r_{i}^{n + 1} = r_{i}^{n} +  v_{i}^{n + \frac{1}{2}} \cdot \Delta t
\end{equation}

\begin{equation}
    v_{i}^{n} = \frac{ v_{i}^{n + \frac{1}{2}} + v_{i}^{n - \frac{1}{2}}}{2}
\end{equation}

\begin{equation}
    f_{i}^{n} = - \sum^{nn}_{j \in \textit{O}_{nn}(i)} r^{n}_{ij} - c \sum^{rn}_{k \in \textit{O}_{rn}(i)} (1 - d_{ik}^{n}) \cdot \frac{r^{n}_{ik}}{d^{n}_{ik}}, \;\;\; r^{n}_{ik} = r^{n}_{i} - r^{n}_{k}
    \label{eq:ivhd_force}
    \vspace{0.4cm}
\end{equation}

These equations can be simplified by the following substitutions:

\begin{equation}
    a \leftarrow \frac{1 - \frac{\lambda \Delta t}{2}}{1 + \frac{\lambda \Delta t}{2}}, \;\;\; b \leftarrow \frac{2k_{nn} \Delta t^2}{1 + \frac{\lambda \Delta t}{2}}, \;\;\; \Delta r_{i}^{n+1} = r_{i}^{n+1} - r_{i}^{n}
    \label{eq:force_directed_a_b_constraints}
    \vspace{0.4cm}
\end{equation}

Finally, we obtain the following scheme:

\begin{equation}
    v_i = \Delta r_i \leftarrow a \cdot \Delta r_i + b \cdot f_{i} = a \cdot v_i + b \cdot f_i
    \label{eq:force_directed_update}
\end{equation}

\begin{equation}
    r_i \leftarrow r_i + \Delta r_i
\end{equation}

The value of $a \in [0,1]$, assuming slow dissipation, should be close to 1. At the beginning of embedding, it can be slightly greater for faster dissipation of large initial fluctuations. Consequently, the value of $c$ should be very small, not to overwhelm the influence of the forces that come from the closest neighborhood $O_{nn}(i)$. By comparing Eqs. \ref{eq:force_directed_update} and \ref{eq:momentum_update}, one can conclude that the methods are identical, but what is important is that force-directed imposes additional constraints for $a$ and $b$ parameters derived from \ref{eq:force_directed_a_b_constraints}:
\begin{equation}
    \frac{a}{b} = \frac{1 - \frac{\lambda \Delta t}{2}}{2k_{nn} \Delta t}
\end{equation}

To ensure simulation speed and system convergence to the stable solution, we used the following self-adaption scheme to determine the value of parameter $b$: 
\vspace{1cm}
\begin{equation}
    \begin{aligned}
        & \hspace{-2cm} \text{if} \hspace{0.3cm} \text{abs}(\Delta T = \sum_{i=1}^{N} [|\Delta r_{i}^{n+1}|^2 - |\Delta r_{i}^{n}|^2]) > \tau \hspace{0.3cm} \\\\
        \text{then} & \hspace{0.3cm} b \leftarrow (\gamma =     
        \begin{dcases}
            \gamma_1 > 1 & \text{\it if} \hspace{0.3cm} \Delta T < -\tau \\
            \gamma_2 < 1 & \text{\it if} \hspace{0.3cm} \Delta T > \tau
        \end{dcases} 
        ) \cdot b \wedge r_{i}^{n+1} = r_{i}^{n}
    \end{aligned}
\end{equation}
\vspace{1cm}

The quality of visualization depends on the relationship between $rn(i)$ and $nn(i)$ and the value of the parameter $c$ (Eq. \ref{eq:ivhd_force}) i.e. the factors influencing the balance between the stretching and contracting forces. In the implementation of the method, for simplicity, we define constant input values of $nn$ and $rn$ as randomly chosen numbers of connected and disconnected vertices for each $i$, respectively. We assume that a neighbor connected to $i$ cannot be its random neighbor. If the number of connected neighbors is less than $nn$, all of them are selected. The correct choice of $nn$, $rn$ and $c$ is crucial for the quality of 2D reconstruction and strongly depends on the type and size of visualized dataset. Experiments show that $nn$ should be greater than or, at most, equal to $rn$ \cite{minch2020}. Otherwise, long-range interactions start to dominate, producing a sphere of particles in 2D.

\subsection{Binary and Euclidean distances comparison}

As shown in Equation \ref{eq:ivhd_binary_distances} - IVHD omits ordering $nn$ for each $y_{i} \in \mathbf{Y}$ to obtain a good approximation of the data manifold. Thus, for simplicity, the distances between the connected vertices of the $nn$-graph should be the same and as close to 0 as possible. Furthermore, the $nn$-graph should be enhanced with additional edges (\textit{random neighbors}) to ensure rigidity. In this section, we compare the results obtained by embedding with IVHD using Euclidean and binary distances for our calculations.

\begin{figure}[ht]
     \centering
        \subfloat[Euclidean distances.]{\includegraphics[width=0.25\textwidth]{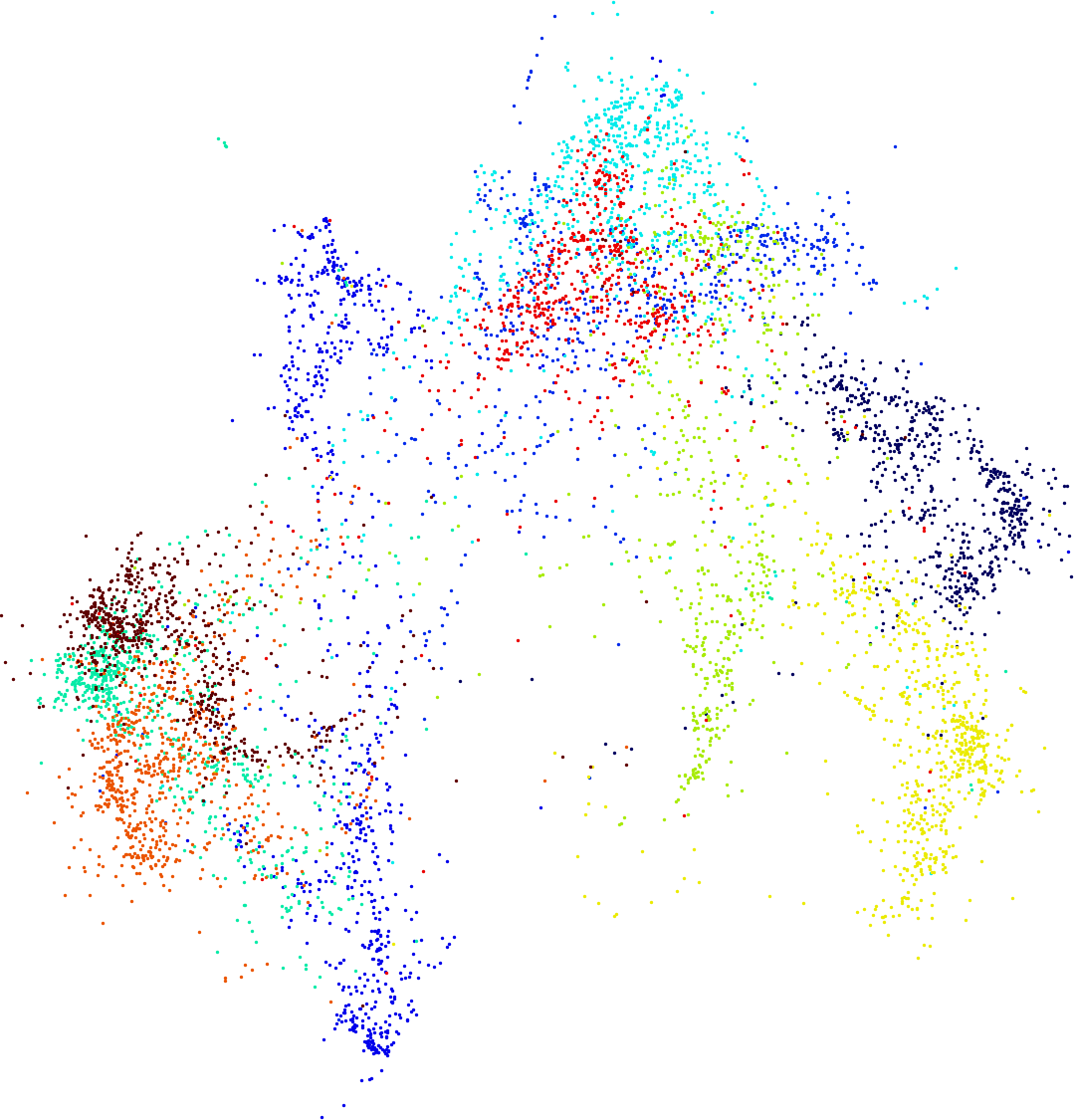}}
        \subfloat[Euclidean distances.]{\includegraphics[width=0.25\textwidth]{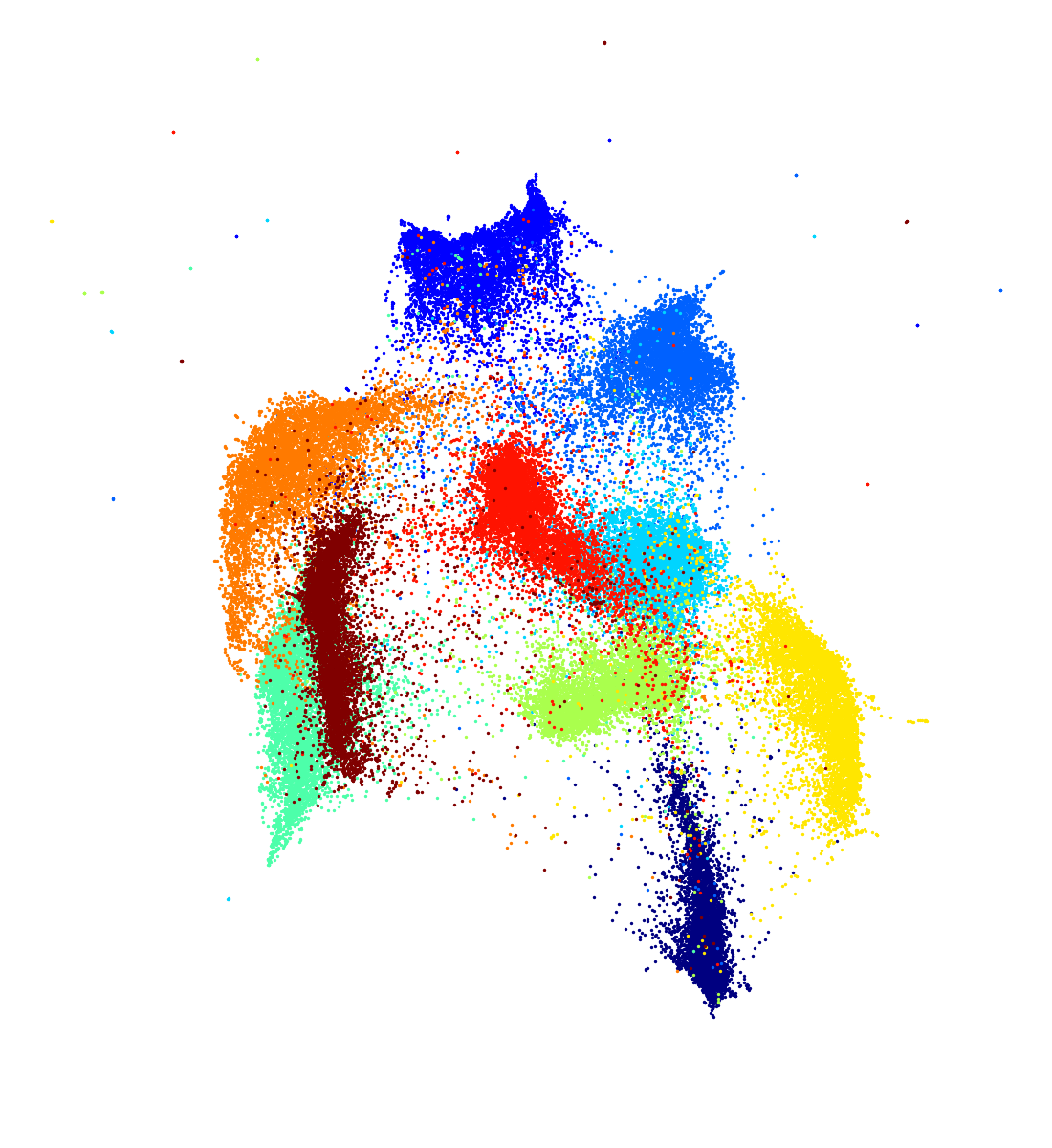}}
        \subfloat[Binary distances.]{\includegraphics[width=0.25\textwidth]{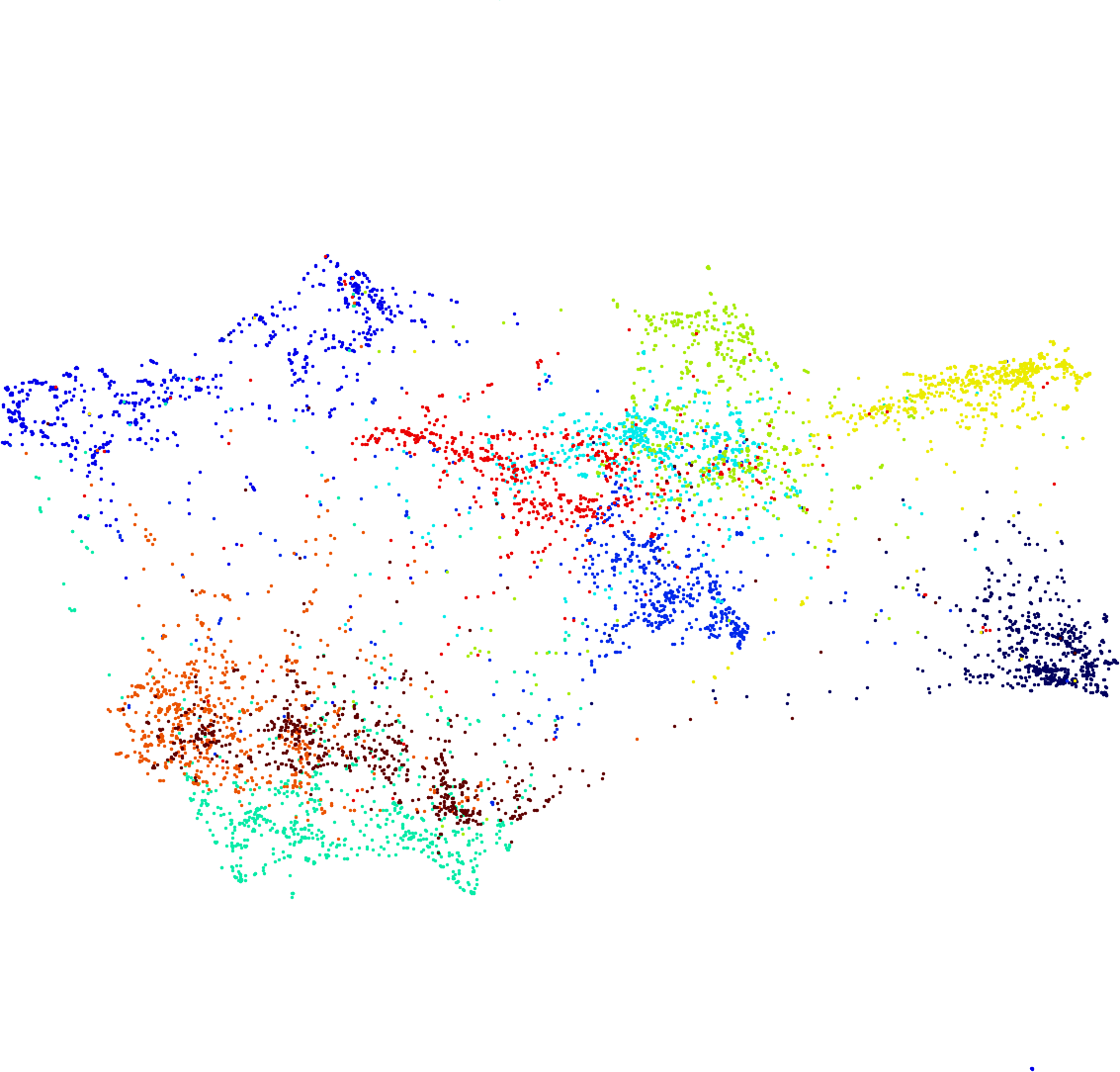}}
        \subfloat[Binary distances.]{\includegraphics[width=0.25\textwidth]{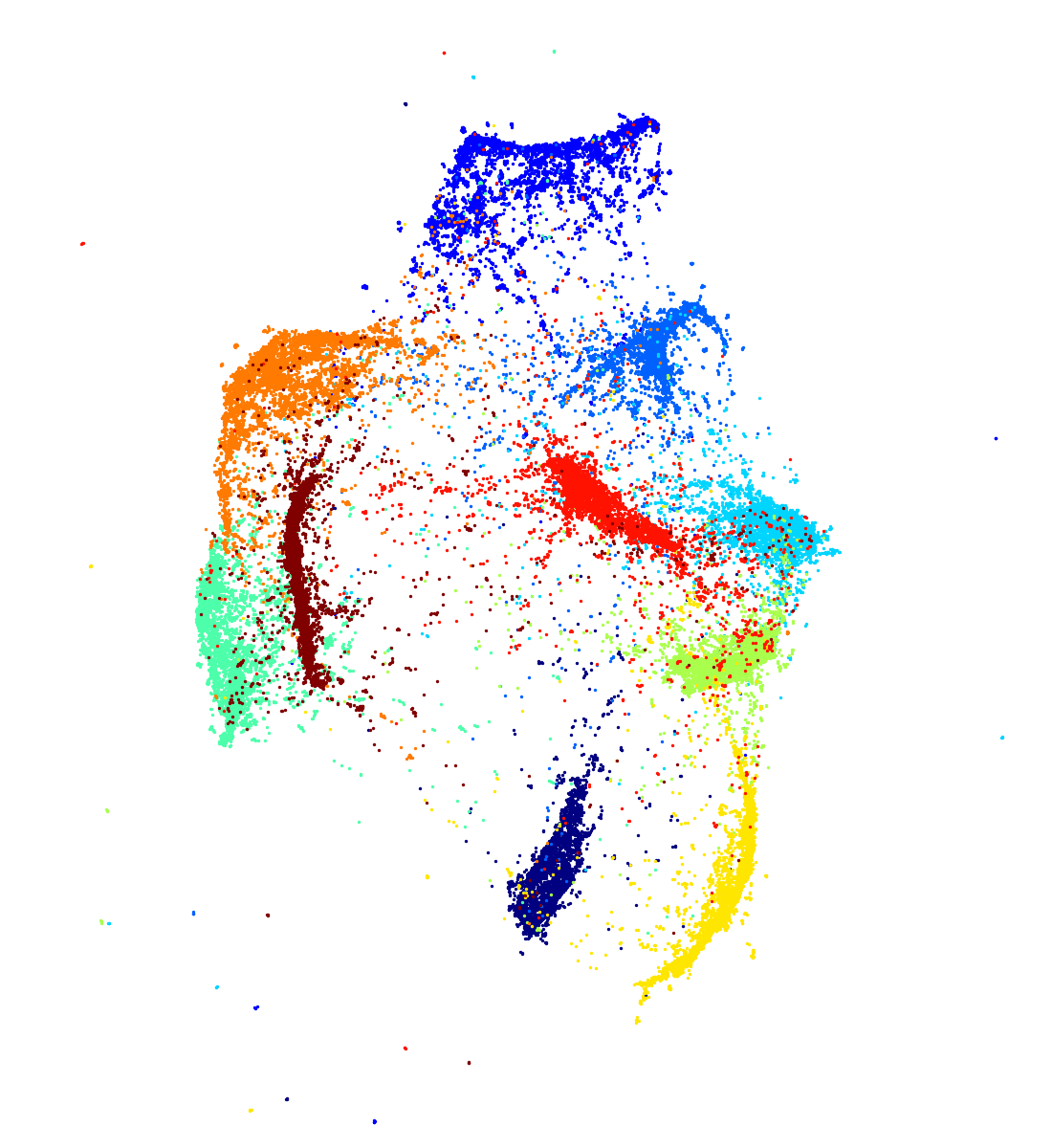}}
        \caption{Visualizations for MNIST dataset and its sub-set ($M$$=$$7$$\cdot$$10^{3}$), for both euclidean and binary distances. }
    \label{fig:ivhd_mnist_euclidean_vs_binary_distance}
\end{figure}
        
We compare the Euclidean and binary distances for the subset of the MNIST dataset (7k samples) and for MNIST after PCA to 100D. The procedure to obtain the appropriate value of the dimensions after reduction is detailed in Section \textit{5.2}. The metrics used are described in detail in Section \textit{5.1}.

\vspace{-2cm}
\begin{figure}[ht]
     \centering
        \subfloat[DR quality.]{\includegraphics[width=0.5\textwidth]{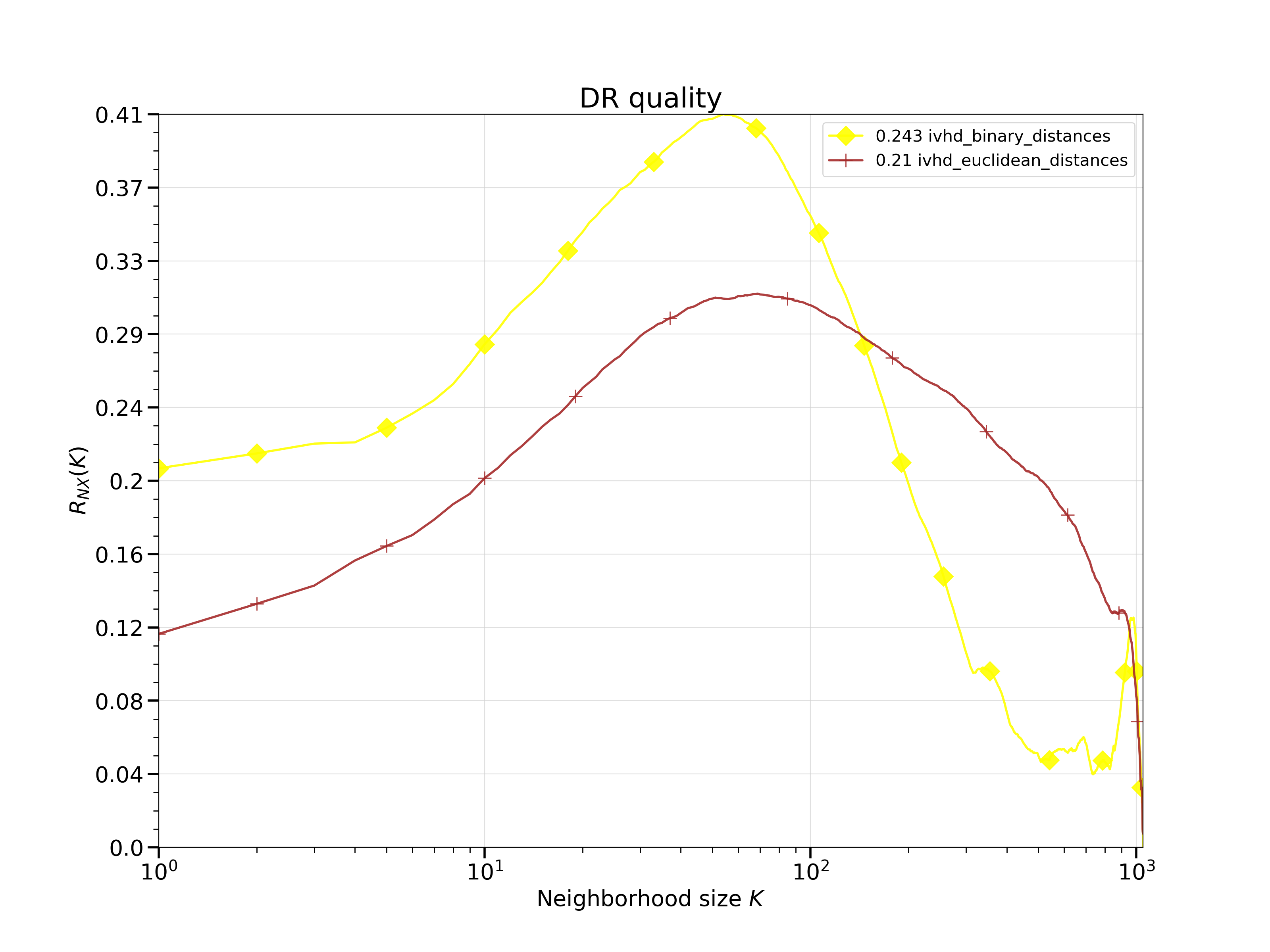}}
        \subfloat[KNN gain.]{\includegraphics[width=0.5\textwidth]{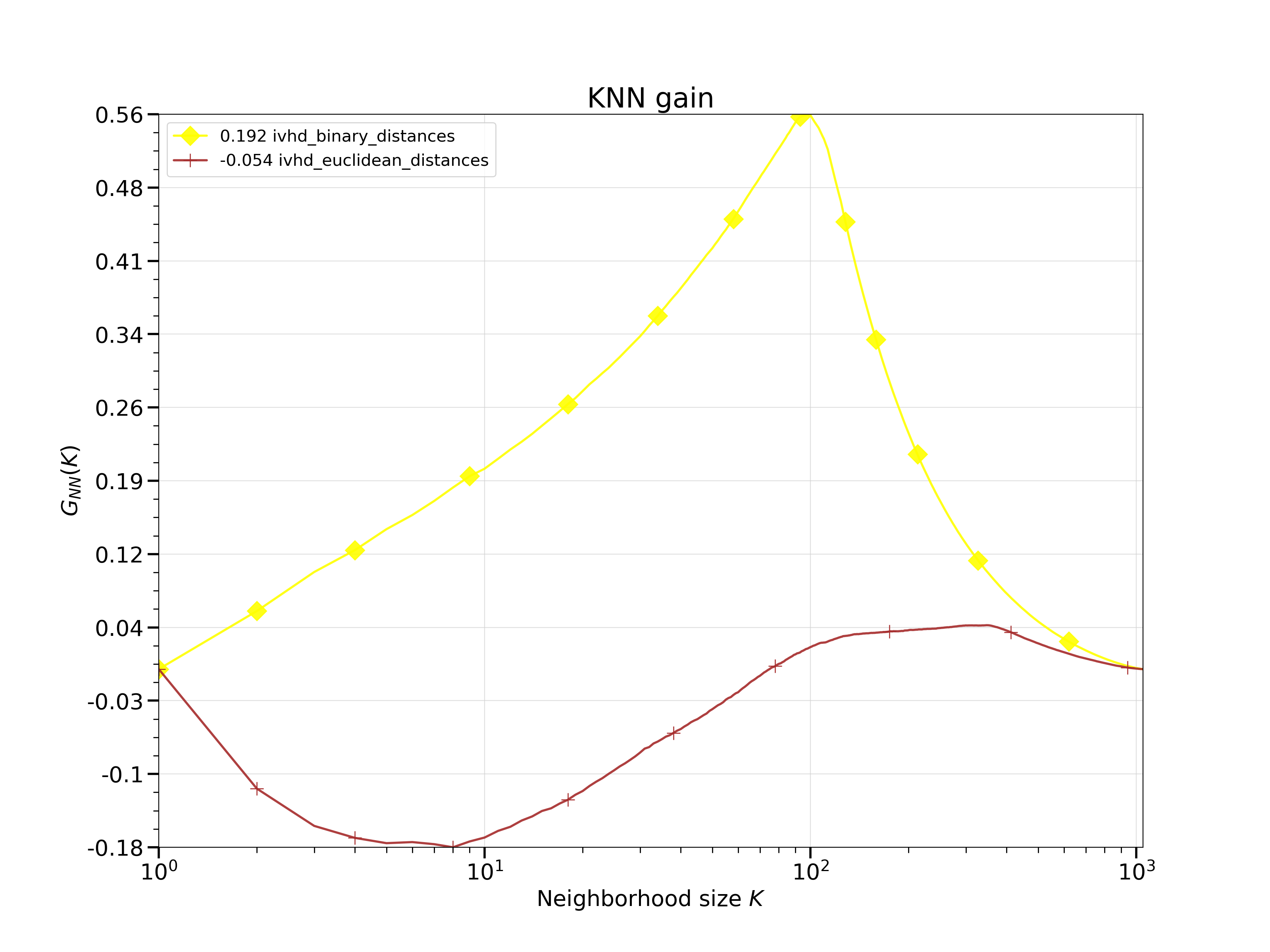}}
        \hfill
        \subfloat[DR quality.]{\includegraphics[width=0.5\textwidth]{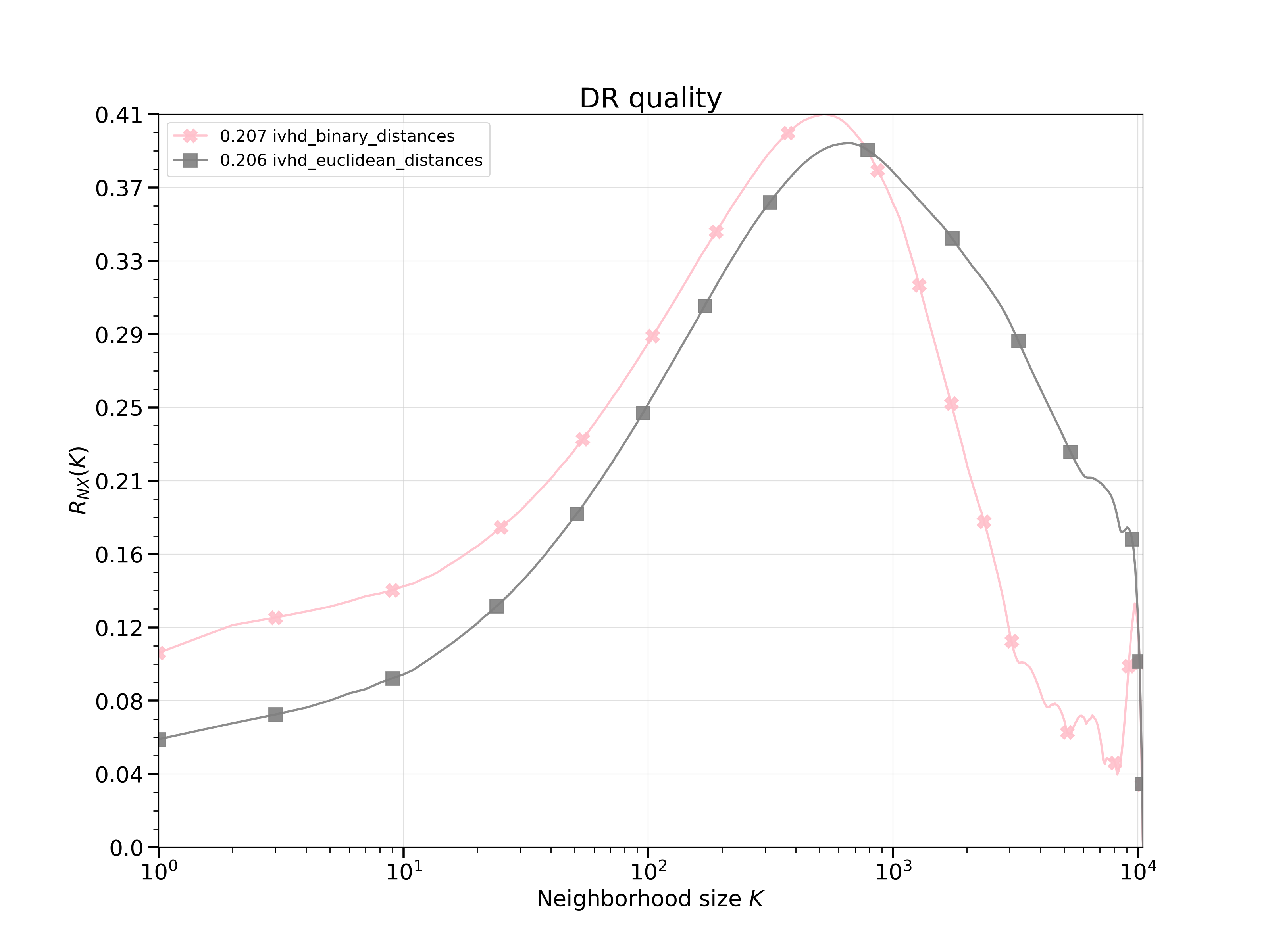}}
        \subfloat[KNN gain.]{\includegraphics[width=0.5\textwidth]{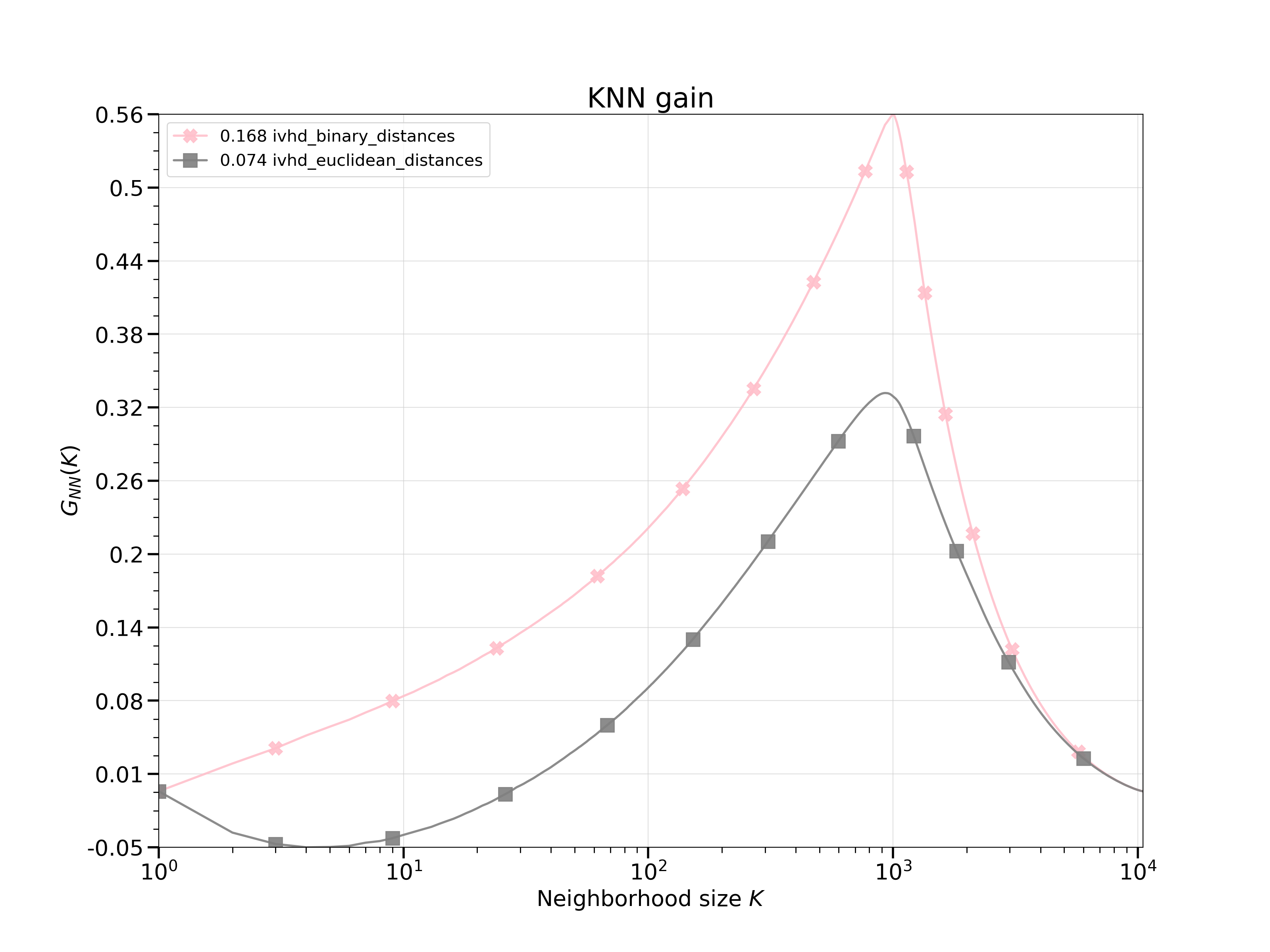}}
        \caption{Metrics obtained for MNIST dataset and its sub-set ($M$$=$$7$$\cdot$$10^{3}$), for both euclidean and binary distances.}
    \label{fig:ivhd_mnist_euclidean_vs_binary_distance_metrics}
\end{figure}

In the visualizations in Fig. \ref{fig:ivhd_mnist_euclidean_vs_binary_distance}, we see that for binary distances, IVHD generates visualizations in which the clusters are more compact. This is intuitive, because for the case of normal distances, the interactions between clusters would be more balanced, resulting in more noise between classes. In addition, this fact is confirmed by the metrics. Both DR quality and KNN gain reach higher values in most of the spectrum of the neighborhood under consideration. For these reasons, we only consider binary distances in further experiments, since IVHD then works more efficiently and better quality visualizations are achieved.

\section{Improvements in IVHD algorithm}
\label{sec:ivhd_improvements}

As shown in the next chapter, IVHD outperforms, in terms of computational time, the state-of-the-art DR algorithms by more than one order of magnitude in the standard DR benchmark data. 
The data embeddings obtained by IVHD are also very effective in reconstructing data separation in large and high-dimensional datasets. 
This is the principal requirement for knowledge extraction, because only multi-scale clusters of data represent basic data \textit{granules of knowledge}. 
Due to the simplicity of the IVHD algorithm, which is, in fact, a clone of the classical MDS algorithm, its efficiency could be further increased by implementing its parallel version in a GPU environment \cite{minch2020} (described in Section \ref{chapter:ivhd-cuda}). It should be mentioned that IVHD produces less impressive results than SNE clones in reconstructing the local neighborhood.  The \textit{crowding problem}, similar to that in SNE, causes most of the data points in 2-D embeddings to gather in the center of the clusters. 

However, an accurate reconstruction of nearest neighbors is a secondary requirement in interactive visualization of big data. The exact $nn$ lists (or their approximates) are well known prior to data embedding. They can be visually presented at any time, for example, as the data points are arranged in $\textbf{Y}$. Furthermore, the exact lists of the nearest neighbors are not reliable in the context of both measurement errors and the \textit{curse of dimensionality's} principle. If a more accurate reconstruction of the location of the data is required, IVHD can be used to generate the initial configuration for DR algorithms based on a much slower SNE. Unlike SNE competitors, IVHD has only three free parameters that must be adapted to the data: $nn$, $rn$, and $c$. Because the method is fast, they can be easily matched interactively, although the \textit{universal} set, that is, $nn=3$, $rn=1$, and $c=0.1$ (or $c=0.01$), works very well for most data (and networks).

Knowing all of these does not mean that further improvements cannot be made to the IVHD method. In this chapter, we will provide a description of the mechanisms that improve the quality of IVHD embeddings. This will be done mainly by modifying the metrics used and the interactions between neighbors at a certain step of the simulation.

\subsection{L1 norm in late stages of embedding}

Using the $\mathcal{l}_{1}$ norm in the late stages of embedding provides a suction effect, which pulls outliers closer to the centers of the clusters. The number of steps performed with the $\mathcal{l}_{1}$ metric cannot be too large, because the embedding would collapse in the centers of the clusters, completely distorting the global embedding structure. 

\begin{figure}[ht]
     \centering
        \subfloat[The baseline IVHD.]{\includegraphics[width=0.4\textwidth]{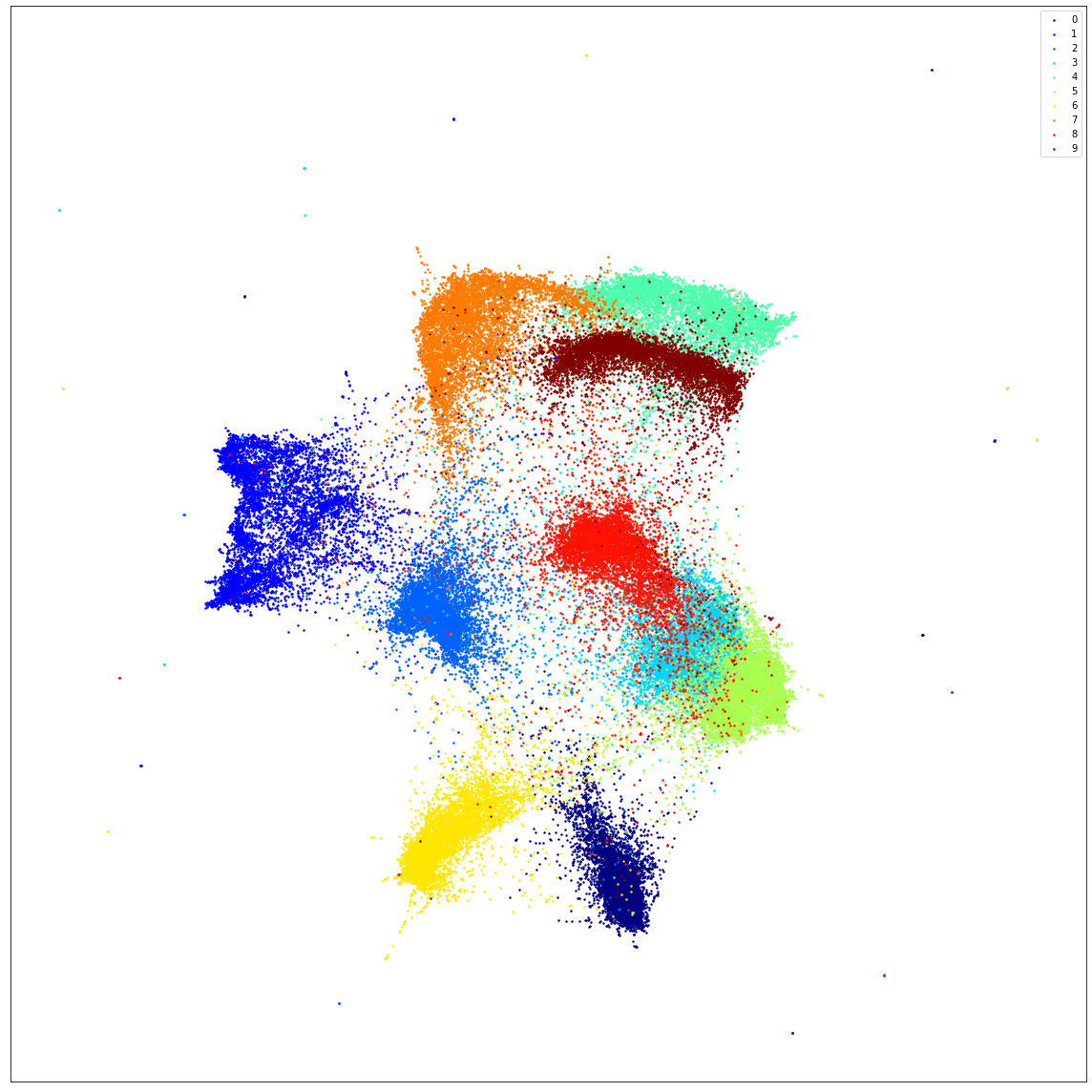}}
        \hspace{0.5cm}
        \subfloat[IVHD with $\mathcal{l}_{1}$ norm.]{\includegraphics[width=0.4\textwidth]{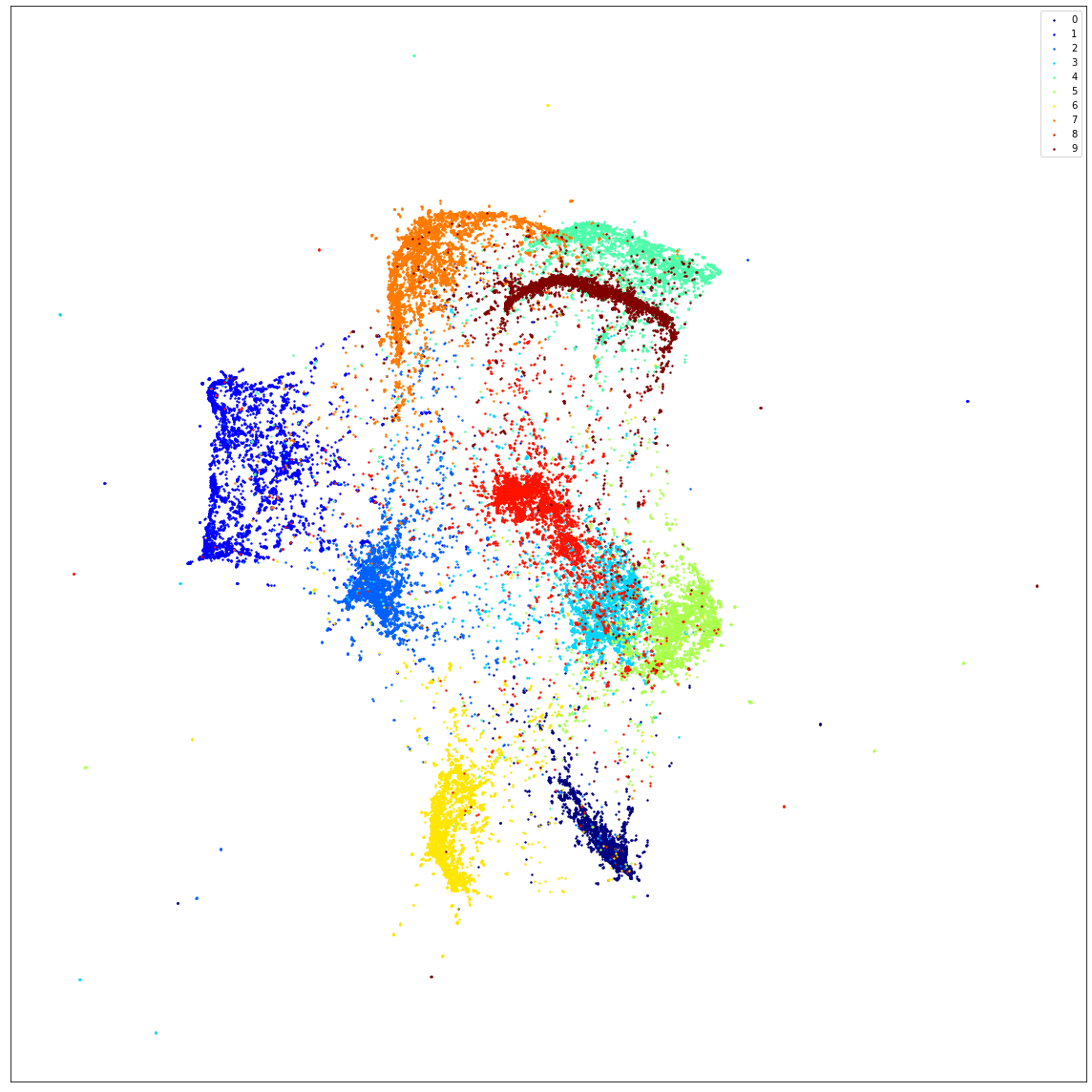}} 
        \caption{IVHD comparison on MNIST dataset, when $\mathcal{l}_{1}$ norm is used for last 50 steps of embedding procedure.}
    \label{fig:ivhd_l1_embedding_mnist}
\end{figure}

\begin{figure}[ht]
     \centering
        \subfloat[The baseline IVHD.]{\includegraphics[width=0.4\textwidth]{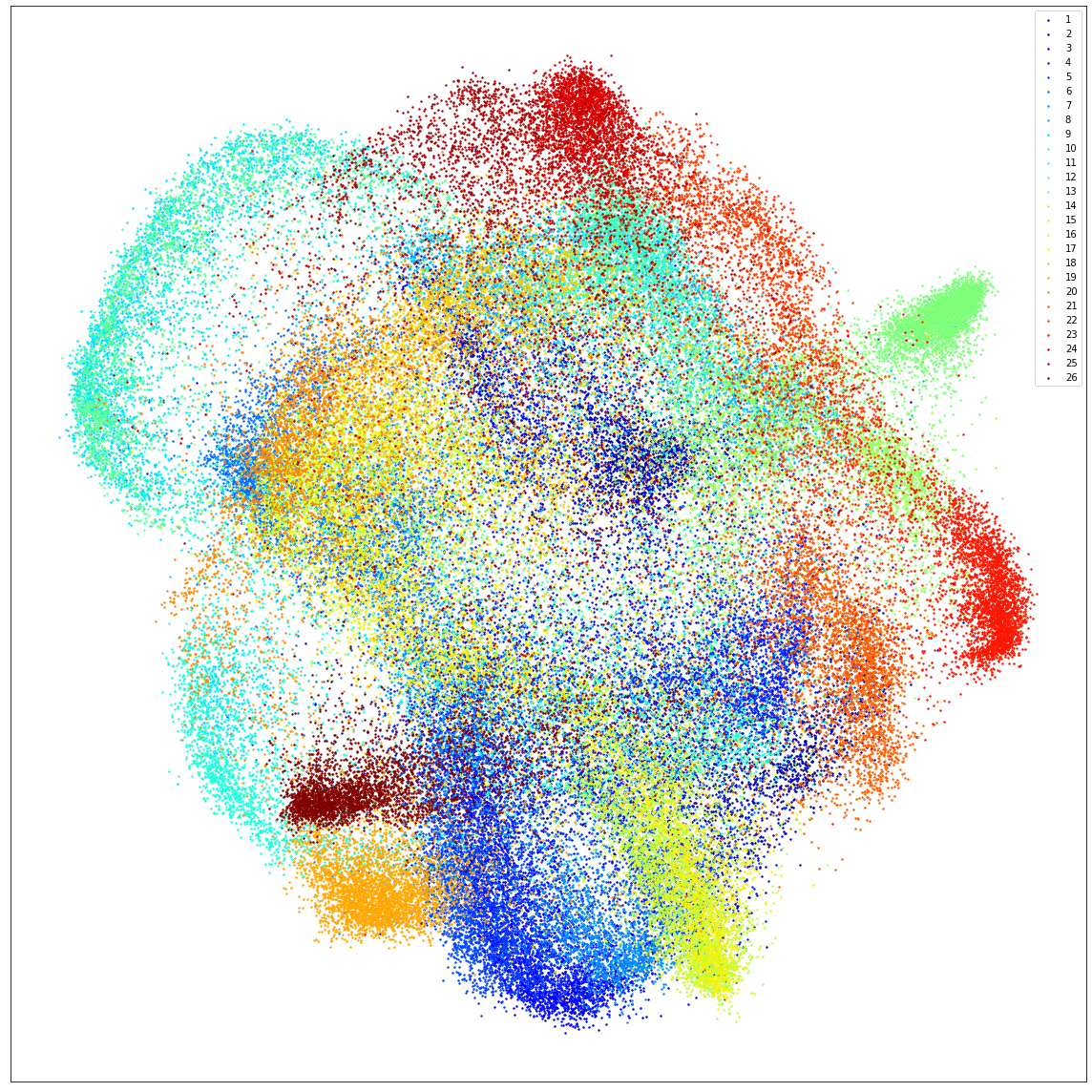}}
        \hspace{0.5cm}
        \subfloat[IVHD with $\mathcal{l}_{1}$ norm.]{\includegraphics[width=0.4\textwidth]{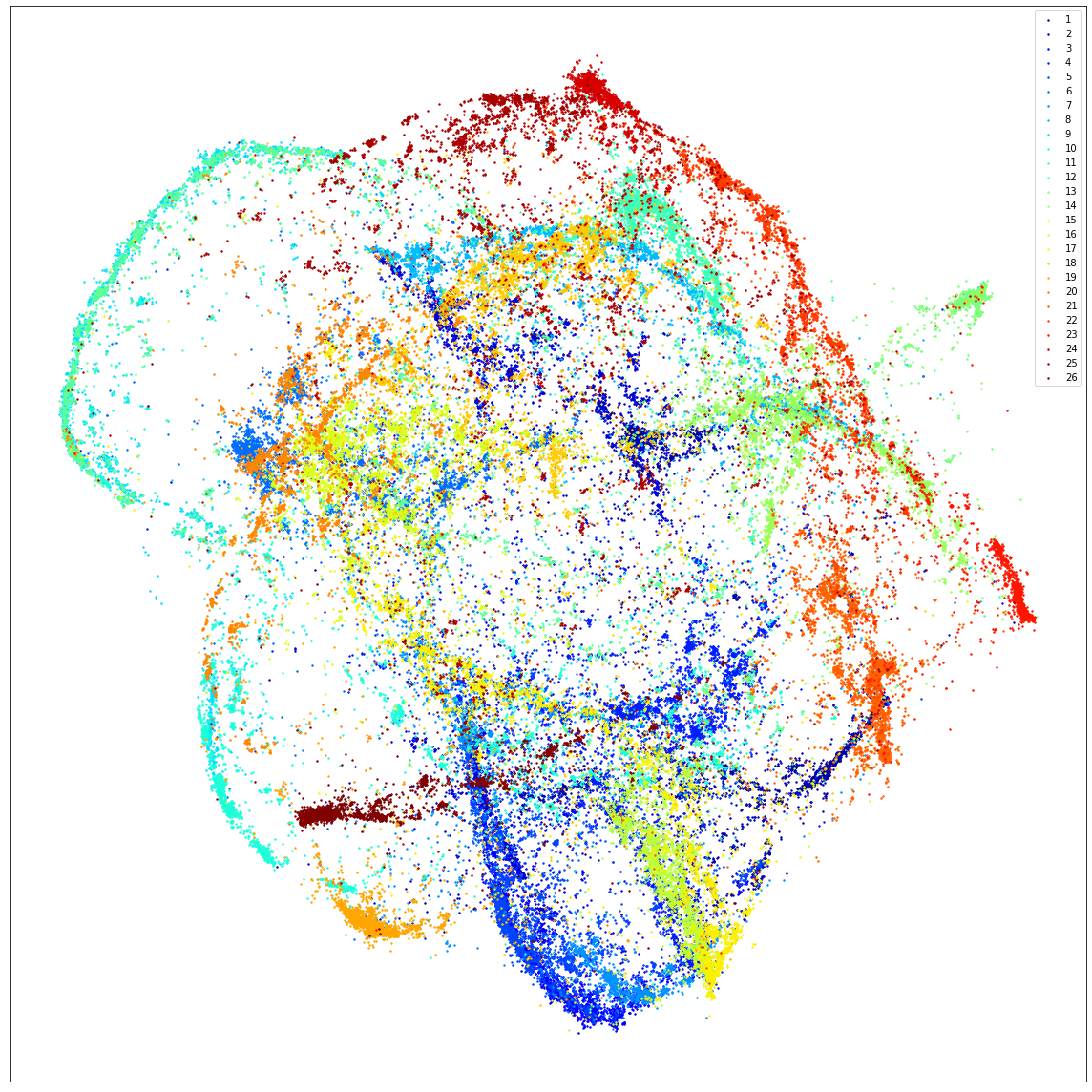}}
        \caption{IVHD comparison on EMNIST dataset, when $\mathcal{l}_{1}$ norm is used for last 50 steps of embedding procedure.}
    \label{fig:ivhd_l1_embedding_emnist}
\end{figure}

Both in Figures \ref{fig:ivhd_l1_embedding_mnist} and \ref{fig:ivhd_l1_embedding_emnist} \textit{suction effect} are clearly visible. Most of the noise is moved to the center of the clusters from the space between. Additionally, the global structure is still maintained and is not distorted (the mutual positioning of classes in relation to each other does not change). This is reflected in the metrics shown in Fig. \ref{fig:chapter_7_dr_quality_knn_gain_mnist_emnist}.

\subsection{Reverse neighbors mechanism}

\begin{wrapfigure}{r}{6.5cm}
    \centering
    \includegraphics[width=6.5cm]{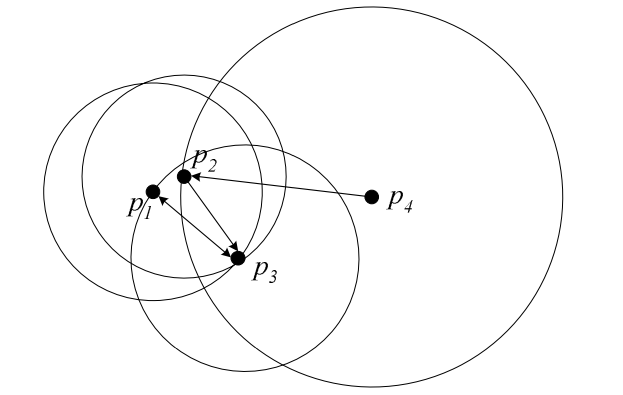}
    \caption{2NN and R2NN examples. Source: \cite{tao_2004}.}
    \label{fig:chapter_7_reverse_neighbors}
\end{wrapfigure}

Let $\mathbf{X}$ be a high-dimensional dataset, and let $q$ be a point in this high-dimensional space. A reverse nearest neighbor (\textit{RNN}) query retrieves all points $p$$\in$$X$ that have $q$ as their nearest neighbor. The set \textit{RNN}(q) of the reverse nearest neighbors of $q$ is called the influence set of $q$. Specifically, $RkNN(q) = \{p \in P \; | \; dist(p,q) \leq dist(p,p_{k})$, where $p_k$ is the \textit{k}-th nearest neighbor to $p \}$. Figure \ref{fig:chapter_7_reverse_neighbors} shows four 2D points, where each point p is associated with a circle that covers its two nearest neighbors. For example, the two NNs of $p_{4}$ $(p_{2}, p_{3})$ are in the circle centered at $p_{4}$. Consequently, $p_{4}$$\in$$R2NN(p_{2})$ and $p_{4}$$\in$$R2NN (p_{3})$. Let $kNN(p)$ be the set of $k$ nearest neighbors to the point $p$. It is important to note that $p \in kNN(q)$ does not necessarily imply $p$$\in$$RkNN(q)$ and vice versa. For example, $2NN(p_{4})=\{p_{2},p_{3}\}$, while $R2NN(p_{4})=\varnothing$ (that is, $p_{4}$ is not contained in the circles of $p_1$, $p_2$ or $p_3$). 

\begin{figure}[ht]
     \centering
        \subfloat[The baseline IVHD.]{\includegraphics[width=0.32\textwidth]{pics/ivhd_improvements/mnist/base_visualization.png}}
        \vspace{0.2cm}
        \subfloat[IVHD with reverse neighbors mechanism.]{\includegraphics[width=0.32\textwidth]{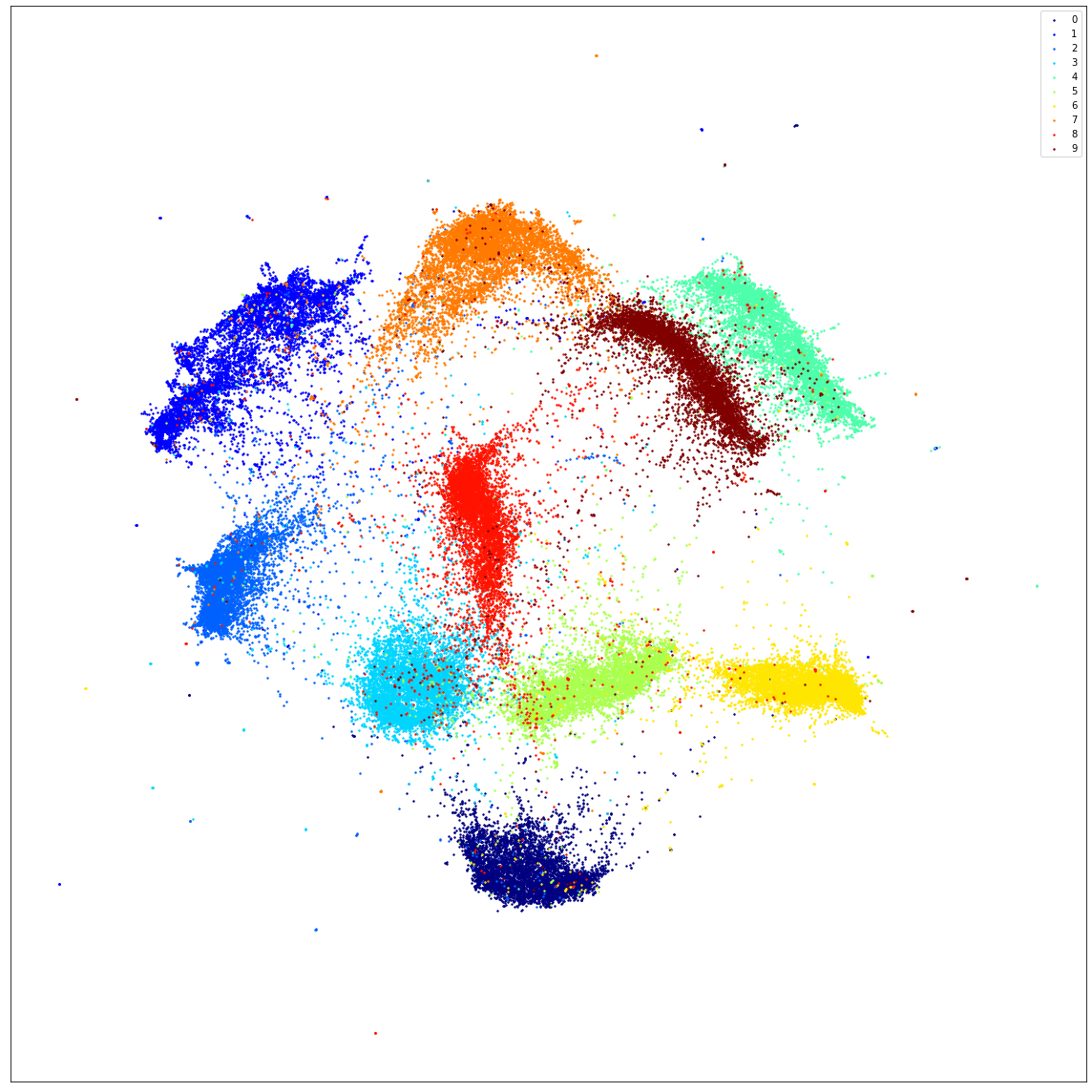}}
        \vspace{0.2cm}
        \subfloat[IVHD with reverse neighbors and $\mathcal{l}_{1}$ norm mechanisms.]{\includegraphics[width=0.32\textwidth]{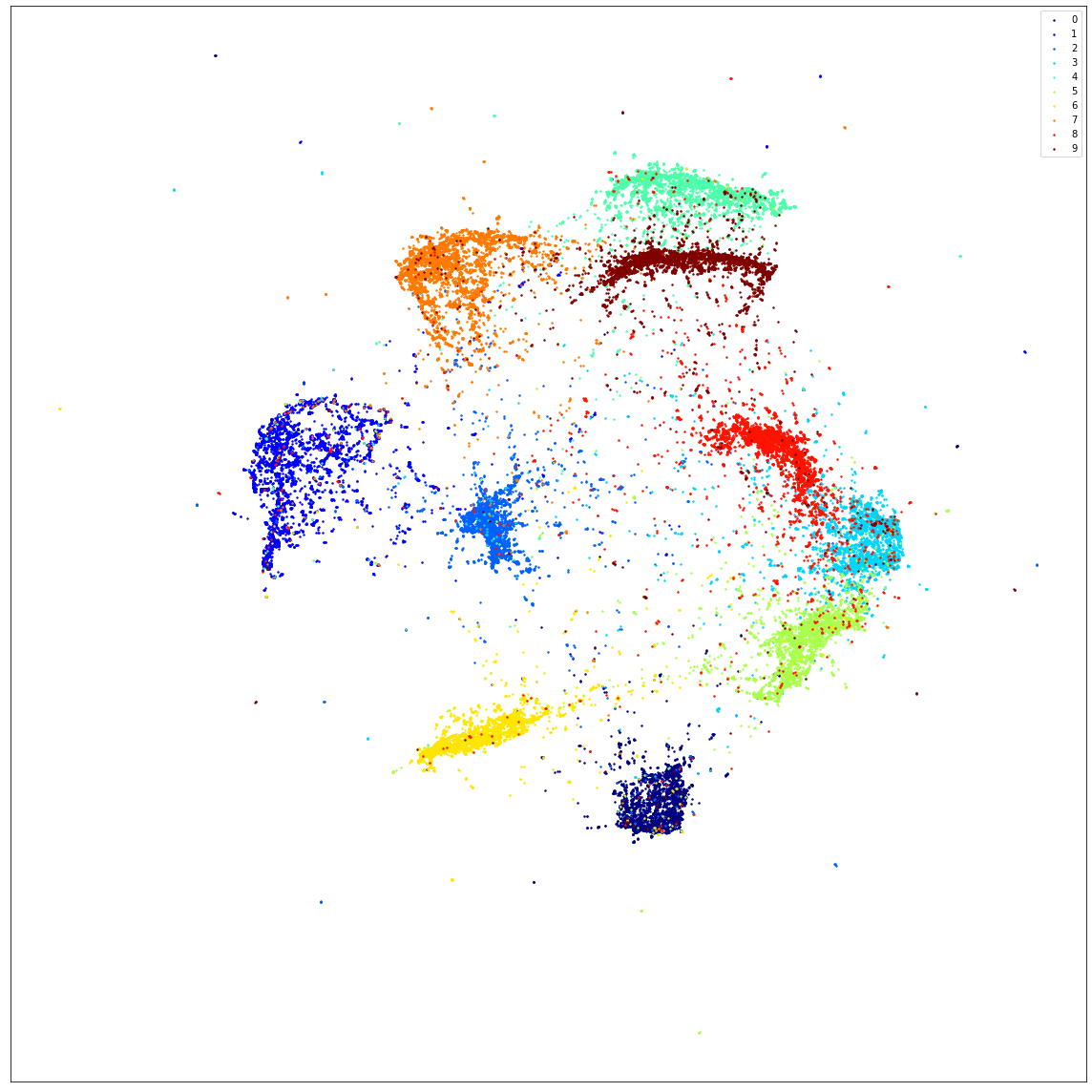}}
        \caption{IVHD comparison on MNIST dataset, when reverse neighbors mechanism is used for last 200 steps of embedding procedure and when it is combined with $\mathcal{l}_{1}$ norm.}
    \label{fig:ivhd_reverse_neighbors_embedding_mnist}
\end{figure}

 From a technical point of view, the RNN mechanism employs a helper nearest neighbors graph, which contains a higher number of neighbors (default: $4 \cdot nn$, where $nn$ is the number of \textit{nearest neighbors} in the \textit{calculation} graph). Using the procedure described in the first paragraph of this section and a helper graph, we determine the reverse nearest neighbors, and we retain only interactions in which the particles are their mutual neighbors or one is the reverse neighbor of the other. In this way, we decrease the number of neighbors (and, as a result, interactions) in the embedding \textit{calculation}, causing the particles to interact with even more specific subsets of particles in the final stage of embedding. As shown in both Figs. \ref{fig:ivhd_reverse_neighbors_embedding_mnist} and \ref{fig:ivhd_reverse_neighbors_embedding_emnist}, the reverse neighbors mechanism used for the last 200 embedding steps reduces the amount of noise as well (but in a different manner, then the norm $\mathcal{l}_{1}$). In the MNIST dataset (Fig. \ref{fig:ivhd_reverse_neighbors_embedding_mnist}) the effect is slightly more noticeable.

\begin{figure}[ht]
     \centering
        \subfloat[DR quality for MNIST.]{\includegraphics[width=0.4\textwidth]{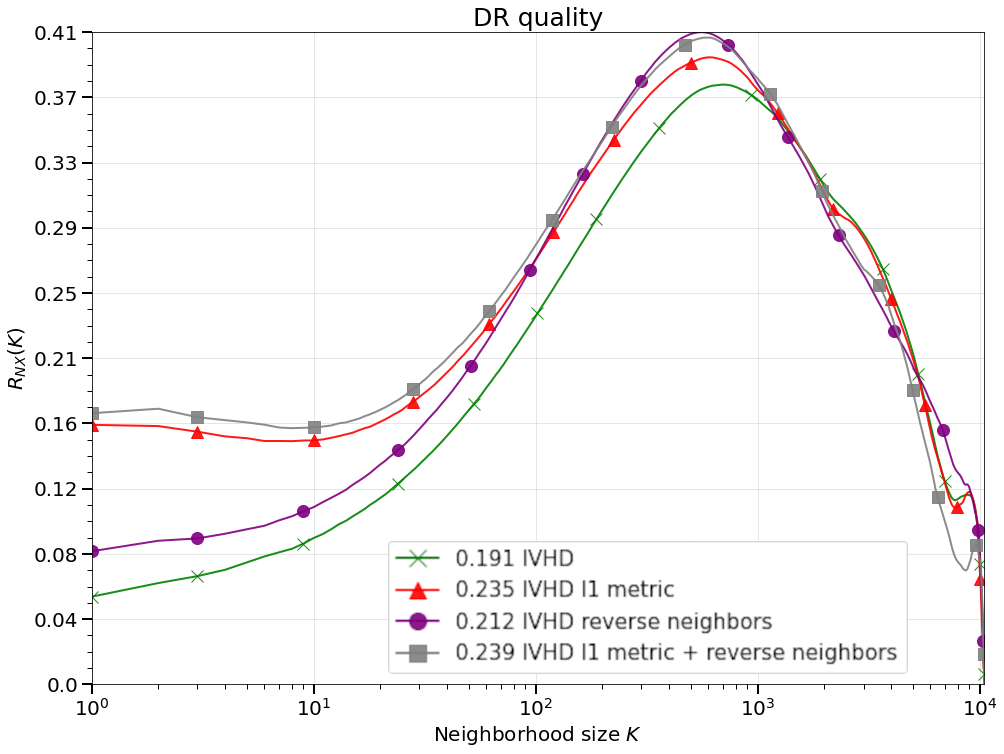}}
        \hspace{0.5cm}
        \subfloat[$k$NN gain for MNIST.]{\includegraphics[width=0.4\textwidth]{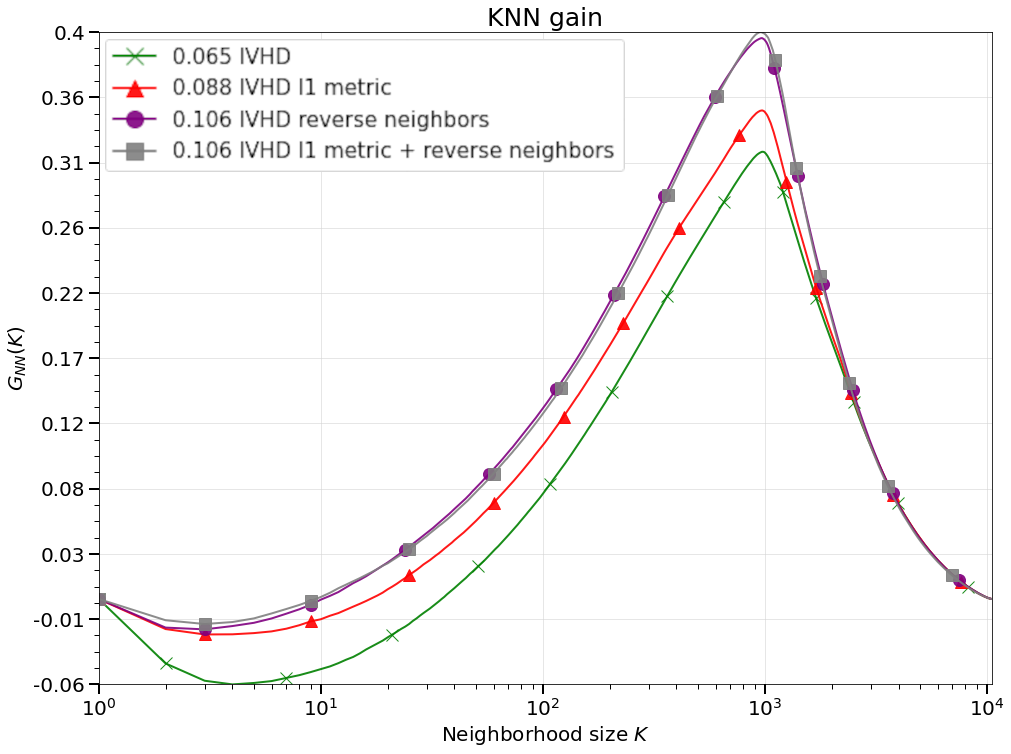}}
        \hfill
        \subfloat[DR quality for EMNIST.]{\includegraphics[width=0.4\textwidth]{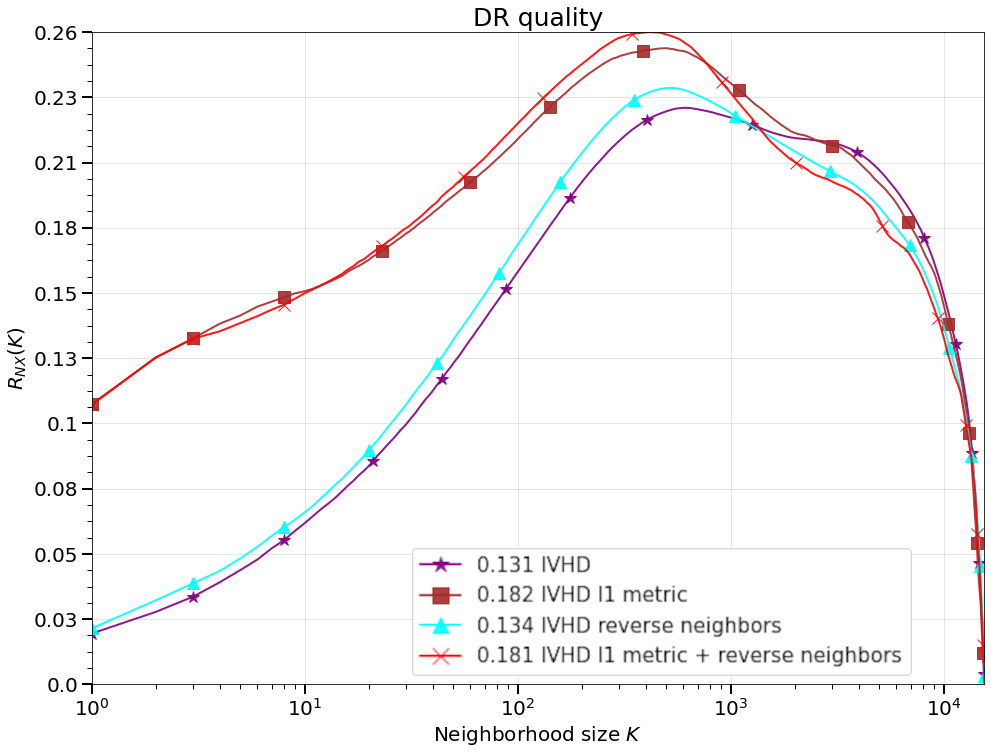}}
        \hspace{0.5cm}
        \subfloat[$k$NN gain for EMNIST.]{\includegraphics[width=0.4\textwidth]{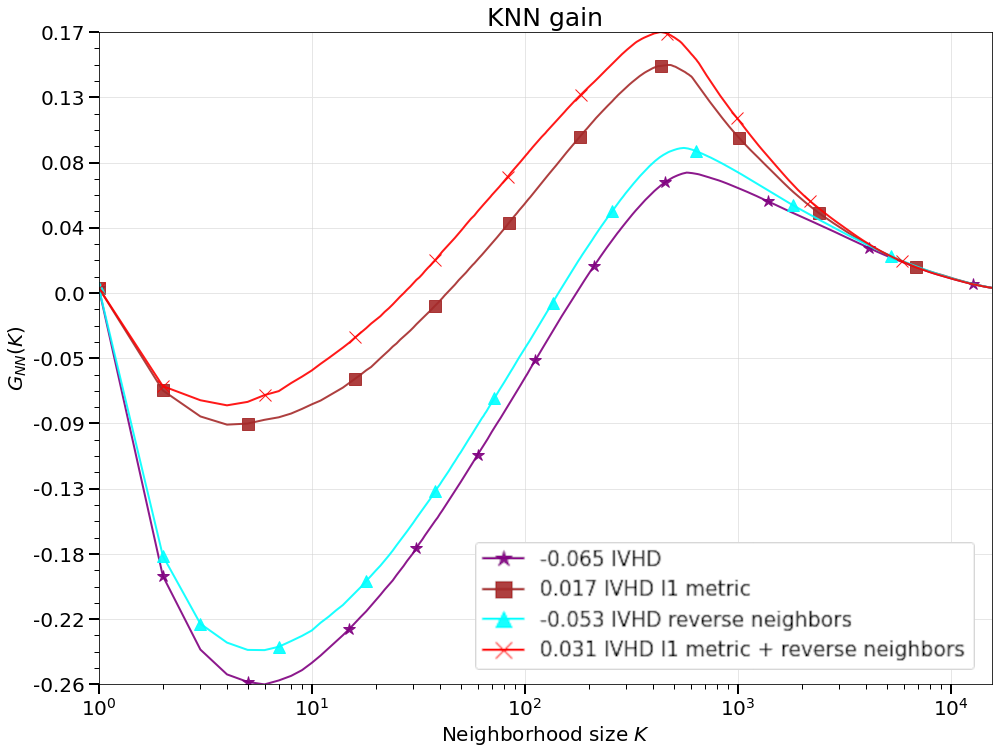}}
        \caption{Metrics obtained for both MNIST and EMNIST dataset. It contains: 1) the baseline IVHD visualization, 2) employed $\mathcal{l}_{1}$ norm mechanism, 3) employed reverse neighbors mechanism and 4) combined $\mathcal{l}_{1}$ norm and reverse neighbors mechanisms.}
    \label{fig:chapter_7_dr_quality_knn_gain_mnist_emnist}
\end{figure}

\begin{figure}[ht]
     \centering
        \subfloat[The baseline IVHD.]{\includegraphics[width=0.32\textwidth]{pics/ivhd_improvements/emnist/base_visualization.png}}
        \vspace{0.2cm}
        \subfloat[IVHD with reverse neighbors mechanism.]{\includegraphics[width=0.32\textwidth]{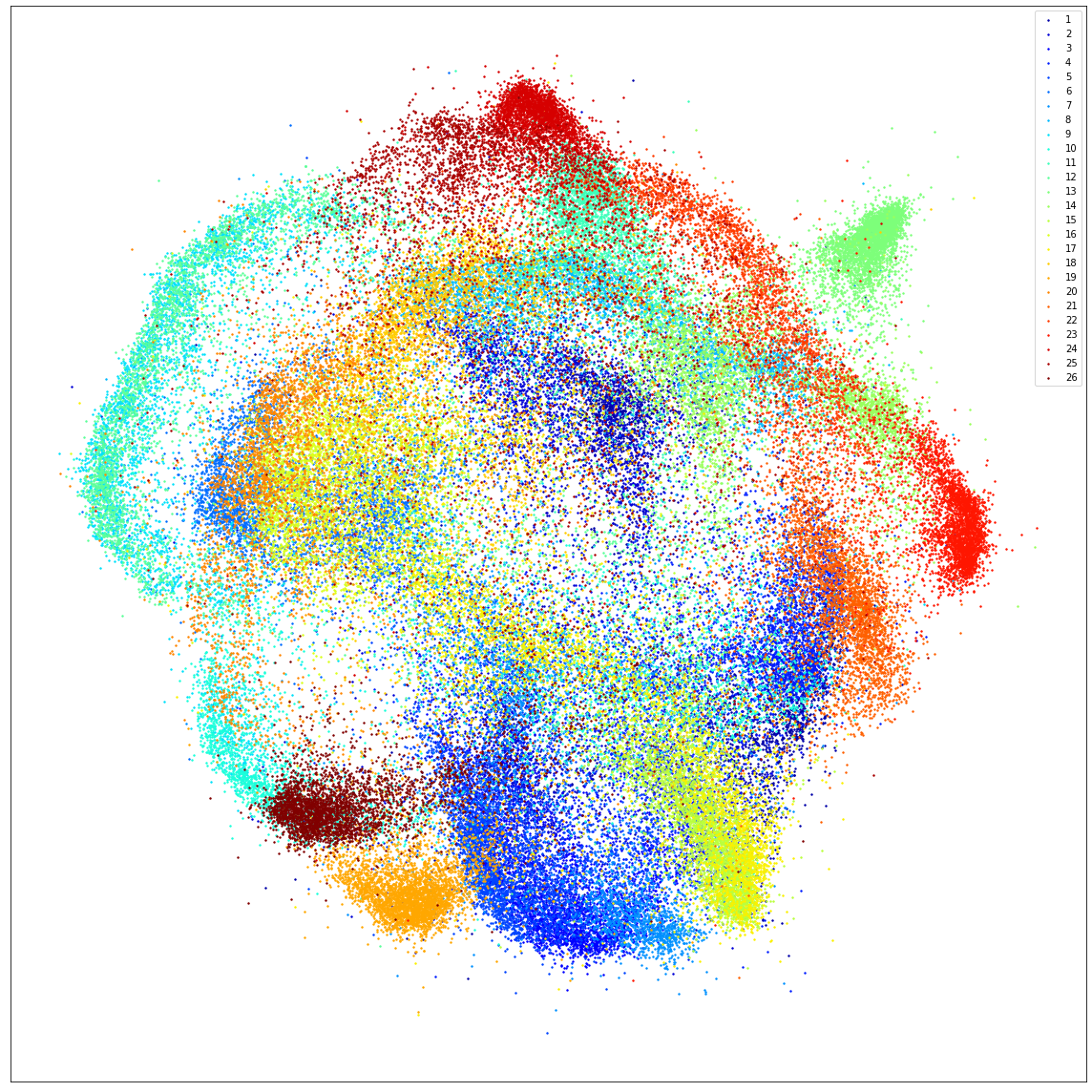}}
        \vspace{0.2cm}
        \subfloat[IVHD with reverse neighbors and $\mathcal{l}_{1}$ norm mechanisms.]{\includegraphics[width=0.32\textwidth]{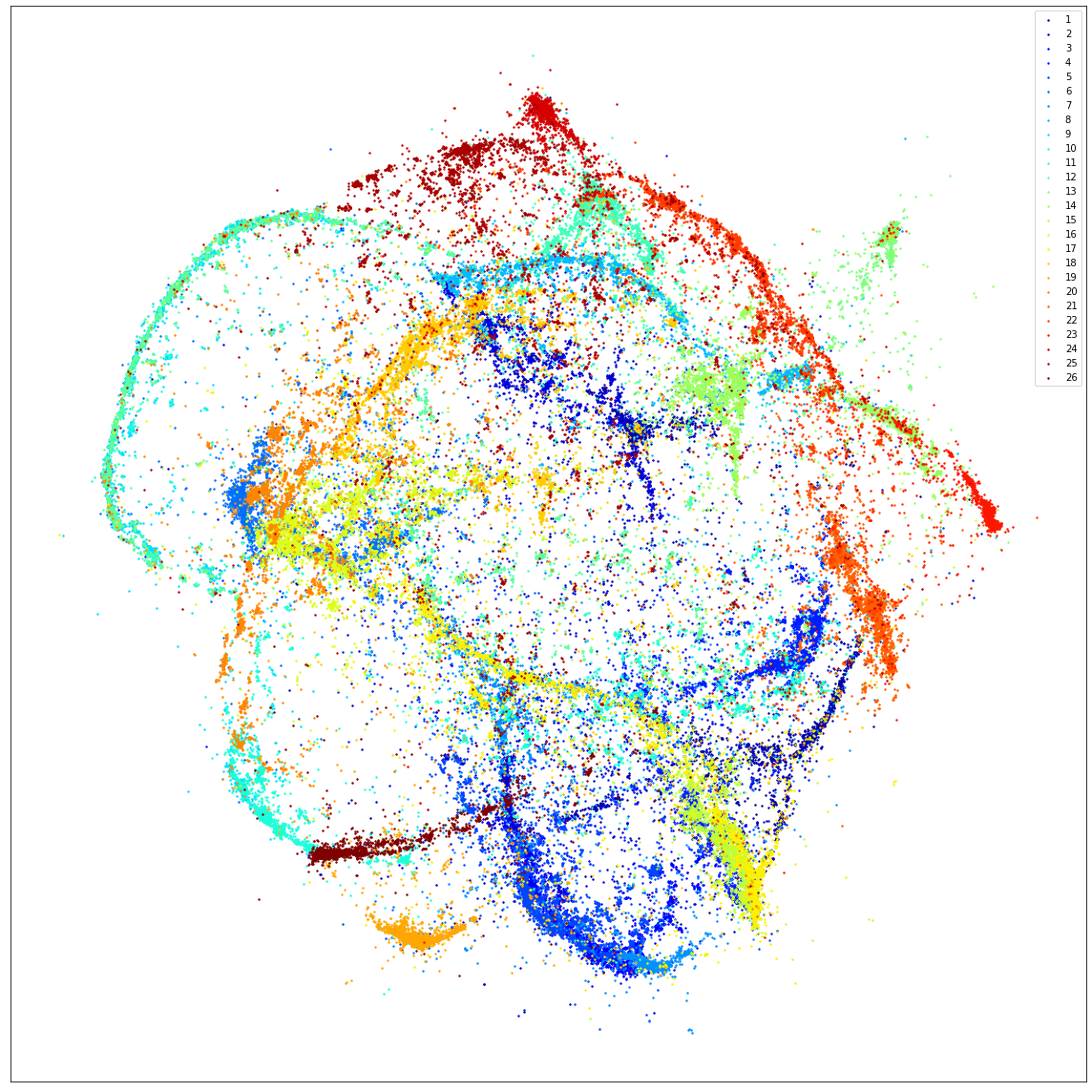}}
        \caption{IVHD comparison on EMNIST dataset, when reverse neighbors mechanism is used for last 200 steps of embedding procedure and when it is combined with $\mathcal{l}_{1}$ norm.}
    \label{fig:ivhd_reverse_neighbors_embedding_emnist}
\end{figure}

We combined both mechanisms to provide the most generic way to deal with the noise that remains in the embedding (Figs. \ref{fig:ivhd_reverse_neighbors_embedding_mnist}c and \ref{fig:ivhd_reverse_neighbors_embedding_emnist}c). As the metrics in Fig. \ref{fig:chapter_7_dr_quality_knn_gain_mnist_emnist} indicate, the combined methods improve the quality of the DR and the kNN gain of the embedding. Changes do not distort the global properties of the visualization, which is confirmed by the proportional increase in both charts. Furthermore, in both cases, the best results were obtained by combining the norm $\mathcal{l}_{1}$ and the reverse neighbors mechanism. Importantly, the improvements do not introduce a large computational overhead. The helper graph is created simultaneously (or read from cache) with the main graph, which is used to calculate interactions between particles. The main operation that causes overhead is the search for reverse neighbors based on the two graphs, but the time of this procedure is negligible compared to the time of the entire embedding. It is worth mentioning that all the improvements added (and tested) to the IVHD method had a crucial assumption, which was that no additional and significantly higher computational load could be introduced into the algorithm.

In summary, the application of the above-described enhancements makes IVHD even more beneficial in terms of data reduction quality of created visualizations. Furthermore, the results have become more comparable to the baseline methods (Chapter \ref{chapter5_experiments}) with no additional computational load. 
\chapter{Experimental setup}

The experimental setup consists of two key components: 1) definition of mutual quality assessment framework and 2) common architectures for running experiments. Multiple methods were implemented in the VisKit visualization library\cite{viskit}, which consists of the C++ and Python APIs. The components of the library are closely interrelated, but it is possible to use them independently. This was done to create the possibility to implement new visualization methods into this framework, without any additional requirements. This framework allows creating a stable environment for: 1) testing and evaluating embedding and optimization methods, 2) adjusting the parameters used by different methods, 3) formulate and instantly verify a number of hypotheses, and 4) integrate the methods in one visualization framework allowing for multiscale analysis of data, employing more accurate methods for finer scales.

\begin{figure}[ht]
    \begin{center}
        \includegraphics[clip,width=0.85\columnwidth]{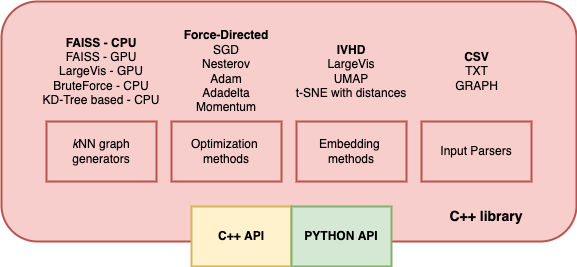}
        \label{pic:library}
        \caption{Viskit library architecture. It consists of multiple visualization methods, that can be freely modified (in terms of implementation) and tested. It also contains different optimization variants, that can be used in IVHD.}
    \end{center}
\end{figure}

\section{Quality assessment of dimensionality reduction}
\label{sec:quality_assessment}
In practice, neither intrinsic dimensionality nor topological properties can be easily identified starting from a set of points. Therefore, the goal of non-linear dimensionality reduction (NLDR) is most often to preserve the structure of the dataset, as indicated, for example, by some kind of neighborhood relations such as proximity or similarity. The research community has focused on the design of new NLDR methods, and the question of a unified quality assessment framework remains largely unanswered. Since most NLDR methods optimize a given objective function, a simplified way to assess quality is to look at the value of the objective function after convergence. Of course, this allows for the comparison of several runs with, for example, different parameter values, but makes the comparison of different methods unfair. Another obvious criterion is the reconstruction error. The first purpose of this chapter is to review some SOTA-based and ranking-based criteria. 

\begin{table}[ht]
\caption{Symbols used to define error measures.}
\vspace{0.4cm}
\begin{tabular}{l l}
    $C_{k}(x_{i})$ & the set of $k$ data vectors that are closest to $x_{i}$ in the original data \\
     & space \\
    $\hat{C_{k}}(x_{i})$ & the set of $k$ data vectors that are closest to $x_{i}$ after projection\\
    $U_{k}(x_{i})$ & the set of data vectors $x_{j}$ for which $x_{j} \in \hat{C_{k}(x_{i}}) \wedge x_{j} \notin C_{k}(x_{i})$ holds \\
    $V_{k}(x_{i})$ & the set of data vectors $x_{j}$ for which $x_{j} \notin \hat{C_{k}(x_{i}}) \wedge x_{j} \in C_{k}(x_{i})$ holds \\
    $r(x_{i}, x_{j}), i \neq j$ & the rank of $x_{j}$ when the data vectors are ordered based on their \\
     & Euclidean distance from the data vector $x_{i}$ in the original data space \\
    $\hat{r}(x_{i}, x_{j}), i \neq j$ & the rank of $x_{j}$ when the data vectors are ordered based on their \\
     & distance from the data vector $x_{i}$ after projection \\
\end{tabular}
\end{table}

\subsubsection{Trustworthiness and continuity}
In principle, errors could simply be measured as the average number of data items that enter or leave the neighborhoods in the projection. Using slightly more informative measures: The trustworthiness of the neighborhoods\cite{venna_2001} is quantified by measuring how far from the original neighborhood the new data points entering a neighborhood come. Distances are measured as rank orders; similar results have also been obtained with Euclidean distances. The trustworthiness of the projected result is defined as:

\begin{equation}
    M_{1}(k) = 1 - \frac{2}{Nk(2N-3k-1)} \sum_{i=1}^{N} \sum_{x_{j} \in U_{k}(x_i)} (r(x_{i}, x_{j}) - k),
    \vspace{0.4cm}
\end{equation}

where the term before the summation scales the values of the measure between zero and one. Preservation of the original neighborhoods (continuity) is measured by the following:
\begin{equation}
    M_{2}(k) = 1 - \frac{2}{Nk(2N-3k-1)} \sum_{i=1}^{N} \sum_{x_{j} \in V_{k}(x_i)} (\hat{r}(x_{i}, x_{j}) - k),
    \vspace{0.4cm}
\end{equation}

When interpreting the visualizations, it is particularly important that small neighborhoods, which have a small value of k, are trustworthy. Although preservation of the original neighborhoods is not as important as the trustworthiness of the projection, it gives insight into the trade-offs made in the projection.

\subsubsection{Shepard and Co-rank diagrams}
In the general situation of non-metric scaling, using the Shepard-Kruskal approach, we have data $Y = \{y_i,\dots,y_N\}$ and a model $f_i(\theta$) with a certain number of free parameters $\theta$. Often this is a non-metric multivariate scaling model in which the model values are dissimilarities. However, linear models and models with an inner product can and have been treated in the same way. We want to choose the parameters so that the rank order of the model best approximates the rank order of the data. To do this, we construct a loss function of the form:

\begin{equation}
    \sigma(\theta, \hat{y}) = \sum_{i=1}^{N} w_{i} (\hat{y_{i}} - f_{i}(\theta))^{2}
    \vspace{0.4cm}
\end{equation}

After we have found the minimum, we can make a scatter plot with the data $Y$ on the horizontal axis and the model values $f$ on the vertical axis. This is what we would also do in a linear or non-linear regression analysis. In summary, \textit{Shepard diagram} shows the change in distances after dimensionality reduction for each pair of points. Similarly, we can define \textit{Co-rank diagram}, which shows the change in mutual distance ranks after dimensionality reduction for each pair of points. The fitness of the co-rank diagram to the identity function is measured with a $R^{2}$ score. In a perfect scenario, the co-rank diagram should depict an identity function, meaning that the relative distance ranks from original dimensionality are accurately preserved in the space of reduced dimensionality.

\begin{figure}[ht]
     \centering
        \subfloat[Shepard diagram.]{\includegraphics[width=2.8in]{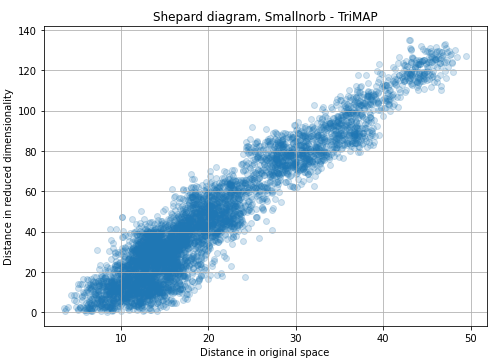}}
        \subfloat[Co-rank diagram.]{\includegraphics[width=2.8in]{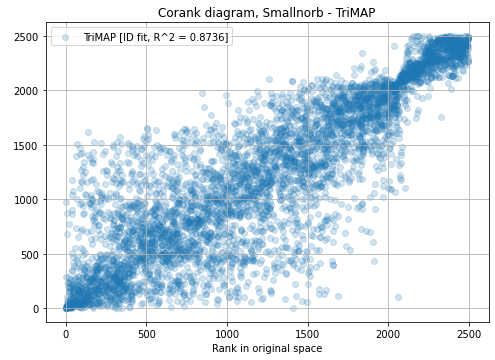}}
        \caption{Example diagrams obtained for Trimap\cite{2019TRIMAP} visualization of smallNORB dataset (described in details in Appendix \ref{appendix_datasets}).}
    \label{fig:chapter_5_shepard_corank_trimap_example}
\end{figure}

\subsubsection{Neighbor hit}
Because of the high computational load required to calculate the precision / recall coefficients, to compare data separability and class purity, we define the following simple metrics, which utilize the mechanisms of neighbor hit (NH) and original neighbor hit (ONH): 

\begin{equation}\label{eq:eq12}
    cf_{\text{\it nn}}=\frac{\sum_{i=1}^{M}nn(i)}{nn{\cdot} M}
    \quad\text{and}\quad
    cf=\frac{\sum_{{\text{\it nn}}=1}^{{\text{\it nn}}_{\max}}cf_{\text{\it nn}}}{nn_{\max}},
    \vspace{0.4cm}
\end{equation}

\noindent where $nn(i)$ is the number of nearest neighbors of $y_i$ in the space $Y$, which belong to the same class as $x_i$. The value of $cf$ is calculated for an arbitrarily defined value of $nn_{\max}$ depending on the number of feature vectors in the class. To reflect a wide range of embedding properties, we use $nn_{\max}$$=$$100$. The value of $cf$$\sim$$1$ for well separated and pure classes, while $cf$$\sim$$1/K$ for random points in the $K$ class. These simple metrics allow us to assess the quality of the embeddings by calculating several $cf_{nn}$ values for a small, medium and greater number of $nn$. Differences in this criterion for confronting methods allow inferring their embedding quality for a very local ($nn$=2), local ($nn$=10) and medium ($nn$=100) reconstruction depth. The stability of $cf_{nn}$ to increase $nn$ means a more compact and circular shape of the classes. For the elongated and mixed classes, the values of $cf_{nn}$ decrease faster with $nn$.

\subsubsection{Rank-based criteria}

Furthermore, we use state-of-the-art quality criteria \cite{lee2009,lee2015} for unsupervised DR methods, measuring the preservation of the high-dimensional neighborhood in the low-dimensional space. The general consensus that emerges from many publications \cite{france2007,chen2009,lee2009,venna2010} is to use the average agreement rate between $k$-ary neighborhoods in high and low dimensions. The rank of $x_{j}$ with respect to $x_{i}$ in a high-dimensional space is written as $\rho_{ij} = |\{ k : \delta_{ik} < \delta_{ij} \;\; or \;\; (\delta_{ik} = \delta_{ij} \;\; and \;\; 1 \leq k < j \leq N) \}|$, where $|A|$ denotes the cardinality of the set $A$. Similarly, the rank of $y_{j}$ relative to $y_{i}$ in the low-dimensional space is $r_{ij} = |\{ k : d_{ik} < d_{ij} \;\; or \;\; (d_{ik} = d_{ij} \;\; and \;\; 1 \leq k < j \leq N) \}|$. Now, let $\textbf{v}_{i}^{k} = \{ j: 1 \leq \rho_{ij} \leq k\}$ and $\textbf{n}_{i}^{k} = \{ j: 1 \leq r_{ij} \leq k\}$ denote the sets of nearest neighbors \textit{k} of $x_i$ and $y_i$ in the high-dimensional and low-dimensional space, respectively. $Q_{NX}(k)$ measures their averaged normalized agreement (also called \textit{neighborhood preservation}):

\begin{equation}
    Q_{NX}(k) = \frac{1}{kN}\sum_{i=1}^{N}|{v}_{i}^{k} \cap {n}_{i}^{k}| \in [0, 1]
    \vspace{0.4cm}
\end{equation}

It varies between 0 and 1, with 0 being an empty intersection and 1 being perfect agreement. Knowing that the random coordinates in $\textbf{Y}$ on average lead to $Q_{NX}(k) = \frac{k}{k-1}$, the useful range of $Q_{NX}(k)$ is $N-1-k$, which depends on $k$. Therefore, to compare or combine the values of $Q_{NX}(k)$ fairly for different neighborhood sizes, the criterion can be rescaled, as in \cite{lee2013}:

\begin{equation}
    R_{NX}(k) = \frac{(N-1) Q_{NX}(k) - k}{N-1-k}
    \vspace{0.4cm}
\end{equation}

This modified criterion indicates an improvement over a random embedding and has the same useful range between 0 (random) and 1 (perfect) for all $k$. $R_{NX}(k)$ is shown, with a logarithmic scale for $k$. This choice is justified by the fact that the size $k$ and the radius $R$ of small neighborhoods with uniform density in a $P$-dimensional space are (locally) related by $k \varpropto R^{P}$ . A logarithmic axis also reflects that errors in large neighborhoods are proportionally less important than in small ones. Eventually, a scalar score is obtained by computing the area under the $R_{NX}(k)$ curve in the logarithmic graph given by:

\begin{equation}
    AUC(R_{NX}(k)) = \frac{\sum_{k=1}^{N-2}\frac{R_{NX}(k)}{k}}{\sum_{k=1}^{N-2}\frac{1}{k}}.
    \vspace{0.4cm}
\end{equation}

The AUC supposedly assesses the average quality of DR on all scales with the most appropriate weights. The higher the AUC, the better the result. \\

When the data come with class labels, unsupervised DR can also be assessed by its performance in classification tasks, reporting the accuracy of a $KNN$ classifier in the LD space:

\begin{equation}
    G_{NN}(k) = \frac{1}{N} \sum_{i=1}^{N} \frac{|j \in {n}_{i}^{k} s.t.c_{i} = c_{j}| - |j \in {v}_{i}^{k} s.t.c_{i} = c_{j}|}{k}
    \vspace{0.4cm}
\end{equation}

Average the gain (or loss, if negative) of neighbors of the same class around each point after DR. Therefore, a positive value is probably correlated with better KNN classification performance. $G_{NN}(k)$ can also be summarized by a global score provided by its Area Under Curve (AUC), where $AUC|G_{NN}(k)| \in [-1, 1]$.

\begin{figure}[ht]
     \centering
        \subfloat[DR quality ($R_{NX}(k)$).]{\includegraphics[width=2.8in]{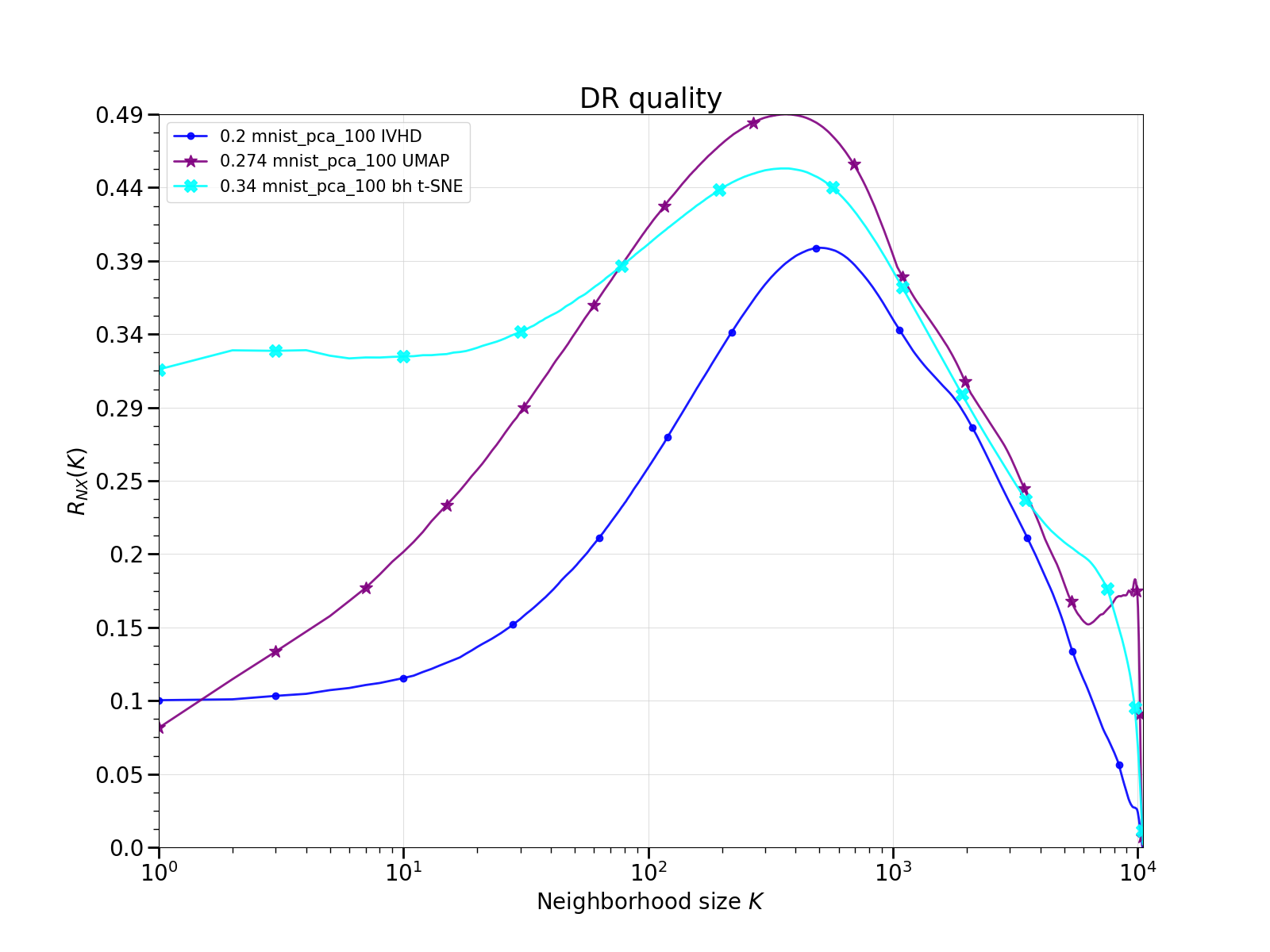}}
        \subfloat[kNN gain ($G_{NN}(k)$).]{\includegraphics[width=2.8in]{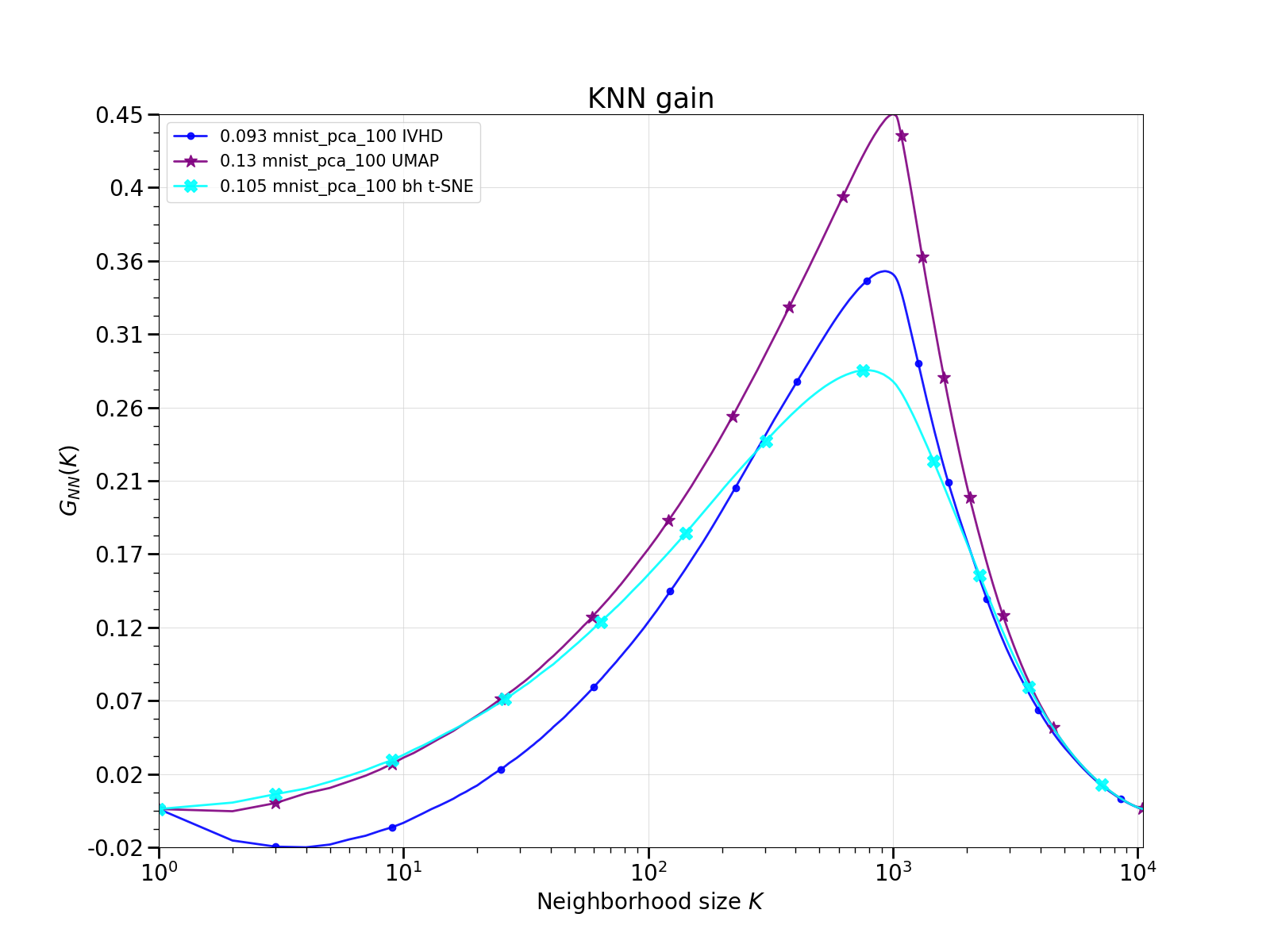}}
        \caption{Example rank-based diagrams obtained for UMAP, t-SNE and IVHD visualizations of MNIST dataset.}
    \label{fig:chapter_5_rank_based_diagrams}
\end{figure}

As shown in Figure 4.3, the DR quality and kNN gain allow you to better observe the results of the data embedding depending on the size of the neighborhood that is being observed. This makes it possible to assess whether the method is better or worse in terms of locality (small neighborhood) or globality (large neighborhood size). However, it is important to remember that the size of the neighborhood (and whether we are talking about locality or globality) depends on the size of the dataset.

\section{Datasets}

The main properties of each baseline dataset are: the number of samples $M$, the dimensionality $N$, and the number of classes $K$. Here, we consider datasets with (1) a large number of samples and relatively low dimensionality, (2) a smaller number of samples but a larger number of features, (3) highly imbalanced data (RCV-Reuters), and (4) skewed data (Small NORB). The most common datasets used in different experiments are described in Table 2. A complete breakdown of the datasets used in this work can be found in the Appendix \ref{appendix_datasets}.

\begin{table}[ht]
\small
\caption{The list of some baseline datasets.}
\vspace{0.2cm}
\begin{tabular}{|p{2.5cm}|c|c|c|p{7.25cm}|}
\hline
\multicolumn{1}{|c|}{Dataset} & \multicolumn{1}{c|}{$N$} & \multicolumn{1}{c|}{$M$} & \multicolumn{1}{c|}{$K$} & \multicolumn{1}{c|}{Short description}\\ \hline
MNIST & 784 & 70 000  & 10 & Well balanced set of grayscale images of handwritten digits.\\ \hline
Fashion-MNIST & 784 & 70 000 & 10 & More difficult MNIST version. Instead of handwritten digits it consists of apparel images.\\ \hline
Small NORB & 2048 & 48 600 & 5 & It contains stereo image pairs of 50 uniform-colored toys under 18 azimuths, 9 elevations, and 6 lighting conditions.\\ \hline
RCV-Reuters & 30 & 804 409 & 8 & Corpus of press articles preprocessed to 30D by PCA.\\ \hline
\end{tabular}
\end{table}

\vspace{-0.5cm}

\subsubsection{Preprocessing}
\label{sec:dataset_preprocessing}

\begin{wrapfigure}{r}{7.5cm}
    \centering
    \includegraphics[width=7.5cm]{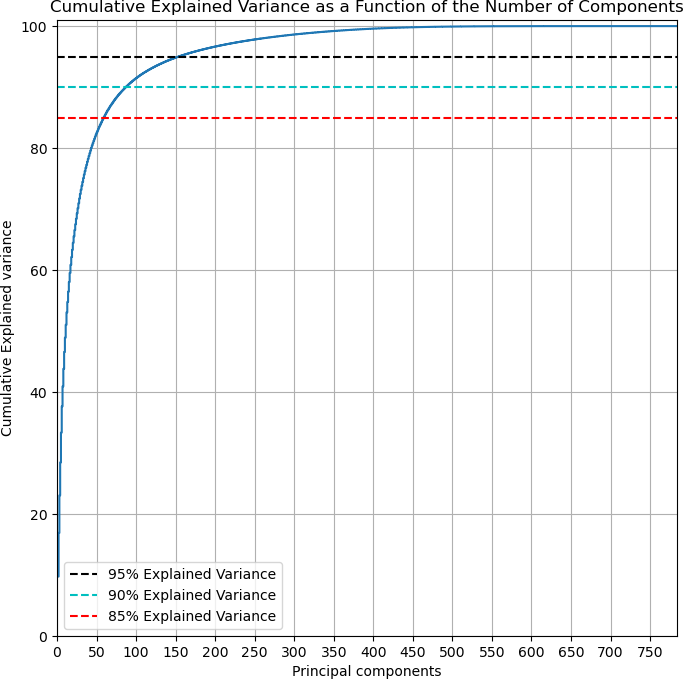}
    \caption{Cumulative explained variance by number of principal components for MNIST dataset.}
    \label{fig:chapter_5_mnist_cumulative_explained_variance}
\end{wrapfigure}

Using PCA, we can introduce explained variance, which measures the proportion to which a mathematical model accounts for the variability (dispersion) of a given data set. Often variability is quantified as variance; then the more specific term explained variance can be used. In simpler words, the explained variance tells us how much information can be attributed to each of the principal components. Using this, we can draw a graph showing the relation between the total explained variance and the number of principal components used. We can observe in Figure \ref{fig:chapter_5_mnist_cumulative_explained_variance} that we can decrease the number of dimensions to $\sim$100 with PCA, still having 90\% of information. We use this method when we do not necessarily need large (multidimensional) datasets. When we mention that the dimensionality of MNIST is 100-D, this means that it has been preprocessed using this methodology.

In addition, we prepared a set of artificially generated "simple" data sets whose internal data structure gives insight into the quality of embeddings created by different methods. They are described in detail in the Appendix \ref{appendix_datasets}. The Mammoth data set was preprocessed, and labels were added based on the Z-axis value (colors were assigned by the height of the skeleton).

\section{Baseline methods}

Although there are many algorithms for visualizing multivariate data and the topic has been intensively researched for years, there are currently two methods that are used as a starting point for ongoing research: UMAP\cite{umap} and t-SNE\cite{tsne}. Most of the concepts that are currently being tested and developed originate from them. Furthermore, LargeVis\cite{largevis} is an interesting approach to visualization of the $NN$-graph, which is what IVHD is doing as well. For these reasons, these are the first three baseline methods (already described in Chapter 2) used for comparison with IVHD. In addition, recent research has contributed to the PaCMAP and TriMAP methods. The first optimizes a low-dimensional embedding using three types of point pairs: neighbor pairs, mid-near pairs, and furthest pairs. The second method uses triplet constraints to create a low-dimensional embedding of a set of points.

\subsubsection{TriMap}

DR method, which focuses on preserving the global structure of data in an embedding. The similarities between points appear to be insufficient to capture the global structure. Instead, TriMap uses a higher-order structure to construct the embedding using triples:

\begin{equation}
    (i, j, k) \Leftrightarrow \text{point i is closer to point j than point k}.
\end{equation}

The key idea of TriMap is derived from semi-supervised metric learning\cite{amid2016}: Given an initial low-dimensional representation for the data points, the triplet information from the high-dimensional representation of the points is used to improve the quality of the embedding. Similarly, the TriMap is initialized with a low-dimensional PCA embedding, and this embedding is then modified using a set of carefully selected triplets from the high-dimensional representation. TriMap chooses a subset $\mathcal{T} = \{(i, j, k)\}$ of triplets and assigns a weight $\omega_{ijk} \geq 0$ to each triplet: a higher value of $\omega_{ijk}$ implies that the pair $(i, k)$ is located much farther away than the pair $(i, j)$. Trimap defines the loss of the triplet $(i,j,k)$ as follows:

\begin{equation}
    C_{ijk} = \omega_{ijk} \frac{s(\textbf{y}_{i}, \textbf{y}_{k})}{s(\textbf{y}_{i}, \textbf{y}_{j}) + s(\textbf{y}_{i}, \textbf{y}_{k})}, \;\;\; s(\textbf{y}_{i}, \textbf{y}_{j}) = (1 + ||\textbf{y}_{i} - \textbf{y}_{j}||^{2})^{-1},
    \vspace{0.2cm}
\end{equation}

where $s(y_i, y_j)$ is a similarity function between $y_i$ and $y_j$. The choice of $s$ is motivated by the good performance of the Student's t distribution for similarities in low dimensions in the t-SNE method. Note that the loss of the triplet $(i, j, k)$ approaches zero as $||y_{i} - y_{j}||$ decreases and $||y_{i} - y_{k}||$ increases. To reflect the relative similarities in high dimension, the weight of the triplet $(i, j, k)$ is defined as:

\begin{equation}
    \hat{\omega}_{ijk} = d_{ik}^{2} - d_{ij}^{2} \geq 0,
\end{equation}

in which $d_{ij}$ is any distance measure between $x_{i}$ and $x_{j}$ of high dimension. As the default distance, TriMap considers the squared Euclidean distance with scaling introduced in \cite{zelnik2004}: $d_{ij}^2$$=$$\frac{||x_{i} - x_{j}||^2}{\sigma_{ij}}$, where $\sigma_{ij}$$=$$\sigma_{i}\sigma_{j}$ and $\sigma_{i}$ are set to the average Euclidean distance between $x_{i}$ and the set of nearest neighbors of $x_{i}$ from neighbors of the 4- to 6-th grade. This choice of $\sigma_{ij}$ adaptively adjusts the scaling according to the density of the data. We change the final weights by subtracting the minimum weight value calculated on all triplets and applying a tempered log transformation, $\omega_{ijk} = \log_{t}(1 + \hat{\omega_{ijk}} - \omega_{min})$, where $\omega_{min} = min_{(i', j', k') \in \mathcal{T}} \hat{\omega}_{i'j'k'}$. Then, the tempered logarithm\cite{naudts2002} is used:

\begin{equation}
    \log_{t}(u) = \frac{1}{1 - t}(u^{1-t} - 1), \;\; t \neq 1.
\end{equation}

Transformation $\log_{t}$ smooths the weights and prevents triplets with large weights from dominating total loss. Note that the limit case $t$$\rightarrow$$1$ recovers the standard $\log$. The default value of $t$ is $0.5$. 

\begin{algorithm}
    \caption{TriMAP scheme.} \label{alg:trimap}
    \SetAlgoLined
    \SetKwInOut{Input}{Input}
    \Input{data matrix $\mathbf{X}$.}
    \textbf{procedure} TriMAP: \\
    \hskip1cm 1. Find a set of triplets $\mathcal{T} = \{(i, j, k)\}$.\\
    \hskip1cm 2. Initialize $\textbf{Y}$ using PCA.\\
    \hskip1cm 3. Minimize the $C_{TriMAP}$ by applying full batch  gradient descent \\
    \hskip1cm with momentum using the delta-bar-delta method.\\
    \hskip1cm 4. Return $\textbf{Y}$.\\
\end{algorithm}

To construct the embedding, TriMAP considers a small subset of all possible triplets $(i, j, k)$ for which the closest point $j$ belongs to the set of nearest neighbors of the point $i$ and the farther point $k$ is among the points that are more distant from $i$ than $j$, chosen uniformly at random. For each point, it considers its nearest neighbors $m$$=$$10$ and the sample $m'$$=$$5$ triplets per nearest neighbor. This yields $m \times m' = 50$ nearest-neighbor triplets per point. Furthermore, we also add random triplets $r$$=$$5$ $(i, j, k)$ per each point $i$ where $j$ and $k$ are sampled uniformly at random and their order can change based on their proximity to $i$. Thus, the overall complexity of the optimization step is linear in number of points $n$. The computational complexity is dominated by the nearest-neighbor search. The final loss is defined as the sum of the losses of the triplets sampled in $\mathcal{T}$:

\begin{equation}
    C_{TriMap} = \sum_{(i,j,k) \in \mathcal{T}} C_{ijk}.
    \vspace{0.4cm}
\end{equation}

The loss is minimized by using the full-batch gradient descent with momentum by using the delta-bar-delta method. The whole procedure is designed in a way that samples the informative triplets from the high-dimensional representation of a set of points and assigns weights to these triplets to reflect the relative similarities of these points. Although TriMap can also be used for the triplet embedding task, it focuses mainly on the DR.

\begin{figure}[ht!]
    \centering
    \includegraphics[width=\textwidth]{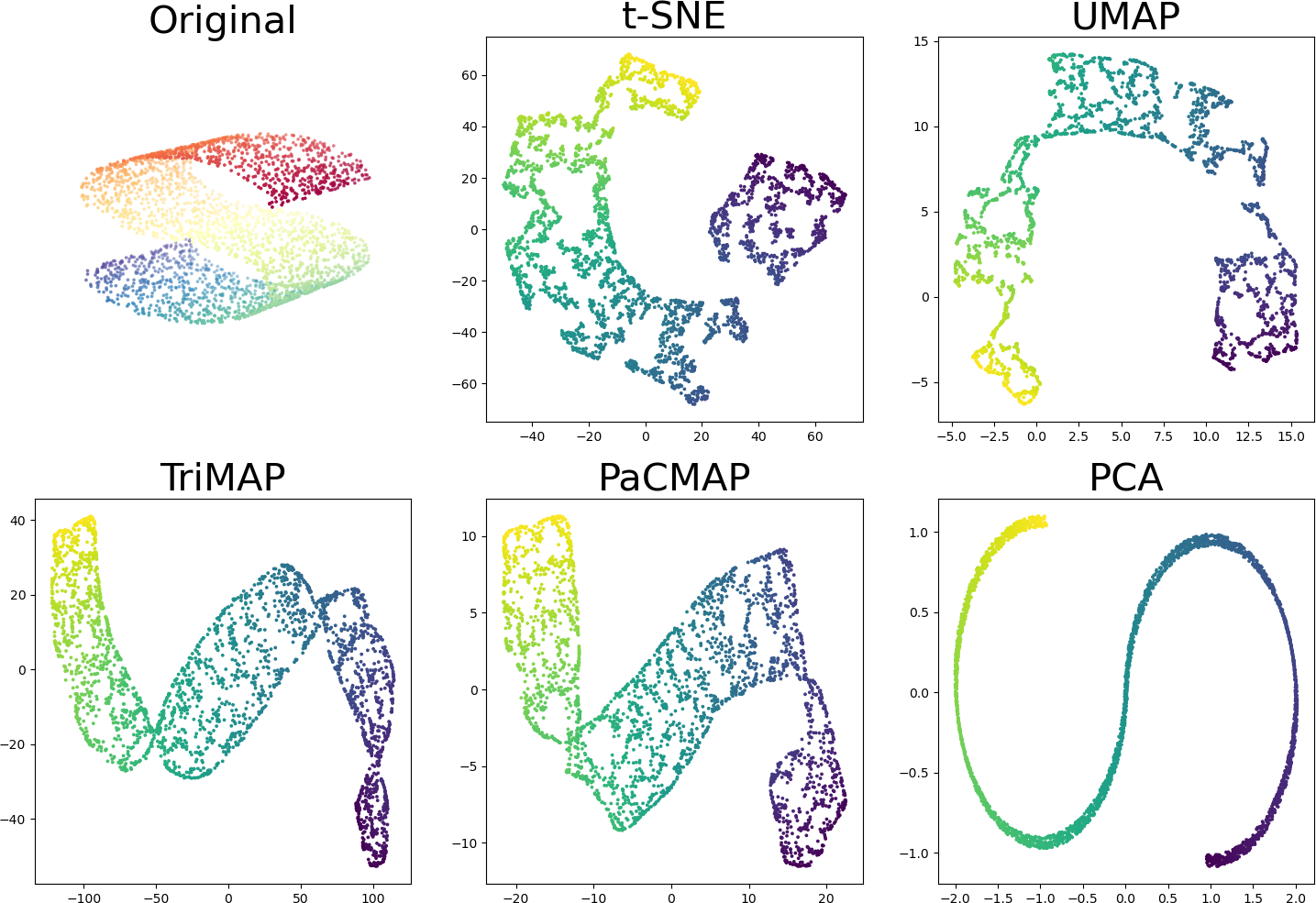}
    \caption{2-D visualizations of the S-curve dataset. As shown, both TriMap and PaCMAP successfully unveils the underlying structure in the original dataset.}
    \label{fig:chapter_5_s_curve_example}
\end{figure}

\subsubsection{PaCMAP}

The goal of Pairwise Controlled Manifold Approximation Projection (PaCMAP) is also to preserve both local and global structure (as in TriMap). It does this by distinguishing between three types of edges in the graph: neighbor pairs (NB), mid-near pairs (MN), and further pairs (FP). The first group consists of the $n_{NB}$ nearest neighbors of each observation in the high-dimensional space. Similarly to TriMap, the following scaled distance metric is used:

\begin{equation}
    d_{ij}^2=\frac{||x_{i} - x_{j}||^2}{\sigma_{ij}}, \; \text{where} \;\; \sigma_{ij}=\sigma_{i}\sigma_{j}
\end{equation}
\vspace{0.1cm}

where $\sigma_{i}$ is the average distance between $i$ and its Euclidean nearest fourth to sixth neighbors. These are used to construct neighbor pairs $(i, j_{t}),t = 1,2,\dots,n_{NB}$. The second group consists of $N \cdot n_{MN}$ mid-near pairs selected by randomly sampling six additional observations from each observation and using the second smallest of them for the mid-near pair. Finally, the third group consists of a random selection of $n_{FP}$ furthest points from each observation. For convenience, the number of pairs of mid-near and other points is determined by the parameter ratio $MN$ and the ratio $FP$ that specify the ratio of these quantities to the number of nearest neighbors, that is, $n_{MN} = MN_{ratio} \cdot n_{NB}$ and $n_{FP} = FP_{ratio} \cdot n_{NB}$. Since the number of neighbors $n_{NB}$ is typically an order of magnitude smaller than the total number of observations, random sampling effectively chooses non-nearest neighbors as mid-near and further pairs. The weights associated with neighbor, near-neighbor and other pairs in iteration $t$ are set to the following default values:
\begin{itemize}
    \item for $t \in [\tau_{1}, \tau_{2})$: $w_{NB} = 2$, $w_{MN}(t) = 1000 \cdot (1 -\frac{t-1}{\tau_{2}-1}) + 3 \cdot \frac{t-1}{\tau_{2}-1}$, $w_{FP} = 1$;
    \item for $t \in [\tau_{2}, \tau_{3})$: $w_{NB} = 3$, $w_{MN} = 3$, $w_{FP} = 1$;
    \item for $t \in [\tau_{3}, n_{iterations}]$: $w_{NB} = 1$, $w_{MN} = 0$, $w_{FP} = 1$;
\end{itemize}

PaCMAP also uses three distinct cost functions for each type of pair:

\begin{equation}
    C_{NB} = \frac{\hat{d_{ij}}}{10 + \hat{d_{ij}}}, \;\; C_{MN} = \frac{\hat{d_{ik}}}{10000 + \hat{d_{ik}}}, \;\;  C_{FP} = \frac{1}{1+\hat{d_{il}}},
    \vspace{0.4cm}
\end{equation}

where $\hat{d_{ab}} = ||y_a - y_b||^2 + 1$. The final cost function, where the weights are dynamically updated is defined as follows:

\vspace{0.5cm}
\begin{equation}
    \begin{aligned}
        C_{PaCMAP} & = w_{NB} \cdot \sum_{\text{i,j are neighbors}} \frac{\hat{d_{ij}}}{10 + \hat{d_{ij}}} + w_{FP} \sum_{\text{i,l are furthest neighbors}} \frac{1}{1+ \hat{d_{il}}} \\\\
                      & + w_{MN} \cdot \sum_{\text{i,k are mid-near neighbors}} \frac{\hat{d_{ik}}}{10000+\hat{d_{ik}}}
    \end{aligned}
\end{equation}
\vspace{0.5cm}

\begin{algorithm}
    \caption{Pairwise Controlled Manifold Approximation Projection scheme.} \label{alg:pacmap}
    \SetAlgoLined
    \SetKwInOut{Input}{Input}
    \Input{data matrix $\mathbf{X}$.}
    \textbf{procedure} PaCMAP: \\
    \hskip1cm 1. Find a set of near, mid-near and furthest pairs.\\
    \hskip1cm 2. Initialize $\textbf{Y}$ using PCA.\\
    \hskip1cm 3. Begin three optimization phases that satisfies: \\
    \hskip1cm $\tau_{1} = 1 \leq\tau_{2} \leq \tau_{3} \leq n_{iterations}$.\\ \hskip1cm Default values: $\tau_{1} = 1$, $\tau_{2} = 101$, $\tau_{3} = 201$. \\
    \hskip1cm 4. Set initial weights. \\
    \hskip1cm 5. Run AdamOptimizer for $n_{iterations}$ to minimize the cost \\ \hskip1cm function $C_{PaCMAP}$, while simultaneously adjusting the weights. \\
    \hskip1cm 6. Return $\textbf{Y}$. \\
\end{algorithm}
    
\textbf{Dynamic Optimization.} The optimization process consists of three phases that aim to avoid the local optimum. In the first phase, the goal is to improve the initial distribution of embedded points to one that preserves both global and local structure to some extent but mainly global structure. This is achieved by heavily weighting mid-near pairs. During the first phase, PaCMAP gradually reduces the weights on the pairs mid-near, allowing the algorithm to gradually shift from the global structure to the local structure. In the second phase, the goal is to improve the local structure while preserving the global structure captured in the first phase by assigning a small (but not zero) weight to mid-near pairs. Together, the first two phases attempt to avoid local optima using a process that reveals similarities to simulated annealing and the \textit{early exaggeration} technique used by t-SNE. However, early exaggeration places more emphasis on neighbors rather than midpoints, while PaCMAP focuses on midpoint pairs first and neighbors later. Finally, in the third phase, the focus is on improving the local structure by reducing the weight of the center pairs of points to zero and the weight of neighbors to a smaller value, emphasizing the role of the repulsive force that helps separate possible clusters and make their boundary clearer.

\section{Experiments}
\label{chapter5_experiments}

In this section, we present visualizations of small artificially generated datasets that provide an easy way to evaluate the specific properties of the baseline DR algorithms. Additionally, we present hardware on which calculations were performed for specific cases.

\subsection{Hardware}

All CPU implementations of the baseline embedding were executed in two environments, depending on the scale of the dataset processed. Mid-scale datasets were visualized on Macbook Pro 2,3 GHz 8-Core Intel Core i9, 16 GB 2667 MHz DDR4. Large-scale datasets were processed in HPC Prometheus environment: Intel(R) Xeon(R) CPU E5-2680 v3 @ 2.50GHz, 64 GB RAM. The codes were compiled using Apple Clang version 13.1.6 and GCC-10.1.

All GPU/CUDA implementations of the baseline embedding methods were executed on the separated remote server with: CPU Intel Xeon E5-2620 v3, GPU Nvidia
GeForce GTX 1070 (1920 CUDA cores, 8GB GDDR5), 252 GB RAM, OS: Ubuntu 18.04.3, architecture x86, 64. Codes were compiled using GCC-7.4 and CUDA Toolkit 10.0.

\subsection{Synthetic datasets}

To evaluate the DR properties of the baseline methods, we begin with experiments on the synthetic datasets described in Table \ref{tab:artificially_generated_datasets}. In this way, we can make a preliminary assessment of which methods have more advantages in terms of locality and globality.

\begin{figure}[ht!]
     \centering
        \includegraphics[width=\textwidth]{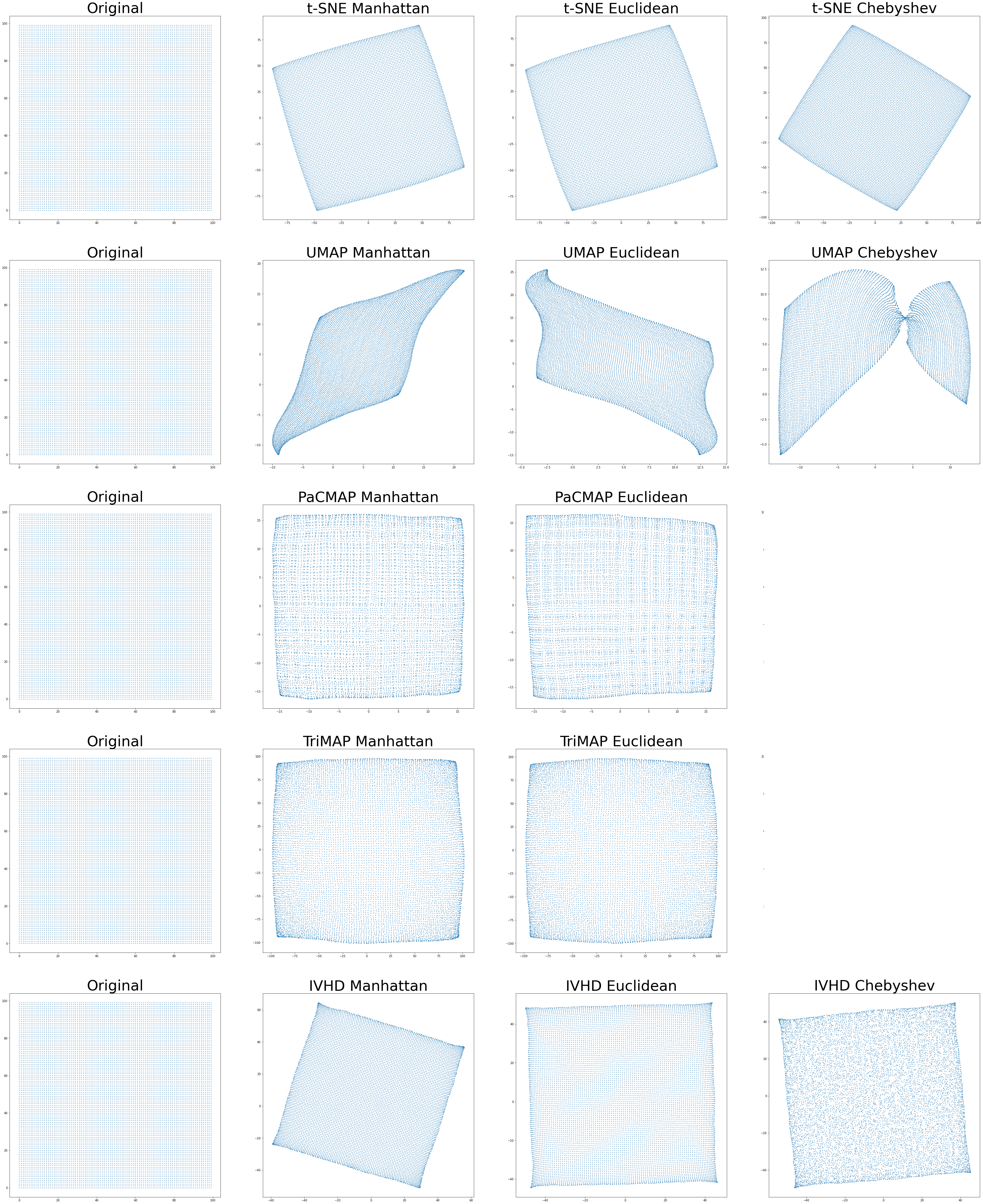}
        \caption{Methods comparison for embedding 2-dimensional grid using various metrics for calculating distance.}
    \label{fig:chapter_5_results_grid_metrics}
\end{figure}

In Figure \ref{fig:chapter_5_results_grid_metrics}, we conducted an experiment that presented how a two-dimensional mesh with evenly spaced points was embedded by all baseline methods while using different metrics to calculate distances. Both the PaCMAP and TriMAP methods do not provide the possibility of using precalculated distances or the Chebyshev norm in the calculation process. It can be seen that both t-SNE and IVHD achieve the best embedding, which preserves the global structure and shape of the mesh without introducing any distortion. UMAP introduce distortions and that curl the mesh during the embedding process. In addition, we can see that all methods cause compression of points, that are located in the corners of the mesh. The overall shape is best preserved by IVHD. The mesh edges in the embedding were arranged as straight lines, and the data points are mostly evenly spaced and also arranged as straight lines, just like in the original mesh.

\begin{figure}[ht!]
     \centering
        \includegraphics[width=\textwidth]{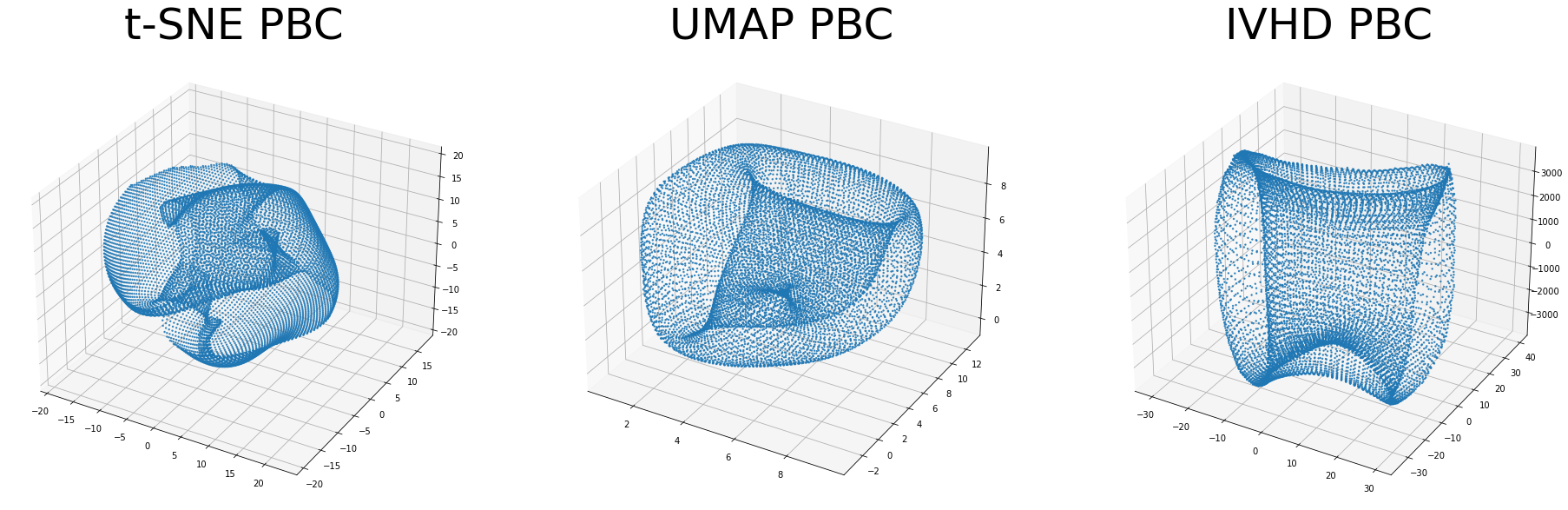}
        \caption{Methods comparison for embedding 2-dimensional grid into 3-dimensional space using periodic boundary condition.}
    \label{fig:chapter_5_results_pbc}
\end{figure}

In Figure \ref{fig:chapter_5_results_pbc}, we demonstrated how t-SNE, UMAP and IVHD embed 2-dimensional grid into 3-dimensional space, when using periodic boundary condition (PBC). Theoretically, in this process one should obtain a torus-like structure. In PBC, the geometry of the unit cell satisfies perfect two-dimensional tiling, and when an object passes through one side of the unit cell, it re-appears on the opposite side with the same velocity. As we can see, IVHD is the only method, that generates torus-like structure among given set of methods. TriMAP, LargeVis and PaCMAP were not considered in this experiment because their implementations do not allow passing precalculated graph as input to the algorithm.

\begin{figure}[ht!]
     \centering
        \includegraphics[width=\textwidth]{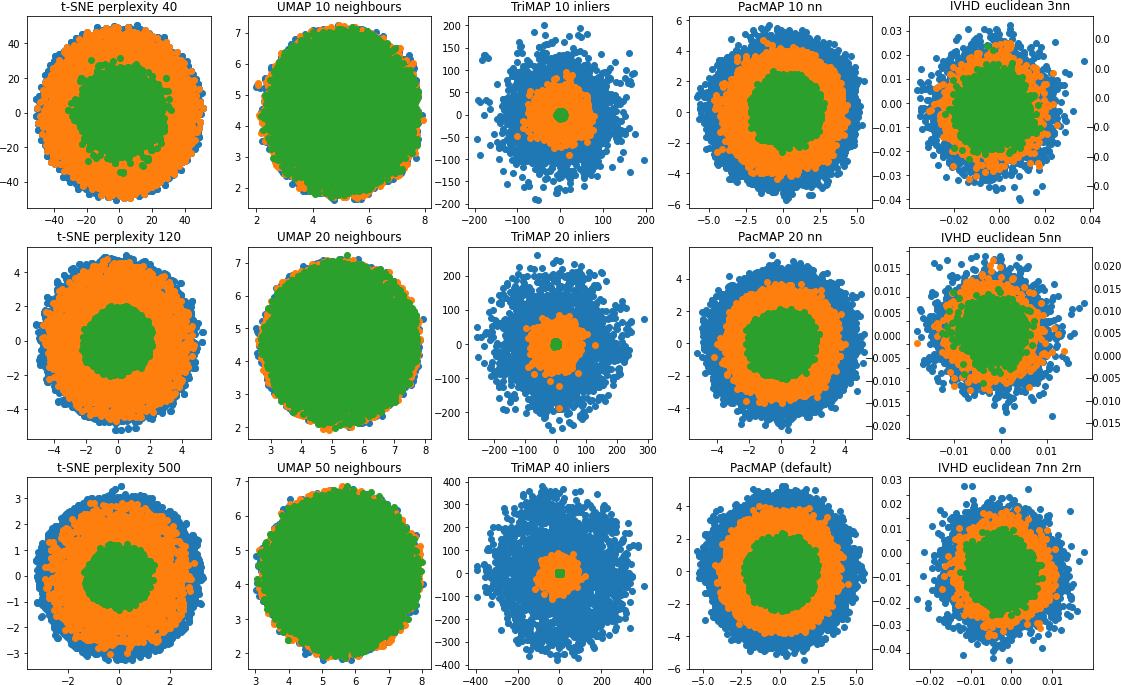}
        \caption{Methods comparison for 15k-sample ball inside of 2 spheres. Both ball and spheres were generated in 30-D space.}
    \label{fig:chapter_5_results_ball_in_2_spheres}
\end{figure}

In Figure \ref{fig:chapter_5_results_ball_in_2_spheres} we show that t-SNE, TriMAP, PaCMAP, and IVHD manage to obtain the global structure of the data in which we can see the ball and the two spheres that surround it. UMAP, on the other hand, is completely unable to separate classes.

\begin{figure}[ht!]
     \centering
        \subfloat[DR quality ($R_{NX}(k)$).]{\includegraphics[width=2.8in]{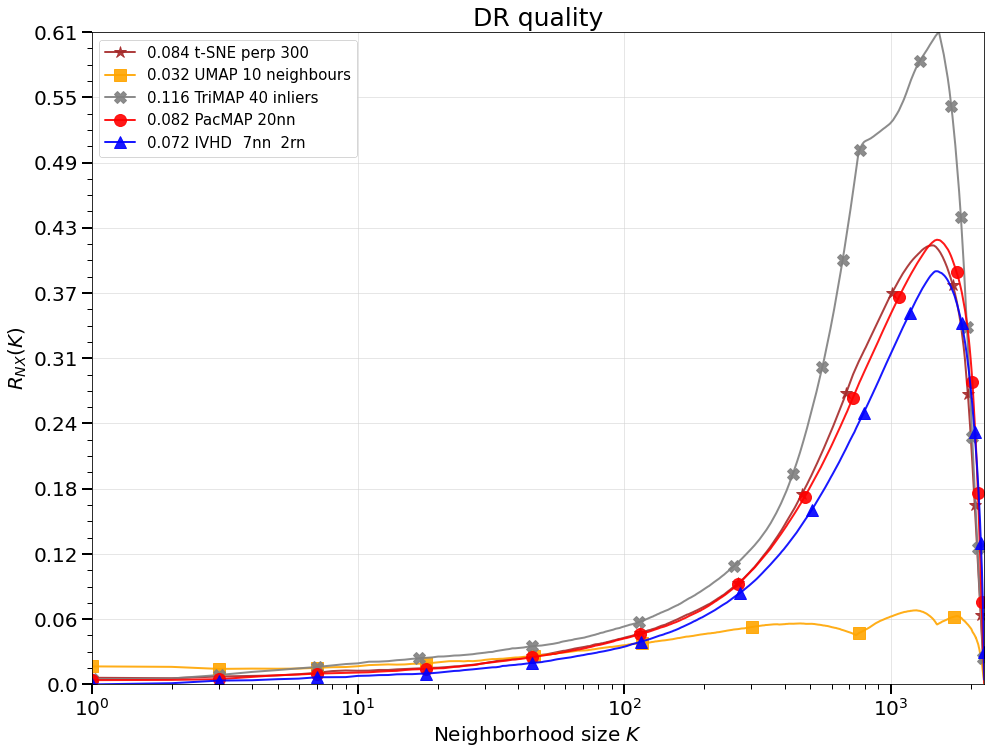}}
        \subfloat[kNN gain ($G_{NN}(k)$).]{\includegraphics[width=2.8in]{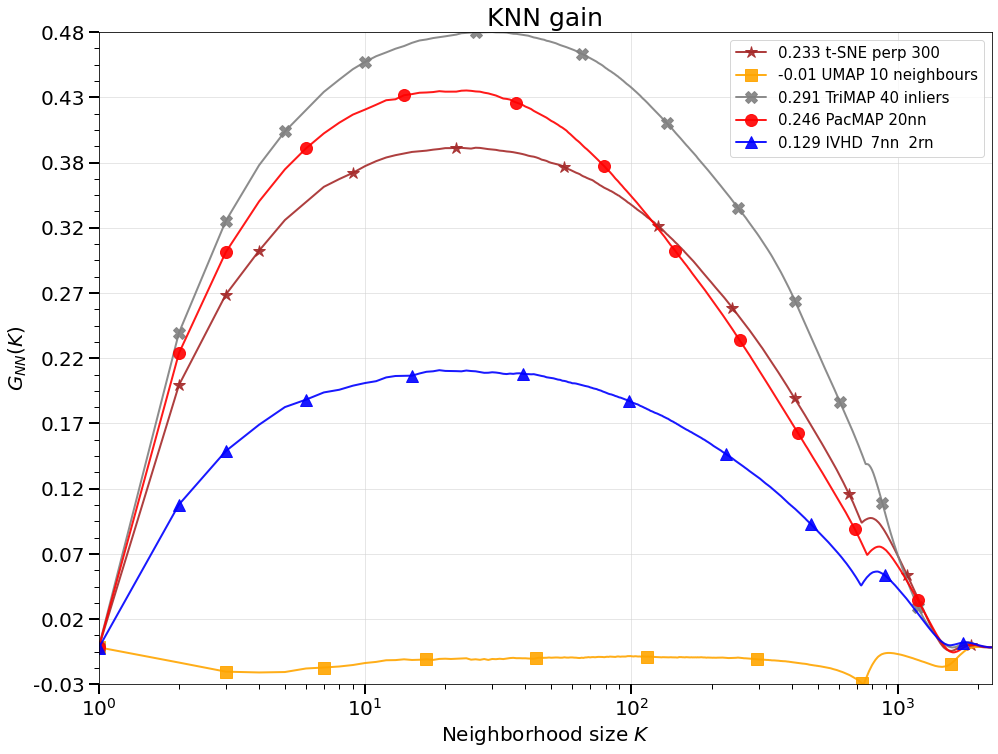}}
        \caption{KNN gain and DR quality obtained for comparison of different DR methods for "ball in 2 spheres" dataset.}
    \label{fig:chapter_5_diagrams_ball_in_2_spheres}
\end{figure}

It is worth mentioning that different parameterizations were tested for each dataset. Here, the best results obtained for each method are presented (both in terms of visual aspect and in terms of metrics obtained).

\begin{figure}[ht!] 
     \centering
        \includegraphics[width=\textwidth]{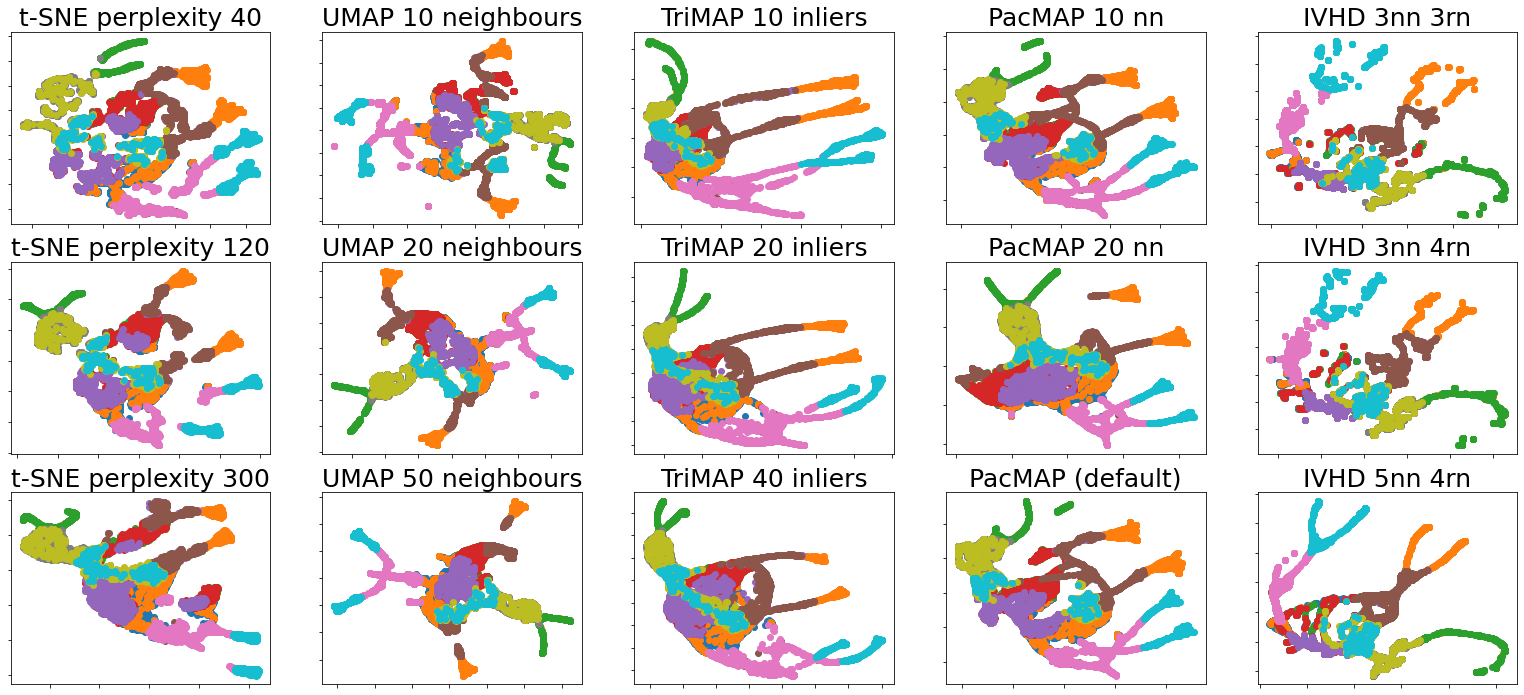}
        \caption{Methods comparison for Mammoth dataset.}
    \label{fig:chapter_5_results_mammoth}
\end{figure}

The mammoth data set (Figure \ref{fig:chapter_5_results_mammoth}) is a 3D point cloud of a mammoth skeleton. Its main research value is that, with this data set, the preservation of local and global properties can be perfectly verified. Ideally, we would like the location of subsequent mammoth body parts to be logically correct (e.g., mammoth tusks should be attached to the head). Visualization of original mammoth data set is presented in Appendix \ref{appendix_datasets}.

PaCMAP, TriMap and IVHD methods achieve a good balance of local and global properties in mammoth dataset embedding. The next mammoth body parts are properly embedded in the low-dimensional 2-D space. For t-SNE and UMAP, it is evident that the overall structure of the embedding is not as clear and obvious, where classes are incorrectly distributed in space.

\begin{figure}[ht!]
    \centering
        \subfloat[DR quality ($R_{NX}(k)$).]{\includegraphics[width=2.8in]{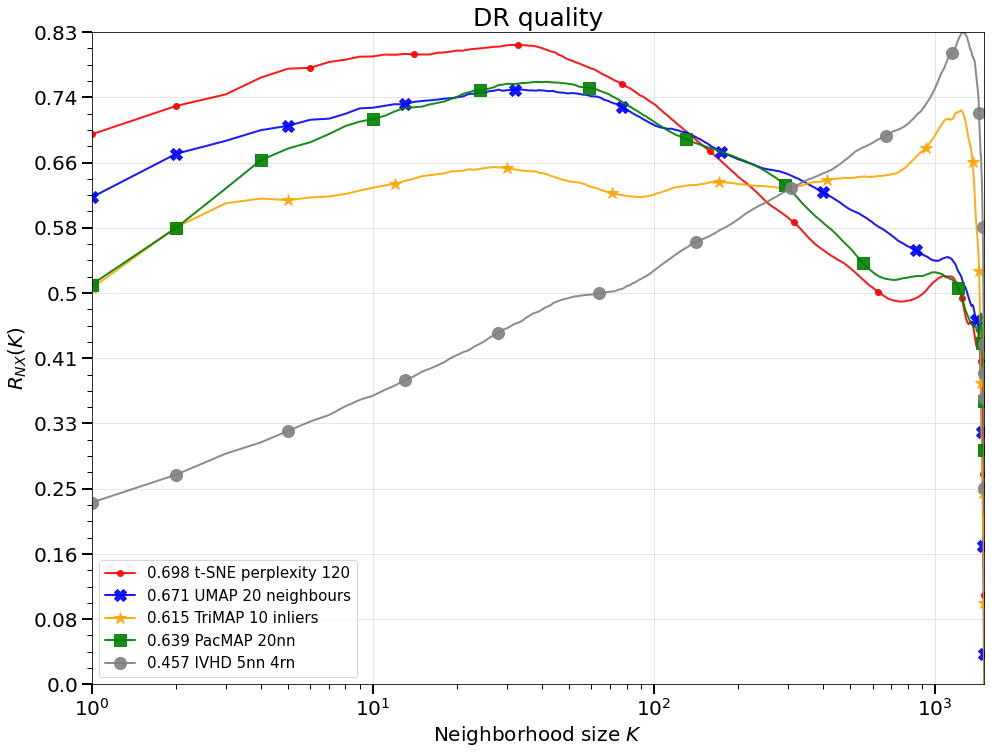}}
        \subfloat[kNN gain ($G_{NN}(k)$).]{\includegraphics[width=2.8in]{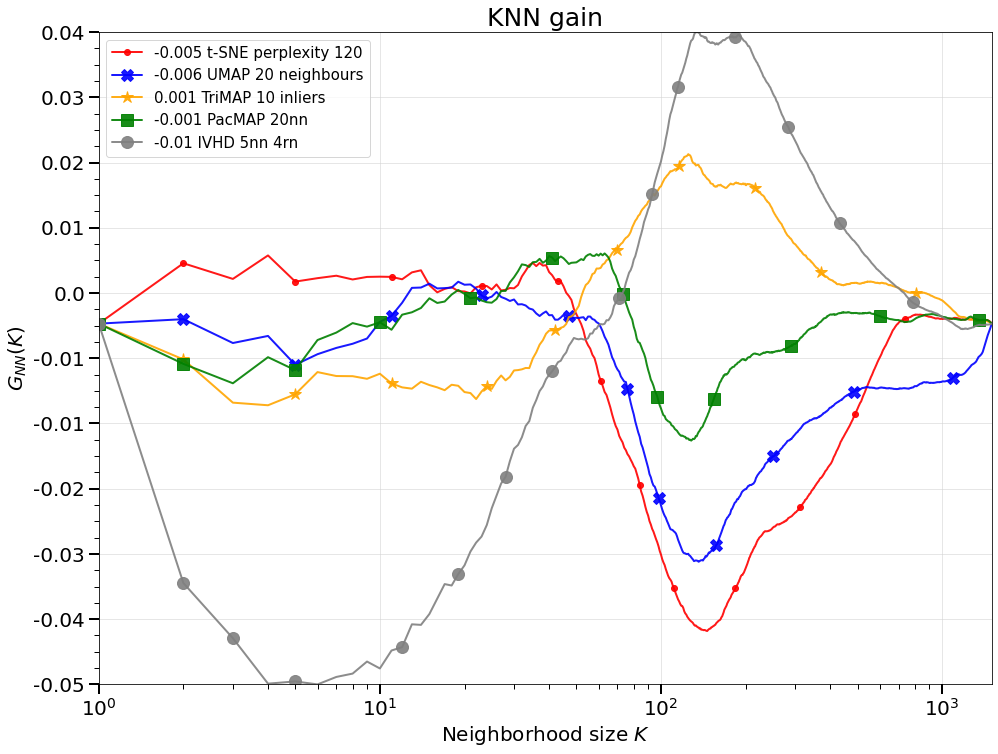}}
        \caption{KNN gain and DR quality obtained for comparison of different DR methods for "Mammoth" dataset.}
    \label{fig:chapter_5_diagrams_mammoth}
\end{figure}

\begin{figure}[ht!]
    \centering
        \includegraphics[width=\textwidth]{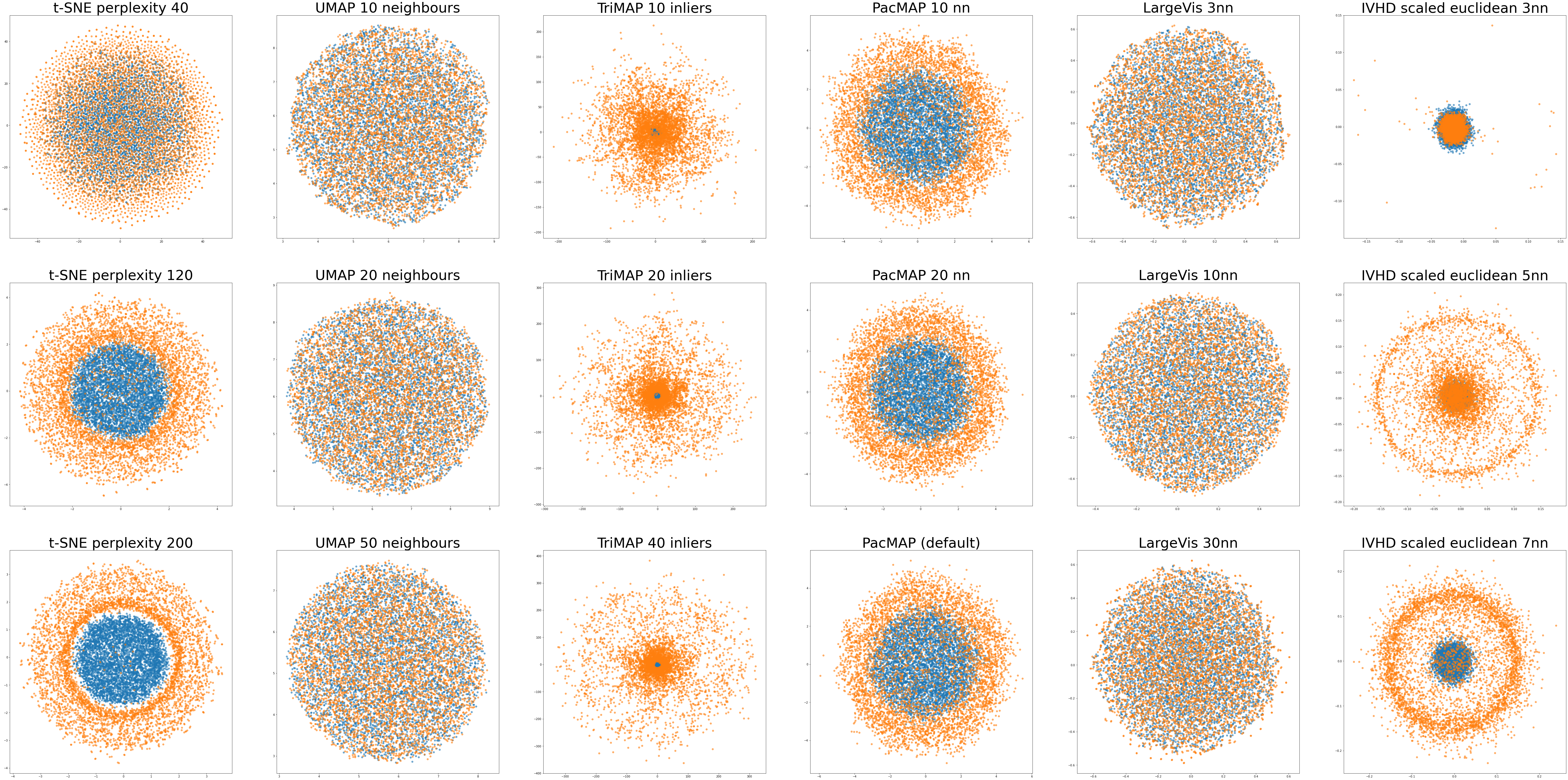}
        \caption{Methods comparison for ball inside a sphere dataset.}
    \label{fig:chapter_5_results_ball_inside_1_sphere}
\end{figure}

Similarly to Figure \ref{fig:chapter_5_results_ball_in_2_spheres}, in Figure \ref{fig:chapter_5_results_ball_inside_1_sphere}, we can see that t-SNE and PaCMAP manage to maintain the global and local structure of the data. What is interesting is that IVHD is the only method that was capable of separate classes by inserting a "space" between the ball and the sphere. Classes in other methods are highly congested. Furthermore, we see that TriMap was unable to split the classes as clearly as in \ref{fig:chapter_5_results_ball_in_2_spheres} (although both datasets are fairly similar). UMAP was again unable to separate classes at all.

\begin{figure}[ht!]
    \centering
        \subfloat[DR quality ($R_{NX}(k)$).]{\includegraphics[width=2.8in]{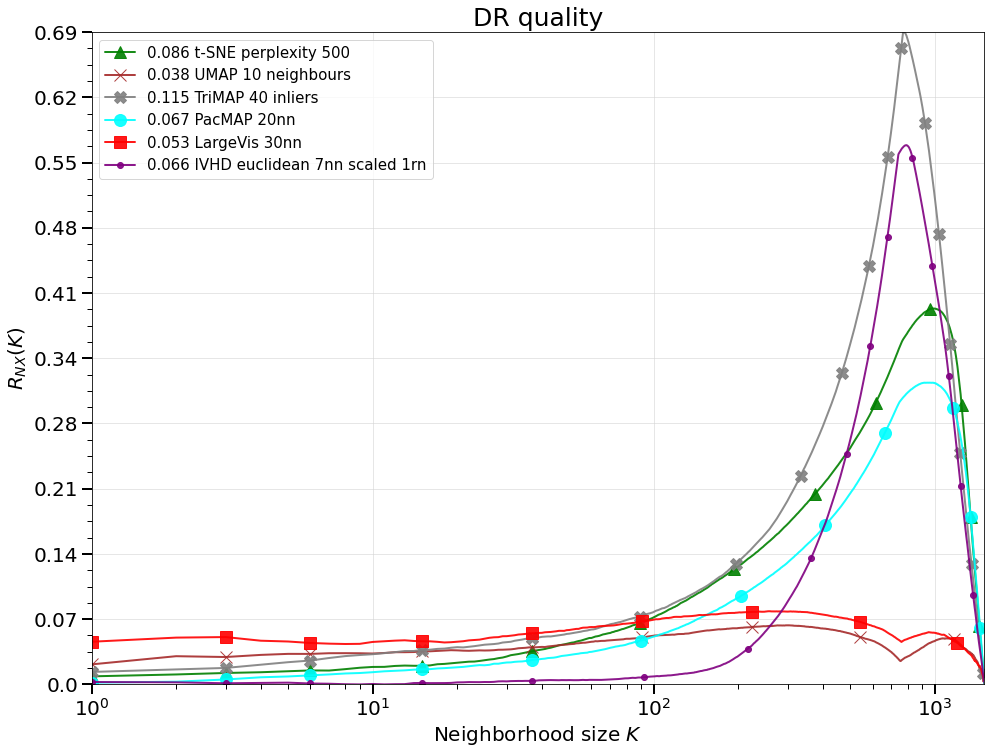}}
        \subfloat[kNN gain ($G_{NN}(k)$).]{\includegraphics[width=2.8in]{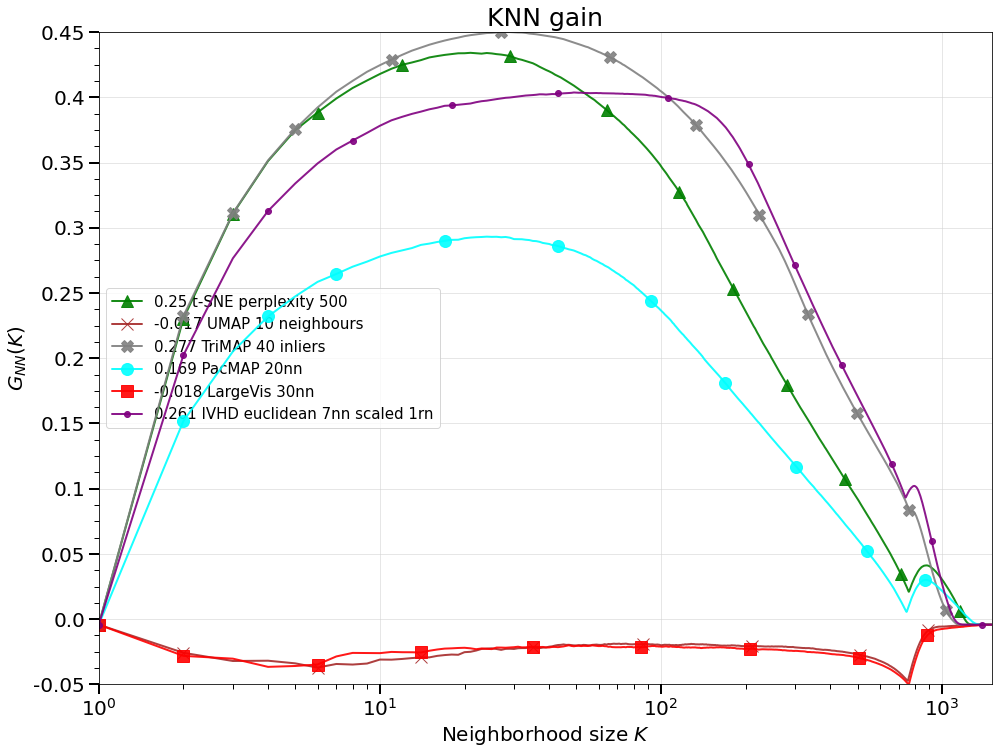}}
        \caption{KNN gain and DR quality obtained for comparison of different DR methods for ball inside 1 sphere dataset.}
    \label{fig:chapter_5_diagrams_ball_inside_1_sphere}
\end{figure}

The KNN gain and the DR quality achieved for this data set (Figure \ref{fig:chapter_5_diagrams_ball_inside_1_sphere}) support the conclusions described above. By far, the worst performer was UMAP. In terms of DR quality, IVHD and TriMap are creating visualizations of the highest quality. 

Using Figs. \ref{fig:chapter_5_results_ball_inside_1_sphere} and \ref{fig:chapter_5_results_ball_in_2_spheres} helps us visualize \textit{crowding phenomenon} observed during dimension reduction by different data embedding methods. Using the "ball inside the sphere" dataset as our framework, we assume that a high-dimensional ball contains two classes of points. Class 1 is inside the ball $B_{2}^{n}(1)$ and class 2 is inside a spherical shell outside class 1. When $\{X_{i}^{N}\}_{i=1} \subseteq \mathbb{R}^2$ is two-dimensional, a direct visualization (Table \ref{tab:artificially_generated_datasets} in Appendix \ref{appendix_datasets}) shows the correct relation between the two classes. When $\{X_{i}^{N}\}_{i=1} \subseteq \mathbb{R}^5$ is five-dimensional and assigned to $\mathbb{R}^2$ using the DE method, its geometrical structure is distorted. Then, we see a severe crowding phenomenon: the second class is pushed towards the center. This is due to the difference in norm concentration between high and low dimensions: a ball in higher dimensions has a volume that grows faster with radius $Vol(B_{2}^{n}(r)) \sim r^{n}$ \cite{fractal_geometry}, where $B_{2}^{n}(r)$ is a ball in $\mathbb{R}^n$ with radius $r$. As a result, high-dimensional points are more likely to be distributed near the surface. When mapped to low dimensions, since there is less room near the surface, the points are pushed towards the center.

\subsection{Mid-scale baseline datasets}

The second part of evaluating the DR properties of the baseline methods is to perform experiments on mid-scale ($N$$<$$150000$) datasets. Analyzes of the datasets that are not covered in the following section are found in the Appendix \ref{appendix_visualizations_of_mid_scale_datasets}. It is worth mentioning that the most efficient method in terms of computational time is IVHD. It is the fastest and requires the least amount of memory. The timings obtained for embedding all datasets using baseline methods are presented in the Appendix \ref{appendix_visualizations_timings_and_methods_parametrization}.

\begin{table}
\caption{Precision values $cf_{nn}$ and trustworthiness $T_{nn}$ for the FMNIST and smallNORB data sets (and various embedding methods).}
\vspace{0.2cm}
\small
\centering
\begin{tabular}{l|ccccc|ccccc|}
\cline{2-11}
& \multicolumn{5}{c|}{\textbf{FMNIST}} & \multicolumn{5}{c|}{\textbf{smallNORB}} \\ \cline{2-11} 
& \multicolumn{1}{c|}{\textbf{$cf_{15}$}}      & \multicolumn{1}{c|}{\textbf{$cf_{50}$}} & \multicolumn{1}{c|}{\textbf{$cf_{100}$}} & \multicolumn{1}{c|}{\textbf{$T_{15}$}} & \textbf{$T_{50}$} & \multicolumn{1}{c|}{\textbf{$cf_{15}$}} & \multicolumn{1}{c|}{\textbf{$cf_{50}$}} & \multicolumn{1}{c|}{\textbf{$cf_{100}$}} & \multicolumn{1}{c|}{\textbf{$T_{15}$}} & \textbf{$T_{50}$} \\ \hline
\multicolumn{1}{|l|}{\textbf{t-SNE}}    & \multicolumn{1}{c|}{\textbf{0.956}} & \multicolumn{1}{c|}{0.950}              & \multicolumn{1}{c|}{0.947}               & \multicolumn{1}{c|}{\textbf{0.987}}    & 0.976             & \multicolumn{1}{c|}{0.951}              & \multicolumn{1}{c|}{0.845}              & \multicolumn{1}{c|}{0.754}               & \multicolumn{1}{c|}{\textbf{0.973}}    & 0.946             \\ \hline
\multicolumn{1}{|l|}{\textbf{UMAP}}     & \multicolumn{1}{c|}{0.951}                   & \multicolumn{1}{c|}{0.951}              & \multicolumn{1}{c|}{0.949}               & \multicolumn{1}{c|}{0.978}             & 0.975             & \multicolumn{1}{c|}{0.911}              & \multicolumn{1}{c|}{0.845}              & \multicolumn{1}{c|}{0.768}               & \multicolumn{1}{c|}{0.966}             & 0.938             \\ \hline
\multicolumn{1}{|l|}{\textbf{TriMap}}   & \multicolumn{1}{c|}{0.944}                   & \multicolumn{1}{c|}{0.944}              & \multicolumn{1}{c|}{0.943}               & \multicolumn{1}{c|}{0.967}             & \textbf{0.977}    & \multicolumn{1}{c|}{0.914}              & \multicolumn{1}{c|}{0.863}              & \multicolumn{1}{c|}{0.813}               & \multicolumn{1}{c|}{0.967}             & \textbf{0.948}    \\ \hline
\multicolumn{1}{|l|}{\textbf{PaCMAP}}   & \multicolumn{1}{c|}{\textbf{0.956}}          & \multicolumn{1}{c|}{\textbf{0.955}}     & \multicolumn{1}{c|}{\textbf{0.955}}      & \multicolumn{1}{c|}{0.974}             & 0.973             & \multicolumn{1}{c|}{\textbf{0.951}}              & \multicolumn{1}{c|}{\textbf{0.918}}     & \multicolumn{1}{c|}{\textbf{0.852}}               & \multicolumn{1}{c|}{0.968}             & 0.939             \\ \hline
\multicolumn{1}{|l|}{\textbf{LargeVis}} & \multicolumn{1}{c|}{0.944}                   & \multicolumn{1}{c|}{0.942}              & \multicolumn{1}{c|}{0.941}               & \multicolumn{1}{c|}{0.961}             & 0.952             & \multicolumn{1}{c|}{0.922}              & \multicolumn{1}{c|}{0.867}              & \multicolumn{1}{c|}{0.795}               & \multicolumn{1}{c|}{0.965}             & 0.913             \\ \hline
\multicolumn{1}{|l|}{\textbf{IVHD}}     & \multicolumn{1}{c|}{0.894}                   & \multicolumn{1}{c|}{0.893}              & \multicolumn{1}{c|}{0.892}               & \multicolumn{1}{c|}{0.938}             & 0.933             & \multicolumn{1}{c|}{0.904}     & \multicolumn{1}{c|}{0.847}     & \multicolumn{1}{c|}{0.764}     & \multicolumn{1}{c|}{0.922}             & 0.906             \\ \hline
\end{tabular}
\end{table}
Poza 
Based on the visualizations and diagrams presented in Figs. \ref{fig:chapter_5_results_fmnist}, \ref{fig:chapter_5_results_smallnorb}, one can make the following observations considering visualized datasets:

\begin{enumerate}
    \item IVHD applied to Fashion-MNIST in Fig. \ref{fig:chapter_5_results_fmnist} forms separate clusters of different, mostly elongated shapes. Moreover, the generated mapping is fuzzy. On the other hand, the \textit{-MAP} methods are much better at creating rounded and clearly separated clusters. Additionally, in t-SNE some classes are mixed and fragmented. In terms of DR quality, TriMap, UMAP, and PaCMAP are achieving the best results. It is also reflected by using the metrics $cf$ and $T$, which have a very stable value regardless of the number of neighbors considered. IVHD starts to exceed t-SNE and LargeVis, when a very large neighborhood is considered ($k$$>$$1000$). 
    \item All methods were more or less capable of preserving the distance ratio between both (high- and low-dimensional) spaces (Fig. \ref{fig:chapter_5_results_fmnist}). Trimap seems to have the most condensed points on the diagonal of the Shepard diagram, which indicates the best preservation of the distance ratio.

    \item For the smallNORB dataset, the TriMap and IVHD methods achieve the best results (Fig. \ref{fig:chapter_5_results_smallnorb}) in terms of both the DR quality and the sustained distance ratio shown in the Shepard diagram (Fig. \ref{fig:chapter_5_results_smallnorb}). In terms of the $cf$ metric, PaCMAP achieves the highest score for all neighborhood sizes. Again, it generates separate clusters of different classes, which are mostly elongated in shape. All the embeddings generate similar in terms of global distribution of classes visualization that contains: 1) separated three classes in LD space (brown, green, orange),  2) separation of other two classes (red, blue) into common sub-space, and 3) mixing of the other four classes.
    
    \begin{figure}[ht!]
        \centering
        \subfloat[Methods comparison for FMNIST dataset.]{\includegraphics[width=\textwidth]{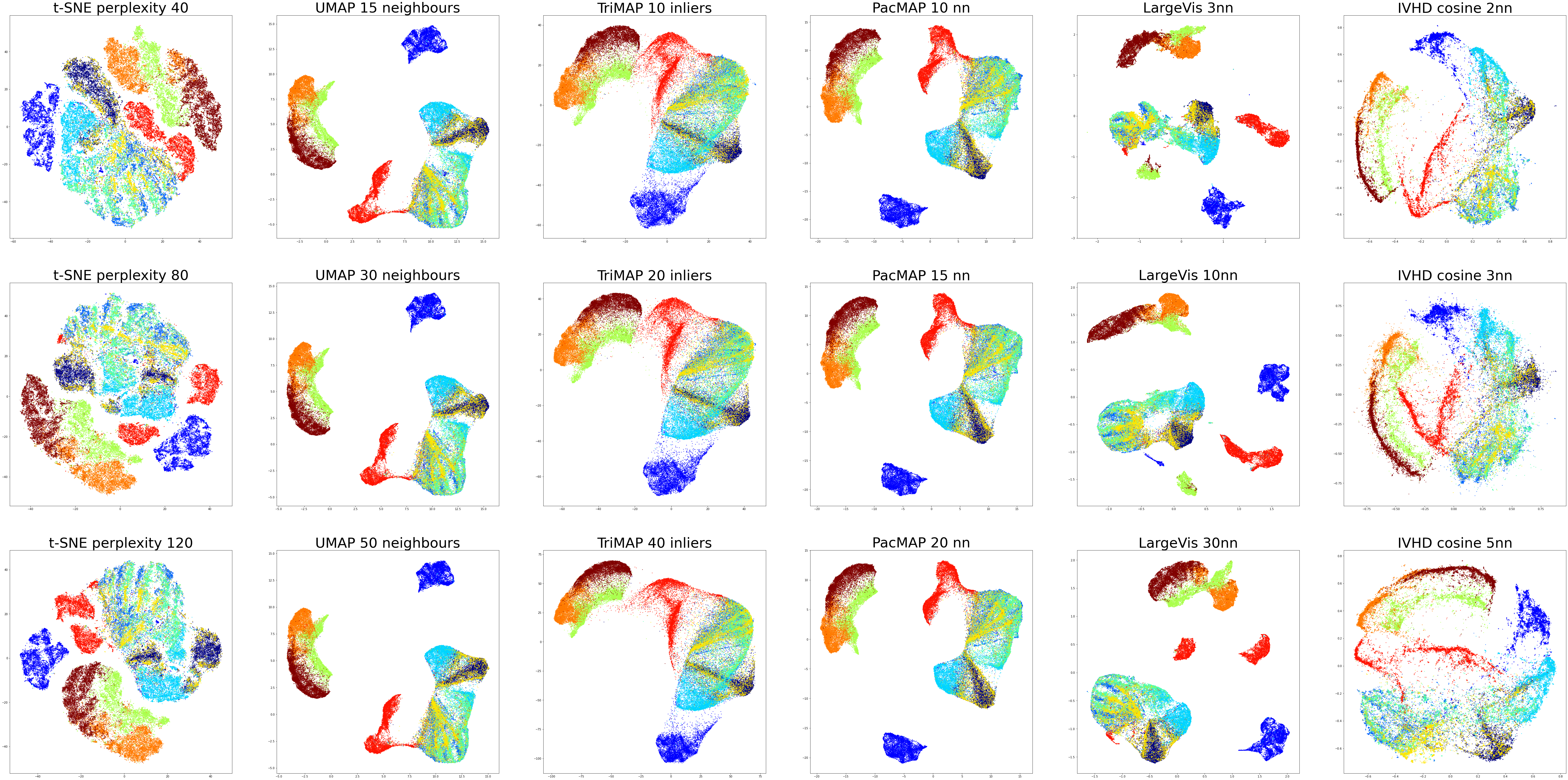}}
        \hfill
        \subfloat[DR quality ($R_{NX}(k)$).]{\includegraphics[width=0.5\textwidth]{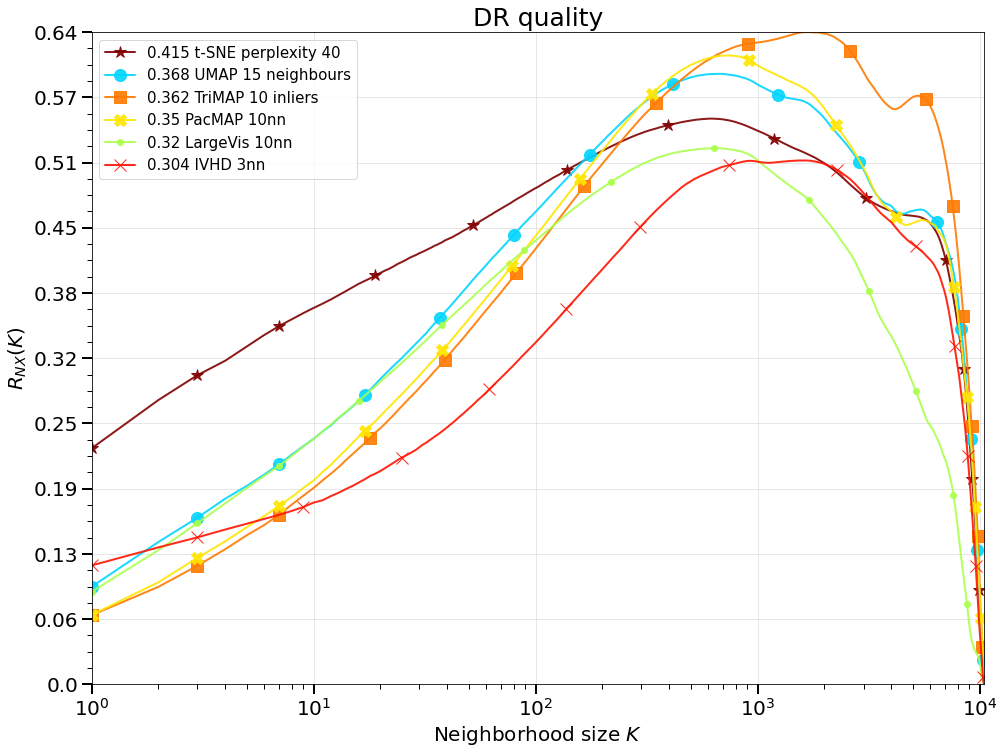}}
        \subfloat[kNN gain ($G_{NN}(k)$).]{\includegraphics[width=0.5\textwidth]{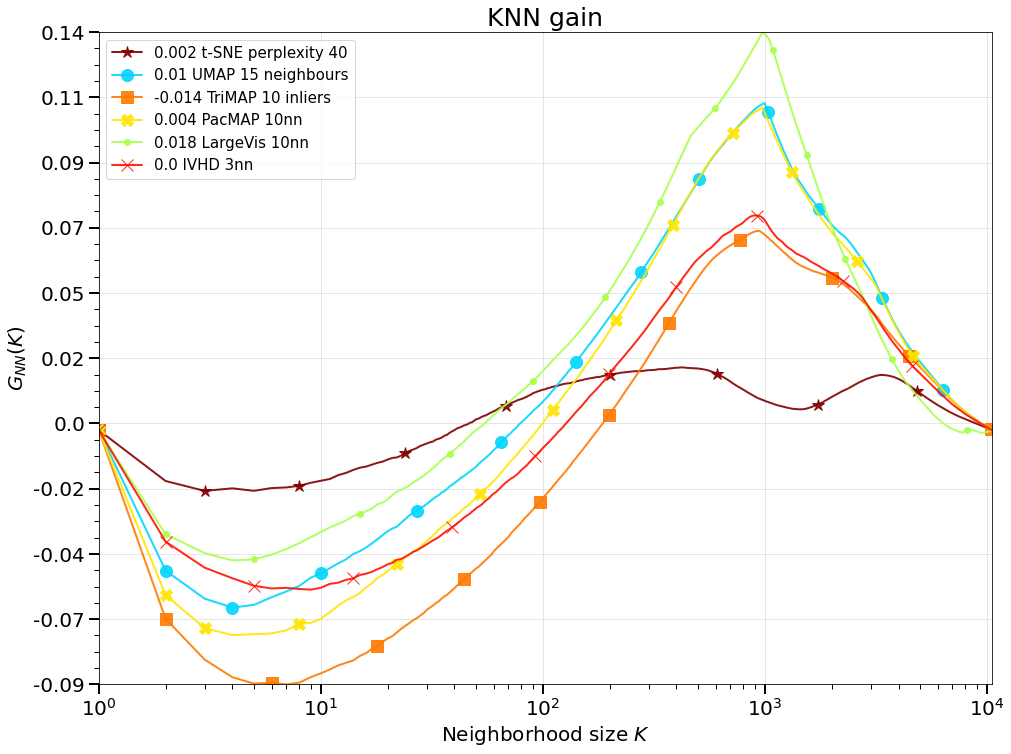}}
        \hfill
        \subfloat[Shepard Diagram ($G_{NN}(k)$).]{\includegraphics[width=\textwidth]{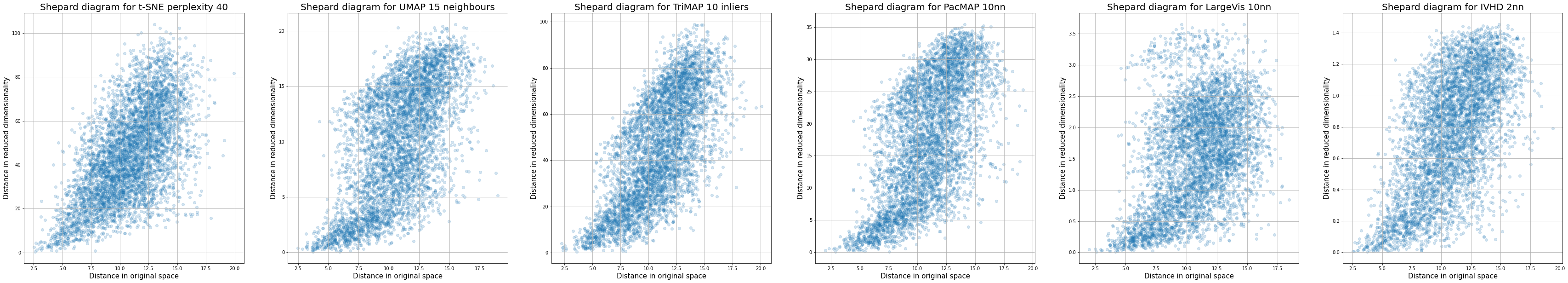}}
        \caption{Visualizations, KNN gain, DR quality and Shepard Diagrams obtained for comparison of different DR methods for FMNIST dataset.}
        \label{fig:chapter_5_results_fmnist}
    \end{figure}
    
    \item All embedding methods split the smallNORB dataset (Fig. \ref{fig:chapter_5_results_smallnorb}) internally, which means that single classes are divided into separate groups. This indicates that all methods managed to find discrepancies between pictures that contain the same class of objects, but with different lighting or rotation.
\end{enumerate}

\begin{figure}[ht!]
    \centering
    \subfloat[Methods comparison for smallNORB dataset.]{\includegraphics[width=\textwidth]{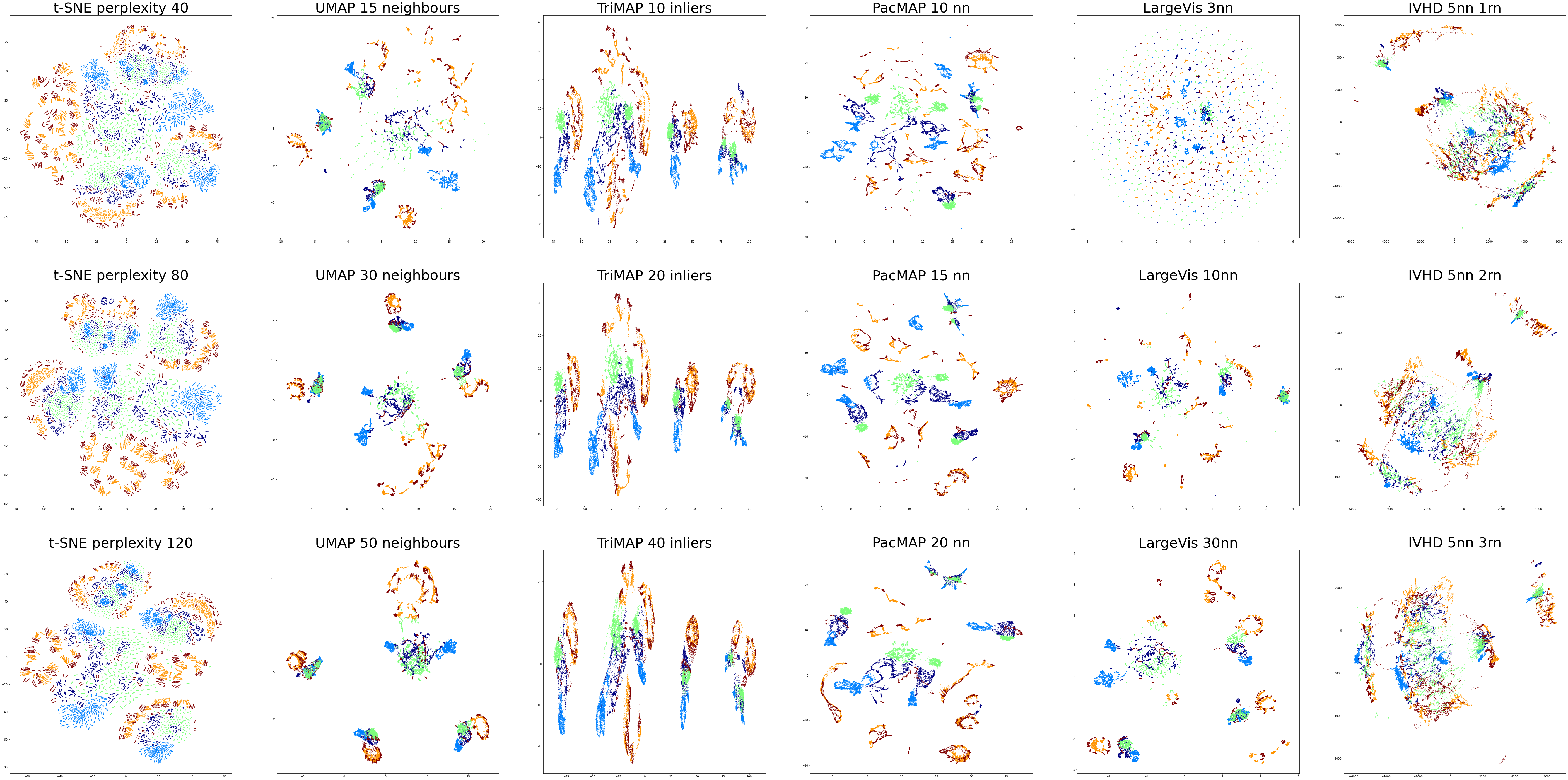}}
    \hfill
    \subfloat[DR quality ($R_{NX}(k)$).]{\includegraphics[width=0.5\textwidth]{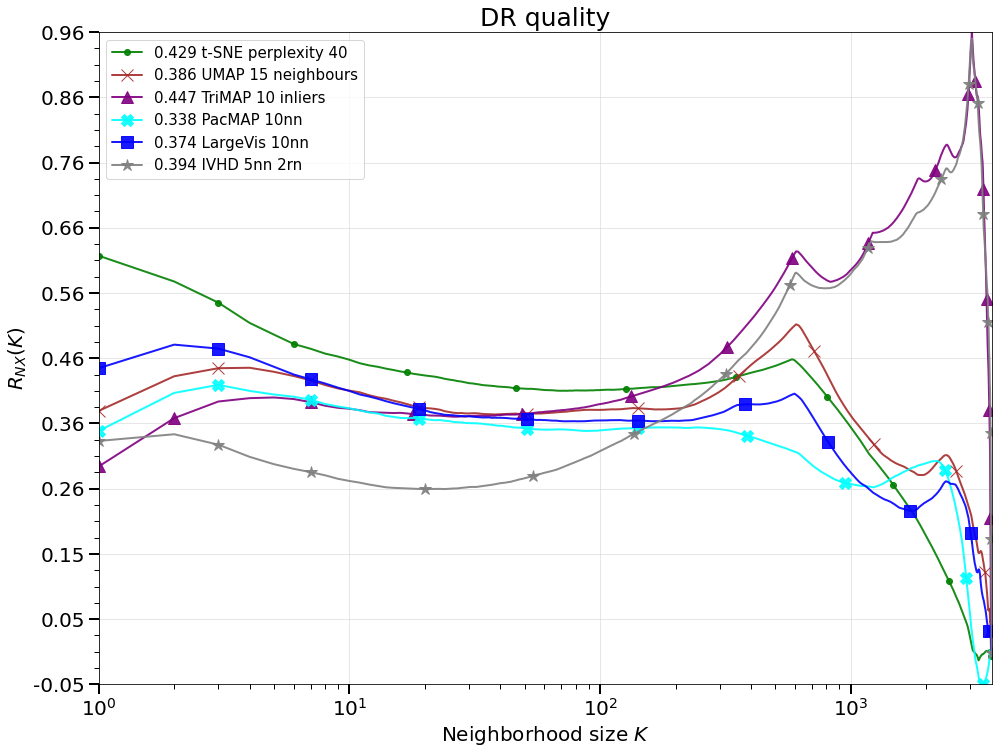}}
    \subfloat[kNN gain ($G_{NN}(k)$).]{\includegraphics[width=0.5\textwidth]{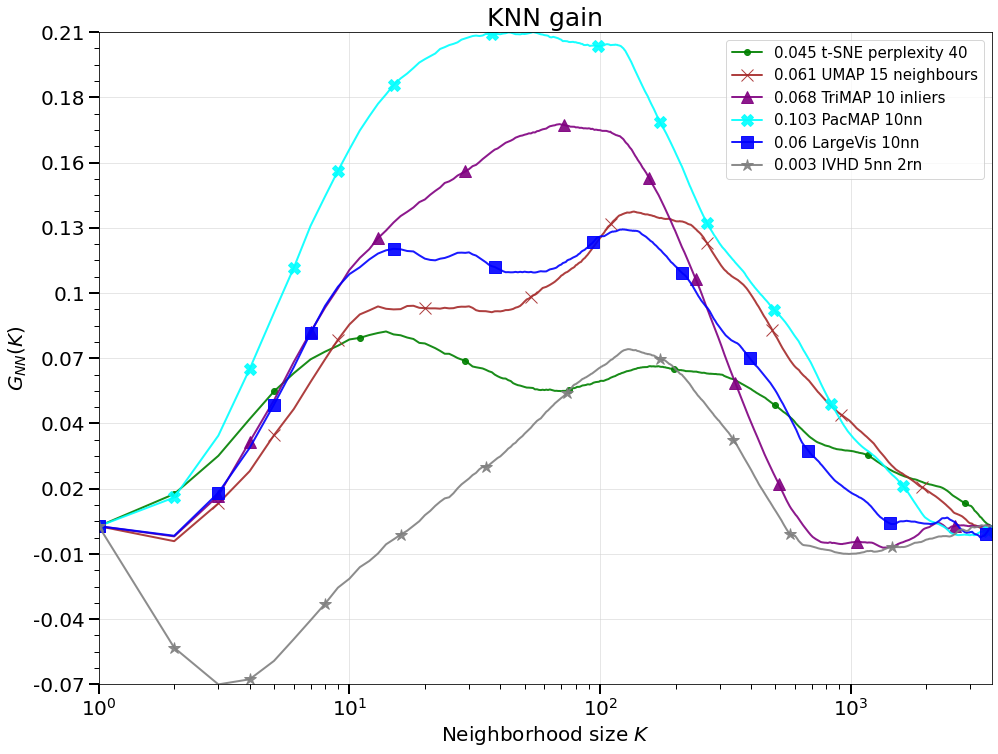}}
    \hfill
    \subfloat[Shepard Diagram ($G_{NN}(k)$).]{\includegraphics[width=\textwidth]{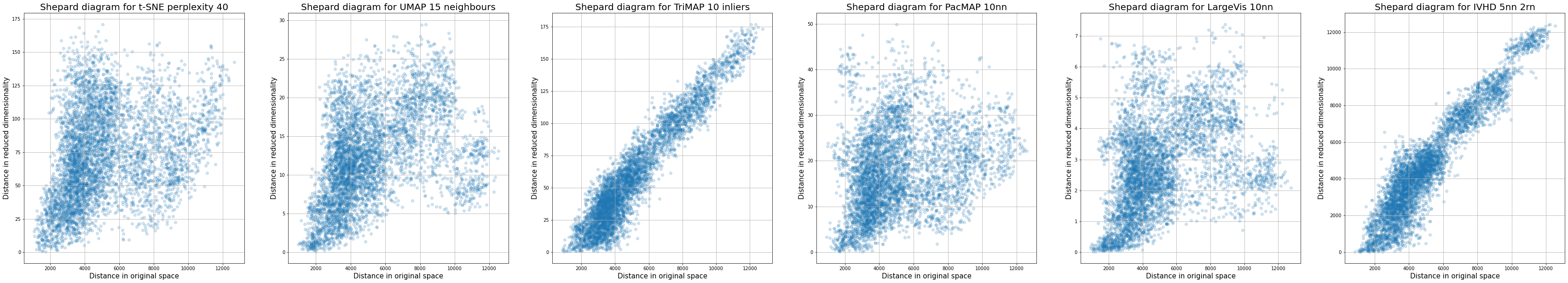}}
    \caption{Visualizations, KNN gain, DR quality and Shepard Diagrams obtained for comparison of different DR methods for smallNORB  dataset.}
    \label{fig:chapter_5_results_smallnorb}
\end{figure}

As shown above, the quality and fidelity of the embeddings strongly depend on the structure of a specific dataset, user expectations, and their data visualization requirements. In general, the local structure of the $source$ data is better reconstructed by the baseline algorithms. This is not a surprise because the IVHD algorithm does not use the correct ordering of \textit{k}NN neighbors and the original distances between a sample $y_i$ and its \textit{k}NN neighbors.

The results obtained for the large-scale baseline datasets ($M$$>$$10^{6}$) were created using CUDA implementations (\textit{IVHD-CUDA}) and are described in Chapter \ref{chapter:ivhd-cuda}. This was due to much faster computation times and the fact that the reference methods did not generate visualizations for such large sets in a relatively reasonable time (<12h). Furthermore, the t-SNE, TriMAP, and PaCMAP CPU implementations generated errors (probably due to insufficient memory resources). It is worth mentioning that UMAP is the only method that contains benchmarks for large-scale datasets (although the resources needed for the calculations are much higher than for CUDA algorithms).

\chapter{GPU-Embedding of kNN-Graph Representing Large and High-Dimensional Data}
\label{chapter:ivhd-cuda}

Despite the fact that there are many algorithms for the visualization of high-dimensional data and that this topic has been extensively studied for years, there are only a few implementations of modern data embedding algorithms in the GPU/CUDA environment \cite{bhsne-gpu,atSNE,dzwinel2014}. This was the root cause of the creation of IVHD-CUDA \cite{minch2020}, which was compared with publicly available CUDA implementations, which generated the best embeddings of large datasets: BH-SNE-CUDA \cite{bhsne-gpu} and Anchor-t-SNE \cite{atSNE} (At-SNE). All algorithms consist of two  stages:  (1)  generate  of  a  weighed kNN-graph; (2) run a proper embedding procedure which is based on: definition of a loss function and its minimization.

\section{\textit{k}NN graph generation}
\textit{k}NN-graph approximates a $n$D non-Cartesian manifold immersed in $\mathbb{R}^N$ sampled by the feature vectors $y_i$. Here, we consider the $k$NN-graph construction procedure shipped by the FAISS library \cite{faiss}. Its authors claim that it is currently the fastest available \textit{k}NN search algorithm implemented on GPU. The FAISS \textit{k}NN-search procedure merges a very efficient and well-parallelized exact \textit{k}NN algorithm and indexing structures that allow an approximate search. To achieve a high-efficiency search with good accuracy, we employ the IVFADC \cite{IVFADC} indexing structure. It uses two levels of quantization combined with vector encoding for compressing high-dimensional vectors. The main idea behind the use of indexes is to divide the input space and divide all input samples into a number of clusters represented by their centroids. To handle a query, the algorithm compares it with the centroid centers, picks the centroid that is the most similar to the query vector, and performs an exact search within the set of samples belonging to this centroid. In addition to the efficient \textit{k}NN algorithm, its CUDA implementation is extremely well optimized \cite{faiss}. We use the same FAISS \textit{k}NN-graph generation procedure in both the baseline and IVHD algorithms.

\section{BH-SNE-CUDA, At-SNE, IVHD-CUDA}

\subsubsection{BH-SNE-CUDA}

BH-SNE (Barnes-Hut t-SNE \cite{bhsne}) is an approximation of the t-SNE method that can reduce the computational complexity of DE from $O(M^2)$ to $O(Mlog M)$ using the Barnes-Hut approximation (described in Section 2.4.1) but at the cost of increasing algorithmic complexity and, consequently, decreasing parallelization efficiency \cite{bhsne}. Its GPU version, the BH-SNE-CUDA algorithm, does not introduce any changes to the BH-SNE \cite{bhsne} algorithm and just matches the instructions and data flows to the GPU architecture \cite{bhsne-gpu}.

\subsubsection{AtSNE-CUDA}

The AtSNE-CUDA algorithm (Anchor-t-SNE \cite{atSNE}) was created to deal with issues related to t-SNE, such as sensitivity to initial conditions and inadequate reconstruction of the global data structure. The authors of the AtSNE-CUDA method claim that their method is 50\% faster than BH-SNE-CUDA and has lower memory requirements. As shown in \cite{atSNE}, it still generates good quality embeddings, although some $k$NN quality scores are better for the classic t-SNE or LargeVis algorithms \cite{largevis}. \\

\textbf{Loss function:} Similar to all t-SNE clones, AtSNE minimizes the regularized KL divergence. Information about the local structure of the data can be obtained from the set of approximated $k$NN neighbors of each $y_i$. Meanwhile, to reconstruct the global structure of the data, the $x_{i}$ points receive 'pulling forces' from the so-called \textit{anchor points} generated from the original data. In this way, \textit{anchor points} are responsible for maintaining the mutual positions and shape of classes in $X$. Consequently, the hierarchical embedding \cite{atSNE} optimizes both the positions of the \textit{anchor points} and regular data samples.

Let $A$ represent the set of \textit{anchor points} in the high-dimensional space and $B$ its low-dimensional embedding. The probabilities $P(Y)$ and $Q(X)$ denote the high- and low-dimensional distributions of \textit{anchor points}, respectively. To preserve both global and local information, the following loss function is minimized:

\begin{equation}
    \label{eq:eq3}
    \begin{array}{c}
    E(.) = \sum_{i} KL(P(Y) || Q(X)) + \sum_{i} KL(P(A)||Q(B)) + \sum_{i} || b_i - \frac{\sum_{y_{k} \in C_{b_{i}}}y_{k}}{|C_{b_{i}}|} ||,
    b_{i} \in C_{b_{i}}
  \end{array}
\end{equation}

\noindent where $C_{b_i}$ denotes the set of points whose cluster center of the K-means is $b_{i}$. The last term is a regularization term explained in detail in \cite{atSNE}. \\

\textbf{Optimization:} The AtSNE-CUDA algorithm preserves global information by minimizing the term $KL(P(A)||Q(B))$ in Eq.\ref{eq:eq3}, while the first term of the loss function is responsible for preserving the local structure of the data. The reason is that the gradient of the loss function is more complicated than that defined by Eq. \ref{eq:ivhd_force_1}, AtSNE-CUDA uses the Hierarchical Optimization Algorithm \cite{atSNE}. This algorithm consists of two optimization layers: global and local. Optimization proceeds in a top-down manner. The main idea is to optimize the two-layer layout alternately: 1) fix the layout of the ordinary points and optimize the layout of the anchor points; 2) fix the anchor point layout and optimize the layout of ordinary points.

\subsubsection{IVHD-CUDA}

In IVHD-CUDA, the force-directed optimization scheme is used, similar in spirit to the synchronous $momentum$ and $Nesterov$ methods, described in detail in \cite{ivhd3}. Let $connection$ mean two "particles" $x_i$ and $x_j$ (the nearest or random neighbors) in a 2D embedded space $X$ joined by an edge. The main IVHD-CUDA implementation loop consists of the following steps: 1) calculating "forces" for each $connection$ where the "force" is proportional to the gradient dependent on the two "particles'" positions and their velocities; 2) summing up the forces and 3) updating positions and velocities for all particles simultaneously. The simplified pseudocode is presented below as Algorithm \ref{alg:alg1}.

\begin{algorithm}
 \label{alg:alg1}
 \SetAlgoLined
 \SetKwInOut{Input}{Input}
 
 \Input{\textit{k}NN-graph precalculated with FAISS library.}
 \vspace{0.1cm}
 initialization\;
 \While{running}
 {
  allocate subsets of force components to threads;
 
  \For{$thread\ in\ threads$}{
        calculate force components for each connection\;
    }
    join threads\;
    
    allocate subsets of samples to threads\;

    \For{$thread\ in\ threads$}{
        sum force/gradient components\;
        calculate new velocities\;
        update particle positions\;
    }
}
\caption{Simplified IVHD-CUDA scheme.}
\end{algorithm}

Because the steps are executed sequentially, this allows one to avoid the race condition between CUDA threads. Each connection receives two different memory addresses where the calculated values can be stored, one $positive$ and one $negative$. If a particle $i$ attracts particle $j$ ($positive$ value), then at the same time a particle $j$ attracts particle $i$ ($negative$ value). In the first step, each thread calculates a subset of the force / gradient components between the particles. During the same time, they execute the same instructions, and thus no branching is required. However, this approach has a minor bottleneck. To save the result for each $connection$, a given thread must execute instructions from $write$ to completely different memory addresses that represent two samples. Meanwhile, minimizing the number of $read/write$ operations is always beneficial, as they are very time-consuming.

In the second step, each thread is assigned to a set of particles to process. As precautions against processing particles that interact with different numbers of particles (which might cause branch divergence), we sort the particles before the first iteration by considering the number of force components to calculate. After sorting, it is guaranteed that all threads in a warp will process samples with the same number of $connections$. The percentage share of each component of the DE procedure (without $k$NN graph generation), e.g., for 70.000 feature vectors with approximately three neighbors each ($nn$=2 and $rn$=1), is as follows: (1) calculation of the force component (78.97\%), (2) update of positions (20.6\%), (3) initialization (0.43\%). In the IVHD-CUDA implementation, the global register and constant memory are used. While $connections$, the force components and positions are stored in global memory, the algorithm parameters are stored in constant memory. A register memory is used for auxiliary and temporary calculations.

\subsubsection{Data embedding comparison}

In this section, we briefly compare IVHD-CUDA with the most efficient and robust publicly available GPU-implemented data embedding codes: BH-SNE-CUDA and AtSNE-CUDA. The most important parameters of the methods are collected in Table \ref{tab:parameters}. For the BH-SNE-CUDA and AtSNE-CUDA  methods, we use default parameters (appx. 15 parameters) proposed in \cite{atSNE,bhsne} and submitted to the GitHub repository with the respective GPU codes \cite{atsne-git,bhsne-gpu}. IVHD parameters are also selected by default, for example, $rn=1$ while $nn$ is automatically fitted as a minimal value producing the largest connected component, which approximates (with a given high accuracy) the $k$NN-graph. The value of $c\in[0.01,0.1]$ from Eq. \ref{eq:ivhd_cost_function} is the only parameter that can be interactively adjusted. 

All datasets used in this section are described in Appendix \ref{appendix_datasets}. There you can find the exact dimensionality of the dataset and a brief description of what is in the dataset.

\begin{figure}
    \begin{center}
    \subfloat[MNIST.]{%
      \includegraphics[clip,width=0.33\columnwidth]{pics/gpu-embedding-of-kNN-graphs/ivhd-cuda/mnist_vis.png}%
      \label{pic:mnist_vis}
    }
    \subfloat[Small NORB.]{%
      \includegraphics[clip,width=0.33\columnwidth]{pics/gpu-embedding-of-kNN-graphs/ivhd-cuda/small_norb_vis.png}%
      \label{pic:small_norb_vis}
    }
    \subfloat[Fashion MNIST.]{%
      \includegraphics[clip,width=0.33\columnwidth]{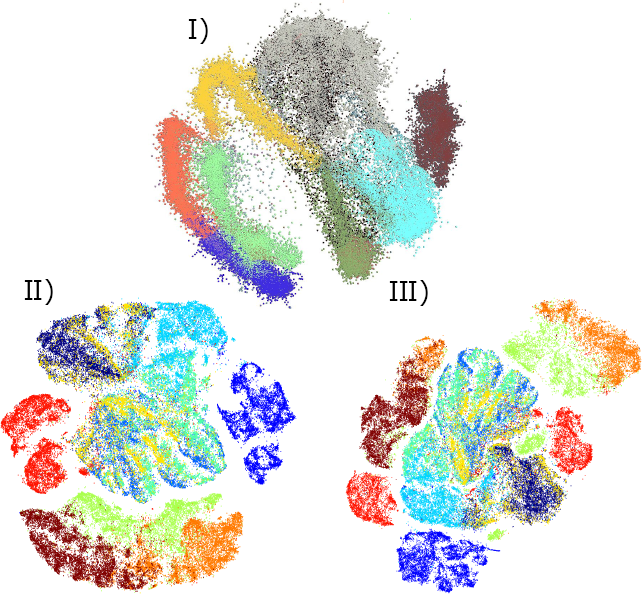}%
      \label{pic:fmnist_vis}
    }
    \caption{Visualization of datasets using: I) IVHD-CUDA, II) BH-SNE-CUDA, III) AtSNE.}
    \label{pic:all1}
    \end{center}
\end{figure}

Based on the visualizations presented in Figs. \ref{pic:all1}, \ref{pic:all2}, one can make the following observations:

\begin{enumerate}
    \item In Fig. \ref{pic:mnist_vis}, IVHD clearly reflects the global structure of MNIST classes and their separation. However, due to the "crowding effect", the classes become very dense in the centers and sparse between them. On the other hand, the results of graph visualization presented in \cite{dzwinel2017ivga}, demonstrate that the "crowding effect" can be controlled by tuning IVHD-CUDA parameters. Consequently, even the fine-grained neighborhood can be preserved with astonishing high precision (see \cite{dzwinel2017}).
    
    \item For the smallNORB dataset (see Fig.\ref{pic:small_norb_vis}), IVHD-CUDA was able to visualize three large clusters clearly separable with a fine-grained data structure. For BH-SNE-CUDA and AtSNE-CUDA the quality of embeddings is much worse and more fragmented. Changes in the parameter values of the baseline algorithms ($perplexity$) do not improve the quality of the visualization. 
    \begin{figure}[ht]
        \subfloat[YAHOO]{%
          \includegraphics[clip,width=0.5\columnwidth]{pics/gpu-embedding-of-kNN-graphs/ivhd-cuda/yahoo_vis.png}%
          \label{pic:yahoo_vis}
        }
        \subfloat[RCV-Reuters]{%
          \includegraphics[clip,width=0.5\columnwidth]{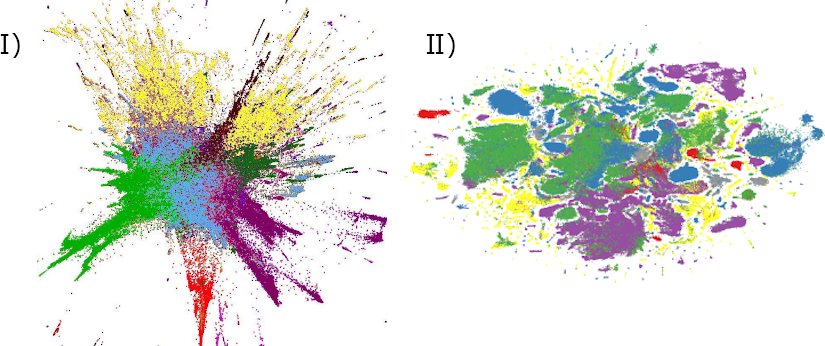}%
          \label{pic:reuters_vis}
        }
        \caption{Visualization of datasets using: I) IVHD-CUDA, II) AtSNE-CUDA.}
        \label{pic:all2}
    \end{figure}

    \item IVHD applied to Fashion-MNIST (see Fig.\ref{pic:fmnist_vis}) creates separate clusters of various shapes, mainly elongated. Moreover, the generated
    mapping is fuzzy. It is clearly reflected in the decreasing values of $cf_{nn}$, particularly for $nn$ = 100. However, both AtSNE-CUDA and BH-SNE-CUDA are much better at creating oblate and clearly separated clusters. As shown in Table \ref{tab:accuracy_for_methods}, the values $cf_{nn}$ are more stable than those for the IVHD method. Nevertheless, unlike the IVHD result, some classes reproduced by the baseline methods are mixed and fragmented.
    
    \item Similar conclusions can be drawn by visualizing the RCV and YAHOO 2D embeddings (see Figs.\ref{pic:yahoo_vis}, \ref{pic:reuters_vis}, respectively). IVHD-CUDA generates more fuzzy output than AtSNE-CUDA, but the samples from the same class are closer together than those generated by the baseline method. Meanwhile, AtSNE-CUDA is capable of separate clearly visible but fragmented clusters. BH-SNE-CUDA was too slow to visualize these large datasets in a reasonable time budget.
\end{enumerate}

As shown in Table \ref{tab:accuracy_for_methods}, IVHD-CUDA is the fastest method for the chosen baseline datasets. Unfortunately, we were not able to control the run-times of AtSNE-CUDA and BH-SNE-CUDA implementations; thus comparisons for various time budgets are not possible. As shown above, the quality and fidelity of the embeddings strongly depend on the structure of a specific dataset, user expectations, and their data visualization requirements. In general, the local structure of the $source$ data is better reconstructed by the two SNE baseline algorithms. This is not a surprise because the IVHD algorithm does not use the correct ordering of \textit{k}NN neighbors and original distances between a sample $y_i$ and its \textit{k}NN neighbors. Moreover, the value of $k$ ($nn$ in this dissertation) is extremely small compared to AtSNE-CUDA and BH-SNE. However, despite this drastic approximation, IVHD properly preserves the class structure and its relative locations. For fine-grained structures of classes (such as in the Small NORB dataset), IVHD outperforms its competitors in both efficiency and, slightly, in the $cf$ accuracy. Moreover, unlike AtSNE-CUDA and BH-SNE-CUDA, IVHD can visualize separated and not fragmented classes. The same can be observed for the highly imbalanced RCV dataset. For MNIST and Fashion MNIST datasets, the dominance of t-SNE-based methods in reproducing the local "microscopic" data structure is visible, mainly due to the strong "crowding effect" seen for IVHD embeddings. However, the "macroscopic" view of the classes is more convincing for IVHD visualization, although due to the high $perplexity$ parameter ($perplexity=50$), the baseline algorithms are also tuned for better coarse-grained visualization.

\begin{table}
\small
\caption{The precision values $cf_{nn}$ for various data sets and embedding methods. The timings show the overall embedding time ($time$), $k$NN-graph generation time ($time_{gg}$), net embedding time ($time_{emb}$). The results are averages over 5-10 simulations.}
\begin{center}
\begin{tabular}{llllllll}
\hline
\multicolumn{1}{|l|}{Dataset} & 
\multicolumn{1}{l|}{Algorithm} & 
\multicolumn{1}{l|}{$time$ {[}s{]}} &
\multicolumn{1}{l|}{$time_{gg}$ {[}s{]}} &
\multicolumn{1}{l|}{$time_{emb}$ {[}s{]}} & 
\multicolumn{1}{l|}{$cf_{2}$} & 
\multicolumn{1}{l|}{$cf_{10}$} & 
\multicolumn{1}{l|}{$cf_{100}$} \\ \hline

\multicolumn{1}{|l|}{MNIST}         &
\multicolumn{1}{l|}{\begin{tabular}[c]{@{}l@{}}BH-SNE-CUDA\\ AtSNE\\ IVHD-CUDA\end{tabular}} &
\multicolumn{1}{l|}{\begin{tabular}[c]{@{}l@{}}32.588\\ 15.980\\ \textbf{7.326}\end{tabular}} &
\multicolumn{1}{c|}{\begin{tabular}[c]{@{}c@{}} \textbf{5.813}\end{tabular}} &
\multicolumn{1}{l|}{\begin{tabular}[c]{@{}l@{}}26.775\\ 10.167\\ \textbf{1.261}\end{tabular}} &
\multicolumn{1}{l|}{\begin{tabular}[c]{@{}l@{}}0.94\\ 0.944\\ \textbf{0.946}\end{tabular}} & \multicolumn{1}{l|}{\begin{tabular}[c]{@{}l@{}}0.938\\ \textbf{0.943}\\ 0.936\end{tabular}} & \multicolumn{1}{l|}{\begin{tabular}[c]{@{}l@{}}0.933\\ \textbf{0.938}\\ 0.924\end{tabular}} \\ \hline

\multicolumn{1}{|l|}{FMNIST} & \multicolumn{1}{l|}{\begin{tabular}[c]{@{}l@{}}BH-SNE-CUDA\\ AtSNE\\ IVHD-CUDA\end{tabular}} &
\multicolumn{1}{l|}{\begin{tabular}[c]{@{}l@{}}32.913\\ 17.453\\ \textbf{8.177}\end{tabular}} &
\multicolumn{1}{c|}{\begin{tabular}[c]{@{}l@{}}\textbf{6.734}\end{tabular}} &
\multicolumn{1}{l|}{\begin{tabular}[c]{@{}l@{}}26.179\\ 10.719\\ \textbf{1.443}\end{tabular}} & \multicolumn{1}{l|}{\begin{tabular}[c]{@{}l@{}}0.757\\ 0.76\\ \textbf{0.767}\end{tabular}} & \multicolumn{1}{l|}{\begin{tabular}[c]{@{}l@{}}0.755\\ \textbf{0.757}\\ 0.726\end{tabular}} & \multicolumn{1}{l|}{\begin{tabular}[c]{@{}l@{}}\textbf{0.738}\\ 0.737\\ 0.670\end{tabular}} \\ \hline

\multicolumn{1}{|l|}{Small NORB} & 
\multicolumn{1}{l|}{\begin{tabular}[c]{@{}l@{}}BH-SNE-CUDA\\ AtSNE\\ IVHD-CUDA\end{tabular}} &
\multicolumn{1}{l|}{\begin{tabular}[c]{@{}l@{}} 38.673 \\ 20.521\\ \textbf{16.151}\end{tabular}} &
\multicolumn{1}{c|}{\begin{tabular}[c]{@{}l@{}} \textbf{15.517}\end{tabular}} &
\multicolumn{1}{l|}{\begin{tabular}[c]{@{}l@{}} 23.161 \\ 5.009\\ \textbf{0.634}\end{tabular}} & \multicolumn{1}{l|}{\begin{tabular}[c]{@{}l@{}}0.944\\ \textbf{0.97}\\ 0.936\end{tabular}} & \multicolumn{1}{l|}{\begin{tabular}[c]{@{}l@{}}0.919\\ \textbf{0.94}\\ 0.921\end{tabular}} & \multicolumn{1}{l|}{\begin{tabular}[c]{@{}l@{}}0.745\\ 0.73\\ \textbf{0.828}\end{tabular}} \\ \hline

\multicolumn{1}{|l|}{RCV-Reuters} & 
\multicolumn{1}{l|}{\begin{tabular}[c]{@{}l@{}}BH-SNE-CUDA\\ AtSNE\\ IVHD-CUDA\end{tabular}} &
\multicolumn{1}{l|}{\begin{tabular}[c]{@{}l@{}}-\\ 220.39\\ \textbf{60.72}\end{tabular}} &
\multicolumn{1}{c|}{\begin{tabular}[c]{@{}l@{}}\textbf{45.302}\end{tabular}} &
\multicolumn{1}{l|}{\begin{tabular}[c]{@{}l@{}}-\\ 175.088\\ \textbf{15.418}\end{tabular}} & \multicolumn{1}{l|}{\begin{tabular}[c]{@{}l@{}}-\\ 0.82\\ \textbf{0.835}\end{tabular}} & 
\multicolumn{1}{l|}{\begin{tabular}[c]{@{}l@{}}-\\ 0.82\\ \textbf{0.828}\end{tabular}} & 
\multicolumn{1}{l|}{\begin{tabular}[c]{@{}l@{}}-\\ \textbf{0.818}\\ 0.803\end{tabular}} \\ \hline

\multicolumn{1}{|l|}{YAHOO} & 
\multicolumn{1}{l|}{\begin{tabular}[c]{@{}l@{}}BH-SNE-CUDA\\ AtSNE\\ IVHD-CUDA\end{tabular}} &
\multicolumn{1}{l|}{\begin{tabular}[c]{@{}l@{}}-\\ 628.63\\ \textbf{70.12}\end{tabular}} &
\multicolumn{1}{c|}{\begin{tabular}[c]{@{}l@{}}\textbf{52.930}\end{tabular}} &
\multicolumn{1}{l|}{\begin{tabular}[c]{@{}l@{}}-\\ 575.7\\ \textbf{18.930}\end{tabular}} & \multicolumn{1}{l|}{\begin{tabular}[c]{@{}l@{}}-\\ \textbf{0.686}\\ 0.668\end{tabular}} & \multicolumn{1}{l|}{\begin{tabular}[c]{@{}l@{}}-\\ \textbf{0.686}\\ 0.662\end{tabular}} & \multicolumn{1}{l|}{\begin{tabular}[c]{@{}l@{}}-\\ \textbf{0.686}\\ 0.653\end{tabular}} \\ \hline

\multicolumn{1}{|l|}{Amazon 20M} & 
\multicolumn{1}{l|}{\begin{tabular}[c]{@{}l@{}}BH-SNE-CUDA\\ AtSNE\\ IVHD-CUDA\end{tabular}} &
\multicolumn{1}{l|}{\begin{tabular}[c]{@{}l@{}}-\\ -\\ \textbf{12218}\end{tabular}} &
\multicolumn{1}{c|}{\begin{tabular}[c]{@{}l@{}}\textbf{11060}\end{tabular}} &
\multicolumn{1}{l|}{\begin{tabular}[c]{@{}l@{}}-\\ -\\ \textbf{1158}\end{tabular}} & \multicolumn{1}{l|}{\begin{tabular}[c]{@{}l@{}}-\\ -\\ \textbf{0.524}\end{tabular}} & \multicolumn{1}{l|}{\begin{tabular}[c]{@{}l@{}}-\\ -\\ \textbf{0.522}\end{tabular}} & \multicolumn{1}{l|}{\begin{tabular}[c]{@{}l@{}}-\\ -\\ \textbf{0.518}\end{tabular}} \\ \hline

\end{tabular}
\label{tab:accuracy_for_methods}
\end{center}
\end{table}

The main advantage of IVHD is that after storing the \textit{k}NN-graph in the disc cache and neglecting the computational time required for its generation, the embedding of IVHD can be more than one order of magnitude faster than the baseline methods. This allows for a very detailed interactive exploration of multiscale data structure by employing broad spectrum of parameter values and various versions of the stress function without the need for the \textit{k}NN-graph recalculation.
 
\begin{figure}[ht]
    \centering
    \subfloat[1M points.]{%
      \includegraphics[clip,width=0.33\columnwidth]{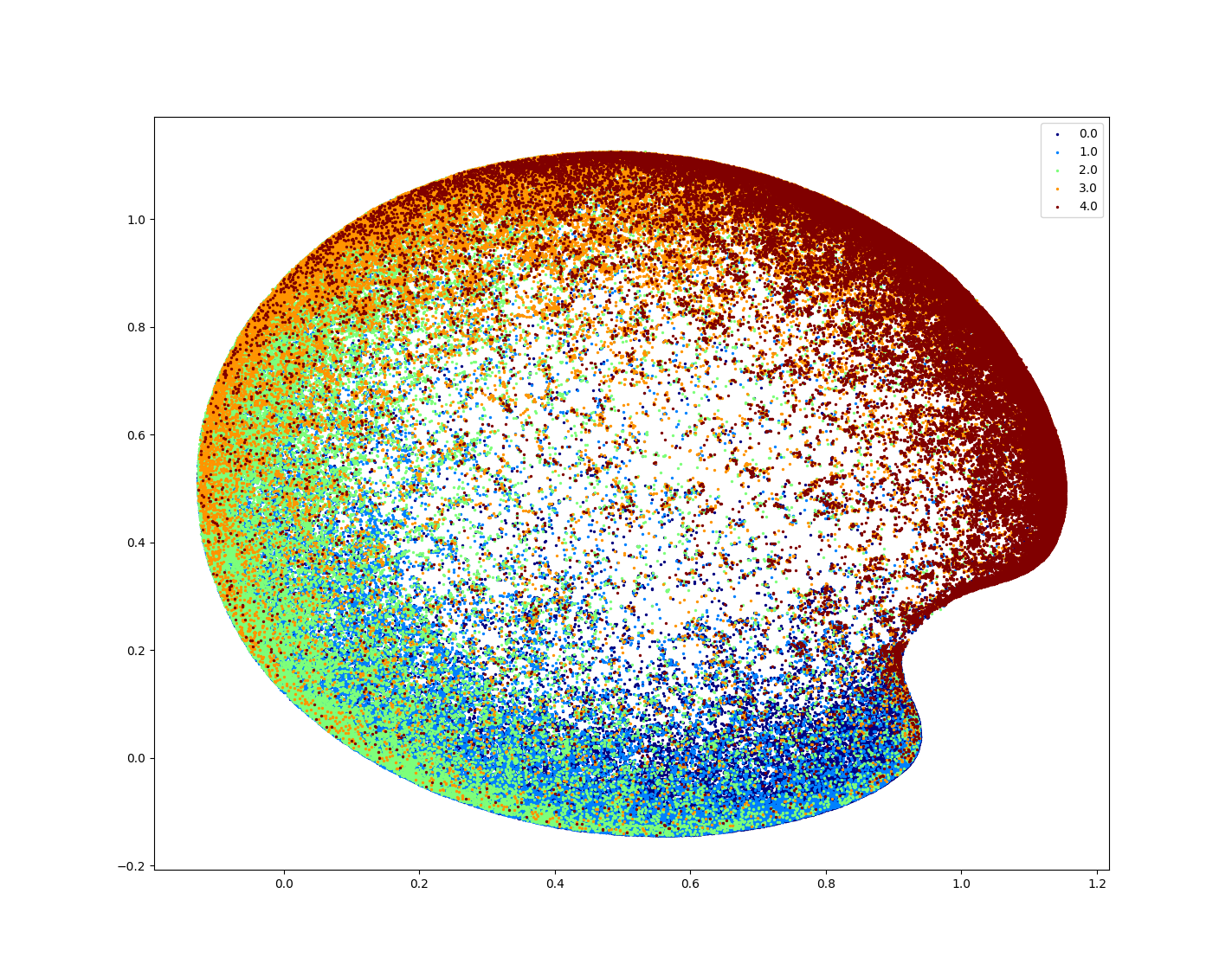}%
      \label{pic:amazon_1M}
    }
    \subfloat[5M points.]{%
      \includegraphics[clip,width=0.33\columnwidth]{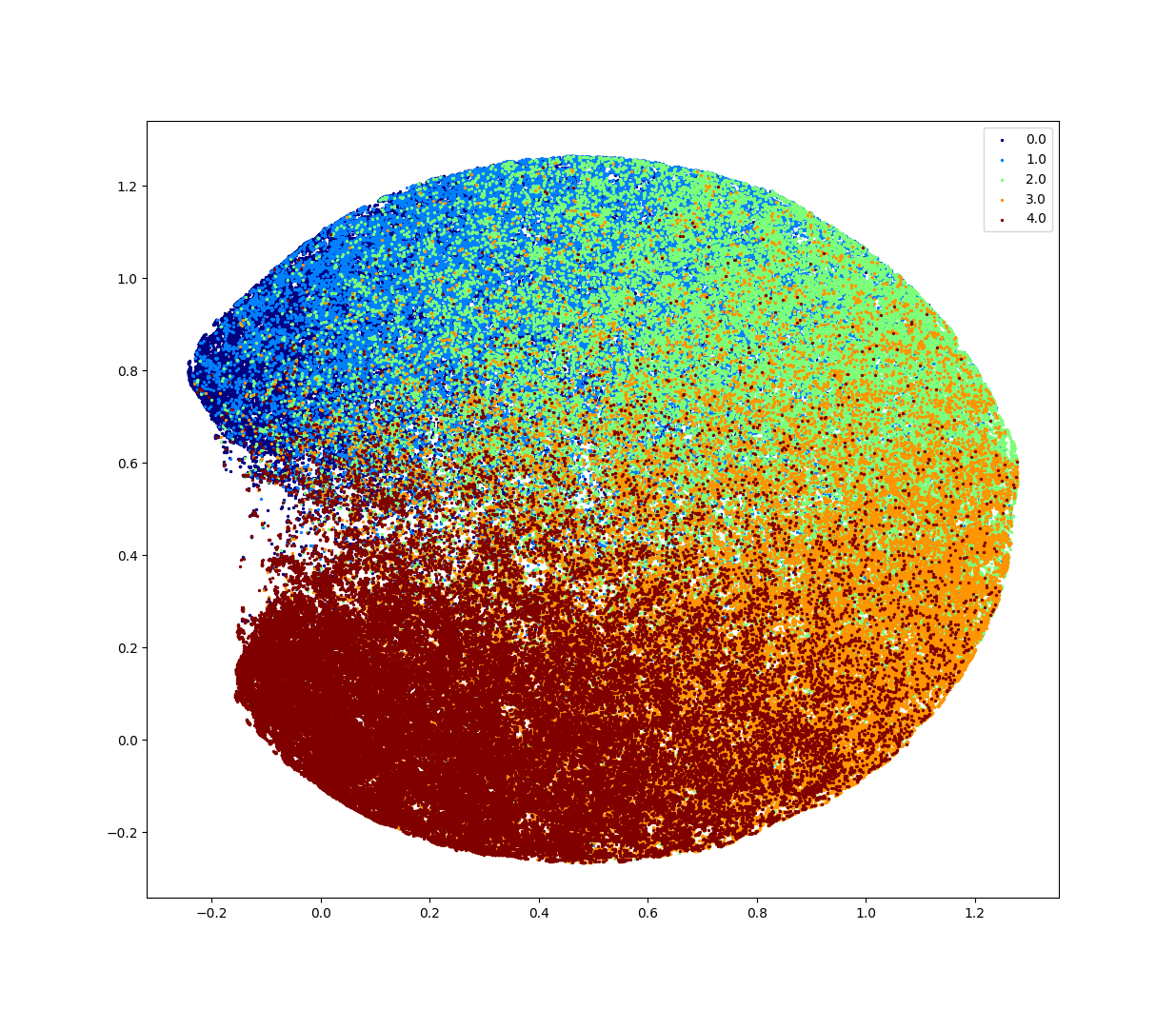}%
      \label{pic:amazon_5M}
    }
    \subfloat[20M points.]{%
      \includegraphics[clip,width=0.33\columnwidth]{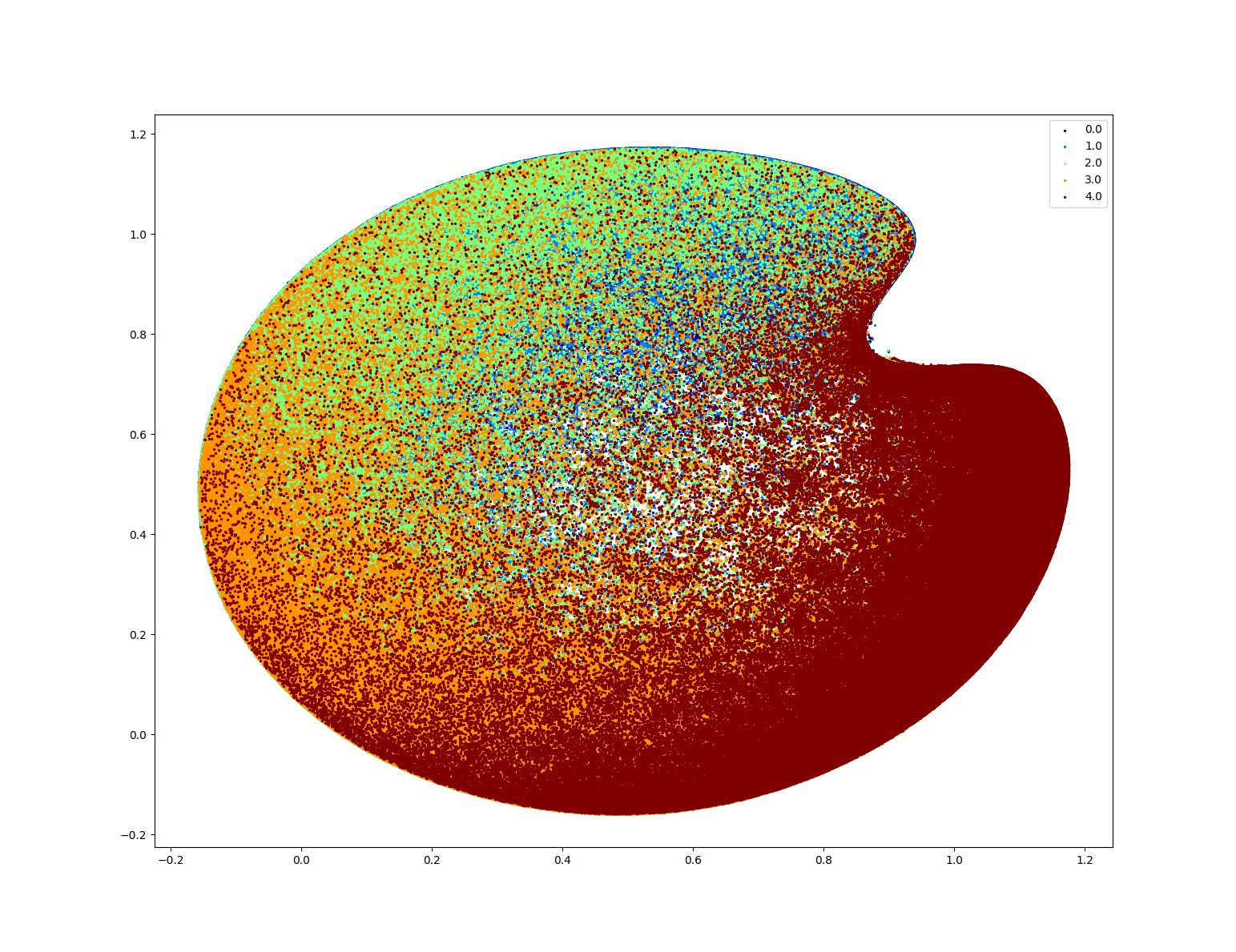}%
      \label{pic:amazon_20M}
    }
    \caption{Visualization of Amazon dataset using IVHD-CUDA.}
    \label{pic:amazon_dataset_results}
\end{figure}

As shown in Fig. \ref{pic:amazon_dataset_results}, IVHD was the only method that was able to generate visualizations for the Amazon20M dataset (and its subsets) in a reasonable period of time. One can see a very clear separation of the 5 classes that appear in this dataset, which contains book reviews from the Amazon platform. Other methods (both those implemented in the CPU and GPU environments) ended with errors or were unable to generate visualizations within 12h, so that the calculation process was interrupted and canceled. This shows that IVHD is the only method that can actually deal with very large datasets containing real-world data in a very fast time using minimal resources, confirming that it is the most efficient and memory-saving method for visualization of high-dimensional data.
\chapter{Meta-platform for supervised visualization}

Supervised learning is a subcategory of machine learning and artificial intelligence. Its goal is to estimate statistical relationships using labeled training data. It has enabled a wide variety of basic and applied discoveries, from the discovery of biomarkers in omics data \cite{vogelstein_2014} to the recognition of objects in images \cite{krizhevsky_2012}. Dimensionality reduction itself is often used as a preliminary step in data mining, both as a pre-processing for classification or regression and for visualization. Most dimensionality reduction techniques to date are unsupervised and do not consider class labels (e.g., PCA \cite{pca}, MDS \cite{mds}, t-SNE \cite{tsne} and its variants \cite{treesne,lion_tsne,atSNE}, Isomap \cite{tenenbaum2001}, LargeVis \cite{largevis}). These methods require large amounts of data and are often sensitive to noise, which can hide important data features. Various attempts at supervised dimensionality reduction methods that take into account auxiliary annotations (e.g., class labels) have been successfully implemented to increase classification accuracy or improve data visualization. Many of these supervised techniques consider labels as a loss function in the form of similarity or dissimilarity matrices, thus creating an undue emphasis on separation between class clusters, which does not realistically reflect local and global relationships in the data. Additionally, these approaches are often sensitive to parameter tuning, which can be difficult to configure without a clear quantitative notion of visual superiority. As for UMAP \cite{umap,parametric_umap}, it can be used for standard unsupervised dimension reduction, but the algorithm itself offers significant flexibility, allowing it to be extended to perform other tasks, including making use of categorical label information to perform supervised dimension reduction and even metric learning.

In this chapter, we describe a novel meta-platform for supervised visualization, which allows each visualization method to be considered as supervised. 

\subsubsection{Meta-platform for supervised high-dimensional visualization}

The idea is based on the use of centroids obtained using any clustering method (such as, for example, k-means) (Fig. \ref{fig:chapter_8_centroids_general_idea}). We generate two types of centroids: 1) local (intra-class) and 2) global (for the entire dataset). Using the centroids thus obtained, parallel layers of neurons of both types are formed. Feed-forward of such a network boils down to calculating the Euclidean distances of a given sample with respect to the centroids to which it is "closest" (you can also use here any classification algorithm, e.g. SVM, which will classify the sample into the appropriate class), and then the activation function is applied. What we achieve in this way is a trained classifier against which we will embed a given sample with the appropriate set of centroids. After this embedding, any method can be used for data visualization, and as a result, we obtain a supervised variation of this method.

\textbf{Experiment with random dataset.} A random 100-D dataset was generated and visualized with t-SNE and PCA. Before visualization, we embedded the data set using a different set of centroids obtained with the k-means method.

\begin{figure}[ht!]
    \centering
    \includegraphics[width=\textwidth]{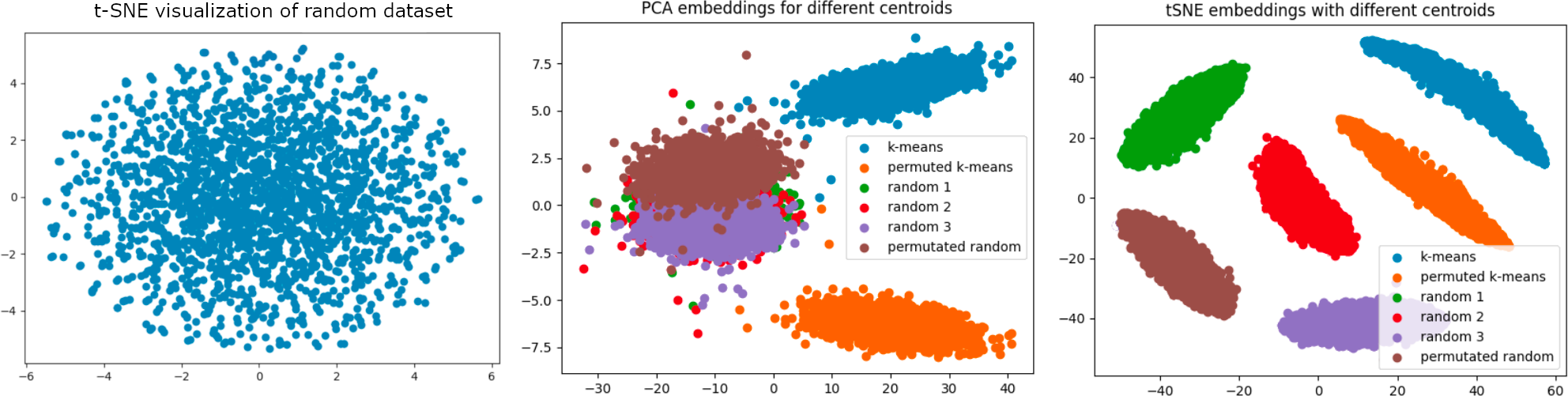}
    \caption{Embedding dataset with different set of centroids. Left: random 100-D dataset visualized with t-SNE. Middle: random dataset embedded with the centroids, visualized with PCA.  Right: random dataset embedded with the centroids, visualized with t-SNE.}
    \label{fig:chapter_8_random_dataset_experiment}
\end{figure}

As shown in Fig. \ref{fig:chapter_8_random_dataset_experiment}, embedding with different sets of centroids causes clusters to separate, when visualized with the PCA or t-SNE method. This is because a dataset embedded with different sets of centroids actually has different bases, and hence the clusters are separate in space. This introduces the need for global centroids (which will provide a common frame-of-reference in the embedding).

\begin{figure}[ht!]
    \centering
    \includegraphics[width=0.8\textwidth]{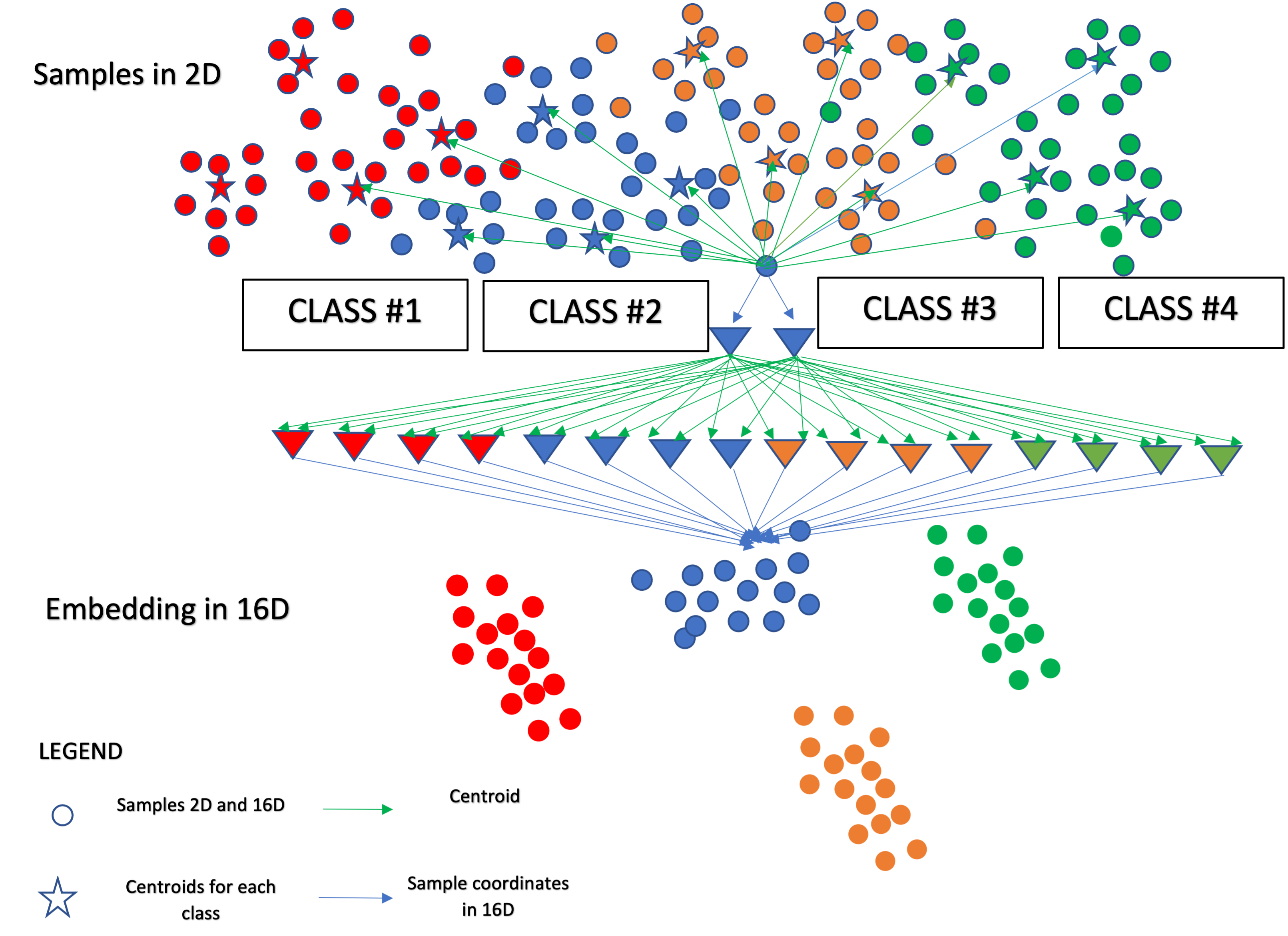}
    \caption{General idea of meta-platform for supervised visualization, which leverages centroids.}
    \label{fig:chapter_8_centroids_general_idea}
\end{figure}

\textbf{Experiment with artificially generated dataset.} A 10-D dataset with overlapped classes. There are 5 classes with 500 points each. As a classification method, we use a "perfect classifier" (actual labels). We do not want the results to be distorted by the classifier. 

\begin{figure}[ht!]
    \centering
    \subfloat[First 3 dimensions.]{\includegraphics[width=0.45\textwidth]{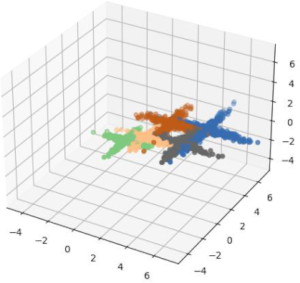}}
    \hspace{0.5cm}
    \subfloat[t-SNE visualization.]{\includegraphics[width=0.45\textwidth]{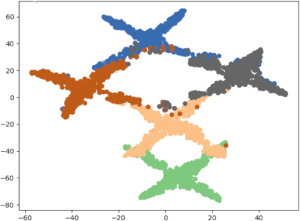}}
    \caption{Artificially generated "X shapes" dataset (first 3 dimensions) and its t-SNE visualization.}
    \label{fig:chapter_8_artificial_dataset_experiment}
\end{figure}

As shown in Fig. \ref{fig:chapter_8_artificial_dataset_experiment}, the dataset contains "X shapes" manifolds immersed in a 10-dimensional space. The overall structure of the dataset can be observed by plotting the first 3 dimensions (Fig. \ref{fig:chapter_8_artificial_dataset_experiment}a). We specifically wanted certain classes to overlap to add realism to the dataset.

\begin{figure}[ht!]
    \centering
    \subfloat[Embedded with only local centroids.]{\includegraphics[width=0.33\textwidth]{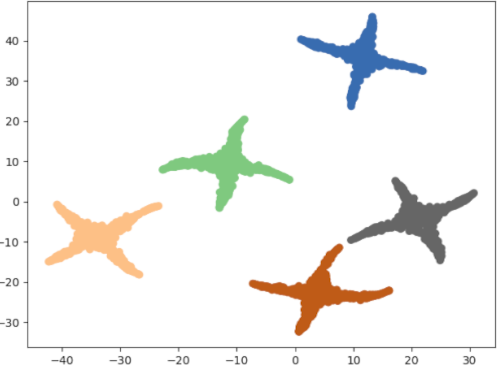}}
    \subfloat[Embedded with only global centroids.]{\includegraphics[width=0.33\textwidth]{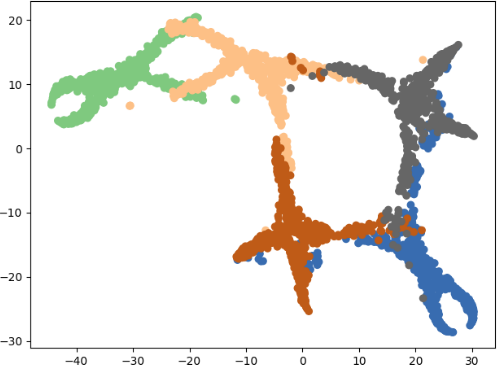}}
    \subfloat[Embedded with both global and local centroids.]{\includegraphics[width=0.33\textwidth]{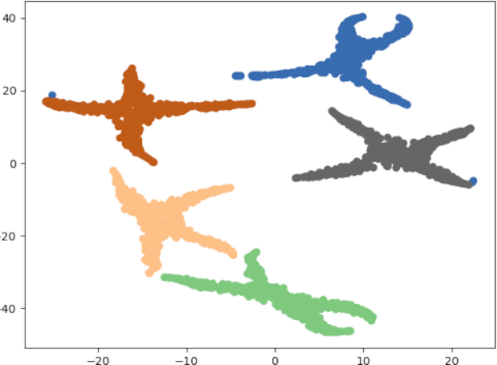}}
    \caption{t-SNE visualization of artificially generated dataset "X shapes", which was preprocessed with different combination of centroids before the visualization.}
    \label{fig:chapter_8_artificial_dataset_experiment_results}
\end{figure}

Based on Fig. \ref{fig:chapter_8_artificial_dataset_experiment_results}, the following conclusions could be drawn:
\begin{enumerate}
    \item For embedding based on distances from centroids belonging to a single recognized class (Fig. \ref{fig:chapter_8_artificial_dataset_experiment_results}a), the classes are separated and retain their internal structure. However, the global structure has nothing to do with their original arrangement.
    \item For embedding based on distances from centroids belonging to the entire dataset (Fig. \ref{fig:chapter_8_artificial_dataset_experiment_results} b), the effect is similar to that of pure t-SNE. Still, sizable areas overlap, global structure is visible, and differences are due to the way the k-means method finds clusters (less emphasis on dense areas).
    \item For embedding based on hybrid distances (Fig. \ref{fig:chapter_8_artificial_dataset_experiment_results}c), visualization preserves the separation of classes, intra-class structure, and also their mutual position in space (both the layout and the fact that yellow and green are closer together than the others).
\end{enumerate}

\textbf{Experiment with local and global centroids.} Knowing how the DR quality and KNN gain metrics work \ref{sec:quality_assessment}, the following observations (which coincide with theory and intuition) can be made:

\begin{enumerate}
    \item With an increase in the number of global centroids, the DR quality should shift to the right (better properties for a bigger neighborhood), and the KNN gain should be lower in a larger part of the "neighborhood size" spectrum. KNN gain gives us information on how the KNN classifier behaves when trained on the embedded dataset. It is intuitive that the classifiers' accuracy should decrease when data is embedded in a more global fashion. Both effects can be observed in the first row of Fig. \ref{fig:chapter_8_artificial_dataset_experiment_lv_vs_gc}. 
    
    \item However, with an increase in the number of local centroids, the DR quality should shift to the left (worse properties for a bigger neighborhood), yet the KNN gain should be higher in a larger part of the "neighborhood size" spectrum. There is less global information when more local centroids are used, so the separation of clusters is becoming more emphasized; thus, the KNN classifier used for metric calculation is obtaining data that are much more "tweaked" to its class separation aspect. Furthermore, both effects can also be observed (in the second row of Fig. \ref{fig:chapter_8_artificial_dataset_experiment_lv_vs_gc}).
    
    \begin{figure}[ht!]
        \centering
        \subfloat[t-SNE visualization (global centroids increased).]{\includegraphics[width=0.5\textwidth]{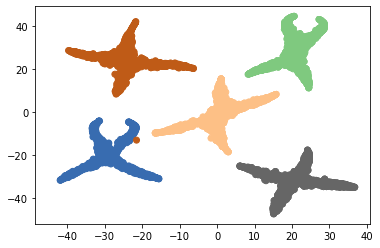}}
        \subfloat[t-SNE visualization (local centroids increased).]{\includegraphics[width=0.5\textwidth]{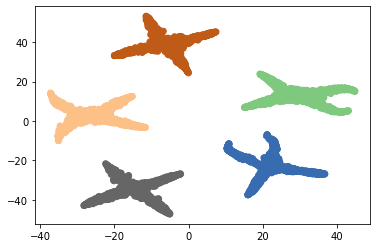}}
        \hfill
        \subfloat[KNN gain.]{\includegraphics[width=0.5\textwidth]{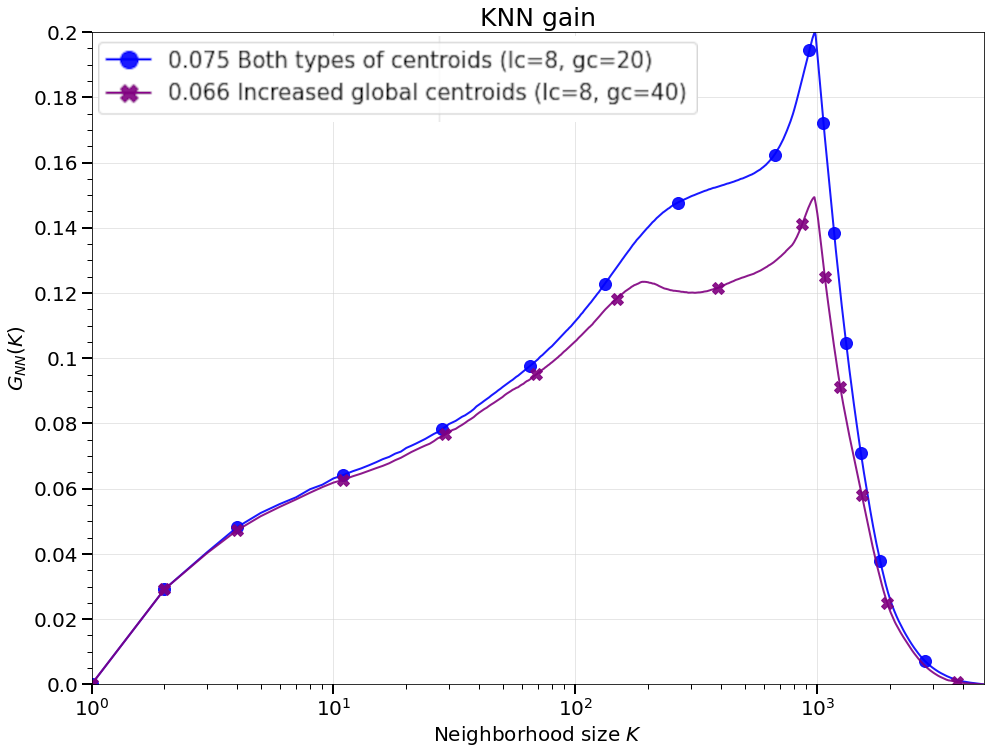}}
        \subfloat[KNN gain.]{\includegraphics[width=0.5\textwidth]{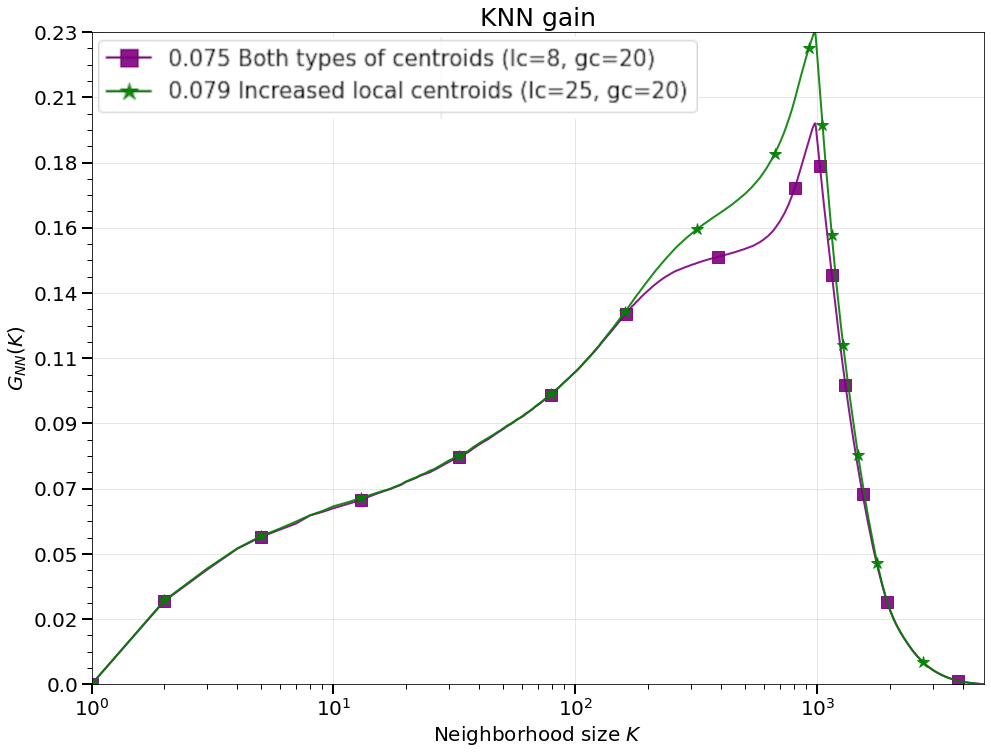}}
        \hfill
        \subfloat[DR quality.]{\includegraphics[width=0.5\textwidth]{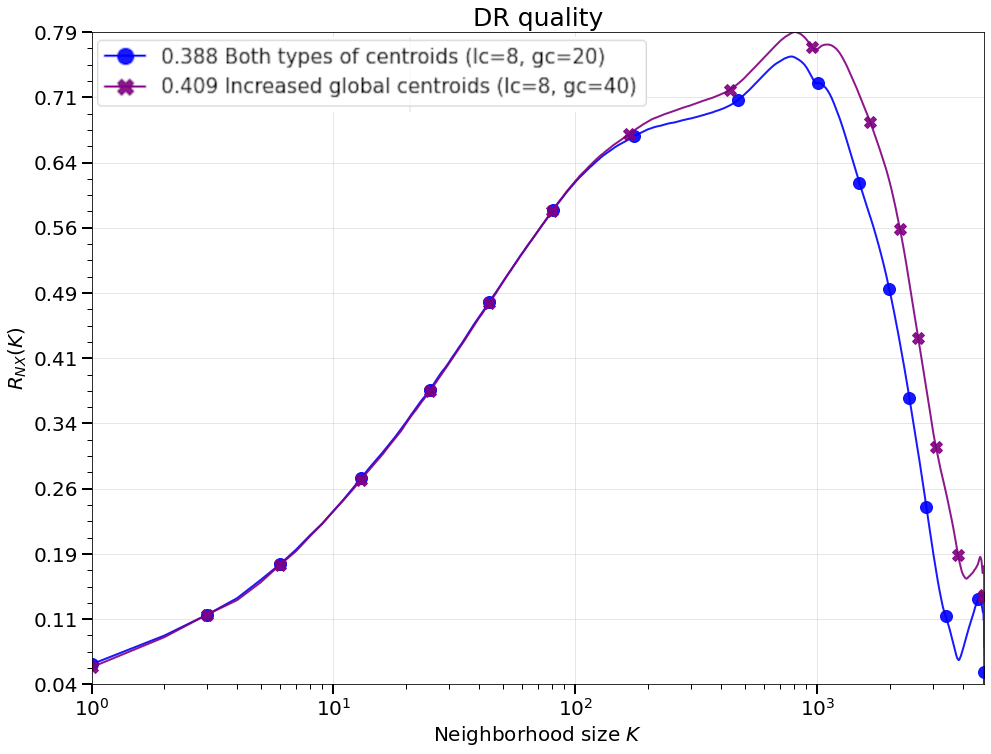}}
        \subfloat[DR quality.]{\includegraphics[width=0.5\textwidth]{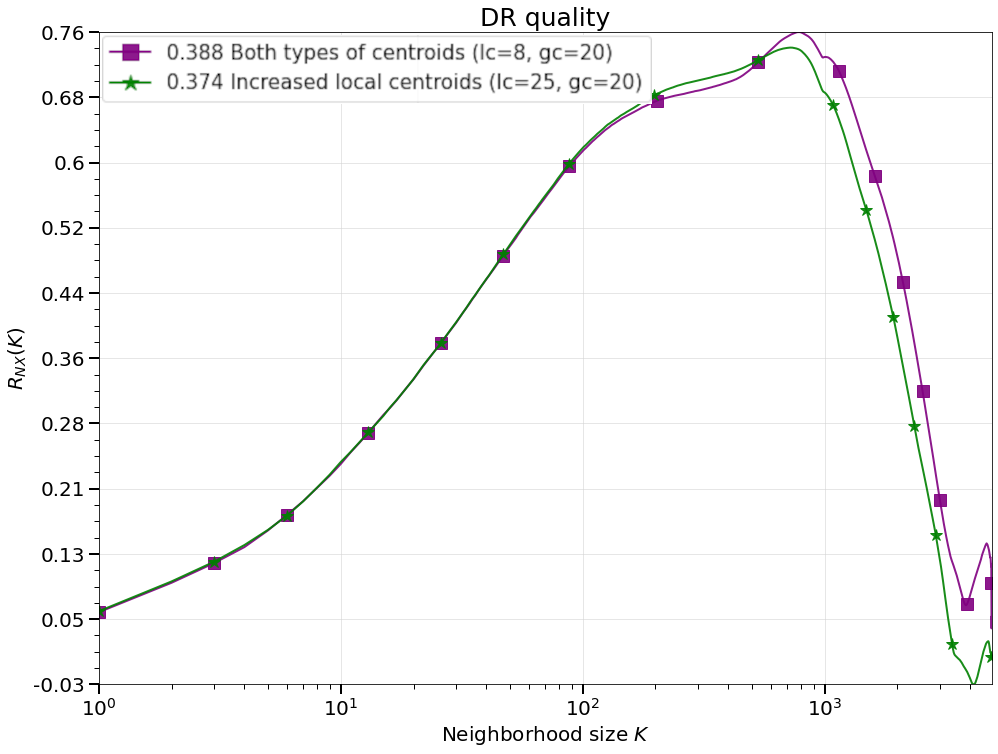}}
        \caption{t-SNE visualization, DR quality and KNN gain of artificially generated dataset "X shapes". In the first column (on the left), we increased the number of global centroids (2 times) in the overall combination of centroids (Fig. \ref{fig:chapter_8_artificial_dataset_experiment_results} c). In the second column (on the right), we increased the number of local centroids (4 times) in overall centroid combination.}
        \label{fig:chapter_8_artificial_dataset_experiment_lv_vs_gc}
    \end{figure}

    \item When comparing Fig. \ref{fig:chapter_8_artificial_dataset_experiment_results} c and Fig. \ref{fig:chapter_8_artificial_dataset_experiment_lv_vs_gc} a, it is also clear that the global geometry of the visualization improved. When analyzing the green class, it is next to the beige class (that is, between gray and brown), and the farthest class is blue. Exactly as in Fig. \ref{fig:chapter_8_artificial_dataset_experiment}b, but with better class separation.
\end{enumerate}

\begin{figure}[ht!]
    \centering
    \includegraphics[width=0.95\textwidth]{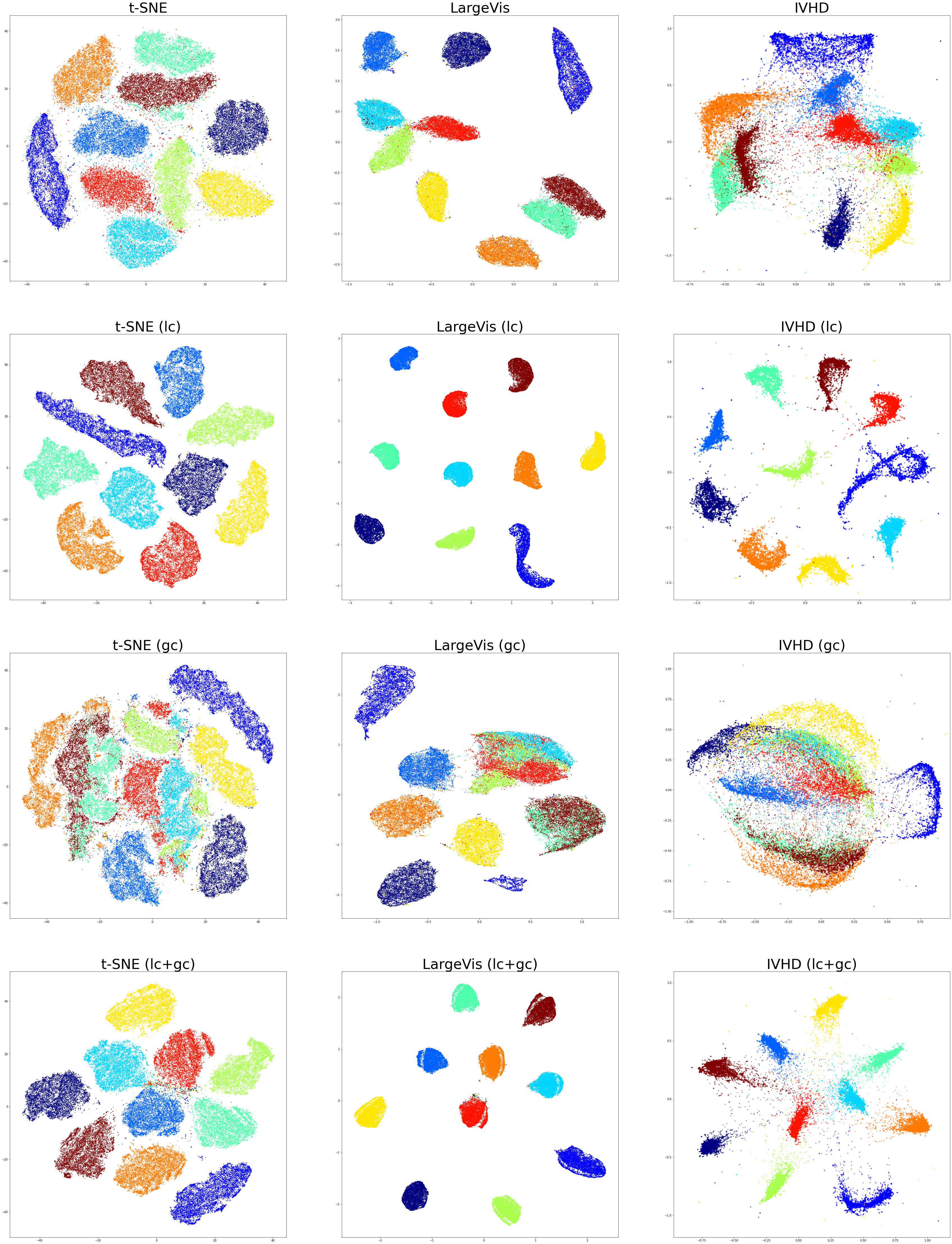}
    \caption{Method comparison on MNIST dataset for both unsupervised and supervised visualization variants.  In next rows of the figure we observe: 1) default visualization of the unsupervised variant of the method, 2) supervised variant, that uses only local centroids, 3) supervised variant, that uses only global centroids and 4) 2) supervised variant, that uses both local and global centroids to perform embedding.}
    \label{fig:chapter_8_mnist_unsupervised_vs_supervised_visualization_comparison}
\end{figure}

\textbf{Experiment with mid-scale dataset.} To verify how this idea behaves with realistic benchmark data, we applied it to the MNIST and FMNIST datasets. To show that it can be used for any unsupervised method, we visualized it with one that is implemented in a GPU environment (AtSNE \cite{atSNE}), one that is implemented in a standard CPU environment t-SNE and IVHD.  To create supervised variants, we used a set-up with 100 global centroids and 100 local centroids. As shown in the following, in Figs. \ref{fig:chapter_8_mnist_unsupervised_vs_supervised_visualization_comparison} and \ref{fig:chapter_8_fmnist_unsupervised_vs_supervised_visualization_comparison}, in both cases, IVHD responds in the most extreme way, separating classes very effectively.  This is a consequence of the fact that IVHD is a method derived from MDS, which in turn is a method that seriously takes into account the aspect of preserving the global structure of the entire data set. 

Regarding performance, generating the set of 100\textit{gc} and 100\textit{lc} and then embedding the MNIST/FMNIST dataset using these centroids is a computational mark-up of about 1 minute (in the local setup described in Section \ref{chapter5_experiments}). It is worth to mention that no optimization was performed and basic Python methods were used for testing, so there is a huge room for improvement and speed-up (although it was not needed at this stage of the research).

It should be noted that the supervised variant is not quite the type of method that is desirable for this type of data analysis. Much more preferable are methods that operate in unsupervised fashion and are efficient enough to analyze datasets in real time (which, as will be shown in the next chapter, can be useful for inspection and interpretation of deep neural networks). The reason for this is that much of the data on which one operates in machine learning is unlabeled and its internal structure is unknown. Then the supervised methods become useless.

\begin{figure}[ht!]
    \centering
    \subfloat[t-SNE DR quality.]{\includegraphics[width=0.45\textwidth]{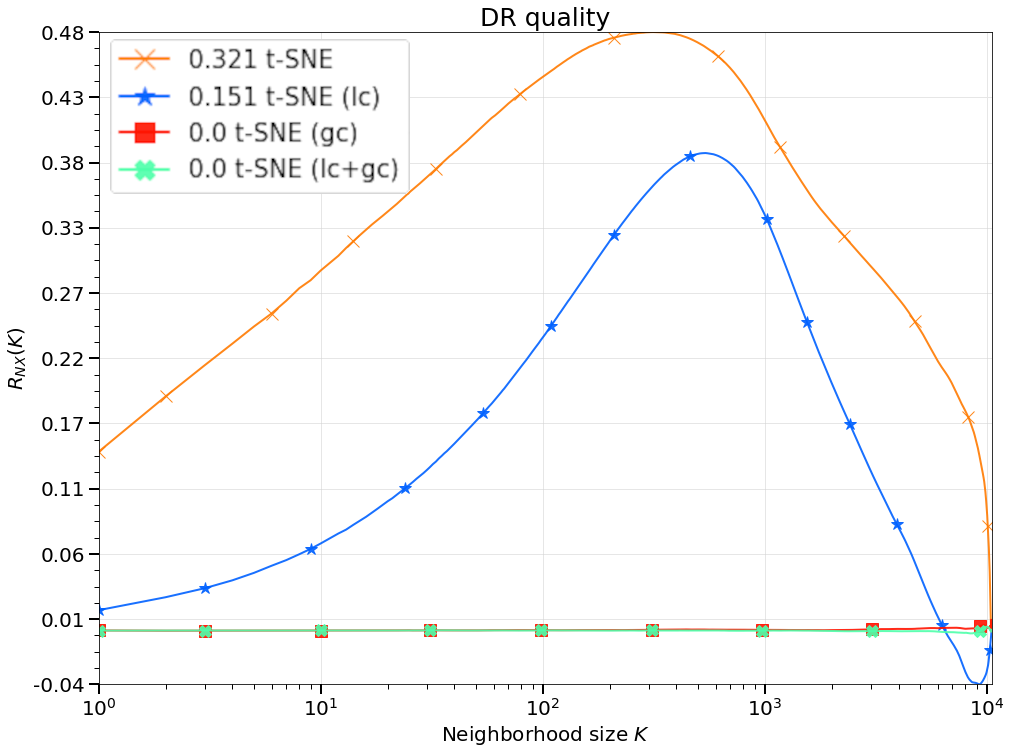}}
    \subfloat[t-SNE KNN gain.]{\includegraphics[width=0.45\textwidth]{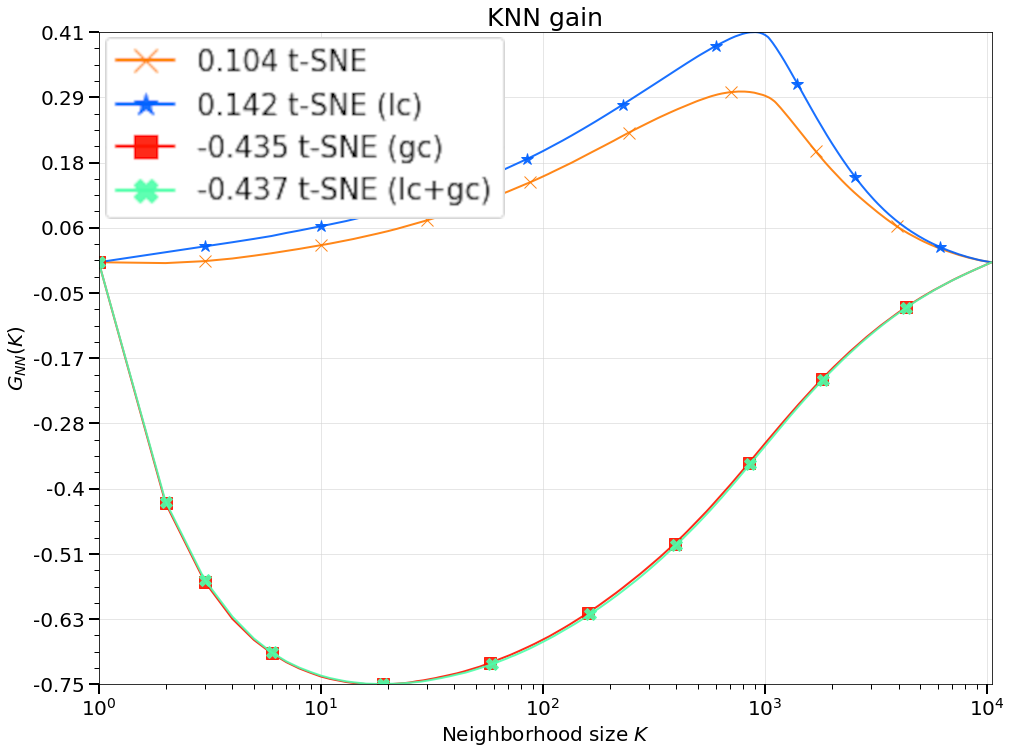}}
    \hfill
    \subfloat[LargeVis DR quality.]{\includegraphics[width=0.45\textwidth]{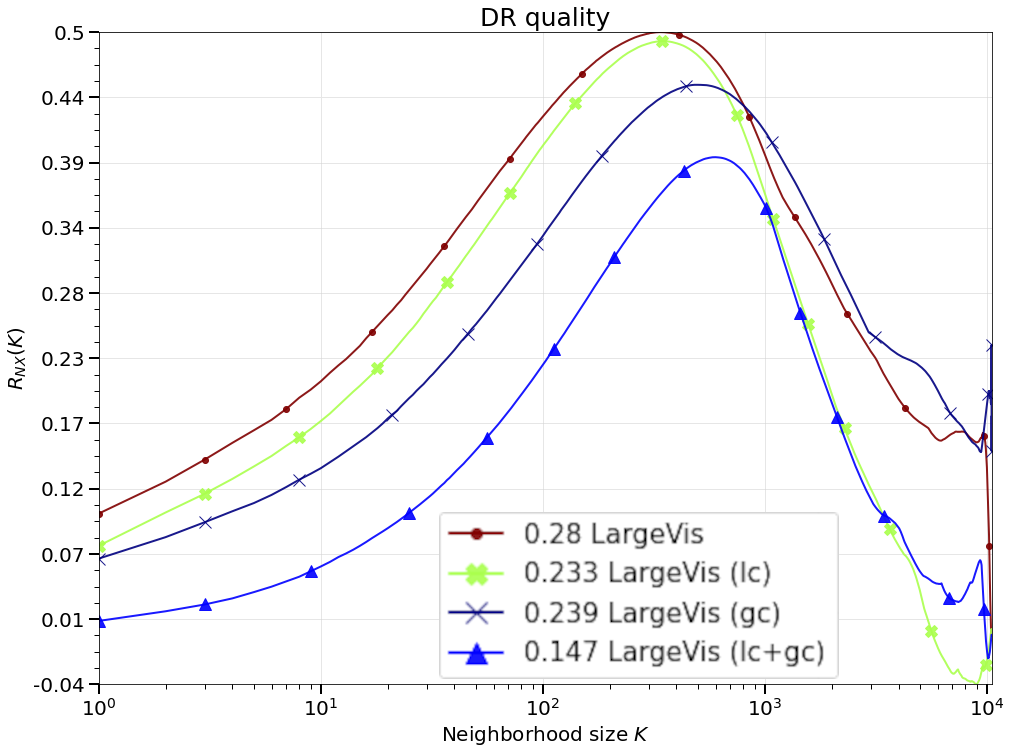}}
    \subfloat[LargeVis KNN gain.]{\includegraphics[width=0.45\textwidth]{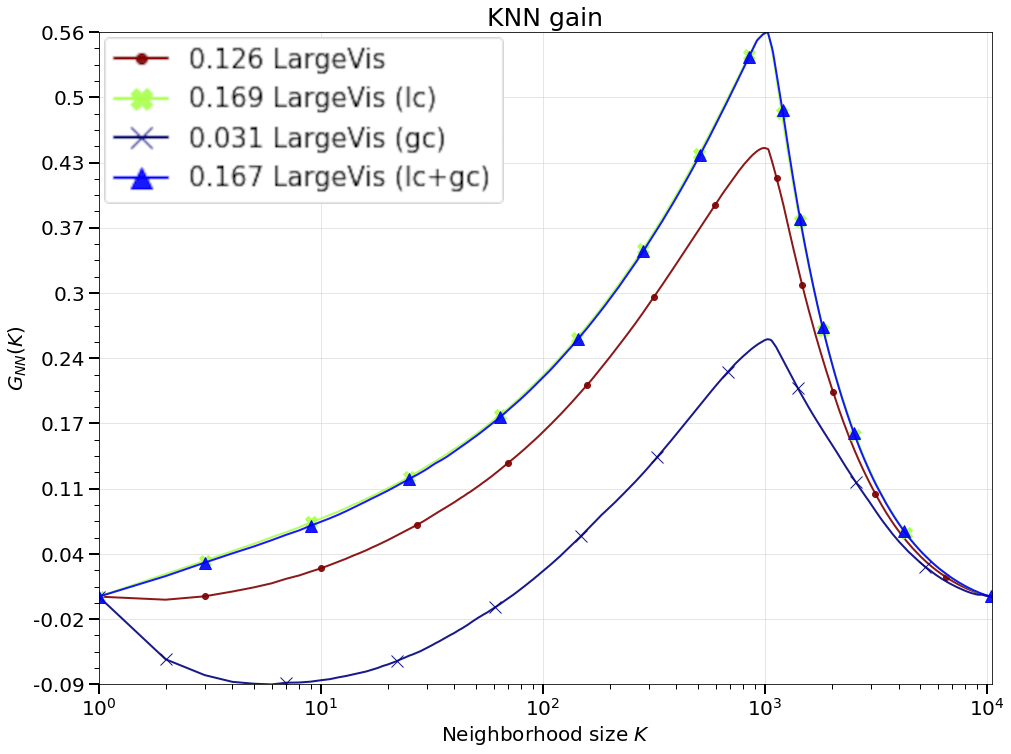}}
    \hfill
    \subfloat[IVHD DR quality.]{\includegraphics[width=0.45\textwidth]{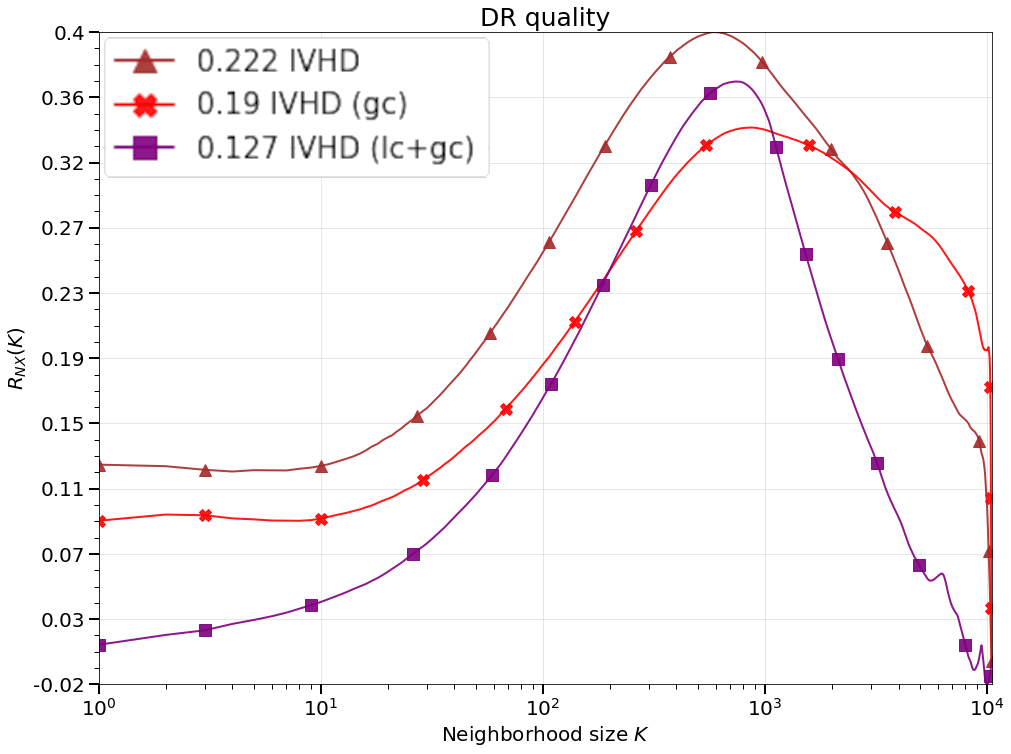}}
    \subfloat[IVHD KNN gain.]{\includegraphics[width=0.45\textwidth]{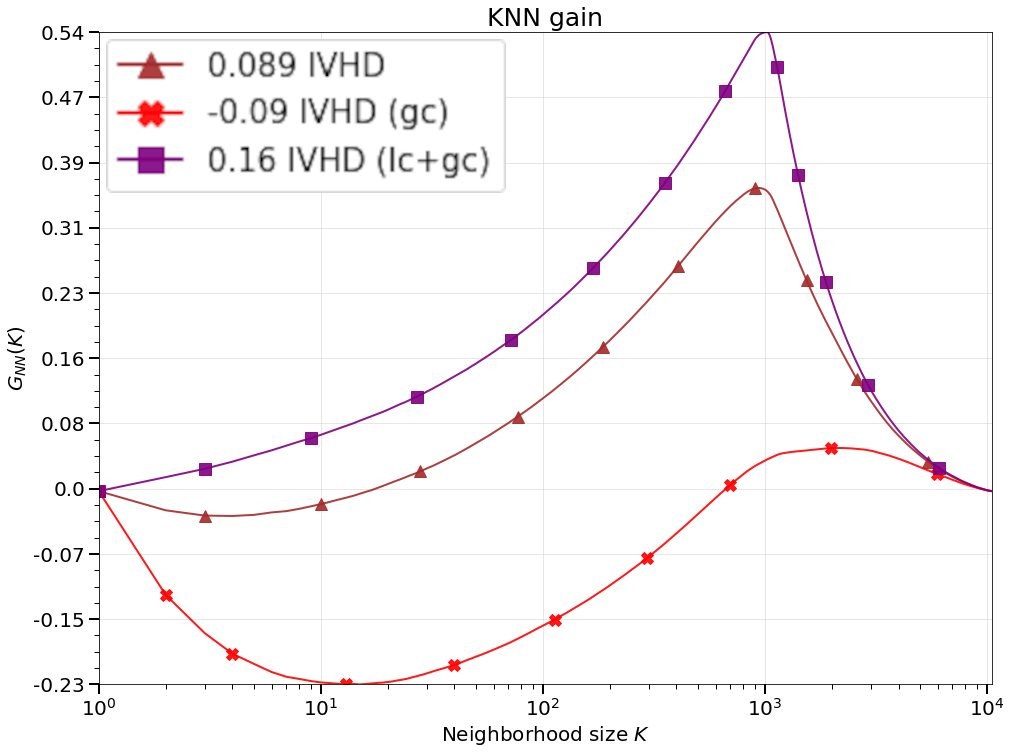}}
    \caption{Metrics obtained for MNIST dataset for both unsupervised and supervised visualization variants for different methods.}
    \label{fig:chapter_8_mnist_unsupervised_vs_supervised_metrics_comparison}
\end{figure}

For both sets of metrics obtained (Figs. \ref{fig:chapter_8_mnist_unsupervised_vs_supervised_metrics_comparison} and \ref{fig:chapter_8_fmnist_unsupervised_vs_supervised_metrics_comparison}), an increase in KNN gain and a slight decrease in DR quality are observed (analogically to Fig. \ref{fig:chapter_8_artificial_dataset_experiment_lv_vs_gc}). It shows that we can use this meta-platform for converting any algorithm from unsupervised to its supervised variant. To verify this thesis, we also visualized the centroid-embedded dataset with AtSNE implemented in the AtSNE CUDA environment (Appendix \ref{appendix_visualizations_timings_and_methods_parametrization}).

\begin{figure}[ht!]
    \centering
    \includegraphics[width=0.95\textwidth]{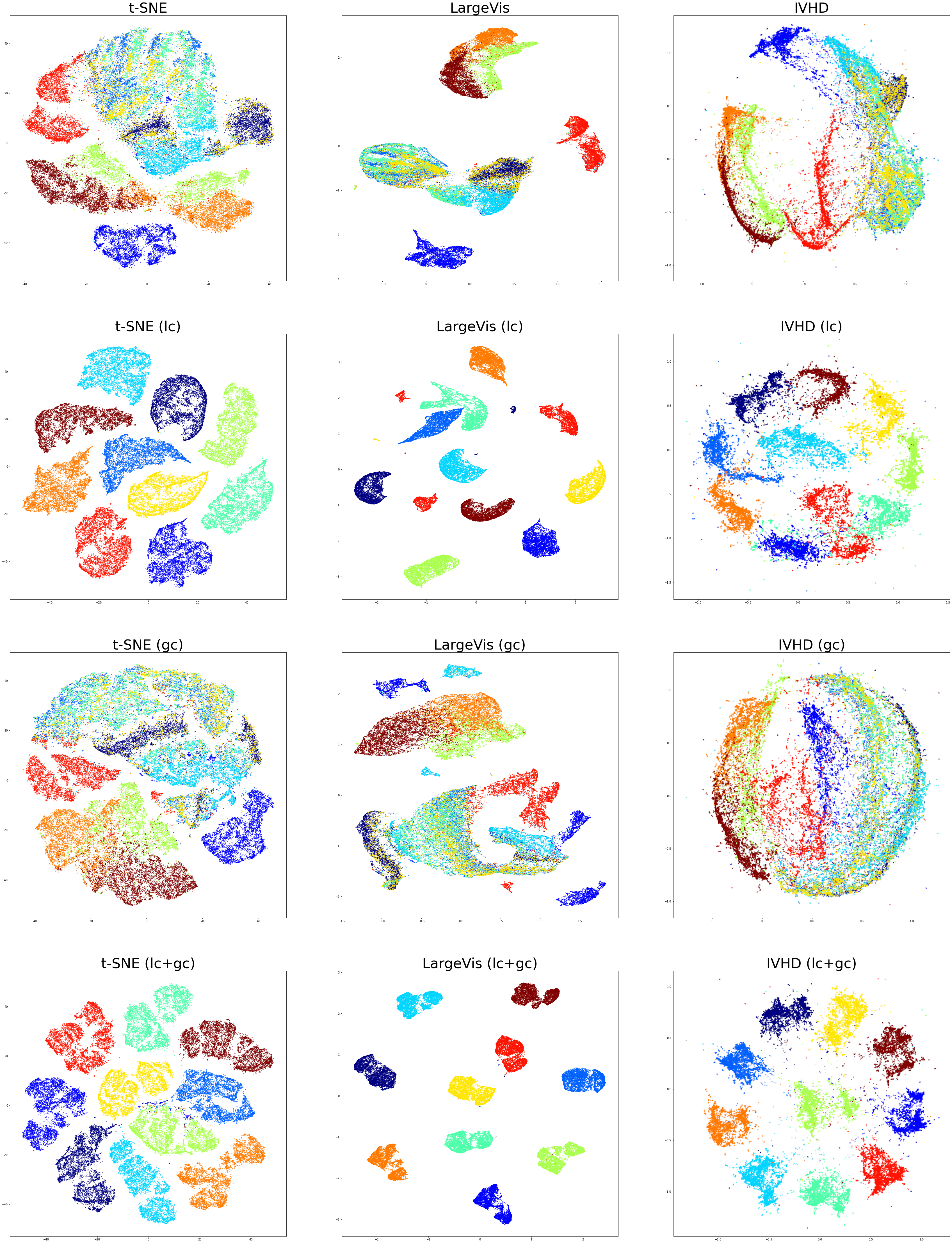}
    \caption{Method comparison on FMNIST dataset for both unsupervised and supervised visualization variants. In next rows of the figures we observe: 1) default visualization of the unsupervised variant of the method, 2) supervised variant, that uses only local centroids, 3) supervised variant, that uses only global centroids and 4) 2) supervised variant, that uses both local and global centroids to perform embedding.}
    \label{fig:chapter_8_fmnist_unsupervised_vs_supervised_visualization_comparison}
\end{figure}

\newpage

\begin{figure}[ht!]
    \centering
    \subfloat[t-SNE DR quality.]{\includegraphics[width=0.45\textwidth]{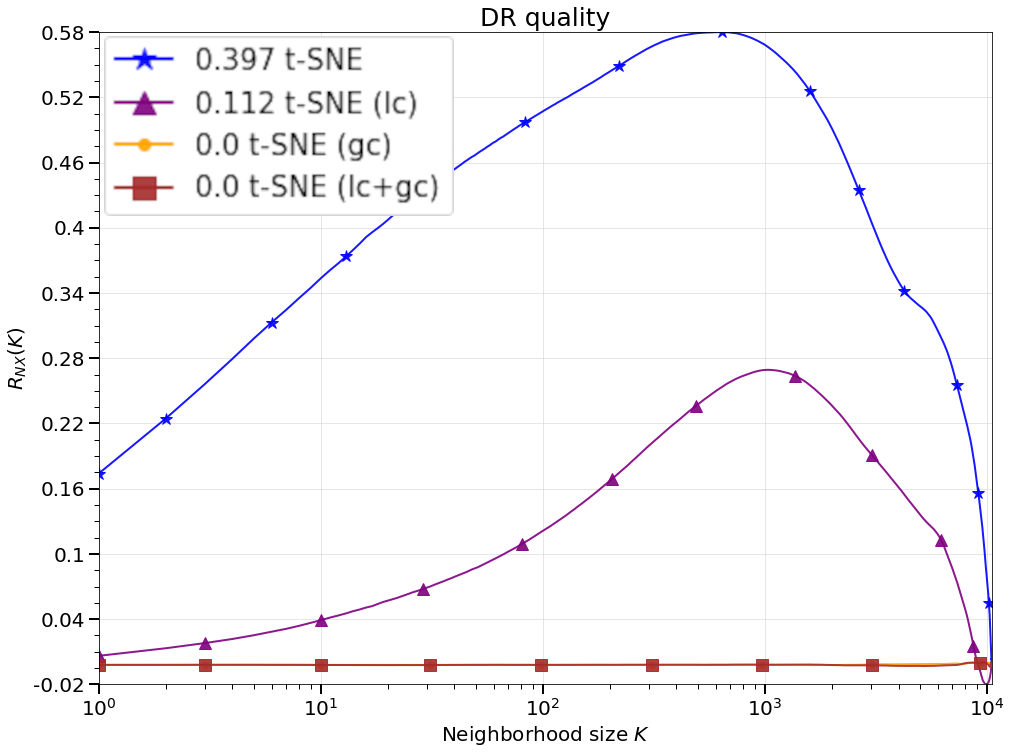}}
    \subfloat[t-SNE KNN gain.]{\includegraphics[width=0.45\textwidth]{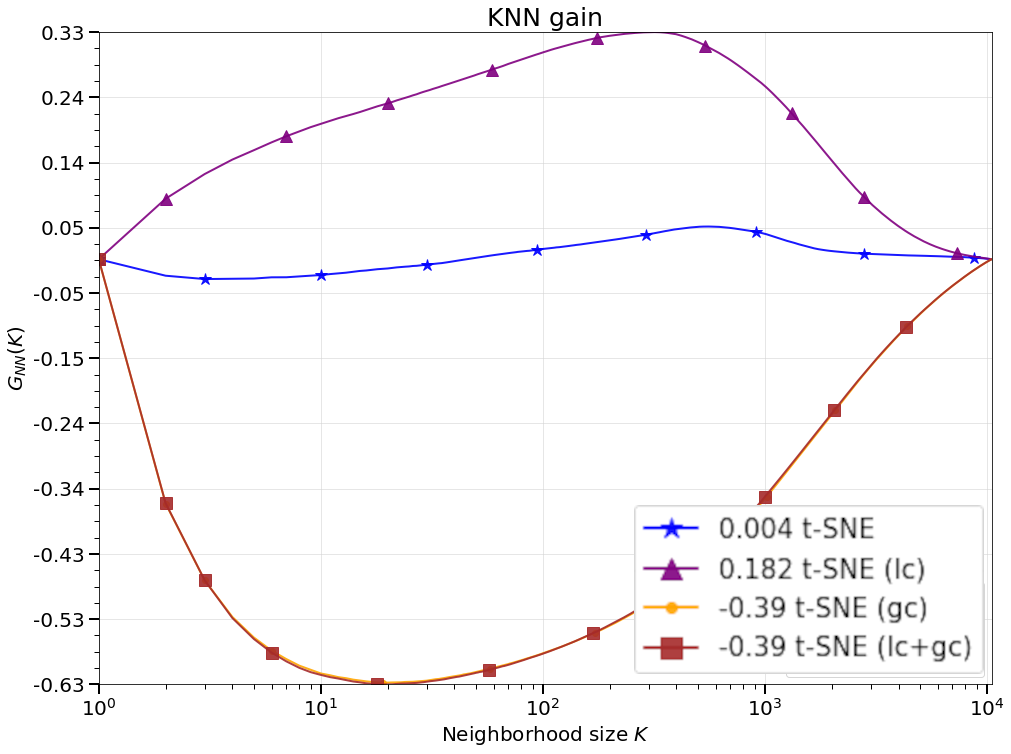}}
    \hfill
    \subfloat[LargeVis DR quality.]{\includegraphics[width=0.45\textwidth]{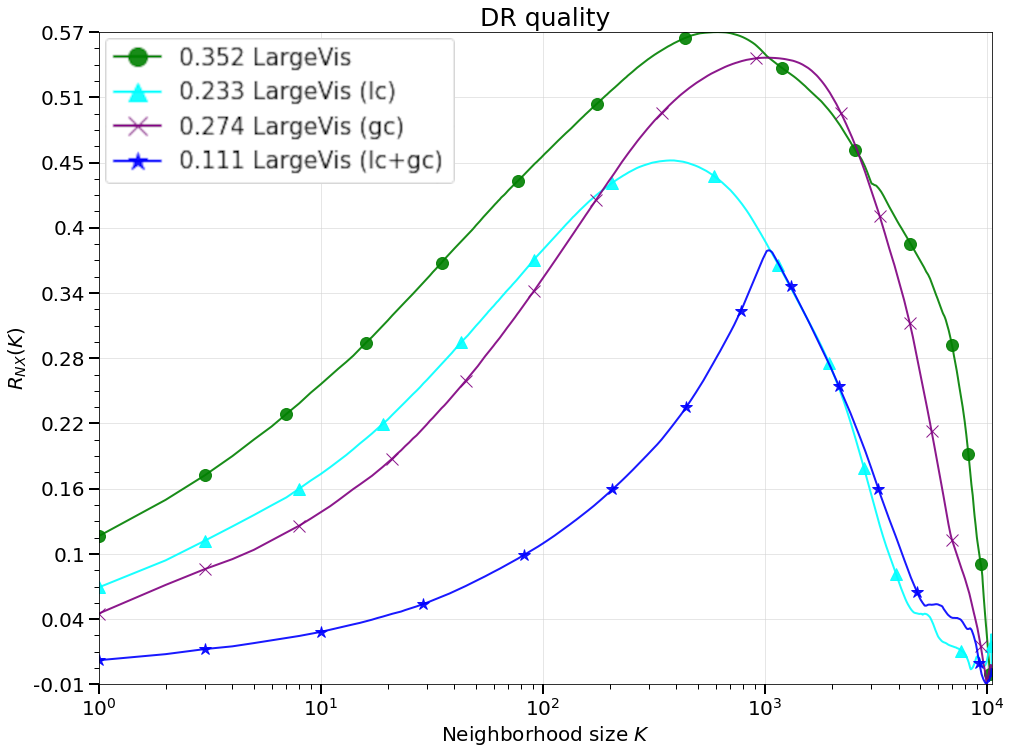}}
    \subfloat[LargeVis KNN gain.]{\includegraphics[width=0.45\textwidth]{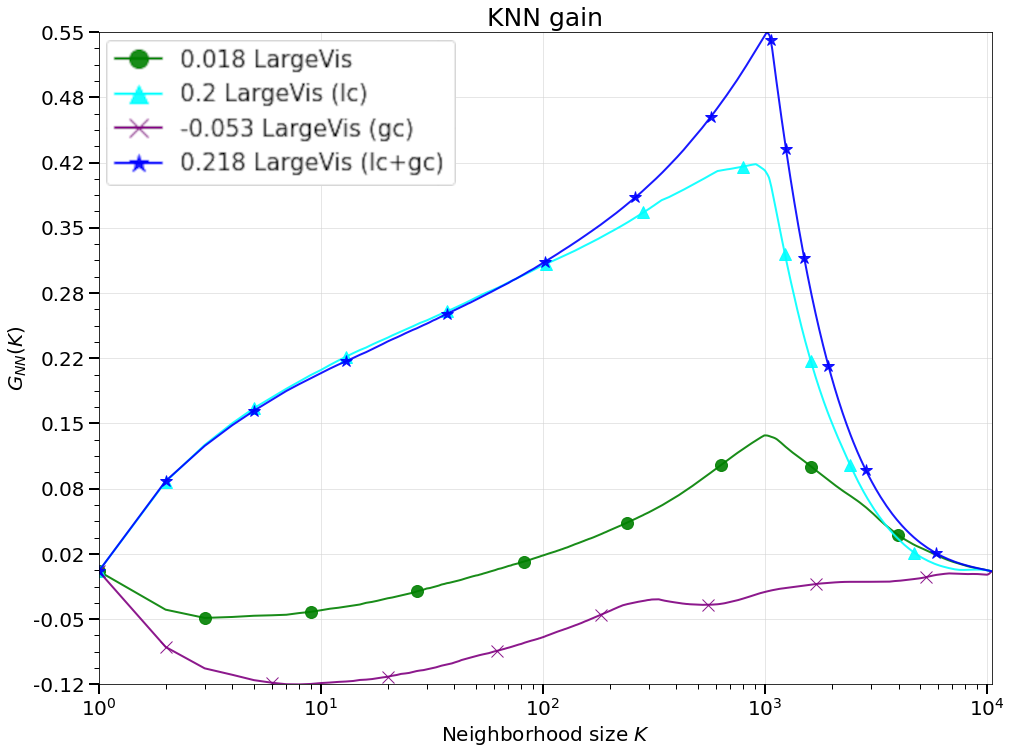}}
    \hfill
    \subfloat[IVHD DR quality.]{\includegraphics[width=0.45\textwidth]{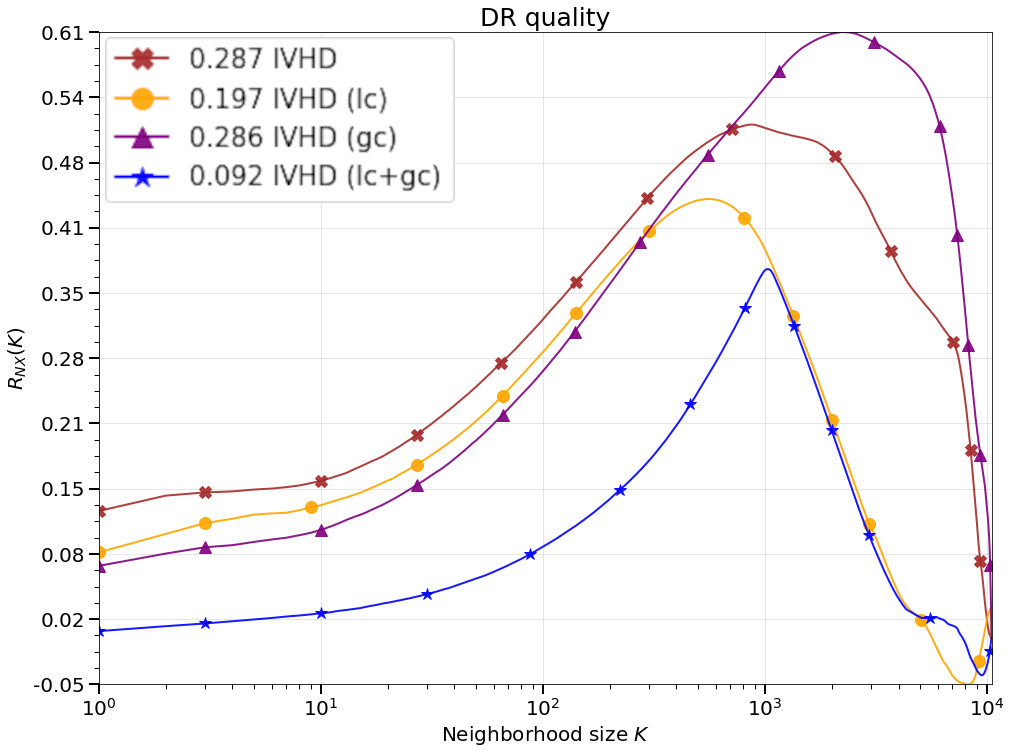}}
    \subfloat[IVHD KNN gain.]{\includegraphics[width=0.45\textwidth]{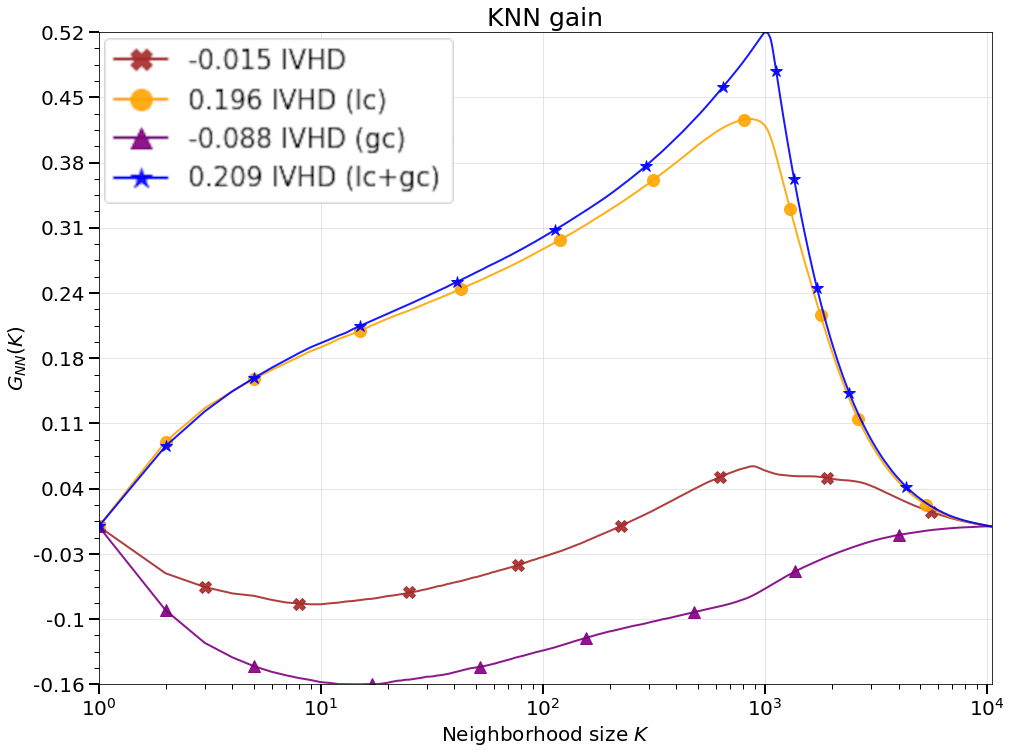}}
    \caption{Metrics obtained for MNIST dataset for both unsupervised and supervised visualization variants for different methods.}
    \label{fig:chapter_8_fmnist_unsupervised_vs_supervised_metrics_comparison}
\end{figure}

The most important feature of the created meta-platform is that it is highly abstracted. We can use arbitrary classifier and unsupervised variant of the embedding method and obtain a supervised-like process of visualization in 2-D (or 3-D) space.
\chapter{Application of HDD visualizations in ANNs inspection and interpretation}

In the field of machine learning, advances in computational power and techniques for building and training artificial neural networks (ANNs) have allowed these models to achieve state-of-the-art performance in many pattern recognition applications \cite{schmidhuber2015}. However, effective ANN training is usually time-consuming and requires considerable knowledge \cite{bengio2012}. Additionally, such networks are typically used as black boxes and it is widely believed that they learn high-dimensional representations of the original observations. In this chapter, we demonstrate the potential of DR techniques to provide insightful visual feedback considering ANNs. Although we focus on multilayer perceptrons and conventional neural networks, this approach is extensible to other types of network (e.g., LSTM or Elman networks [14]). In particular, proposed visualizations address the following two tasks \cite{rauber2016}: 1) exploring the relationships between alternative representations of observations \textit{learned} by ANNs and 2) exploring the relationships between artificial neurons. The projection-based visualization approach proposed in (1) can be found in the machine learning literature \cite{decaf2014,mnih2015,nitish2014,hinton2012,hamel2010}. Using approach (2), which is related to techniques developed for feature-space exploration, we use projections to represent similarities between artificial neurons. It is a novel approach to visualizing the relationships between artificial neurons and classes, which was proposed in \cite{rauber2016}.

\section{Experimental protocol}

Visualization approach that we use is based on hidden-layer activations extracted from a network trained for a given dataset and can be divided into two parts: creating projections of these activations (1) and depicting the relationships between the neurons that originate these activations (2). It is largely inspired by the work presented in \cite{rauber2016}. 

\textbf{Datasets} We include three well-known image classification benchmarks: MNIST \cite{mnist1998}, CIFAR-10 \cite{cifar10}, and SVHN \cite{svhn}. All datasets are described in detail in the Appendix \ref{appendix_datasets}.

\textbf{Neural networks architectures} of two types were considered:

\begin{enumerate}
    \item Multilayer perceptron(MLP): 3072 (784, for MNIST) input neurons, followed by four rectified linear hidden layers of 1000 neurons each. The output layer is softmax with 10 neurons. Dropout \cite{nitish2014} is applied from the first hidden layer, increasing from 0.2 to 0.5 in steps of 0.1 per layer.
    \item Convolutional neural network (CNN): $32 \times 32 \times 3$ input image ($28 \times 28 \times 1$, for MNIST), followed by a convolutional layer with 32 $3 \times 3 \times 3$ (or $3 \times 3 \times 1$) filters, a convolutional layer with 32 $3 \times 3 \times 32$ filters, a $2 \times 2$ max-pooling layer (dropout 0.25), a convolutional layer with 64 $3 \times 3 \times 32$ filters, a convolutional layer with 64 $3 \times 3 \times 64$ filters, a $2 \times 2$ max-pooling layer (dropout 0.25), a fully connected layer with 4096 (or 3136) neurons (dropout 0.5), a fully connected layer with 512 neurons and a softmax output layer with 10 neurons. All fully convolutional and connected layers (except the output) are rectified linear.
\end{enumerate}

Although larger models (in number of layers and parameters) are used for certain difficult classification tasks, the architectures sketched above are fully realistic, typical for image classification tasks, and sufficiently complex to warrant exploration.

\textbf{Training} is performed by mini-batch stochastic gradient descent based on momentum. For MLPs, the batch size is 16, the learning rate is 0.01, the momentum coefficient is 0.9, and the learning decay is $10^{-9}$. For CNNs, the batch size is 32, the learning rate is 0.01, the momentum coefficient is $0.9$, and the learning decay is $10^{-6}$. The initial weights of a neuron in layer $l$ are sampled from a uniform distribution in $[-s, s]$, where $s = [6/(N^{(l-1)} + N^{(l+1)})]^{1/2}$, and the biases start at 0. We manually chose these hyperparameters, together with the architectures mentioned above, based on cross-validation using the predefined validation sets and experiments performed in \cite{rauber2016}. 

\textbf{Activations} for a given layer, the subject of our analysis, are extracted for a random subset of 2000 observations from the test sets, strictly to facilitate visual presentation. This subset is always the same for a given dataset. In two cases, we also extract activations from a random subset of a training set. For CNNs, we only extract activations from fully connected layers.

\textbf{Projections} are created using a fast (approximate) implementation of t-distributed Stochastic Neighbor Embedding (t-SNE) and IVHD. We chose these techniques on the basis of their widespread popularity (t-SNE) and computational speed (IVHD). Both have the proven ability to preserve neighborhoods and clusters in projections (Chapter 4).

To carry out the research demonstrated in this section, we used Python, Keras, Theano \cite{theano}, NumPy, scikit-learn \cite{scikit}, and our methods for feature space exploration. For visualization, we used the Python interface provided with the IVHD method, which made it possible to use all components of this study in one environment.

\subsection{Exploring the relationships between alternative representations of observations learned by ANN}

\textbf{Inter-layer evolution}: Let $\mathbb{A}\{1\},\dots, \mathbb{A}\{T\}$ be a sequence of sets of (high-dimensional) activations, where each activation $\textit{a}\{t\} \in \mathbb{A}\{t\}$ originates from the same observation as a single activation $\textit{a}\{t+1\} \in \mathbb{A}\{t+1\}$. One way to visualize the evolution in such sets of activations is by reducing the dimensionality, applied in such a way that changes in the resulting projections reflect changes in the corresponding high-dimensional data \cite{jackle_2016}. This can be done by creating a projection $\mathbb{A}_{p}\{t\}$ for each activation set $\mathbb{A}_{t}$. When doing this, it is essential to eliminate the variability between projections that does not reflect changes in the high-dimensional data. 

\begin{figure}[ht!]
    \centering
    \subfloat[IVHD $\mathbb{A}_{p}\{1\}$.]{\includegraphics[width=0.25\textwidth]{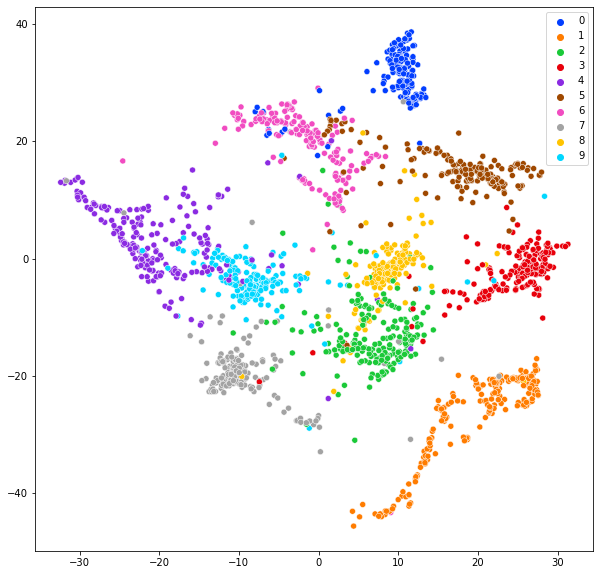}}
    \subfloat[IVHD $\mathbb{A}_{p}\{2\}$.]{\includegraphics[width=0.25\textwidth]{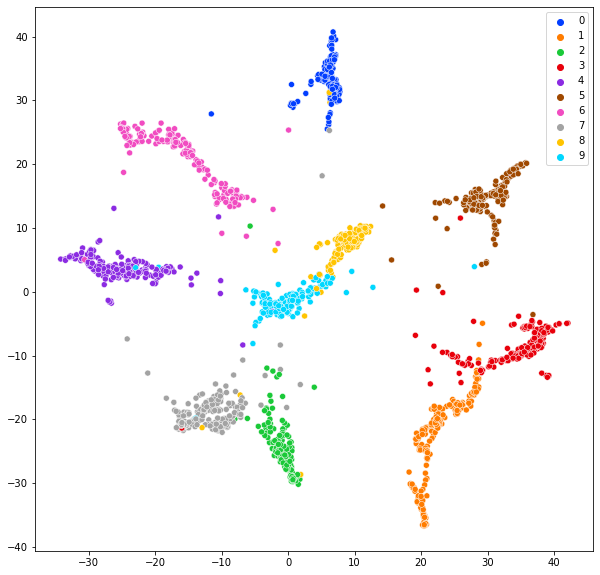}}
    \subfloat[IVHD $\mathbb{A}_{p}\{3\}$.]{\includegraphics[width=0.25\textwidth]{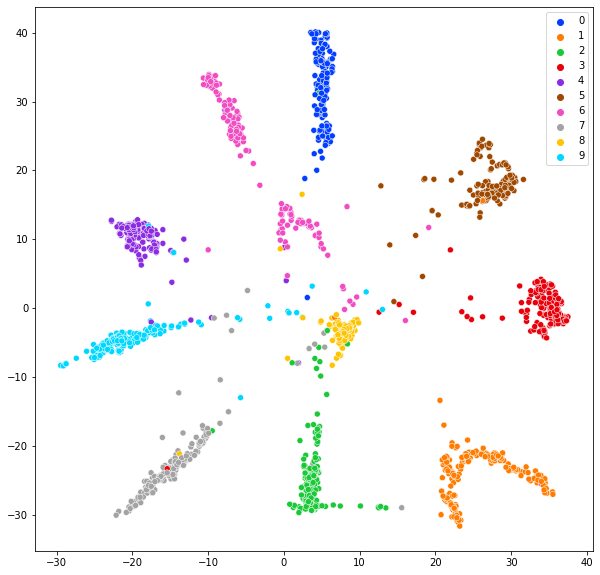}}
    \subfloat[IVHD $\mathbb{A}_{p}\{4\}$.]{\includegraphics[width=0.25\textwidth]{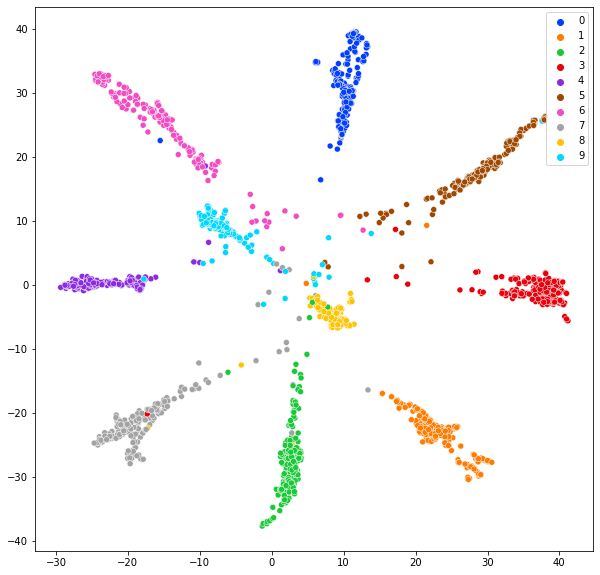}}
    \hfill
    \subfloat[t-SNE $\mathbb{A}_{p}\{1\}$.]{\includegraphics[width=0.25\textwidth]{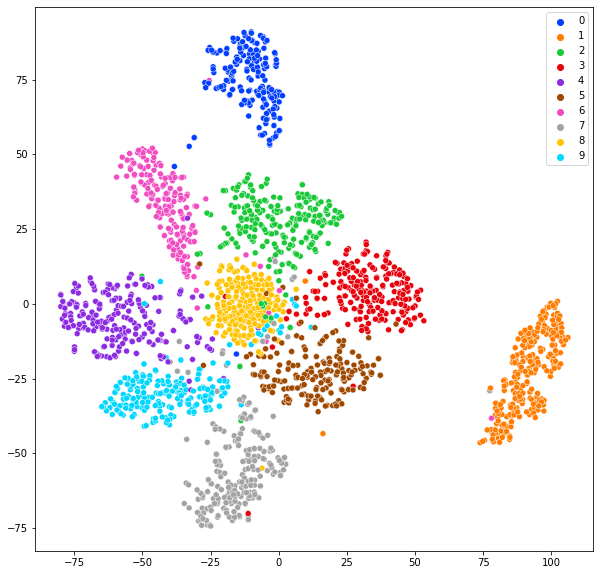}}
    \subfloat[t-SNE $\mathbb{A}_{p}\{2\}$.]{\includegraphics[width=0.25\textwidth]{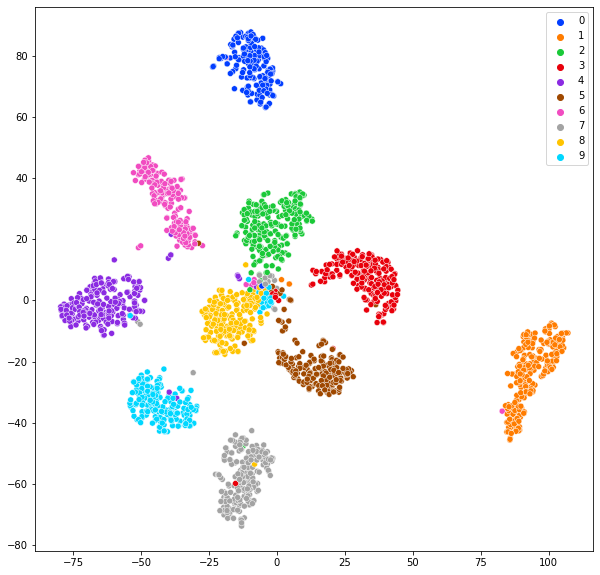}}
    \subfloat[t-SNE $\mathbb{A}_{p}\{3\}$.]{\includegraphics[width=0.25\textwidth]{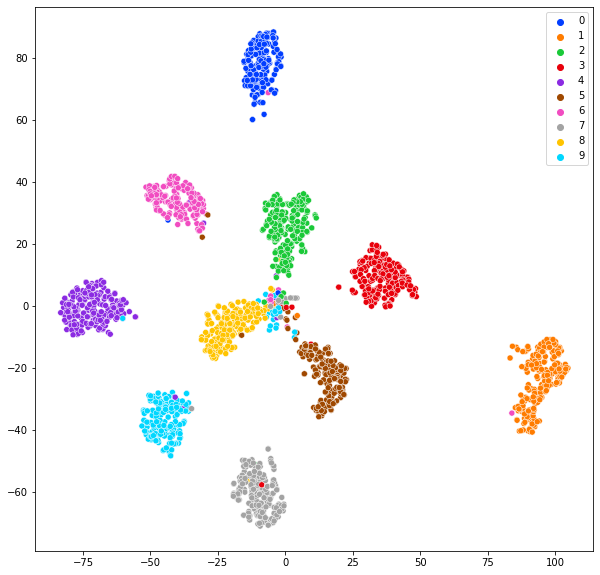}}
    \subfloat[t-SNE $\mathbb{A}_{p}\{4\}$.]{\includegraphics[width=0.25\textwidth]{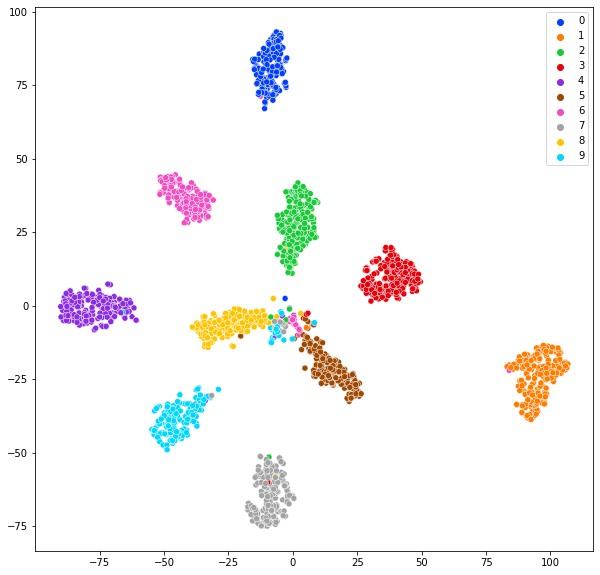}}
    \hfill
    \subfloat[IVHD overall inter-layer evolution.]{\includegraphics[width=0.5\textwidth]{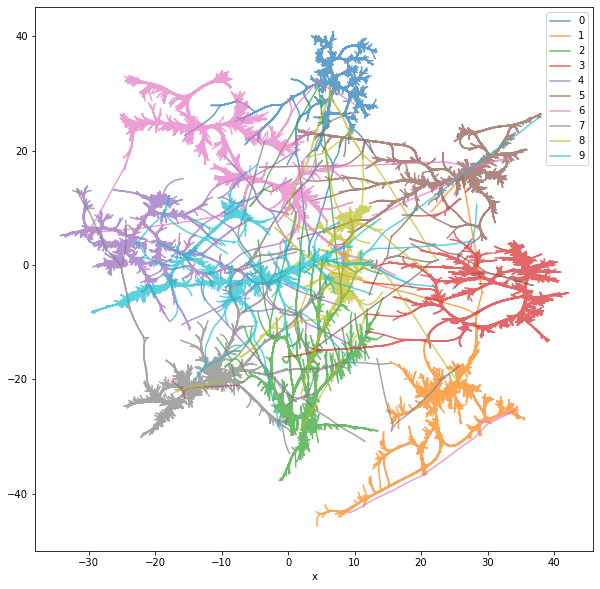}}
    \subfloat[t-SNE overall inter-layer evolution]{\includegraphics[width=0.5\textwidth]{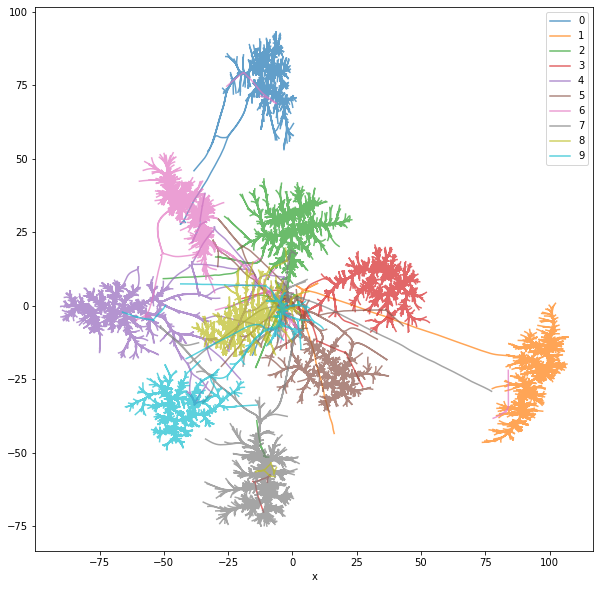}}
    \caption{Inter-layer evolution of MLP after training using MNIST dataset.}
    \label{fig:inter_layer_evolution_mnist_mlp}
\end{figure}

It is important to remember that dimensionality reduction techniques, including t-SNE and IVHD, often produce large changes in global visual cluster placement for small data changes, which registration cannot eliminate \cite{garcia2013}. For iterative techniques, an intuitive solution is to initialize the positioning in $\mathbb{A}_{p}\{t+1\}$ with the previously calculated $\mathbb{A}_{p}\{t\}$. In work \cite{rauber2016}, authors verified that this is a poor alternative, as it significantly biases the projection sequence to show evolution due to initialization in a (presumably) better state with respect to the optimization goal. Instead, a simple strategy is employed: calculate a projection (randomly initialized) $\mathbb{A}_{p}$ of $\mathbb{A}\{1\}$ using t-SNE and use $\mathbb{A}_{p}$ to initialize each $\mathbb{A}\{t\}$.

\begin{figure}[ht!]
    \centering
    \subfloat[IVHD $\mathbb{A}_{p}\{1\}$.]{\includegraphics[width=0.25\textwidth]{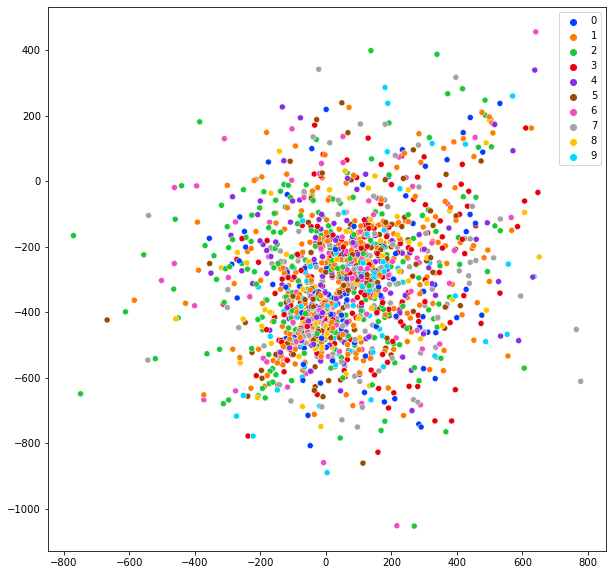}}
    \subfloat[IVHD $\mathbb{A}_{p}\{2\}$.]{\includegraphics[width=0.25\textwidth]{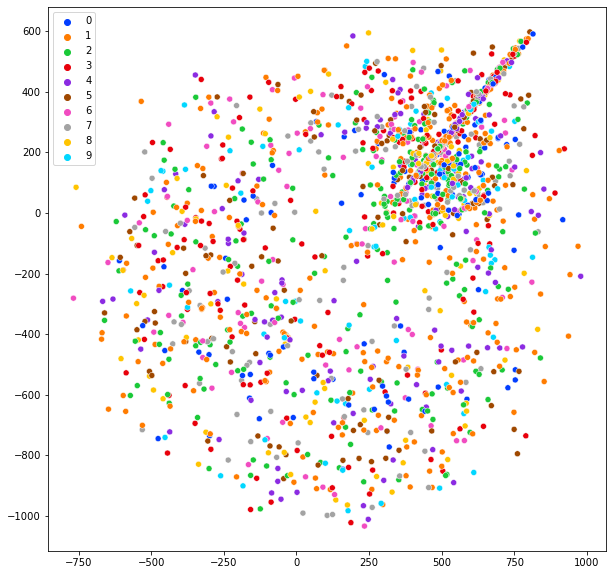}}
    \subfloat[IVHD $\mathbb{A}_{p}\{3\}$.]{\includegraphics[width=0.25\textwidth]{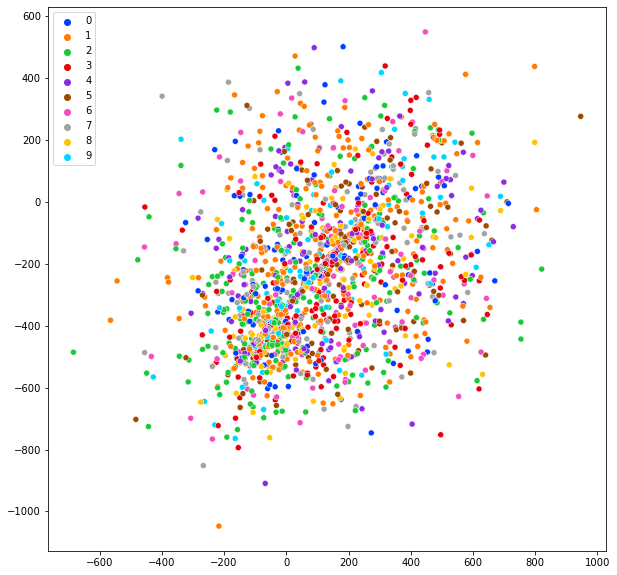}}
    \subfloat[IVHD $\mathbb{A}_{p}\{4\}$.]{\includegraphics[width=0.25\textwidth]{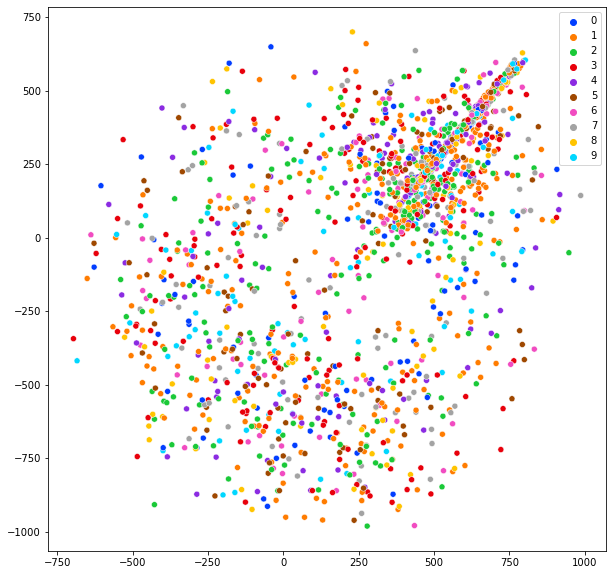}}
    \hfill
    \subfloat[t-SNE $\mathbb{A}_{p}\{1\}$.]{\includegraphics[width=0.25\textwidth]{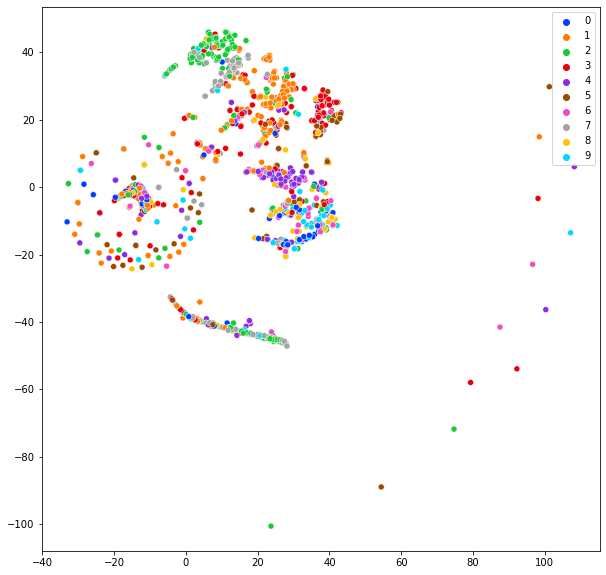}}
    \subfloat[t-SNE $\mathbb{A}_{p}\{2\}$.]{\includegraphics[width=0.25\textwidth]{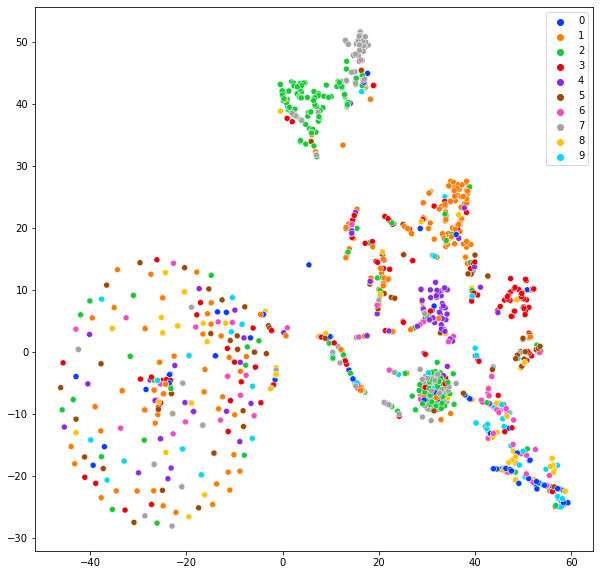}}
    \subfloat[t-SNE $\mathbb{A}_{p}\{3\}$.]{\includegraphics[width=0.25\textwidth]{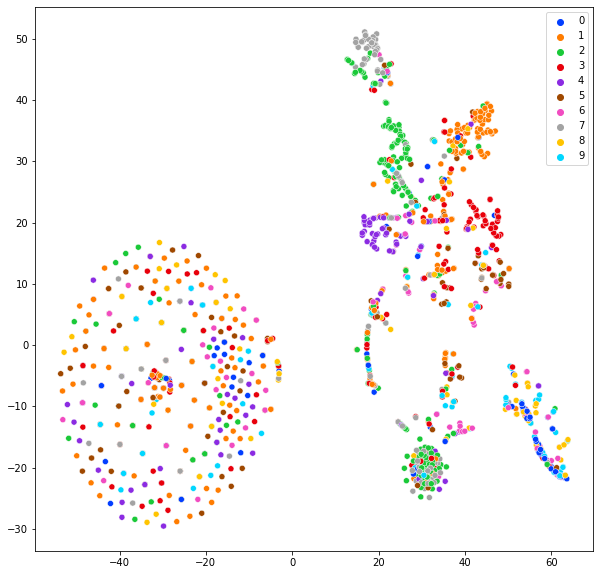}}
    \subfloat[t-SNE $\mathbb{A}_{p}\{4\}$.]{\includegraphics[width=0.25\textwidth]{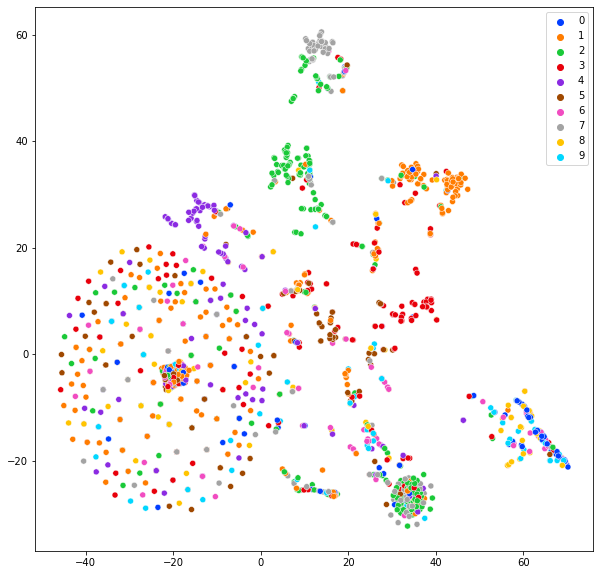}}
    \hfill
    \subfloat[IVHD overall inter-layer evolution.]{\includegraphics[width=0.5\textwidth]{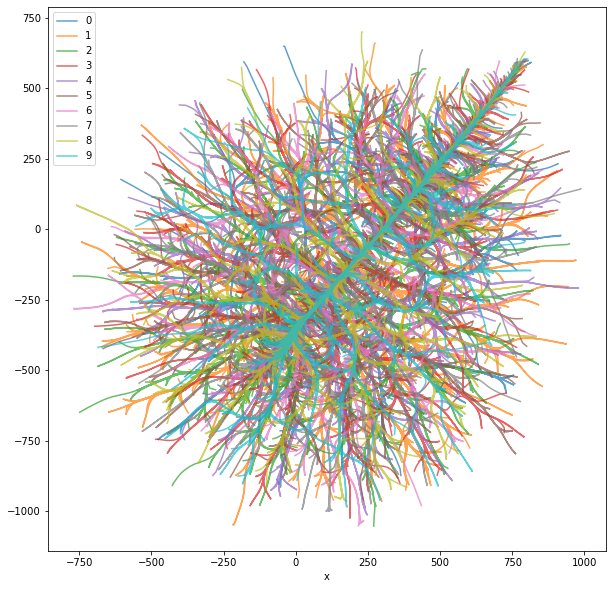}}
    \subfloat[t-SNE overall inter-layer evolution]{\includegraphics[width=0.5\textwidth]{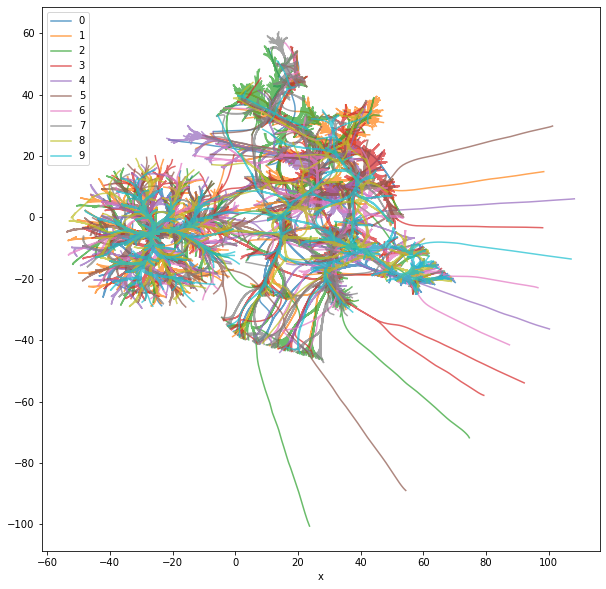}}
    \caption{Inter-layer evolution of MLP after training using SVHN dataset.}
    \label{fig:inter_layer_evolution_svhn_mlp}
\end{figure}

\begin{figure}[ht!]
    \centering
    \subfloat[IVHD $\mathbb{A}_{p}\{1\}$.]{\includegraphics[width=0.25\textwidth]{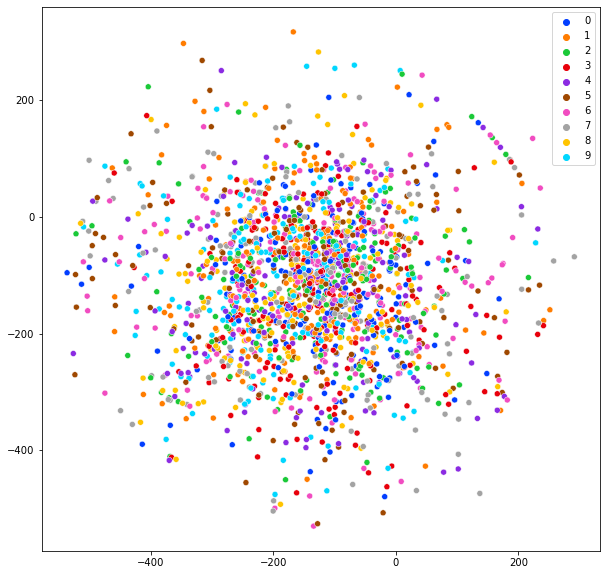}}
    \subfloat[IVHD $\mathbb{A}_{p}\{2\}$.]{\includegraphics[width=0.25\textwidth]{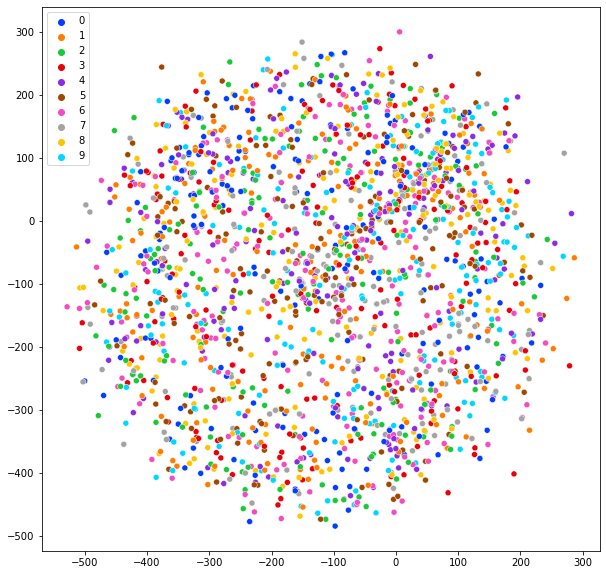}}
    \subfloat[IVHD $\mathbb{A}_{p}\{3\}$.]{\includegraphics[width=0.25\textwidth]{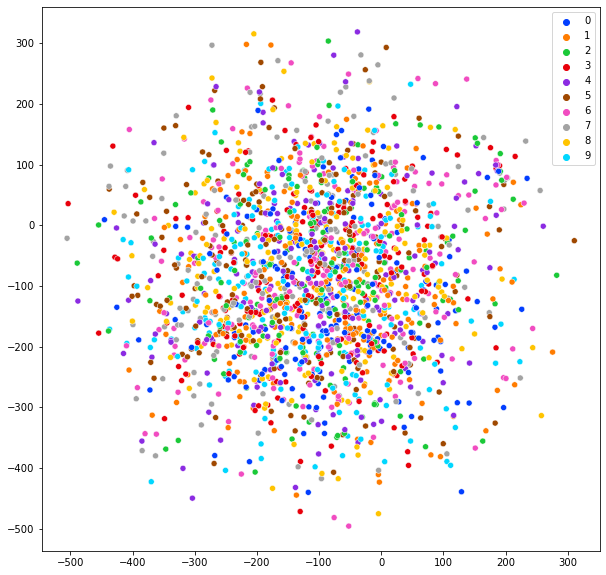}}
    \subfloat[IVHD $\mathbb{A}_{p}\{4\}$.]{\includegraphics[width=0.25\textwidth]{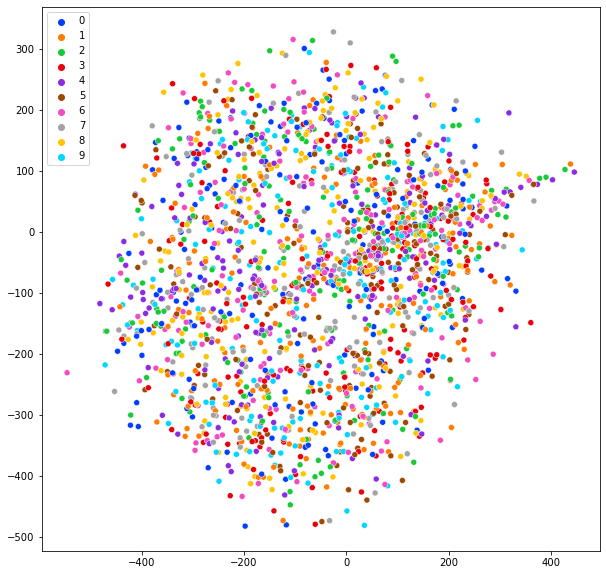}}
    \hfill
    \subfloat[t-SNE $\mathbb{A}_{p}\{1\}$.]{\includegraphics[width=0.25\textwidth]{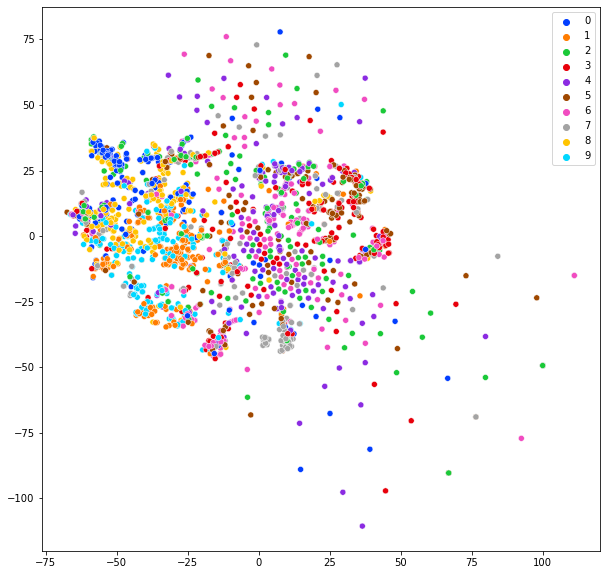}}
    \subfloat[t-SNE $\mathbb{A}_{p}\{2\}$.]{\includegraphics[width=0.25\textwidth]{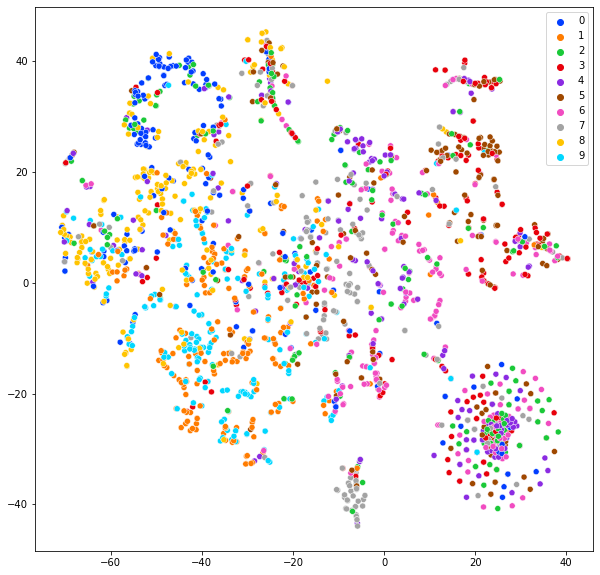}}
    \subfloat[t-SNE $\mathbb{A}_{p}\{3\}$.]{\includegraphics[width=0.25\textwidth]{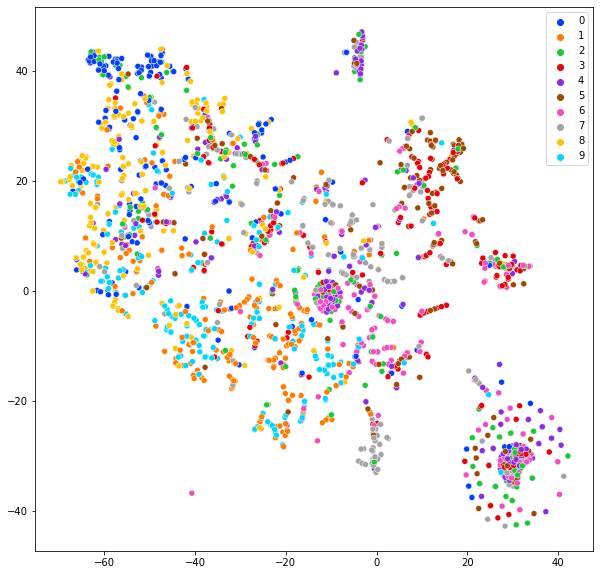}}
    \subfloat[t-SNE $\mathbb{A}_{p}\{4\}$.]{\includegraphics[width=0.25\textwidth]{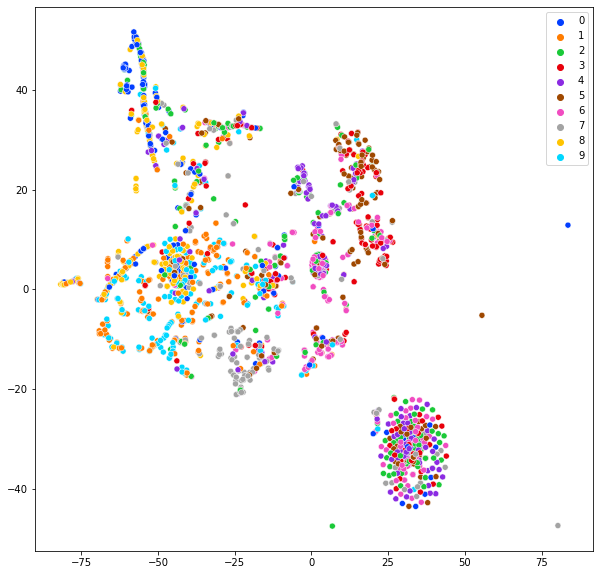}}
    \hfill
    \subfloat[IVHD overall inter-layer evolution.]{\includegraphics[width=0.5\textwidth]{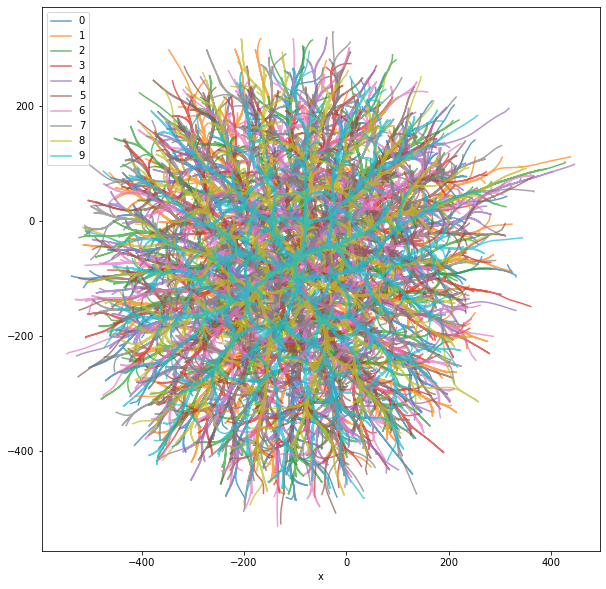}}
    \subfloat[t-SNE overall inter-layer evolution]{\includegraphics[width=0.5\textwidth]{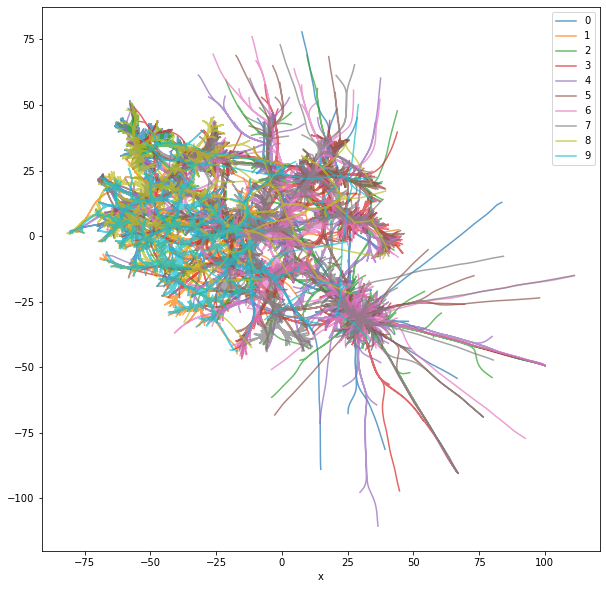}}
    \caption{Inter-layer evolution of MLP after training using Cifar-10 dataset.}
    \label{fig:inter_layer_evolution_cifar_mlp}
\end{figure}

Figures \ref{fig:inter_layer_evolution_mnist_mlp}, \ref{fig:inter_layer_evolution_svhn_mlp}, \ref{fig:inter_layer_evolution_cifar_mlp} present the evolution between layers for the test subsets (after training). The merged image summarizes a sequence of four projections, one per hidden layer, shown as thumb nails. The colors of the paths encode the classes, we can also observe how the activation data "flows" through the four layers of the network, by comparing next-layer projections (and overall inter-layer evolution). For MNIST, the beam shapes show that the visual clusters are quite stable in all layers. Thus, the network has reached a fairly good separation between classes in the early layers. The gradients also show that some visual clusters become more compact in later layers and that some clusters move away from others (e.g., the brightness pattern in the purple cluster). 

\begin{figure}[ht!]
    \centering
    \subfloat[IVHD.]{\includegraphics[width=0.5\textwidth]{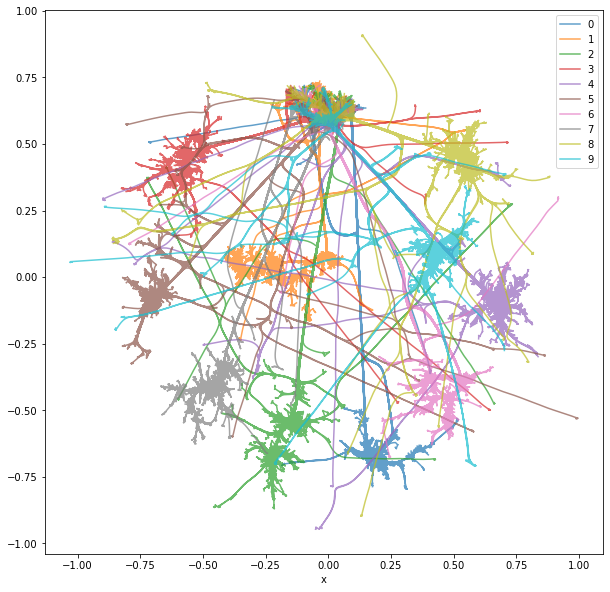}}
    \subfloat[IVHD.]{\includegraphics[width=0.5\textwidth]{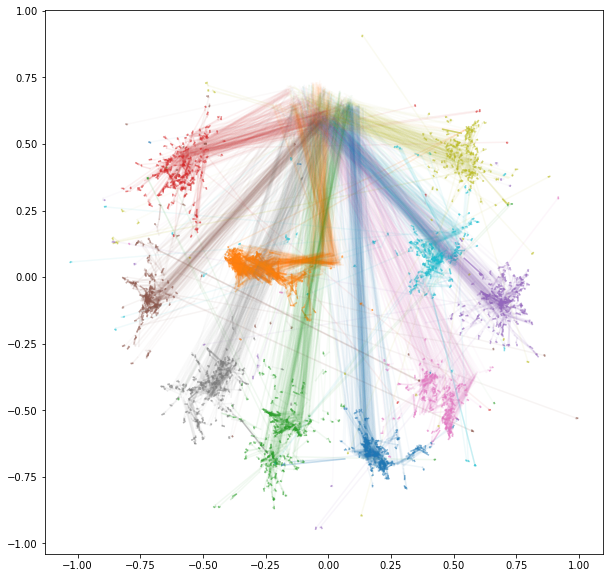}}
    \hfill
    \subfloat[t-SNE.]{\includegraphics[width=0.5\textwidth]{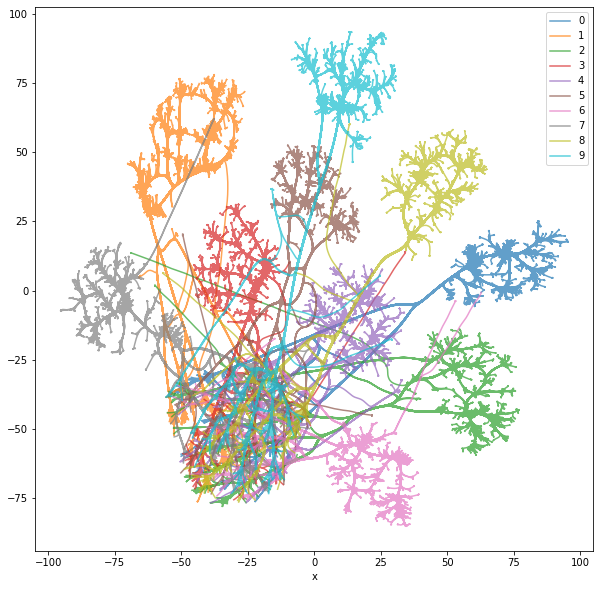}}
    \subfloat[t-SNE.]{\includegraphics[width=0.5\textwidth]{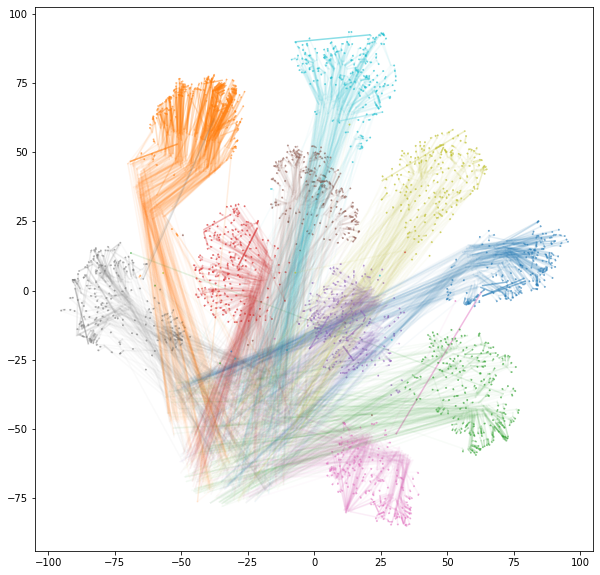}}
    \caption{Inter-epoch evolution, last CNN hidden layer. On the right side: trace plot, where trail parts show later epochs and the points are showing the final position obtained in the last epoch.}
    \label{fig:inter_epoch_evolution_mnist_cnn}
\end{figure}

\textbf{Inter-epoch evolution}: The same idea as described above can be employed to visualize inter-epoch evolution. However, as shown in \cite{rauber2016}, the results in this case are significantly more difficult to interpret. This is due to a combination of large changes in the very first epoch, high intra-visual-cluster variance between epochs, and a much larger number of frames (typically hundreds) to be summarized. For this reason, different strategy is used to visualize inter-epoch evolution. Consider again the sequence $\mathbb{A}\{1\},\dots, \mathbb{A}\{T\}$ of the activation sets. For inter-epoch evolution, $\mathbb{A}\{t\} \subset \mathbb{R}^{k}$ for a fixed \textit{k}, for all \textit{t}. Therefore, we can create a projection for the set $\cup_{k}\mathbb{A}\{t\}$, which contains activations for all epochs. As we compute a single projection, there is no spurious inter-frame variation. Figure \ref{fig:inter_epoch_evolution_mnist_cnn} shows the evolution between epochs for the last CNN hidden layer activations using this strategy, from epochs 0 to 100, in steps of 20 (12K points in total). It is interesting to note how the dimensionality reduction technique placed the points corresponding to earlier epochs in the center of the projection, considering that it does not explicitly receive this information.

\subsection{Exploring the relationships between artificial neurons}

As in \cite{rauber2016}, we introduce a projection of a neuron, where a point represents a neuron. Points are placed in 2D on the basis of the similarity between neurons. The dissimilarity $d_{i,j}$ between neurons $i$ and $j$ is defined as $d_{i,j} = 1- |r_{i,j}|$, where $r_{i,j}$ is the empirical correlation coefficient (Pearson) between neurons $i$ and $j$ in a dataset composed of layer-l activations (recall that each element of an activation vector is a neuron output). This metric captures both positive and negative linear correlations between pairs of neurons. From the pairwise dissimilarity matrix, a projection is computed using multidimensional scaling (MDS \cite{mds}). Although t-SNE is particularly concerned with preserving neighborhood relationships \cite{tsne}, MDS attempts to preserve global pairwise dissimilarities as much as possible, which is more appropriate in this scenario.

\begin{figure}[ht!]
    \centering
    \subfloat[Before training.]{\includegraphics[width=0.4\textwidth]{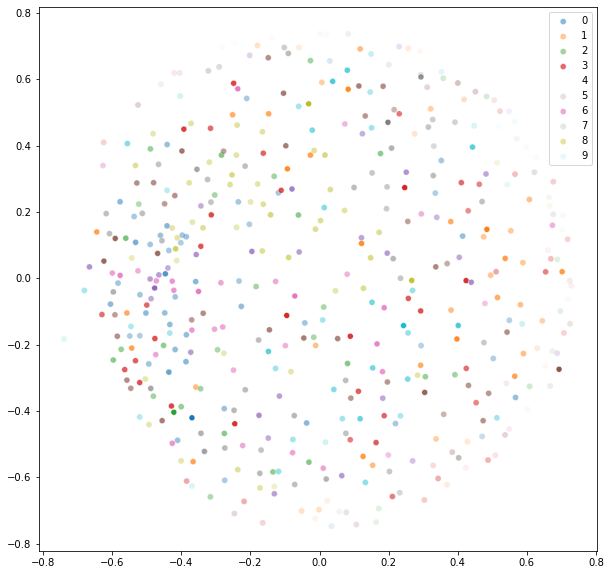}}
    \subfloat[After training.]{\includegraphics[width=0.4\textwidth]{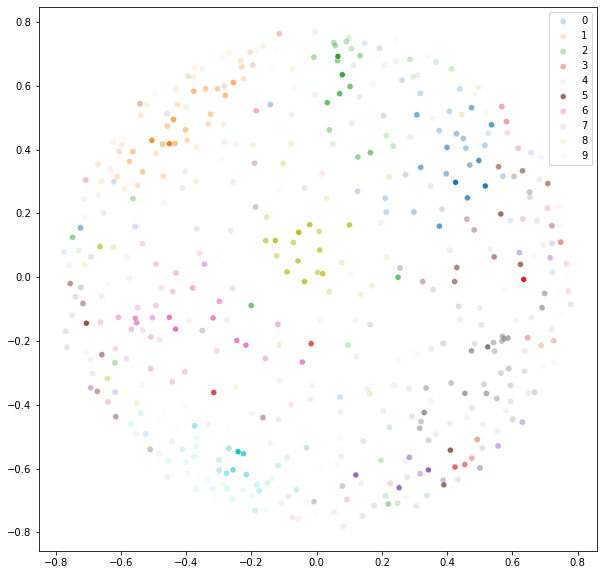}}
    \caption{Activation and neuron projections of last CNN hidden layer activations before (on the left) and after (on the right) training. Neuron projection colors show the neurons’ power to discriminate class 0 vs rest.}
    \label{fig:cifar_discriminative_power}
\end{figure}

\begin{wrapfigure}{r}{5.5cm}
    \centering
    \includegraphics[width=5.5cm]{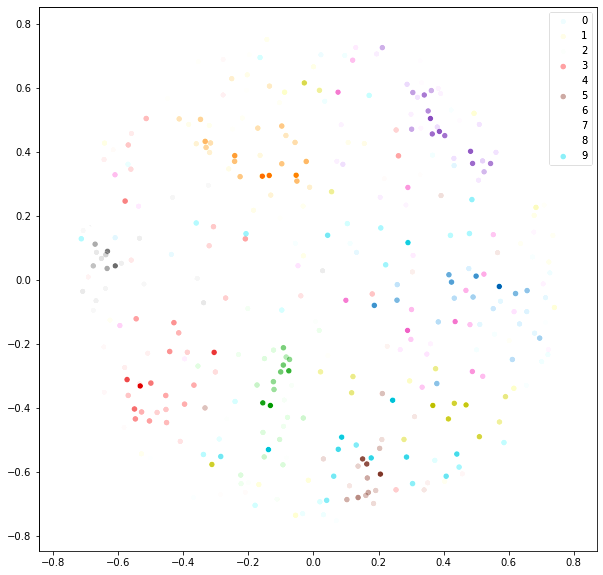}
    \caption{Discriminative neuron map of last CNN hidden layer activations after training, SVHN test subset.}
    \label{fig:svhn_discriminative_neuron_map}
\end{wrapfigure}

Figure \ref{fig:svhn_discriminative_neuron_map} shows the discriminative neuron map for the SVHN test subset, the last activations of the hidden layer, after training. The presence of compact visual clusters shows how the entire set of neurons can be (almost) partitioned into groups with related discriminative functions (specializations), even though the neuron projection is created without any class information. Neuronal activation and projections can be combined to elucidate the role of particular neurons.

\begin{figure}[ht!]
    \centering
    \subfloat[Before training.]{\includegraphics[width=0.4\textwidth]{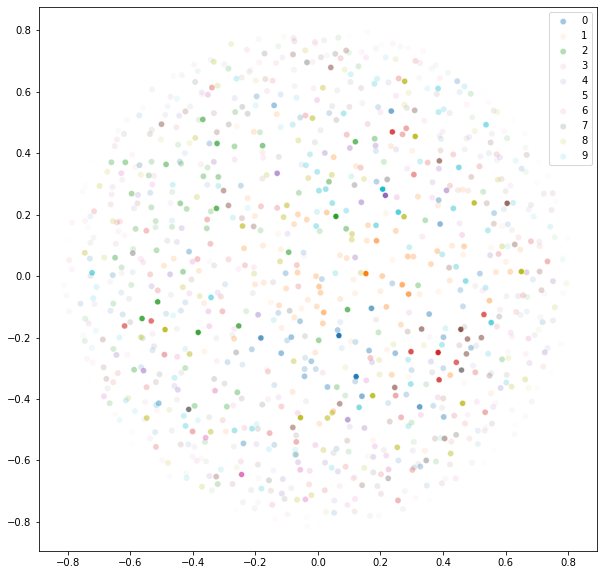}}
    \subfloat[After training.]{\includegraphics[width=0.4\textwidth]{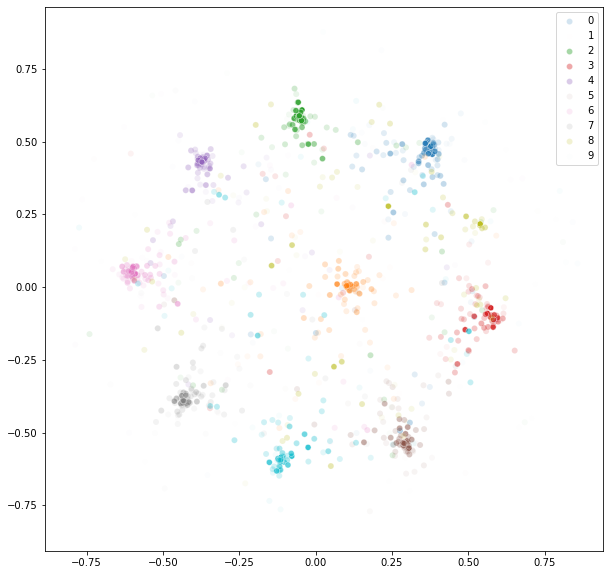}}
    \caption{Activation and neuron projections of last MLP hidden layer activations before (on the left) and after (on the right) training. Neuron projection colors show the neurons’ power to discriminate class 0 vs rest.}
    \label{fig:mnist_mlp_discriminative_power}
\end{figure}

In Figure \ref{fig:mnist_mlp_discriminative_power} the discriminative neuron map for the MNIST test subset is presented (from the last activations of the hidden layer, after training). We can clearly see which neurons are associated to which class. For MNIST dataset, the results presented are the most intuitive and informative. 

In this chapter, we show how dimensionality reduction can be used to visualize the relationships between learned representations (1) and between neurons (2) in artificial neural networks. Concerning the first task, the visualizations support the identification of confusion zones, outliers, and clusters in the internal representations computed by such networks. Separately, it was shown how to visually track inter-layer and inter-epoch evolution of learned representations. Concerning the second task, it was presented how to inspect the relationships between neurons and classes (specialization) and similarity between neurons.

\chapter{Conclusions}

In this work, we explored areas of how data embedding methods can be analyzed using a common framework and showed that \textit{diemensioanlity reduction (DR)} can be perceived as visualization of unweighted $k$NN-graphs, constructed for the source $\mathbf{X}$ data. The framework made it possible to verify how knowledge extracted from different DE algorithms can help to develop the most efficient and memory-saving method for interactive visualization of high-dimensional data. In our study, we focus on different types of data, starting with synthetic datasets that can provide insight into very specific properties of a method and ending with large datasets to verify the effectiveness and precision of each method. 

Firstly, based on recent results posted in a paper currently under review by the Promoter, we claim that t-SNE and UMAP are in fact computationally consistent with each other. Up to minor changes, one could have presented the theoretical foundation of UMAP and implemented it with the tSNE algorithm or vice versa. Furthermore, t-SNE can be simplified step by step to the most simplistic IVHD method.

Second, we show that IVHD with binary distances is the most time\&memory optimal and is capable of analyzing truly large datasets and graphs in a reasonable amount of time \cite{dzwinel2017ivga}, which is not achievable by other baseline methods. For this purpose, a very extensive analysis of methods of visualizing high-dimensional data was performed. We evaluated five well-recognized in the research community data embedding algorithms: t-SNE, LargeVis, UMAP, TriMAP and PaCMAP. The results obtained suggest that while all of these methods generate visualizations at a high level (in terms of metrics such as: DR quality, kNN gain, Shepard diagram), baseline methods are significantly inferior to IVHD in terms of efficiency. Through the use of GPU-CUDA environment, we made it possible to visualize datasets containing millions of data points in a reasonable amount of time. We also proposed several improvements to the IVHD method that did not impose a computational burden, but improved its visualization efficiency. Specifically:

\begin{itemize}
    \item use the $L1$ norm in the final steps of algorithm calculation,
    \item utilize the reverse neighbor mechanism in the final steps of algorithm calculation (along with the norm $L1$),
    \item add auto-adaptation of time step during simulation,
    \item we implemented IVHD in the \textit{viskit} visualization library \cite{viskit}, which contains different DR and optimization methods, which creates the possibility to mix different optimizers "on fly", 
\end{itemize}

For datasets that are not that large, IVHD generates visualizations whose global values do not deviate from the average achieved by other methods, and in some cases the global properties are the best (smallNORB, mammoth dataset, synthetically generated ball in a sphere). However, IVHD has two shortcomings. First, it squeezes classes (crowding) (but it also preserves the neighborhood well \cite{dzwinel2017}). It can be loosened at the end of the simulation by changing the parameters or using another method locally. Second, it generates "dust" between classes. These points are the result of reaching equilibrium of the forces that affect them. One way to improve this was to add a mechanism that leaves only the interaction with reverse neighbors in the last steps of the algorithm.

In the third part of this thesis, we introduced a meta-platform that allows any DR method to be used in a supervised fashion. It is highly abstracted, allowing for an arbitrary classifier and an unsupervised variant of any embedding method to obtain a supervised-like process of visualization in 2-D (or 3-D) space. The supervised variant is not exactly the type of method that is desirable in data visualization research. Much more preferable are methods operating in an unsupervised fashion, which are powerful enough to analyze datasets in real time. This is because much of the data on which machine learning operates is unlabeled and its internal structure unknown. Of course, there is also the problem of the availability of labeled data because sets of such data can still be difficult to access.

Finally, we use our highly efficient IVHD method to see how it can be applied to inspect and interpret the training process in ANN. We recognize that building transparent machine learning systems is a convergent approach to both extracting new domain knowledge and performing model validation. As machine learning is increasingly applied to real-world decision making, the need for a transparent machine learning process will continue to grow. Using IVHD, we show that relationships between learned representations and between neurons can be studied, and efficient and time\&memory optimal high-dimensional data visualization methods can be very useful in this interpretation process. Furthermore, the entire DNN visualization framework can be further improved by efficiently implementing it on modern hardware, for example, GPGPUs (\textit{IVHD-CUDA}).

In future work, we plan to further investigate the visualization and interpretation of DNNs. In particular, we are eager to develop and explore a unified procedure for real-time analysis of the training process, which could greatly contribute to the possibility of optimizing deep network architectures. Furthermore, although dimensionality reduction is among the most scalable methods for high-dimensional data visualization, it still has some problems. For this reason, we plan to improve our CPU implementation of IVHD method, which could be further accelerated by using low-level AVX instructions or more advanced multithreading operations. We also plan to further investigate supervised visualization methods and our meta-platform. Perhaps supervised training variants of DE methods would be possible alongside DNN, which could somehow elevate the real-time visualization of the training process. We hope that these approaches will further improve the performance of large high-dimensional data visualization and DNN applied to sparse, high-dimensional, unstructured data.

\appendix
\chapter{Datasets}
\label{appendix_datasets}

In this appendix, we list and briefly describe the datasets used in this work. These datasets are summarized in Table A.1. The experiments in Chapter 5 and Appendix B were conducted based on popular image and English-language text datasets, namely -MNIST, SmallNORB, 20 Newsgroups, and RCV1. Some basic pre-processing was applied to the datasets (\ref{sec:dataset_preprocessing}). In the case of RCV, only the 10 largest components (classes) were taken into account during visualization to remove the huge granularity. GPU methods were also tested using a YAHOO dataset, which contains questions and answers from the YAHOO service. In addition, a set of small synthetic datasets was generated to allow simple benchmarking of the embedding methods.

\begin{table}[ht]
\small
\caption{The list of baseline datasets.}
\vspace{0.2cm}
    \begin{tabular}{|p{2.5cm}|c|c|c|p{7.25cm}|}
    \hline
    \multicolumn{1}{|c|}{Dataset} & \multicolumn{1}{c|}{$N$} & \multicolumn{1}{c|}{$M$} & \multicolumn{1}{c|}{$K$} & \multicolumn{1}{c|}{Short description}\\ \hline
    MNIST & 784 & 70 000  & 10 & Well balanced set of grayscale images of handwritten digits.\\ \hline
    Fashion-MNIST & 784 & 70 000 & 10 & More difficult MNIST version. Instead of handwritten digits it consists of apparel images.\\ \hline
    Extended-MNIST Letters & 784 & 103 600  & 26 & Merges a balanced set of the uppercase and lowercase letters into a single set.\\ \hline
    Small NORB & 2048 & 48 600 & 5 & It contains stereo image pairs of 50 uniform-colored toys under 18 azimuths, 9 elevations, and 6 lighting conditions.\\ \hline
    20-NG & 5000 & 18846 & 20 & Collection of approximately 20 000 newsgroup documents corresponding to a different topic.\\ \hline
    RCV-Reuters & 30 & 804 409 & 8 & Corpus of press articles preprocessed to 30D by PCA.\\ \hline
    SVHN & 1024 & 63200 & 10 & 32x32 images of real-world numbers obtained from house numbers in Google Street View images.\\ \hline
    CIFAR-10 & 1024 & 60000 & 10 & 32x32 images of real-world objects (e.g. airplanes, birds, dogs).\\ \hline
    YAHOO & 100 & 1.4 million & 10  & Questions and answers from YAHOO. The answers service preprocessed with FastText \cite{fasttext}.\\ \hline
    Amazon20M & 100 & 20 million & 5  & Book reviews from Amazon. The reviews were preprocessed with FastText \cite{fasttext}.\\ \hline
    \end{tabular}
\end{table}

\newpage

\section{Datasets}

\appendixsubsection{MNIST}

The MNIST dataset \cite{mnist1998} is a widely used benchmark for machine learning algorithms. It consists of images of handwritten digits (Figure A.1). In the original dataset, the images are represented by 256 grayscale levels, but in this work we used rescaled pixel intensities to the $[0, 1]$ interval.

\begin{figure}[ht!]
    \centering
    \includegraphics[width=0.8\textwidth]{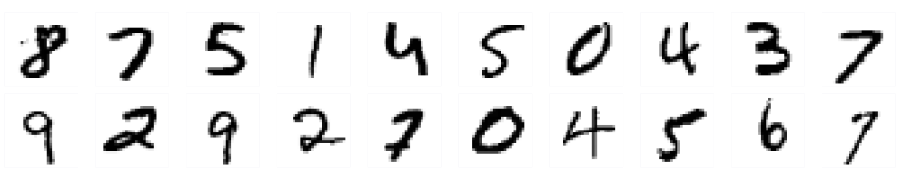}
    \caption{Examples of MNIST dataset.}
    \label{fig:appendix_datasets_mnist}
\end{figure}

\appendixsubsection{Fashion-MNIST}

Fashion-MNIST \cite{fmnist2017} is a dataset of Zalando's product images, consisting of a training set of 60,000 examples and a test set of 10,000 examples. Each example is a 28x28 grayscale image associated with a label from 10 classes. Zalando intends Fashion-MNIST to serve as a direct drop-in replacement for the original MNIST dataset for benchmarking machine learning algorithms. It shares the same image size and structure as the training and testing splits.

\begin{figure}[ht!]
    \centering
    \includegraphics[width=0.6\textwidth]{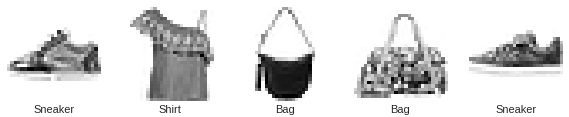}
    \caption{Examples of FMNIST dataset.}
    \label{fig:appendix_datasets_fmnist}
\end{figure}

\appendixsubsection{Extended-MNIST Letters}

The EMNIST dataset \cite{emnist2017} is a set of handwritten character digits converted to a 28x28 pixel image format and a data set structure that corresponds directly to the MNIST dataset. The Letters data set merges a balanced set of upper- and lowercase letters into a single 26-class set.

\begin{figure}[ht!]
    \centering
    \includegraphics[width=0.6\textwidth]{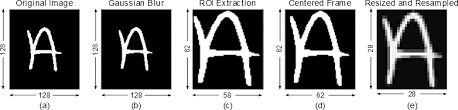}
    \caption{Examples of EMNIST dataset.}
    \label{fig:appendix_datasets_emnist}
\end{figure}

\appendixsubsection{smallNORB}

The smallNORB dataset \cite{smallnorb2005} is a dataset used for the recognition of 3D objects of its shape. It contains images of 50 toys belonging to five generic categories: four-legged animals, human figures, airplanes, trucks, and cars. The objects were imaged by two cameras under six lighting conditions, nine elevations (30 to 70 degrees every 5 degrees) and 18 azimuths (0 to 340 every 20 degrees).

\begin{figure}[ht!]
    \centering
    \includegraphics[width=0.8\textwidth]{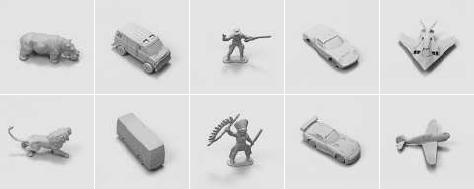}
    \caption{Examples of smallNORB images.}
    \label{fig:appendix_datasets_smallnorb}
\end{figure}

\appendixsubsection{20 News Groups}

The 20 Newsgroups dataset \cite{20newsgroups} is a collection of approximately 20 000 newsgroup documents. The dataset is organized into 20 different newsgroups, each corresponding to a different topic. Some of the newsgroups are very closely related to each other (e.g. \textit{comp.sys.ibm.pc.hardware} and \textit{comp.sys.mac.hardware}), while others are highly unrelated (e.g. \textit{misc.forsale} and \textit{soc.religion.christian}).

\begin{table}[ht!]
\caption{List of the 20 newsgroups, partitioned (more or less) according to topic.}
\vspace{0.2cm}
\begin{tabular}{|l|l|l|}
\hline
\begin{tabular}[c]{@{}l@{}}comp.graphics\\ comp.os.ms-windows.misc\\ comp.sys.ibm.pc.hardware\\ comp.sys.mac.hardware\\ comp.windows.x\end{tabular} & \begin{tabular}[c]{@{}l@{}}rec.autos\\ rec.motorcycles\\ rec.sport.baseball\\ rec.sport.hockey\end{tabular} & \begin{tabular}[c]{@{}l@{}}sci.crypt\\ sci.electronics\\ sci.med\\ sci.space\end{tabular}         \\ \hline
misc.forsale                                                                                                                                        & \begin{tabular}[c]{@{}l@{}}talk.politics.misc\\ talk.politics.guns\\ talk.politics.mideast\end{tabular}     & \begin{tabular}[c]{@{}l@{}}talk.religion.misc\\ alt.atheism\\ soc.religion.christian\end{tabular} \\ \hline
\end{tabular}
\end{table}

\appendixsubsection{RCV-Reuters}

In 2000, Reuters Ltd. made available a large collection of Reuters News stories for use in the research and development of natural language processing, information retrieval, and machine learning systems. This corpus, known as the "Reuters Corpus, Volume 1" or RCV1 \cite{reuters2004}, is significantly larger than the older and well-known Reuters-21578 collection, which is widely used in the text classification community. It is an archive of more than 800,000 manually categorized newswire stories made available by Reuters, Ltd. for research purposes.

\appendixsubsection{Street View House Numbers (SVHN)}

SVHN \cite{svhn} is a real-world image dataset for developing machine learning and object recognition algorithms with a minimal data pre-processing and formatting requirement. It can be seen as similar in flavor to MNIST (e.g., the images are of small cropped digits), but incorporates an order of magnitude more labeled data (over 600,000 digit images) and comes from a significantly harder, unsolved, real-world problem (recognizing digits and numbers in natural scene images). It is obtained from house numbers in Google Street View images.

\begin{figure}[ht!]
    \centering
    \includegraphics[width=0.8\textwidth]{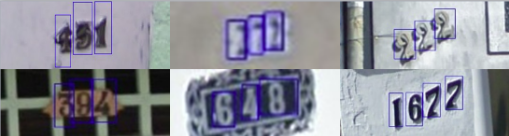}
    \caption{Examples of SVHN images.}
    \label{fig:appendix_datasets_svhn}
\end{figure}

\appendixsubsection{CIFAR-10}

CIFAR-10 \cite{cifar10} consists of 60000 32x32 colored images in 10 classes, with 6000 images per class. The dataset is divided into five training batches and one test batch, each with 10000 images. The 10 different classes represent airplanes, cars, birds, cats, deer, dogs, frogs, horses, ships, and trucks. The classes are mutually exclusive, which means that there is no overlap between different classes (for example, automobiles and trucks).

\begin{figure}[ht!]
    \centering
    \includegraphics[width=\textwidth]{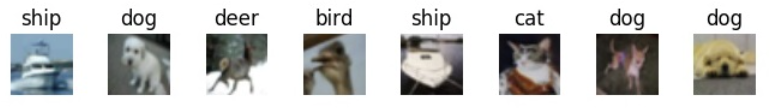}
    \caption{Examples of CIFAR-10 images.}
    \label{fig:appendix_datasets_cifar10}
\end{figure}

There is also a variant of CIFAR-10 called CIFAR-100. It is like CIFAR-10, except that it has 100 classes containing 600 images each, which provides much greater diversity. 

\appendixsubsection{Amazon3M and Amazon10M}

Amazon3M is sampled from all Amazon Reviews (233.1 million), which contains 3 million 100-D vectors. Amazon10M was sampled from Amazon20M. The description of whole Amazon dataset is placed below in the Amazon20M section.

\appendixsubsection{Amazon20M}

This dataset contains product reviews and metadata from Amazon, including 233.1 million reviews spanning May 1996 - October 2018. Reviews include ratings, text, helpfulness votes, descriptions, category information, price, brand, image features, and links. Amazon20M contains reviews from the Books section.

\begin{verbatim}
{
    "reviewerID": "A2SUAM1J3GNN3B",
    "asin": "0000013714",
    "reviewerName": "J. McDonald",
    "helpful": [2, 3],
    "reviewText": "I bought this for my husband who plays the piano.
    He is having a wonderful time playing these old hymns. 
    Great purchase though!",
    "overall": 5.0,
    "summary": "Heavenly Highway Hymns",
    "unixReviewTime": 1252800000,
    "reviewTime": "09 13, 2009"
}
\end{verbatim} Snippet A.1: Sample of the Amazon dataset record (\textit{reviewText} is later preprocessed by FastText \cite{fasttext}.)

\newpage

\appendixsubsection{Synthetic datasets}

We prepared a set of artificially generated "simple" data sets whose internal data structure gives insight into the quality of embeddings created by different methods.

\begin{center}
\begin{table}[ht!]
        \centering
        \begin{tabular}{|p{5cm}|p{8.5cm}|}
        \hline
        \multicolumn{1}{|c|}{Description} & \multicolumn{1}{c|}{2-D and 3-D example view}\\ \hline
        \vspace{-3cm} An N-dimensional ball divided in half. & \hspace{1cm} \includegraphics[width=3cm]{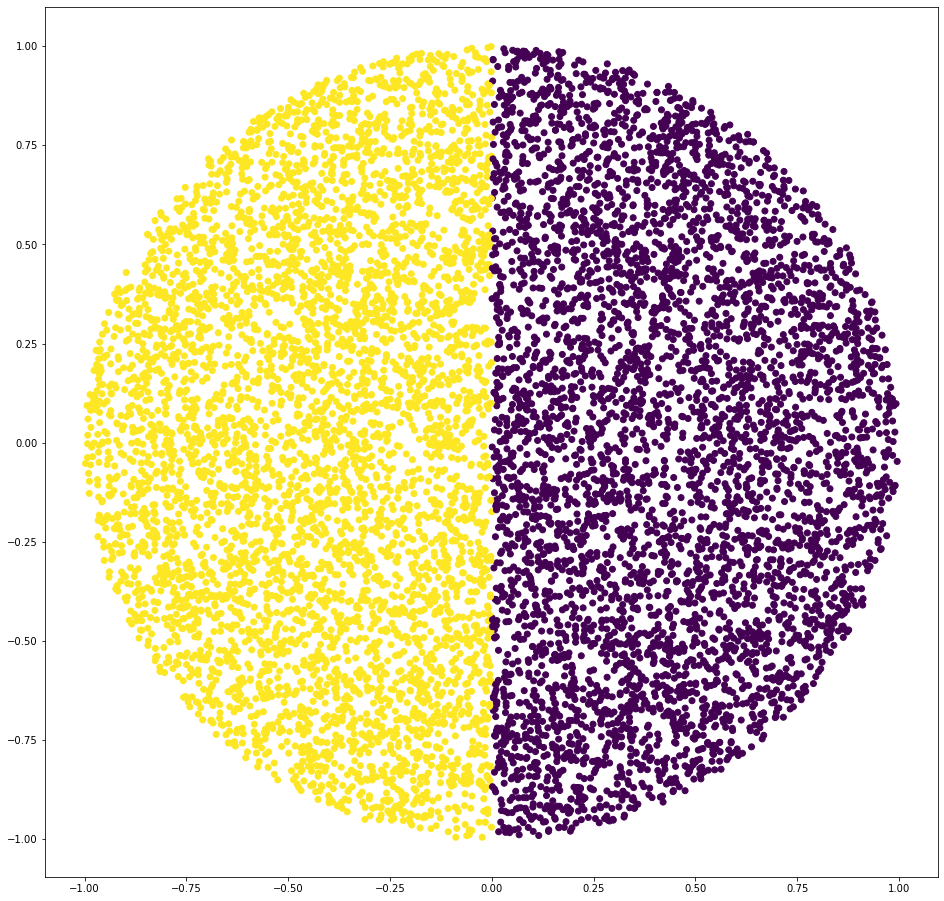} \hspace{0.5cm} \includegraphics[width=3cm]{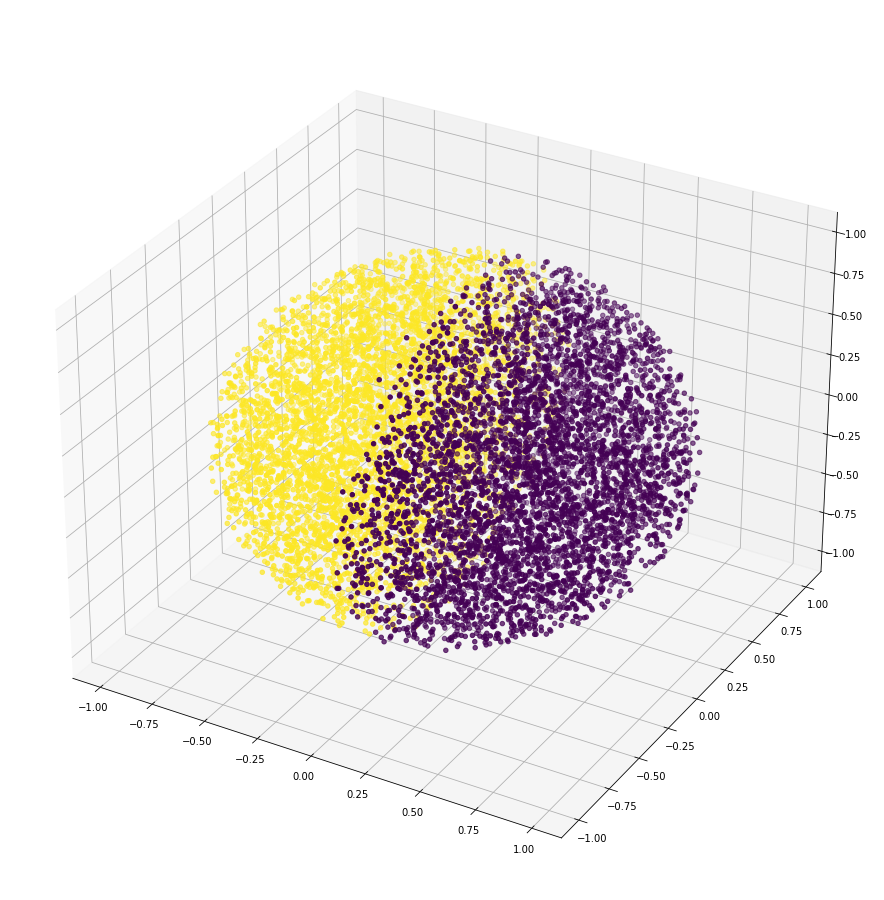} \\ \hline
        \vspace{-3cm}An N-dimensional ball inside a sphere. & \hspace{1cm} \includegraphics[width=3cm]{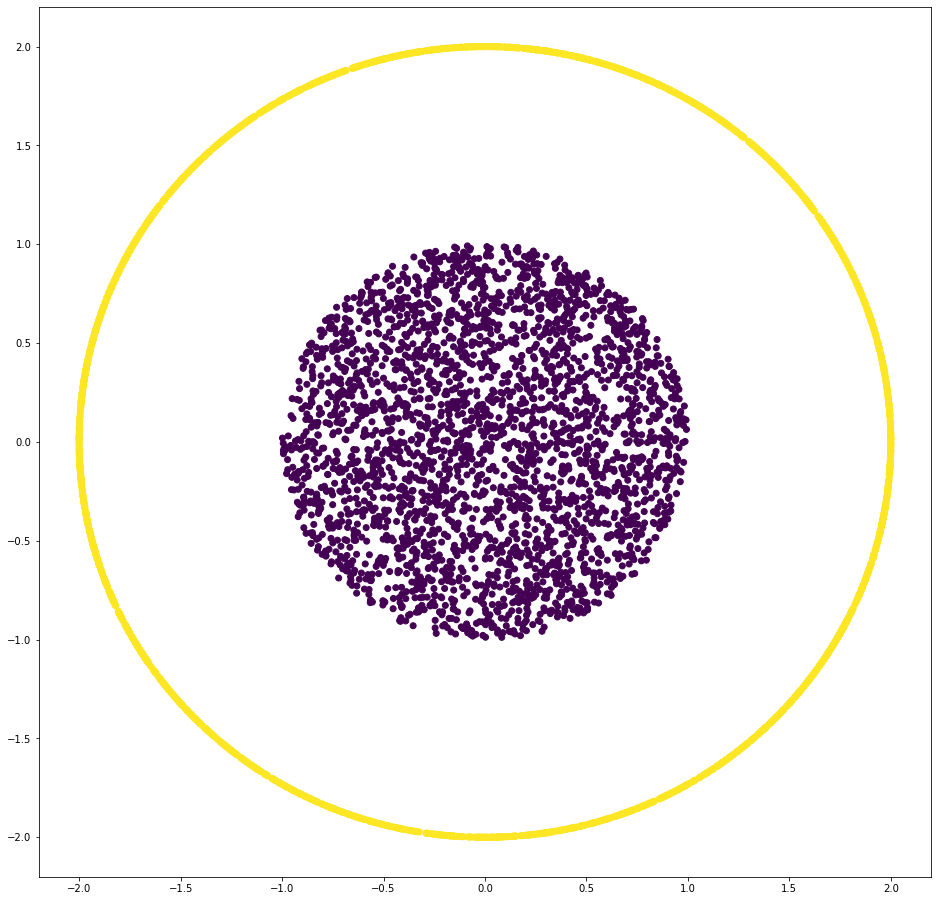} \hspace{0.5cm} \includegraphics[width=3cm]{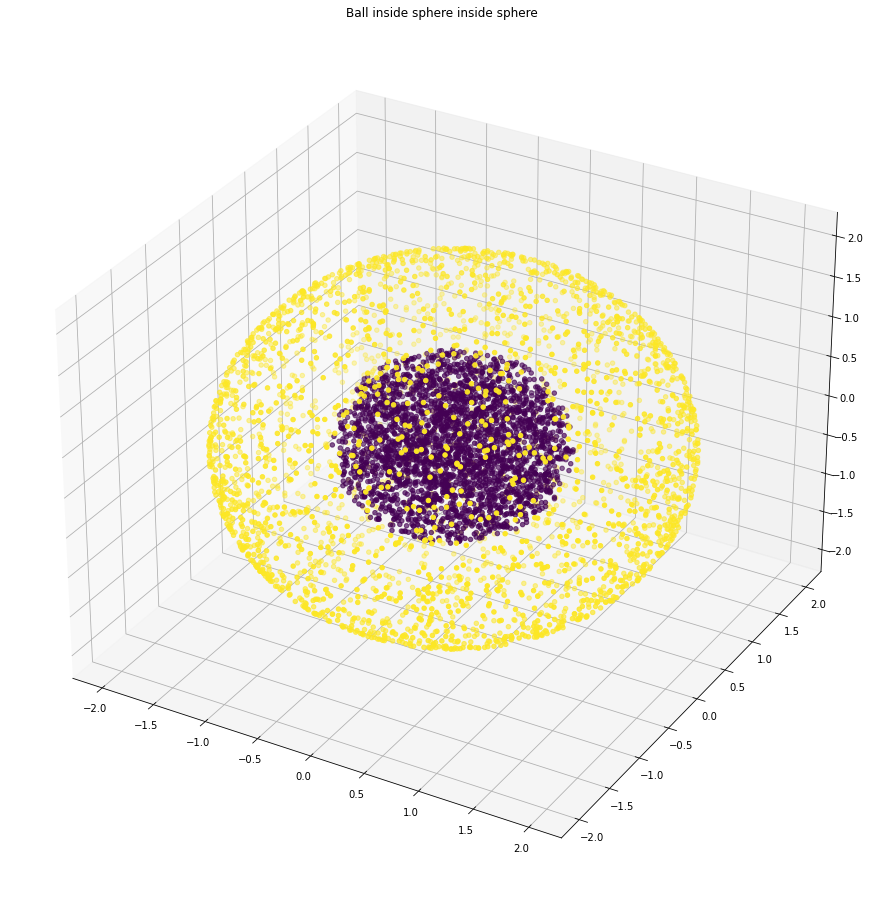}\\ \hline
        \vspace{-3cm} An N-dimensional ball inside two spheres. & \hspace{1cm} \includegraphics[width=3cm]{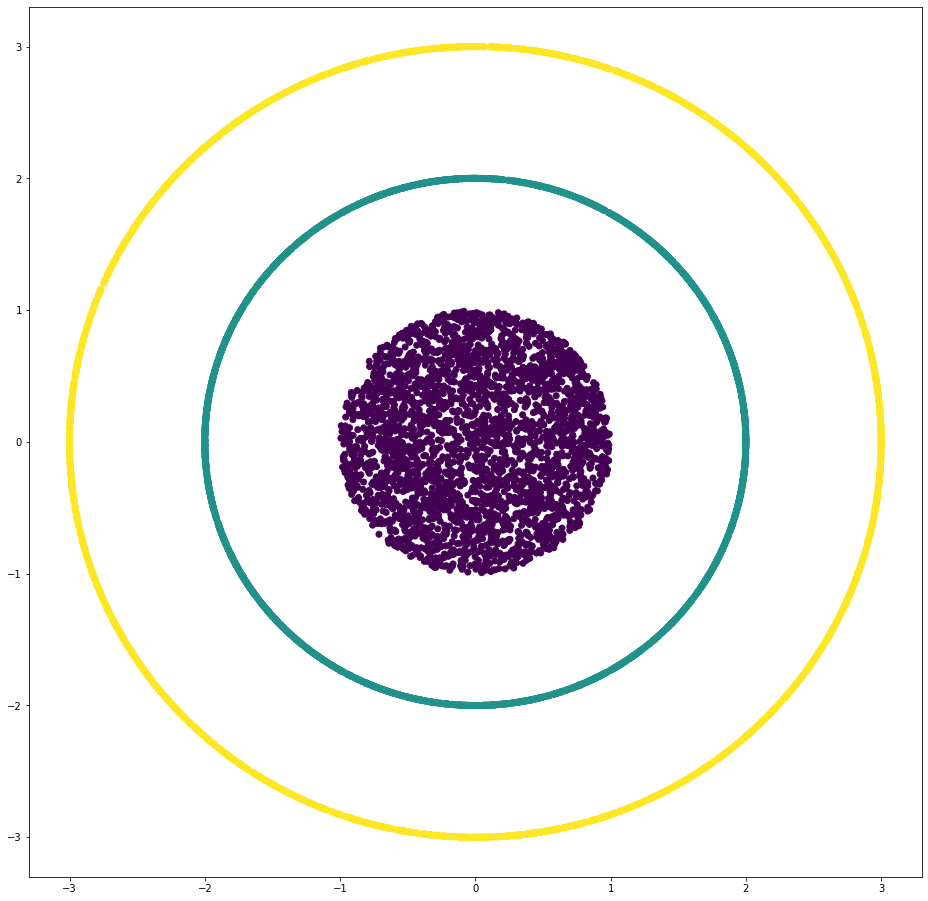} \hspace{0.5cm} \includegraphics[width=3cm]{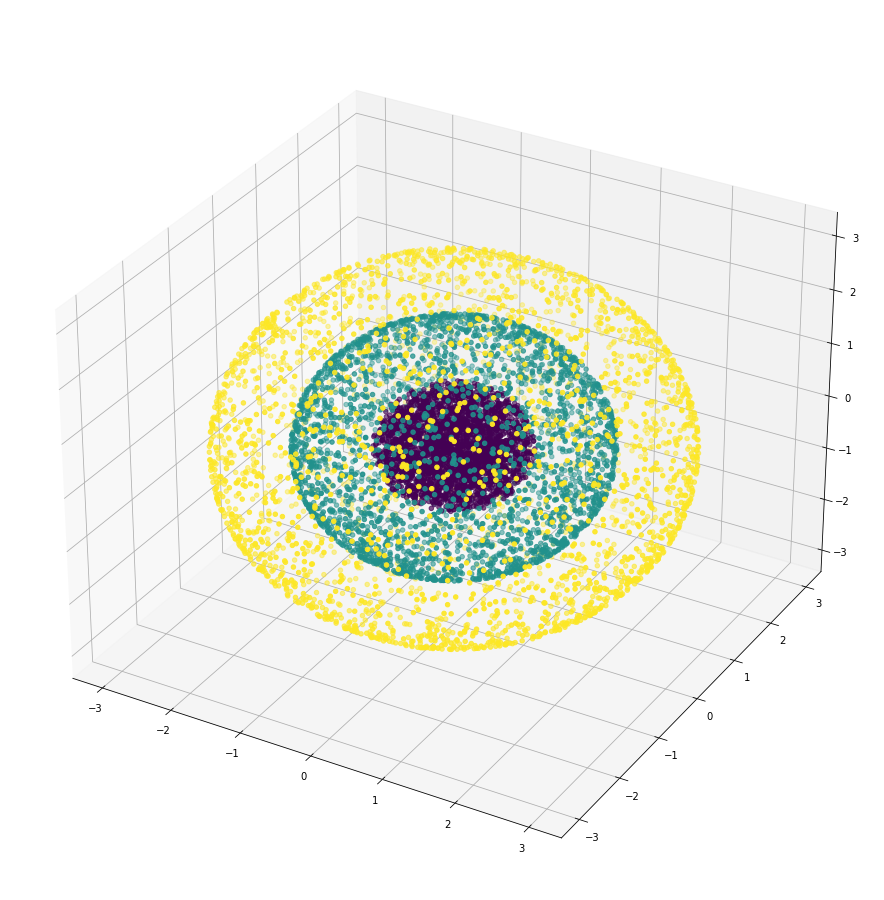}  \\ \hline
        \vspace{-3cm} An N-dimensional ball inside an N-dimensional empty ball. & \hspace{1cm} \includegraphics[width=3cm]{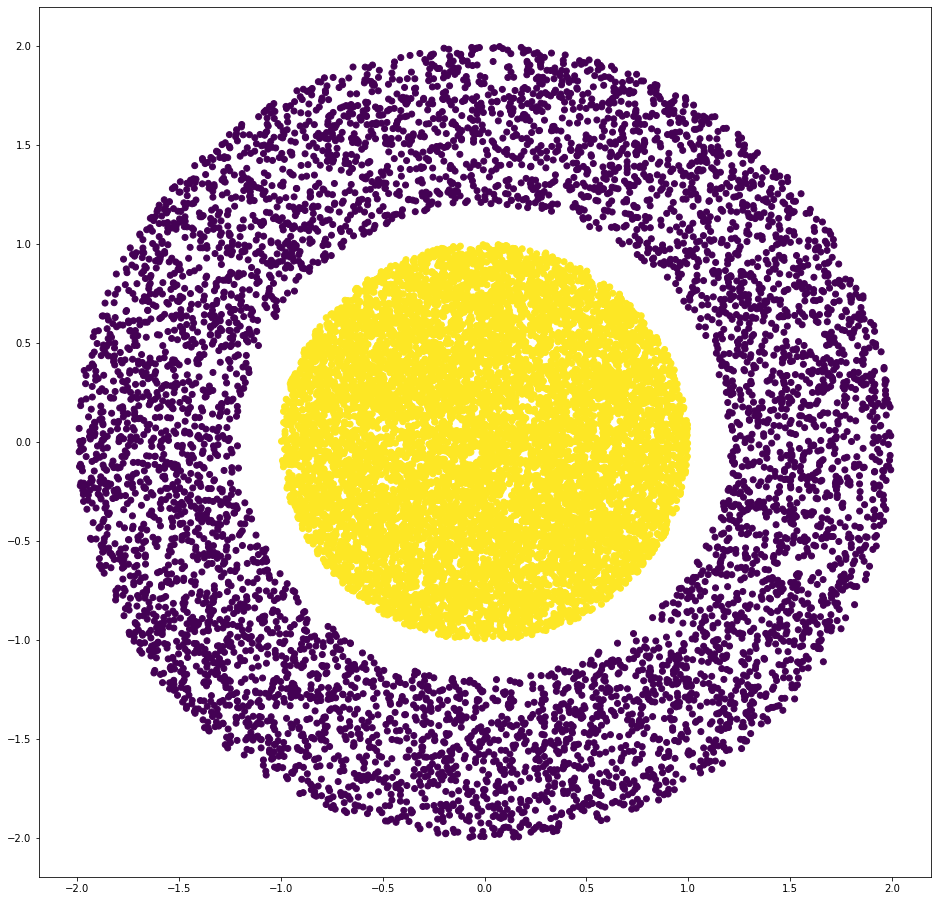} \hspace{0.5cm} \includegraphics[width=3cm]{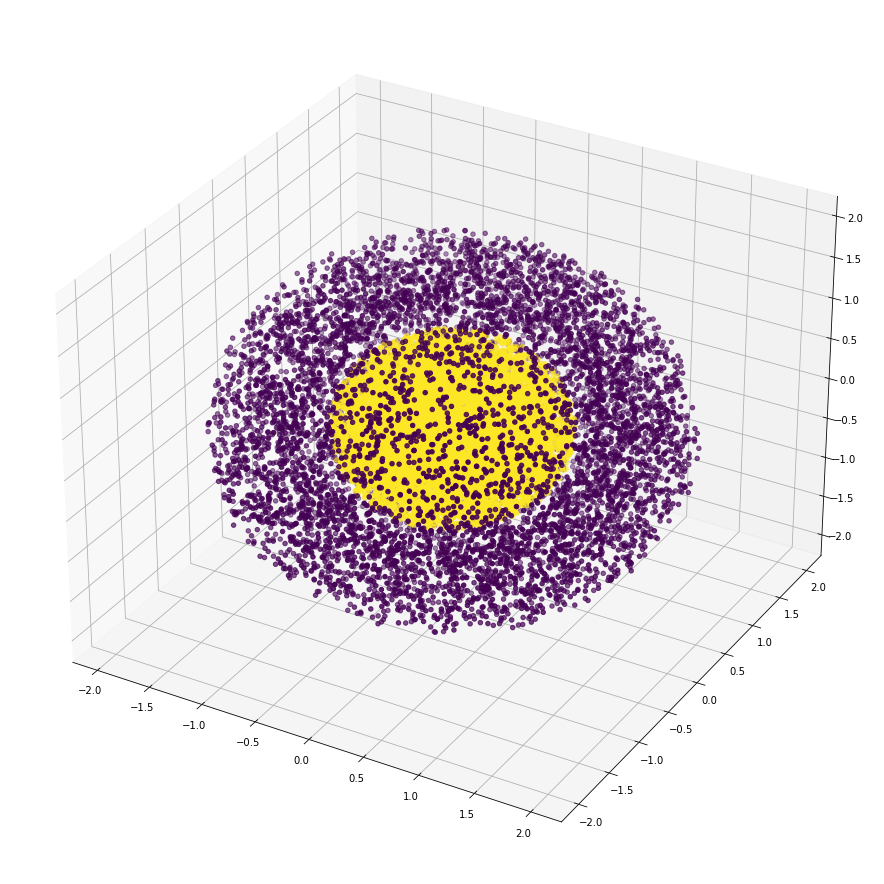}  \\ \hline
        \vspace{-3cm} Mammoth dataset. & \hspace{2.3cm} \includegraphics[width=4cm]{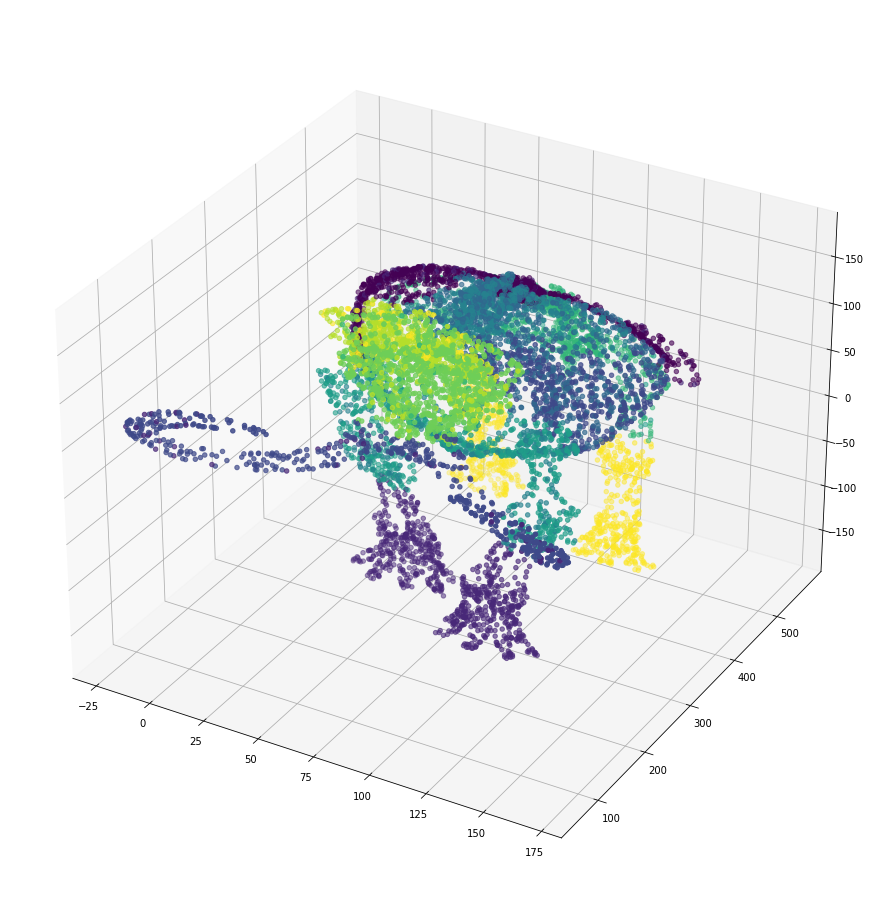} \\ \hline
        \end{tabular}
        \caption{Artificially generated ("synthetic") N-D baseline datasets and its 3-D (and 2-D) example views.}
        \label{tab:artificially_generated_datasets}
    \end{table}
\end{center}

\chapter{Visualizations, timings and methods parametrization}
\label{appendix_visualizations_timings_and_methods_parametrization}

In this appendix, we include additional visualizations (for datasets not presented in the main body of dissertation), timings achieved for both CPU and GPU implementation of different methods and thorough parametrization, that was used for methods.

\begin{table}[ht!]
\small
\caption{Glossary of the parameters of the DE algorithms used in the experiments. All IVHD parameters are displayed. We also include general parameters used for UMAP, t-SNE, TriMAP, and PaCMAP. All algorithms except IVHD require matching a considerably higher number of parameters (about 10) \cite{bhsne-gpu,atsne-git}), so due to brevity only the most important ones are given, that is, $perplexity$, $nn$, $ni$ (number of inlier points for triplet constraints).}
\begin{center}
\begin{tabular}{|l|l|l|l|l|l|l|l|ll}
\hline
Method & Dataset & Perplexity & ni & nn & rn & c    & Metric \\ \hline
IVHD / IVHD-CUDA & MNIST & - & - & 2  & 1  & 0.01 & Euclidean \\ \hline
IVHD / IVHD-CUDA & Fashion-MNIST & - & - & 2  & 1  & 0.01 & Cosine \\ \hline
IVHD / IVHD-CUDA & Small Norb & - & - & 5 & 1  & 0.01 & Cosine \\ \hline
IVHD / IVHD-CUDA & RCV-Reuters & - & - & 3 & 1  & 0.1  & Cosine \\ \hline
IVHD / IVHD-CUDA & YAHOO & - & - & 2  & 1  & 0.1  & Cosine \\ \hline
t-SNE & All & 40, 80, 120 & - & -  & -  & -  & Euclidean \\ \hline
UMAP & All & - & - & 15,30,50  & -  & -  & Euclidean \\ \hline
TriMap & All & - & 12, 20, 40 & - & -  & -  & Euclidean \\ \hline
PaCMAP & All & - & - & 10, 15, 20 & -  & -  & Euclidean \\ \hline
BH-SNE-CUDA & All & 50 & - & 32  & -  & -  & Euclidean \\ \hline
AtSNE-CUDA & All & 50  & - & 100  & -  & -  & Euclidean \\ \hline
\end{tabular}
\label{tab:parameters}
\end{center}
\end{table}

All metrics were calculated and plotted to obtain the best configuration of a given method (from those presented above). The results were compared locally, and one that obtains the best metrics results was selected for the final drawing of the metrics.

\subsubsection{An N-dimensional ball divided in half.}

\begin{figure}[ht!]
     \centering
        \includegraphics[width=\textwidth]{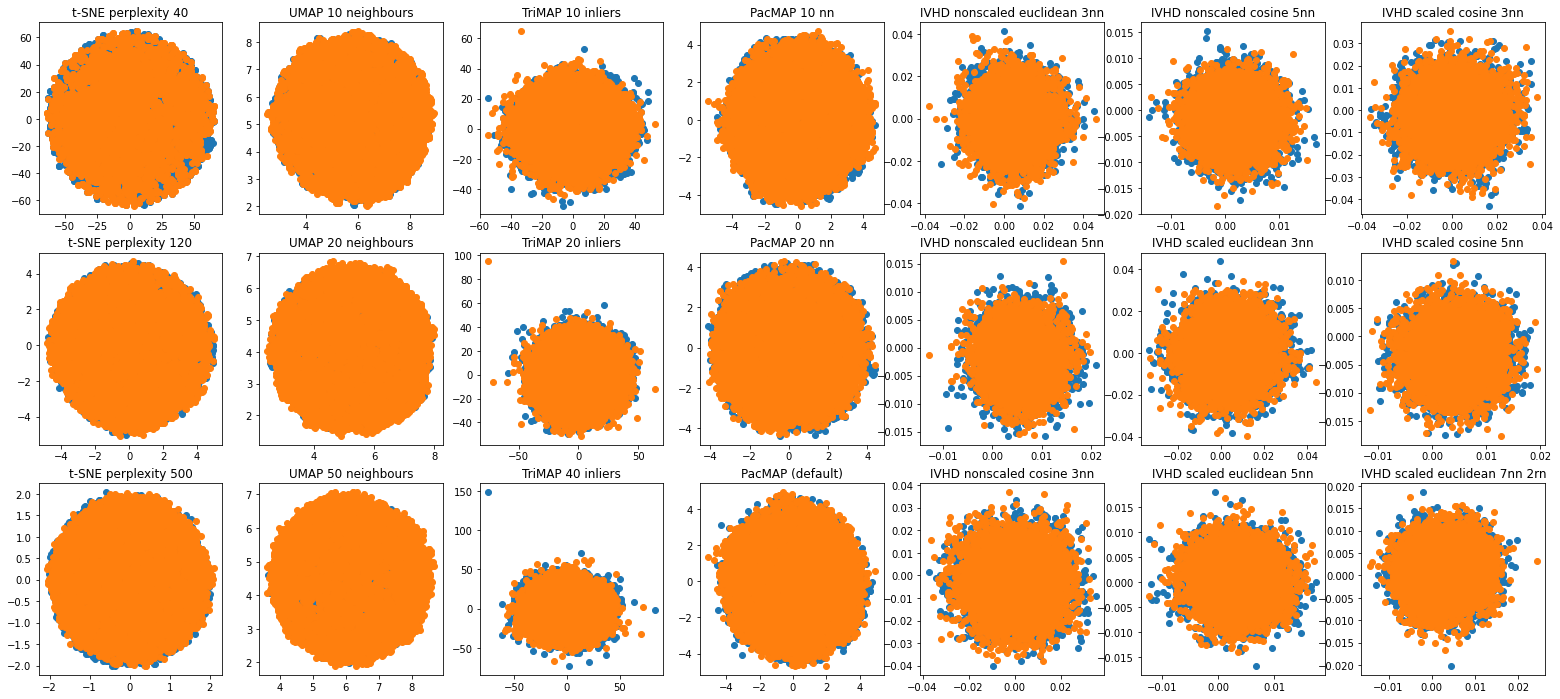}
        \caption{Methods comparison for 10k-sample ball (30-D) splitted into 2 separate classes. }
    \label{fig:appendix_visualization_ball_splitted_into_2_classes_30d}
\end{figure}

\begin{figure}[ht!]
     \centering
        \includegraphics[width=\textwidth]{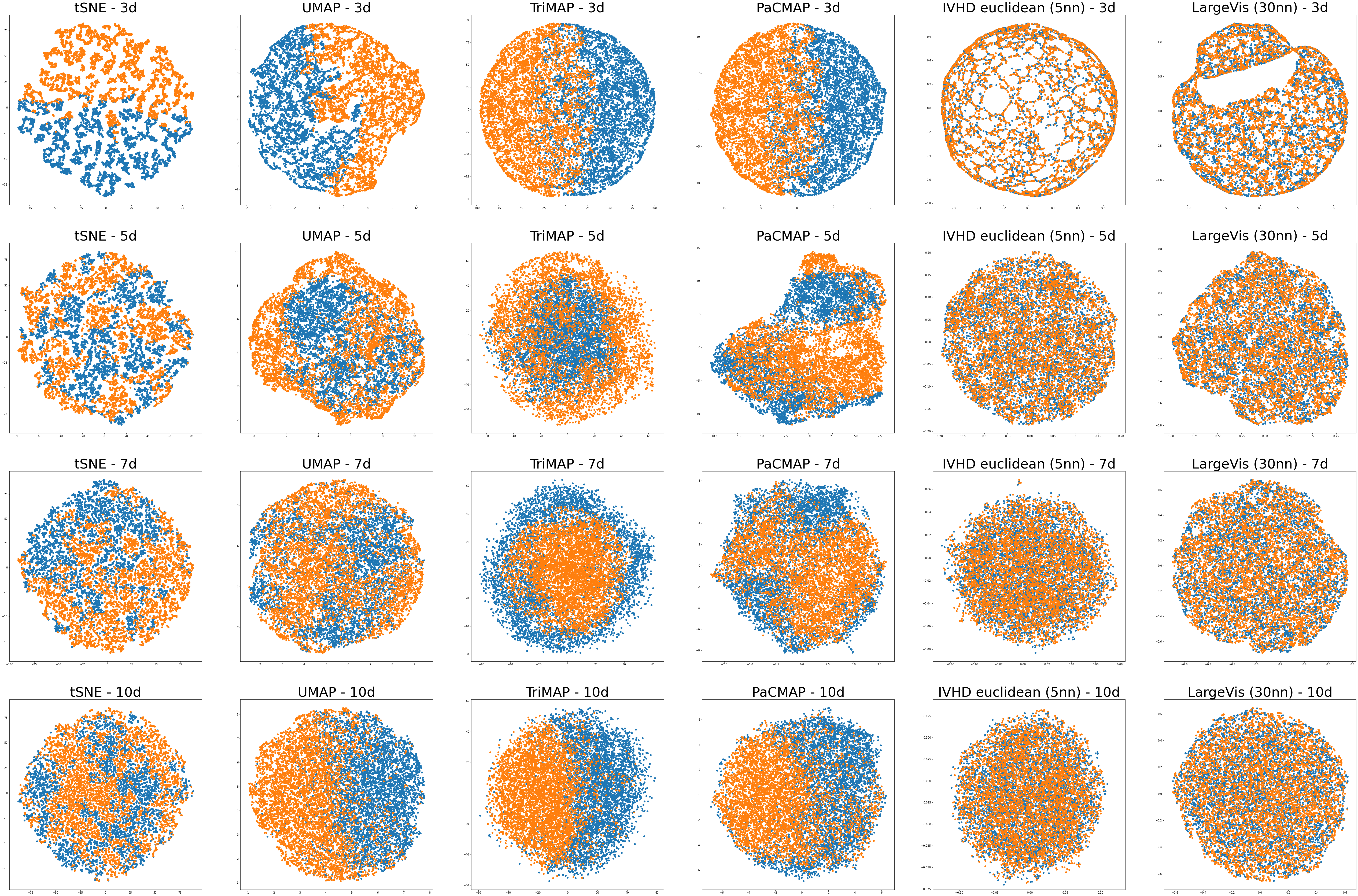}
        \caption{Methods comparison for 10k-sample ball splitted into 2 separate classes. Various dimensionality.}
    \label{fig:appendix_visualization_ball_splitted_into_2_classes_different_dimensionalities}
\end{figure}

All methods were unable to separate the classes for a 30-dimensional sphere. The two classes completely overlap (Fig. \ref{fig:appendix_visualization_ball_splitted_into_2_classes_30d}). Therefore, we performed cross-sectional visualizations for different dimensionalities of the ball (Fig. \ref{fig:appendix_visualization_ball_splitted_into_2_classes_different_dimensionalities}). As you can see, separation is possible only for low-dimensionalities (3D-5D). Any higher dimensionality causes the classes to merge, which is the direct cause of \textit{crowding effect}.

Methods that have coped as such with the separation of low dimensions are UMAP, TriMAP, and PaCMAP (2nd, 3rd, 4th column from the left in Fig. \ref{fig:appendix_visualization_ball_splitted_into_2_classes_different_dimensionalities}). IVHD, as the simplest and fastest method, is unfortunately unable to recognize the features responsible for the separations in such a homogeneous dataset. The same is true for both LargeVis and t-SNE (although t-SNE could separate two classes when embedded from 3-D). For all methods, we observe that they tend to distribute data fairly uniformly around the space, which may potentially contribute to the preservation of local structure and hinder the preservation of global structure. Visually, for the highest dimension (last row in Fig. \ref{fig:appendix_visualization_ball_splitted_into_2_classes_different_dimensionalities}), the TriMap and UMAP methods are methods that create a clear separation of classes in the center of the ball.

\begin{figure}[ht!]
     \centering
        \includegraphics[width=0.8\textwidth]{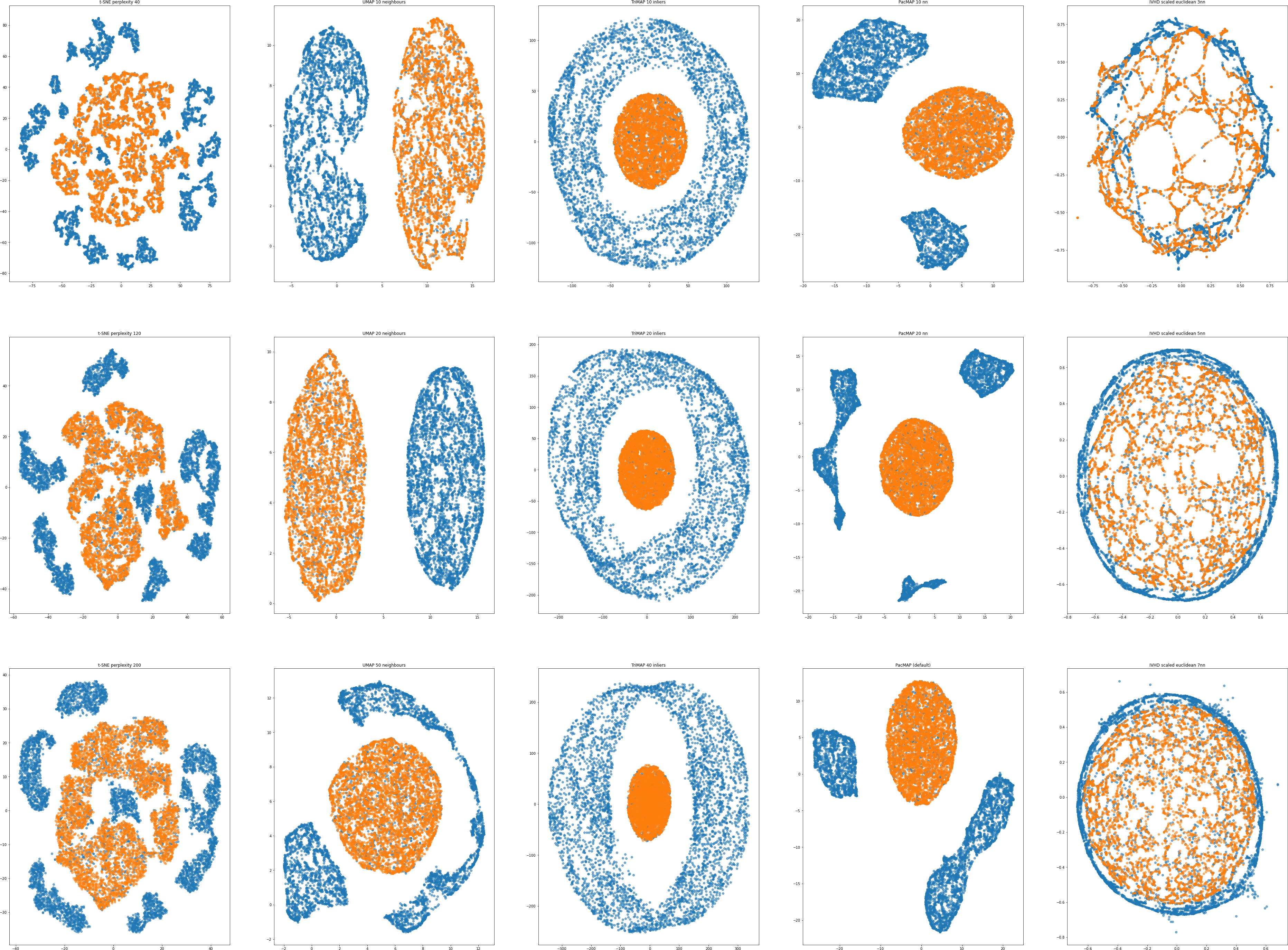}
        \caption{Methods comparison for 10k-sample ball inside empty ball dataset. Both manifolds were generated in 30-D.}
    \label{fig:appendix_visualization_ball_inside_empty_balls}
\end{figure}

In Figures \ref{fig:appendix_visualization_ball_inside_empty_balls}, \ref{fig:appendix_diagrams_ball_inside_empty_balls} we can observe the embedding of the fourth synthetic data set (an N-dimensional ball inside
an N-dimensional ring). The best performance in terms of DR quality is achieved by the TriMAP method. For all the cases, there is a clear separation for two classes, the ball and the empty ball.

\subsubsection{An N-dimensional ball inside of empty ball.}

The best visualization of this dataset was undisputedly created using the TriMap method. There is not only a clear separation of the two classes, but the method was also able to create an inner ball that is bounded by a second hollow ball (a bold sphere). The rest of the methods had trouble rendering this shape. IVHD is the only method, in addition to TriMap, that was able to reflect the overall structure of the classes, but it is not as accurate as in the first case. UMAP and PaCMAP created the internal ball, but its surrounding was chaotic and completely did not reflect the real structure of this dataset.

\begin{figure}[ht!]
    \centering
        \subfloat[DR quality ($R_{NX}(k)$).]{\includegraphics[width=2.8in]{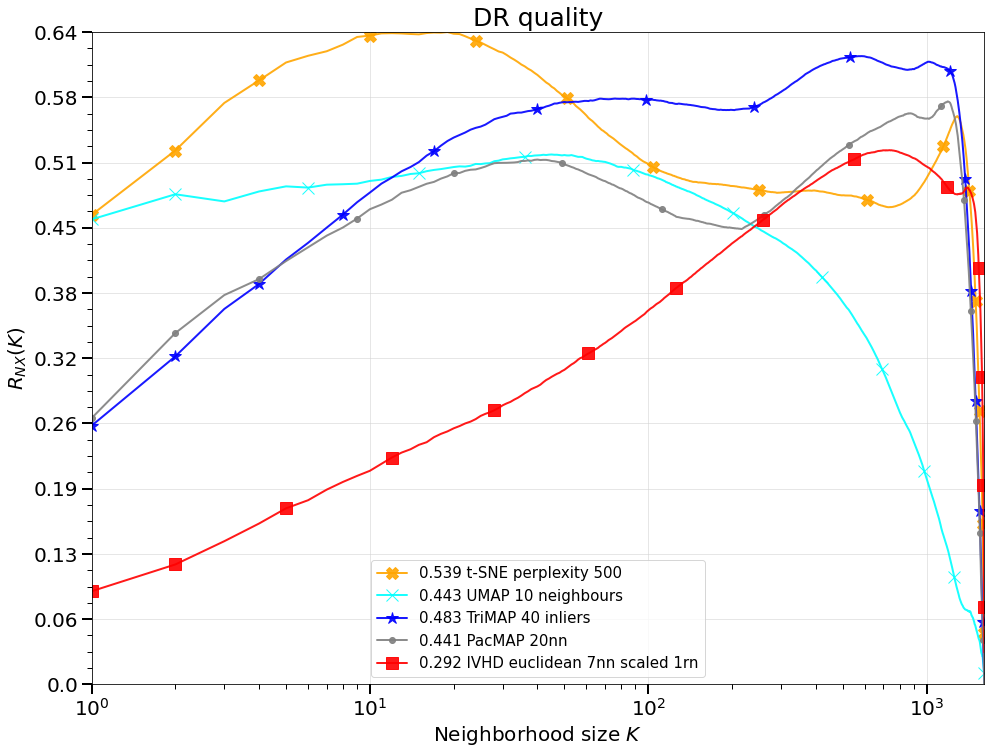}}
        \subfloat[kNN gain ($G_{NN}(k)$).]{\includegraphics[width=2.8in]{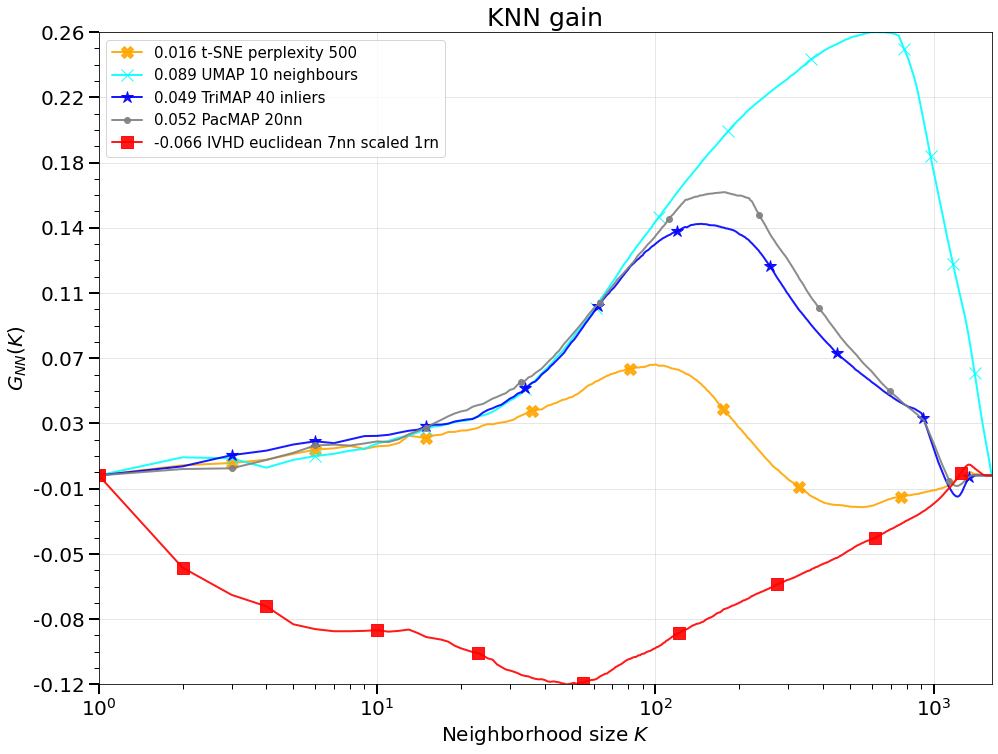}}
        \caption{KNN gain and DR quality obtained for comparison of different DR methods for ball inside empty ball dataset.}
    \label{fig:appendix_diagrams_ball_inside_empty_balls}
\end{figure}

The metrics presented in Fig. \ref{fig:appendix_diagrams_ball_inside_empty_balls} confirm the conclusions described above, in which TriMap achieves the best results (DR quality plot), but UMAP completely fails to preserve quality. It generates very clear two clusters and for this reason it spikes in the kNN gain metric plot.

\subsubsection{Mid-scale datasets}
\label{appendix_visualizations_of_mid_scale_datasets}

Based on the visualizations presented in Figs. \ref{fig:appendix_results_emnist}, \ref{fig:appendix_results_mnist}, \ref{fig:appendix_results_mnist} and the diagrams presented in Figs. \ref{fig:appendix_diagrams_emnist}, \ref{fig:appendix_diagrams_mnist} and \ref{fig:appendix_diagrams_tng} one can make the following observations:

\begin{figure}[ht!]
    \centering
    \includegraphics[width=\textwidth]{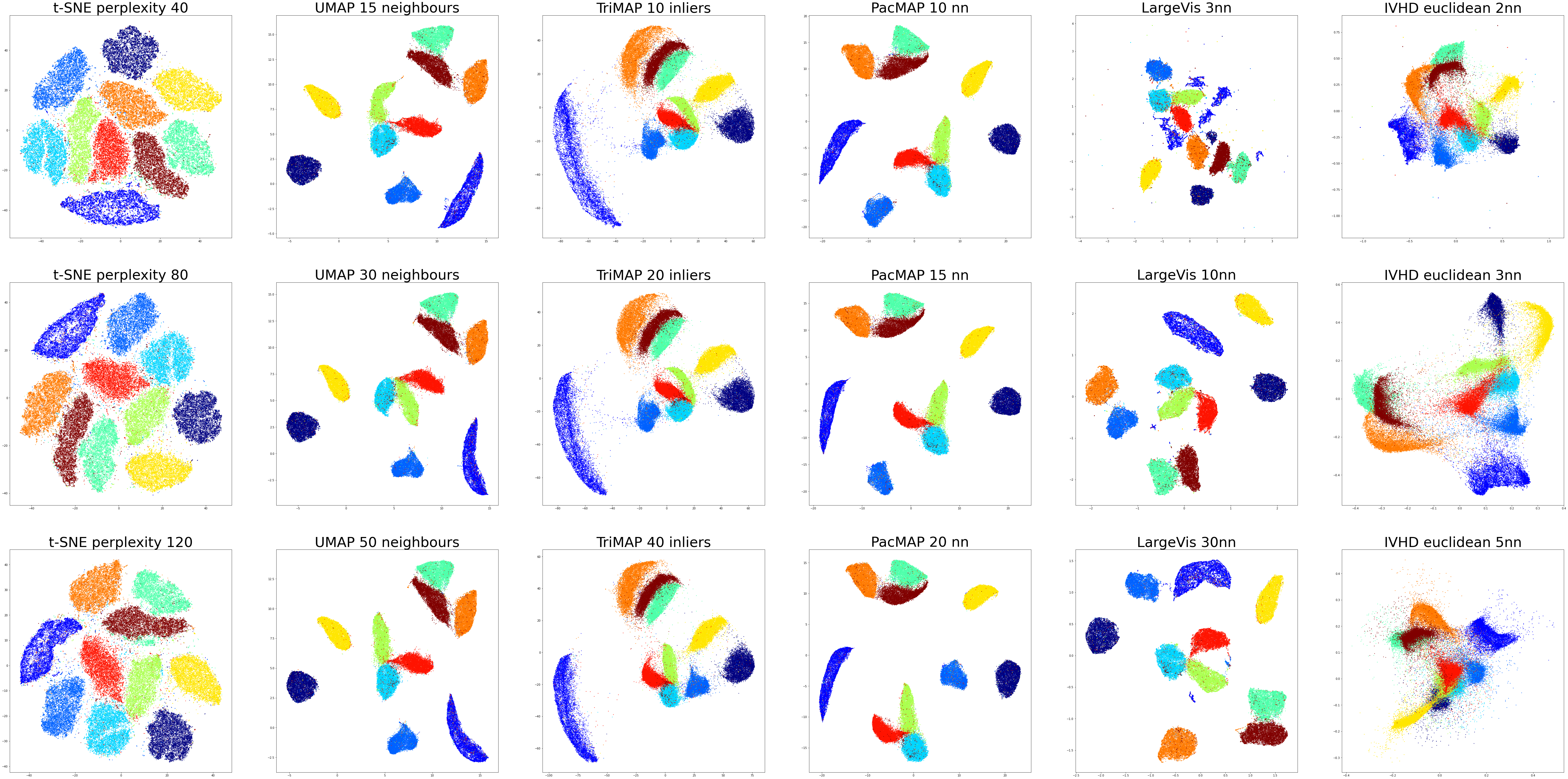}
    \caption{Methods comparison for MNIST dataset.}
    \label{fig:appendix_results_mnist}
\end{figure}

\begin{enumerate}
    \item IVHD applied to both MNIST and EMNIST forms separate clusters of different, mostly elongated shapes (similarly as with FMNIST in Fig. \ref{fig:chapter_5_results_fmnist}). Furthermore, the generated mapping is fuzzy. On the other hand, the LargeVis, PacMAP, t-SNE, and TriMap methods are much better at creating rounded and clearly separated clusters. Additionally, in t-SNE, some classes are mixed and fragmented. In terms of DR quality, LargeVis, UMAP, and PaCMAP are achieving the best results.
    
    \item All methods were more or less capable of preserving the distance ratio between the (high and low dimensional) spaces for the EMNIST and MNIST dataset (Figs. \ref{fig:appendix_diagrams_emnist} and \ref{fig:appendix_diagrams_mnist}). MAP methods seem to have the most condensed points on the diagonal of the Shepard diagram, which indicates the best preservation of the distance ratio. Although in both cases, IVHD obtains the most distorted Shepard diagram.

    \begin{figure}[ht!]
        \centering
        \subfloat[DR quality ($R_{NX}(k)$).]{\includegraphics[width=0.5\textwidth]{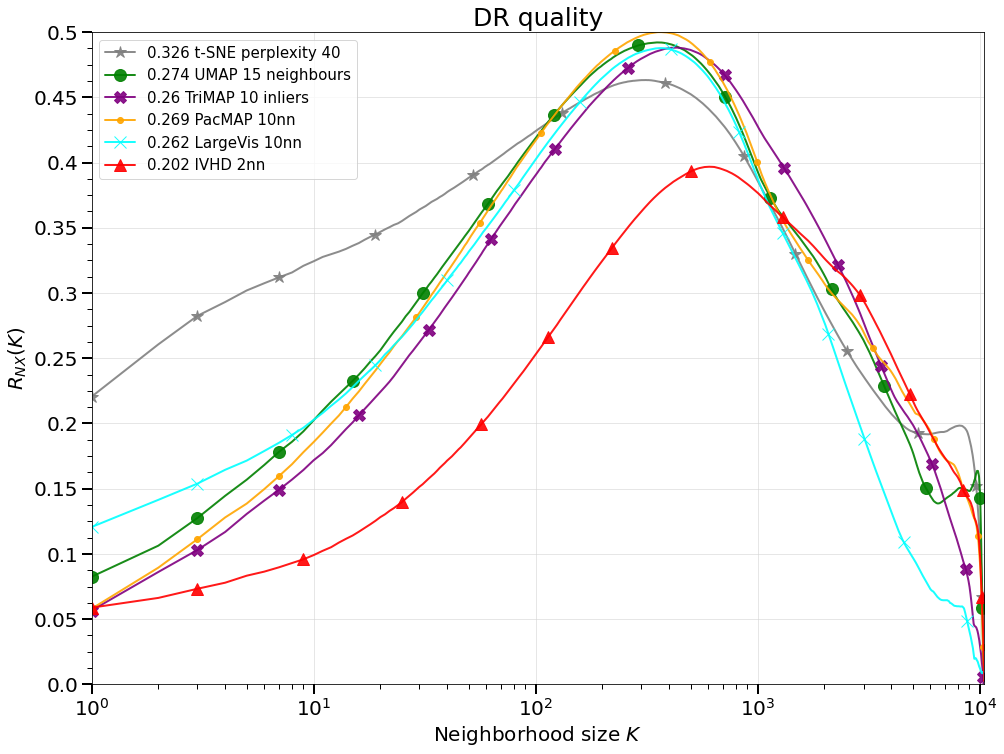}}
        \subfloat[kNN gain ($G_{NN}(k)$).]{\includegraphics[width=0.5\textwidth]{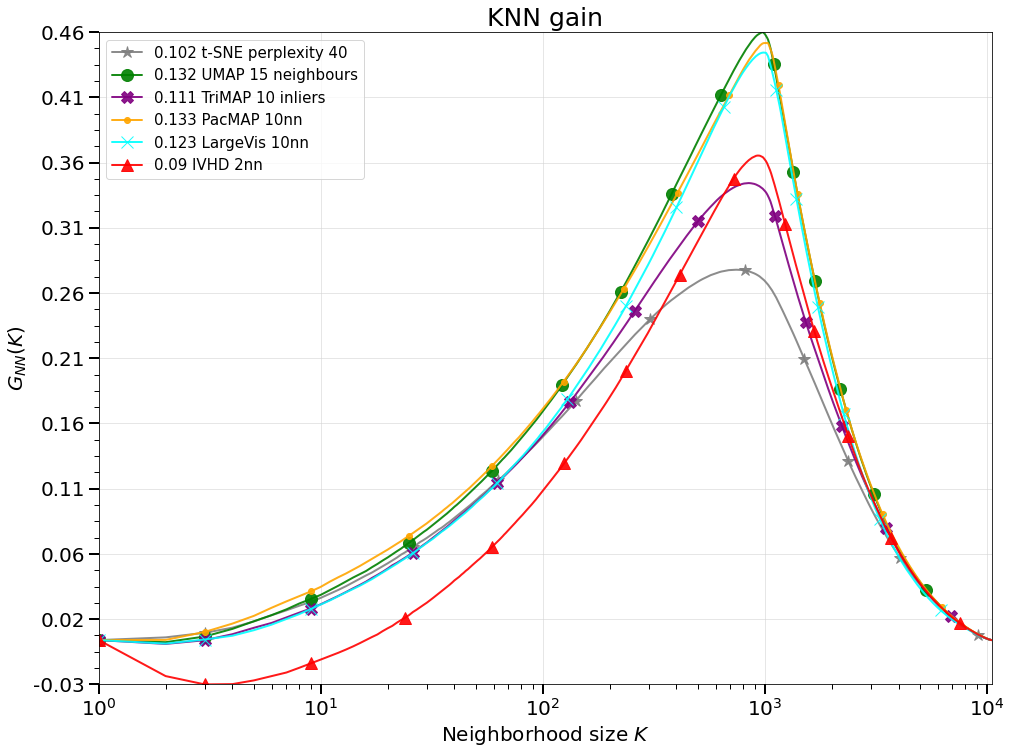}}
        \hfill
        \subfloat[Shepard Diagram ($G_{NN}(k)$).]{\includegraphics[width=\textwidth]{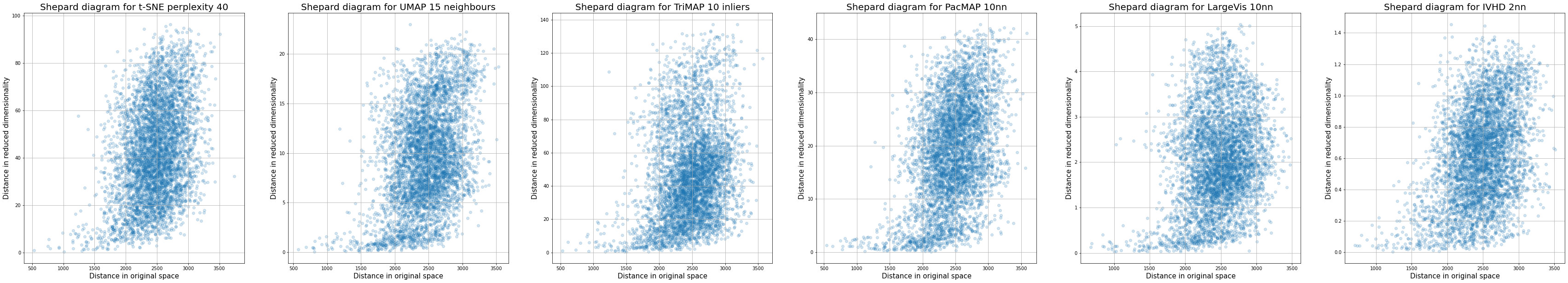}}
        \caption{KNN gain, DR quality and Shepard Diagrams obtained for comparison of different DR methods for MNIST  dataset.}
        \label{fig:appendix_diagrams_mnist}
    \end{figure}

    \item The original IVHD implementation does not separate clusters in the EMNIST data set, as could be with proper adjustments. Therefore, this data set was chosen to be one of the benchmark data sets to show the improvements in IVHD presented in the section \ref{sec:ivhd_improvements}.
    
    \begin{figure}[ht!]
        \centering
        \includegraphics[width=\textwidth]{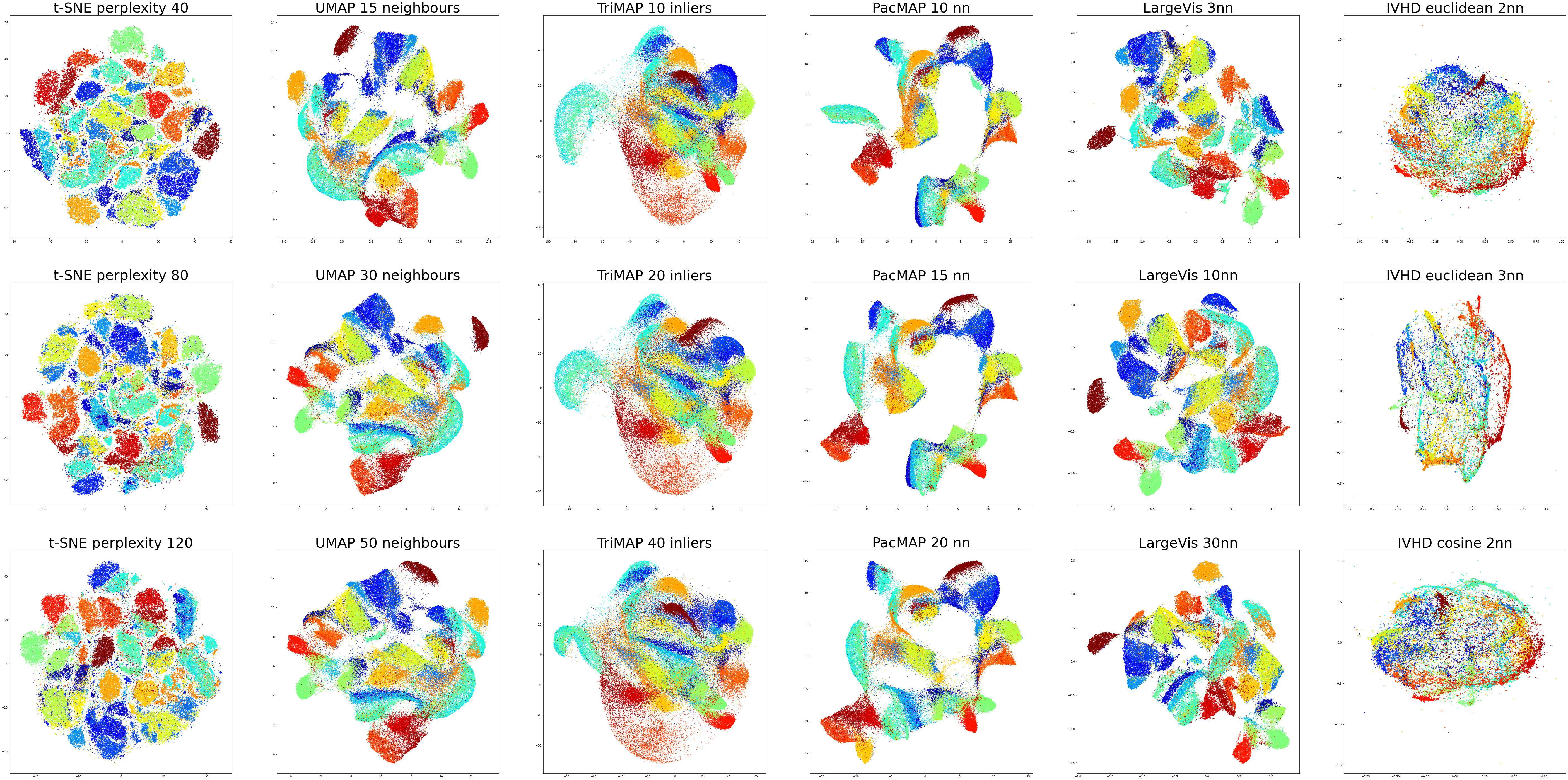}
        \caption{Methods comparison for EMNIST-Letters dataset.}
        \label{fig:appendix_results_emnist}
    \end{figure}

    \item For the 20 newsgroups dataset, we can see that there is a complete separation of classes in the case of the IVHD and LargeVis methods, resulting in very low DR quality and a huge KNN gain. For all methods, the distance ratio shown in the Shepard diagram (Fig. \ref{fig:appendix_diagrams_tng}) is not "linear". This means that there is no preservation of the distance ratio and that the embeddings do not preserve the original data structure. It is quite abject, especially in IVHD and LargeVis case, where we see this extreme cluster separation.
    
    \begin{figure}[ht!]
        \centering
        \subfloat[DR quality ($R_{NX}(k)$).]{\includegraphics[width=0.5\textwidth]{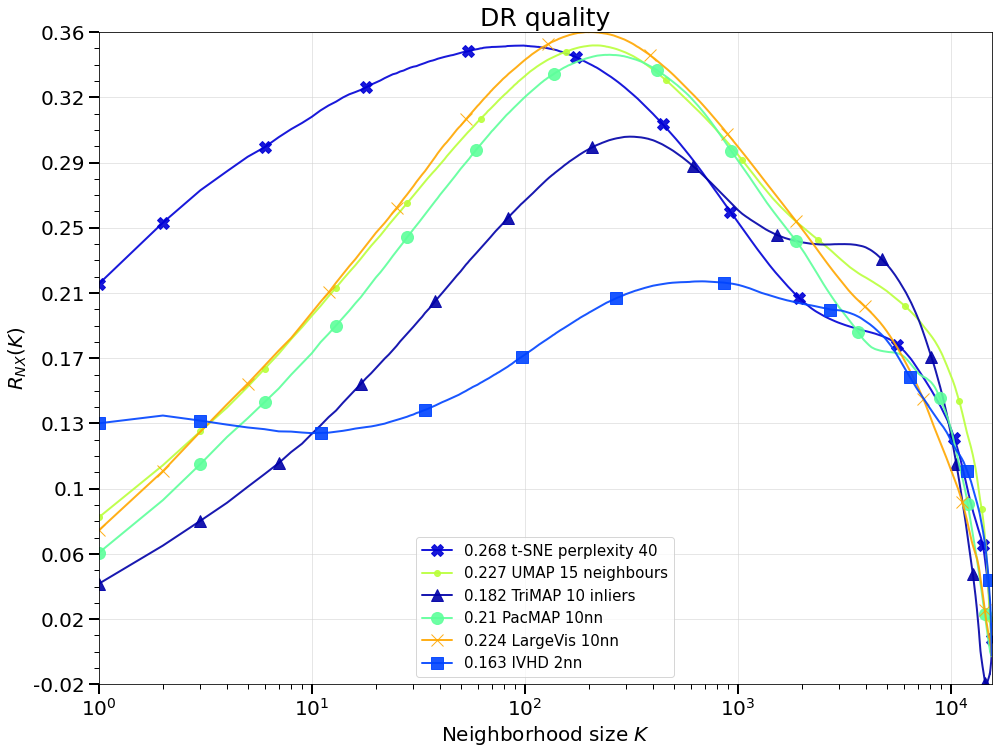}}
        \subfloat[kNN gain ($G_{NN}(k)$).]{\includegraphics[width=0.5\textwidth]{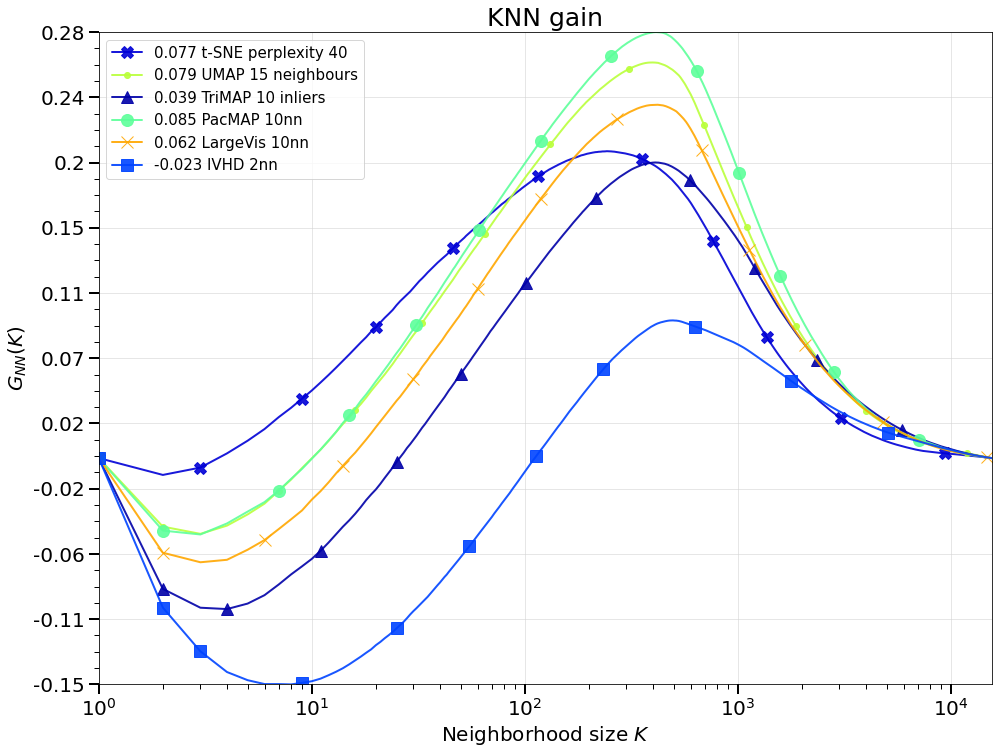}}
        \hfill
        \subfloat[Shepard Diagram ($G_{NN}(k)$).]{\includegraphics[width=\textwidth]{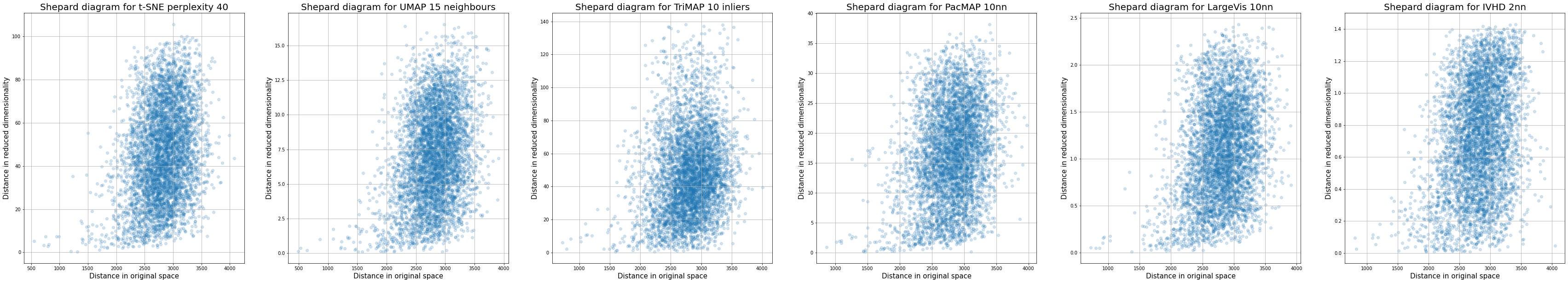}}
        \caption{KNN gain, DR quality and Shepard Diagrams obtained for comparison of different DR methods for EMNIST  dataset.}
        \label{fig:appendix_diagrams_emnist}
    \end{figure}

    \item In terms of DR quality, we can verify that PacMAP has the highest quality in neighborhood size of $50$$-$$1000$. For KNN gain, of course, LargeVis and IVHD obtain the highest score, both being the methods with the highest level of cluster separation. It is important to remember that it does not mean that embedding preserves the overall data structure (as mentioned, DR quality gives us that information).
\end{enumerate}

\begin{figure}[ht!]
    \centering
    \includegraphics[width=\textwidth]{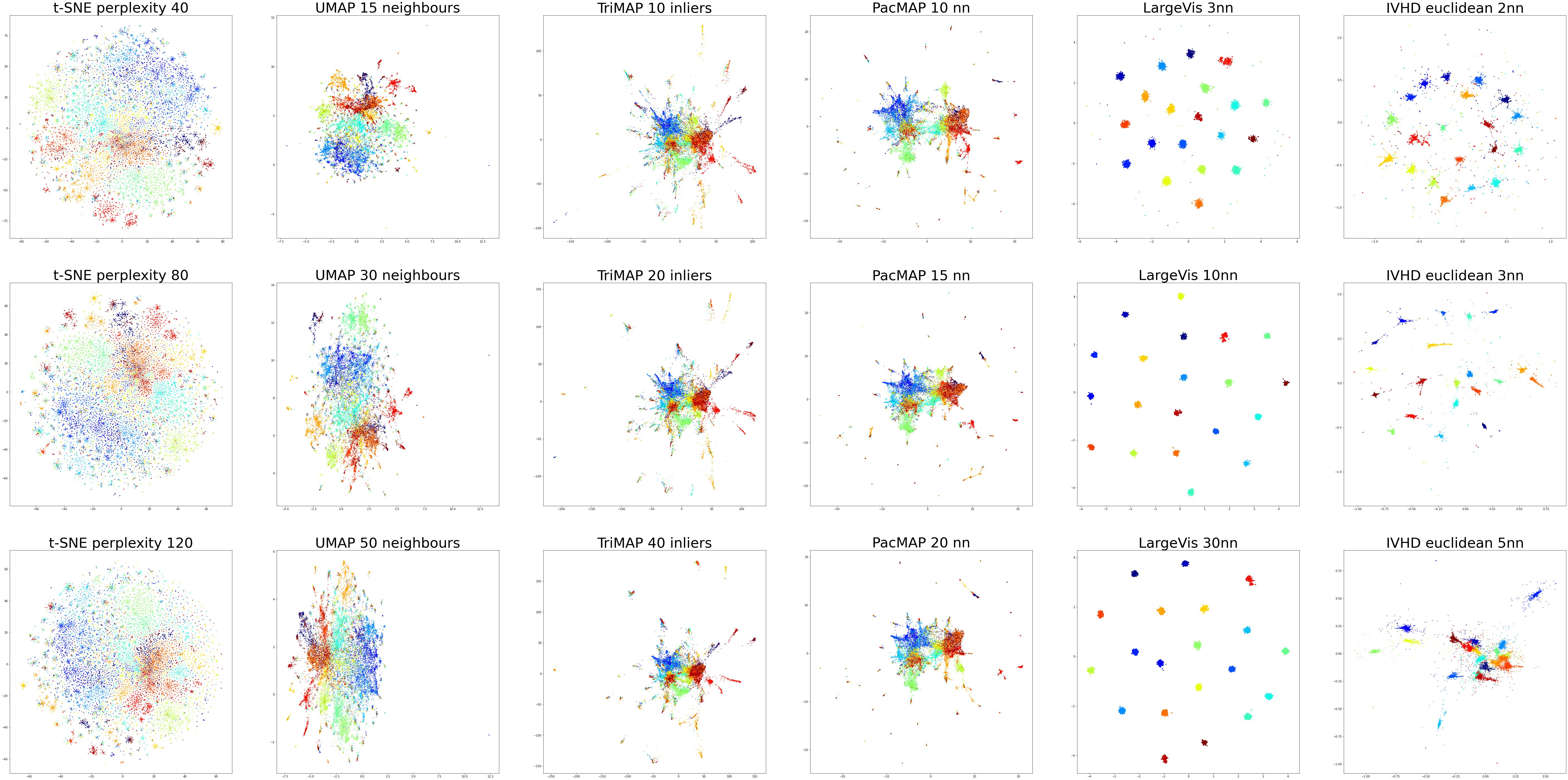}
    \caption{Methods comparison for for 20-News Groups dataset.}
    \label{fig:appendix_results_tng}
\end{figure}

\begin{figure}[ht!]
    \centering
    \subfloat[DR quality ($R_{NX}(k)$).]{\includegraphics[width=0.5\textwidth]{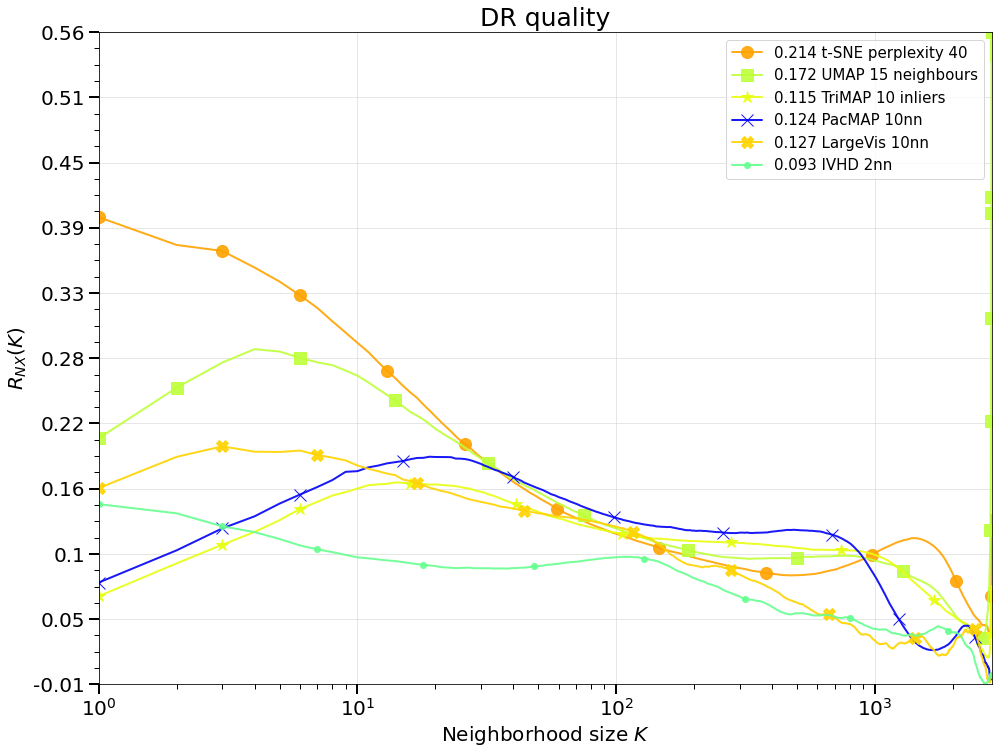}}
    \subfloat[kNN gain ($G_{NN}(k)$).]{\includegraphics[width=0.5\textwidth]{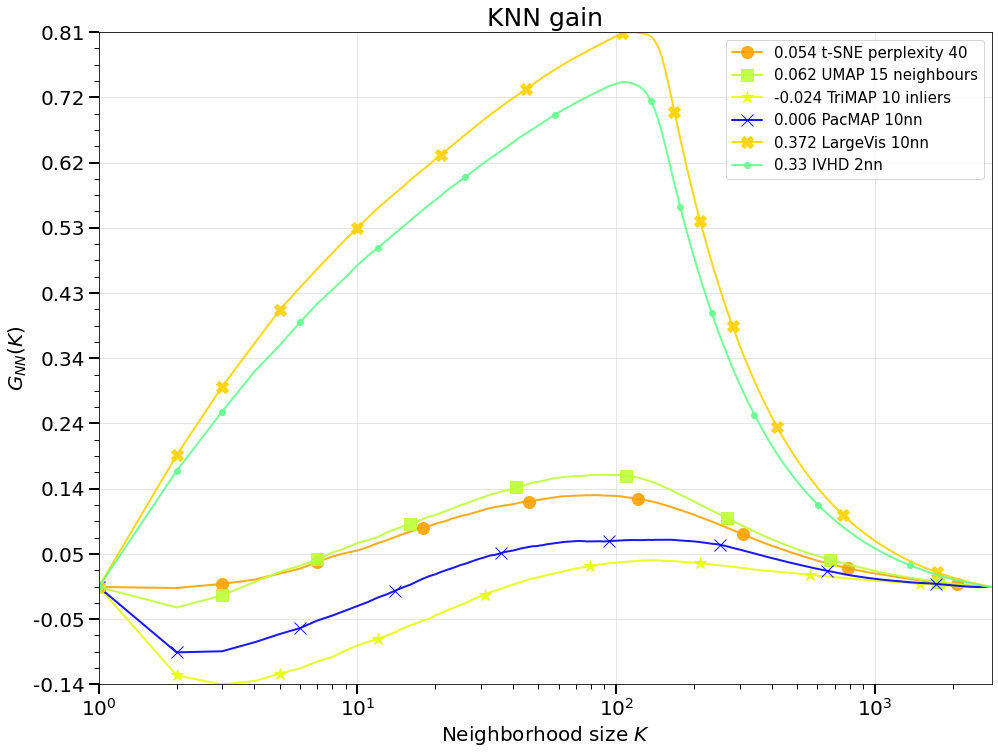}}
    \hfill
    \subfloat[Shepard Diagram ($G_{NN}(k)$).]{\includegraphics[width=\textwidth]{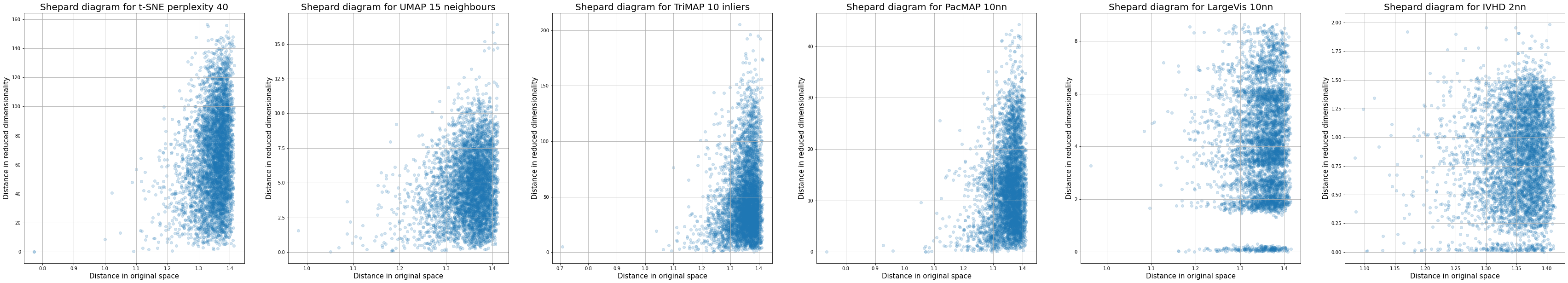}}
    \caption{KNN gain, DR quality and Shepard Diagrams obtained for comparison of different DR methods for 20-News Groups  dataset.}
    \label{fig:appendix_diagrams_tng}
\end{figure}

\bibliographystyle{dissertation_bib}
\bibliography{BIBLIOGRAPHY}

\end{document}